\documentclass{article}

\PassOptionsToPackage{numbers}{natbib}
\usepackage[final]{neurips_2024}

\usepackage[utf8]{inputenc} %
\usepackage[T1]{fontenc}    %
\usepackage{hyperref}       %
\usepackage{url}            %
\usepackage{booktabs}       %
\usepackage{amsfonts}       %
\usepackage{nicefrac}       %
\usepackage{microtype}      %
\usepackage{xcolor}         %
\usepackage{minitoc}
\usepackage{bbm}

\usepackage{amsthm,amssymb}
\usepackage{float,leqno,amssymb,epsf,graphicx}
\usepackage{euscript}
\usepackage{mathtools,thmtools}

\usepackage{subcaption}
\usepackage{rotating}

\newcommand{\pa} {\partial}

\renewcommand{\i}{\ifmmode\mathit{\mathchar"7010 }\else\char"10 \fi}
\renewcommand{\j}{\ifmmode\mathit{\mathchar"7011 }\else\char"11 \fi}
\newcommand{\R}{\mathbb{R}}

\newcommand{\Z}{\mathbb{Z}}
\newcommand{\C}{\mathbb{C}}

\newcommand{\bv}{{\mathbf{v}} }

\renewcommand{\i}{\ifmmode\mathit{\mathchar"7010 }\else\char"10 \fi}
\renewcommand{\j}{\ifmmode\mathit{\mathchar"7011 }\else\char"11 \fi}

\newcommand{\abs}[1]{\left|#1\right|}

\newcommand{\cF}{\EuScript{F}}

\newcommand{\map}{\mathcal{L}}
\newcommand{\cB}{\mathcal{B}}

\newcommand{\er}{\EuScript{E}}
\newcommand{\sol}{\EuScript{S}}
\newcommand{\bX}{\EuScript{X}}

\newcommand{\bN}{\EuScript{N}}

\newcommand{\bE}{\EuScript{E}}

\DeclareMathOperator{\sign}{sgn}

\title{\textsc{Poseidon}: Efficient Foundation Models for PDEs}

\author{
Maximilian Herde$^{1,*}$ \And Bogdan Raoni\'c$^{1,2,*}$ \And Tobias Rohner$^1$  \And Roger Käppeli$^1$  \And Roberto Molinaro$^1$ \And Emmanuel de B\'ezenac$^1$  \And Siddhartha Mishra$^{1,2}$
\AND \textnormal{$^1$Seminar for Applied Mathematics, ETH Zurich, Switzerland}\\
$^2$ETH AI Center, Zurich, Switzerland\\
Correspondence to \texttt{herdem@ethz.ch}
}

\begin{document}

\doparttoc %
\faketableofcontents %

\maketitle

\begin{abstract}
    We introduce \textsc{Poseidon}, a foundation model for learning the solution operators of PDEs. It is based on a multiscale operator transformer, with time-conditioned layer norms that enable continuous-in-time evaluations. A novel training strategy leveraging the semi-group property of time-dependent PDEs to allow for significant scaling-up of the training data is also proposed. \textsc{Poseidon} is pretrained on a diverse, large scale dataset for the governing equations of fluid dynamics. It is then evaluated on a suite of 15 challenging downstream tasks that include a wide variety of PDE types and operators. We show that \textsc{Poseidon} exhibits excellent performance across the board by outperforming baselines significantly, both in terms of sample efficiency and accuracy. \textsc{Poseidon} also generalizes very well to new physics that is not seen during pretraining. Moreover, \textsc{Poseidon} scales with respect to model and data size, both for pretraining and for downstream tasks. Taken together, our results showcase the surprising ability of \textsc{Poseidon} to learn effective representations from a very small set of PDEs during pretraining in order to generalize well to unseen and unrelated PDEs downstream, demonstrating its potential as an effective, general purpose PDE foundation model. Finally, the \textsc{Poseidon} model as well as underlying pretraining and downstream datasets are open sourced, with code being available at \href{https://github.com/camlab-ethz/poseidon}{https://github.com/camlab-ethz/poseidon} and pretrained models and datasets at \href{https://huggingface.co/camlab-ethz}{https://huggingface.co/camlab-ethz}. 
\end{abstract}

\def\thefootnote{*}\footnotetext{Equal contribution}\def\thefootnote{\arabic{footnote}}

\section{Introduction} 
Partial Differential Equations (PDEs) \cite{Evansbook} are referred to as the \emph{language} of physics as they mathematically model a very wide variety of physical phenomena across a vast range of spatio-temporal scales. \emph{Numerical methods} such as finite difference, finite element, spectral methods etc. \cite{NAbook} are commonly used to approximate or \emph{simulate} PDEs. However, their (prohibitive) computational cost, particularly for the so-called many-query problems \cite{MQ}, has prompted the design of various \emph{data-driven} machine learning (ML) methods for simulating PDEs, \cite{scimlrev1,scimlrev2} and references therein. Among them, \emph{operator learning} algorithms have gained increasing traction in recent years. 

These methods aim to learn the underlying PDE solution operator, which maps function space inputs (initial and boundary conditions, coefficients, sources) to the PDE solution. They include algorithms which approximate a \emph{discretization}, on a fixed grid, of the underlying solution operator. These can be based on convolutions \cite{ZhuZab1,BG1}, graph neural networks \cite{worral,bat1,bat2} or transformers \cite{Cao1,VIDON,LOCA,GNOT,OFormer}. Other operator learning algorithms are \emph{neural operators} which can directly process function space inputs and outputs, possibly sampled on multiple grid resolutions \cite{NO,Reno}. These include  DeepONets \cite{chenchen,deeponets}, Fourier Neural Operator \cite{FNO}, SFNO \cite{SFNO}, Geo-FNO \cite{geofno}, Low-rank NO \cite{LNO} and Convolutional Neural Operator \cite{CNO}, among many others.

\begin{figure}[htbp]
    \centering
    \includegraphics[width=\textwidth]{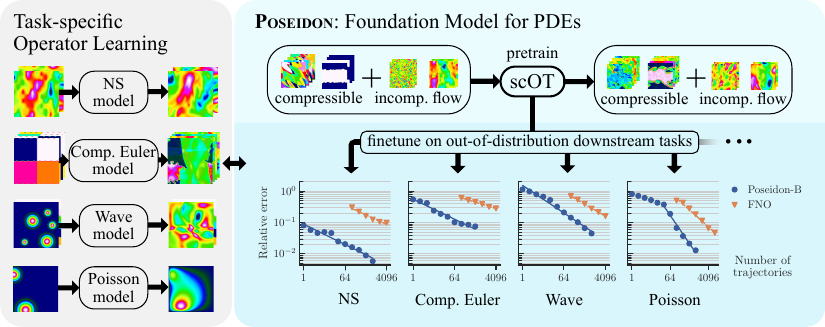}
    \caption{As opposed to PDE-specific operator learning, our pretrained model {\textsc{Poseidon}} is up to multiple orders of magnitude more sample efficient than a task-specific neural operator while also being able to transfer to unseen physics during finetuning.}
    \label{fig:1}
\end{figure}

However, existing operator learning methods are not \emph{sample efficient} as they can require a very large number of training examples to learn the target solution operator with desired accuracy (see Figure \ref{fig:1} or Figure 3 of \cite{CNO}). This impedes their widespread use as 
\emph{task-specific} training data is very expensive to generate either with numerical simulations or measurements of the underlying physical system.  

\emph{How can the number of training samples for PDE learning be significantly reduced?} In this context, can we learn from language modeling and computer vision where a similar question often arises and the current paradigm is to build \emph{foundation models} \cite{FM1}. These \emph{generalist} models are \emph{pretrained}, at-scale,  on large datasets drawn from a diverse set of data distributions. They leverage the intrinsic ability of neural networks to learn \emph{effective representations} from pretraining and are then successfully deployed on a variety of \emph{downstream} tasks by \emph{finetuning} them on a few task-specific samples. Examples of such models include highly successful large language models \cite{gpt,llama}, large multi-modal models\cite{gemini,4M} and foundation models for robotics \cite{rt2}, chemistry \cite{fmchem}, biology \cite{fmbio}, medicine \cite{medgemini} and climate \cite{climax}.  

Despite very recent preliminary attempts \cite{mahoney,icon,UPT,MPP,DPOT,prosepde,shen2024upsefficientlybuildingfoundation}, the challenge of designing such foundation models for PDEs is \emph{formidable}, given the sheer variety of PDEs (linear and nonlinear, steady and evolutionary, elliptic, parabolic, hyperbolic and mixed etc.), the immense diversity of data distributions, wide range of underlying spatio-temporal scales and the paucity of publicly available high-quality datasets. In particular, the very feasibility of designing PDE foundation models rests on the fundamental and unanswered science question of \emph{why pretraining a model on a (very) small set of PDEs and underlying data-distributions can allow it to learn effective representations and generalize to unseen and unrelated PDEs and data-distributions via finetuning?}

The investigation of this open question motivates us here to present the \textsc{Poseidon} family of PDE foundation models. \textsc{Poseidon}, see Figures  \ref{fig:1} and \ref{fig:2},  is based on i) scalable Operator Transformer or scOT, a \emph{multiscale vision transformer} with (shifted) windowed or Swin attention \cite{swinv1,swinv2}, adapted for operator learning, ii) a novel all2all training strategy for efficiently leveraging \emph{trajectories} of solutions of time-dependent PDEs to scale up the volume of training data and iii) an open source  large-scale pretraining dataset, containing a set of novel solution operators of the compressible Euler and incompressible Navier-Stokes equations of fluid dynamics. We evaluate \textsc{Poseidon} on a challenging suite of 15 downstream tasks, comprising of well-established benchmarks in computational physics that encompass linear and nonlinear, time-dependent and independent and elliptic, parabolic, hyperbolic and mixed type PDEs. All of these tasks are \emph{out-of-distribution} with respect to the pretraining data. Moreover, nine out of the 15 tasks even involve PDEs (and underlying physical processes) which are not encountered during pretraining. 

Through extensive experiments, we find that i) \textsc{Poseidon} shows impressive performance across the board and outperforms baselines on the downstream tasks, with significant gains in accuracy and order of magnitude gains in sample efficiency. For instance, on an average (median) over the downstream tasks, \textsc{Poseidon} requires a mere 20 samples to attain the same error level as the widely-used FNO does with 1024 samples. ii) These gains in accuracy and sample efficiency are also displayed on tasks which involve PDEs not encountered during pretraining, allowing us to conclude that \textsc{Poseidon} can \emph{generalize to unseen and a priori unrelated physical processes and phenomena} with a few task-specific training examples and iii) \textsc{Poseidon} scales with model and dataset size, both for the pretraining as well as for downstream tasks and iv) through case studies, we elucidate possible mechanisms via which \textsc{Poseidon} is able to learn \emph{effective representations} during pretraining, which are then leveraged to generalize to unrelated PDEs downstream. Taken together, these results provide the first positive answers to the afore-mentioned fundamental question of the very feasibility of PDE foundation models and pave the way for the further development and deployment of \textsc{Poseidon} as an \emph{efficient general purpose PDE foundation model.} Finally, we also open source the \textsc{Poseidon} model and the entire pretraining and downstream task datasets within the \textsc{PDEgym} database. 
\section{Approach}
\label{sec:2}
\textbf{Problem Formulation.} 
We denote a generic time-dependent PDE as, 
\begin{equation}
\label{eq:pde}
\begin{aligned}
&\pa_t u(x,t) + \map\left(u,\nabla_x u, \nabla^2_x u , \ldots \right) = 0, \quad \forall  x \in D \subset \R^d, t \in (0,T), \\
&\cB(u) = 0, \quad \forall (x,t) \in \pa D \times (0,T), \quad  
u(0,x) = a(x), \quad x \in D 
\end{aligned}
\end{equation}
Here, with a function space $\bX \subset L^p(D;\R^{n})$ for some $1 \leq p < \infty$, $u \in C([0,T];\bX)$ is the solution of \eqref{eq:pde}, $a\in \bX$ the initial datum and $\map,\cB$ are the underlying differential and boundary operators, respectively. Note that \eqref{eq:pde} accommodates both PDEs with high-order time-derivatives as well as PDEs with (time-independent) coefficients and sources by including the underlying functions within the solution vector and augmenting $\map$ accordingly (see {\bf SM}~\ref{sec:dataft} for examples). 

Even \emph{time-independent} PDEs can be recovered from \eqref{eq:pde} by taking the \emph{long-time limit}, i.e., $\lim_{t\rightarrow \infty} u= \overline{u}$, which will be the solution of  the (generic) time-independent PDE, 
\begin{equation}
\label{eq:tipde}
\begin{aligned} 
\map\left(\overline{u}(x),\nabla_x \overline{u}, \nabla^2_x \overline{u}, \ldots \right) &= 0, \quad \forall x \in D, \quad 
\cB(\overline{u}) = 0, \quad \forall x \in \pa D.
\end{aligned}
\end{equation}

Solutions of the PDE \eqref{eq:pde} are given in terms of the underlying \emph{solution operator} $\sol:[0,T]\times \bX\mapsto \bX$ such that $u(t)=\sol(t,a)$ is the solution of \eqref{eq:pde} at any time $t \in (0,T)$. Given a data distribution $\mu \in {\rm Prob}(\bX)$, the \emph{underlying operator learning task (OLT)} is,
\begin{itemize}%
\item [{\bf OLT}:] \emph{Given any initial datum $a \sim \mu$, find an approximation $\sol^{\ast} \approx \sol$ to the solution operator $\sol$ of \eqref{eq:pde}, in order to generate the entire solution trajectory $\{\sol^{\ast}(t,a)\}$ for all $t \in [0,T]$}.
\end{itemize}
It is essential to emphasize here that the learned operator $\sol^\ast$ has to generate the \emph{entire solution trajectory for \eqref{eq:pde}, given only the initial datum (and boundary conditions)}, as this is what the underlying solution operator $\sol$ (and any numerical approximation to it) does.

\begin{figure}[tb]
    \centering
    \includegraphics[width=\textwidth]{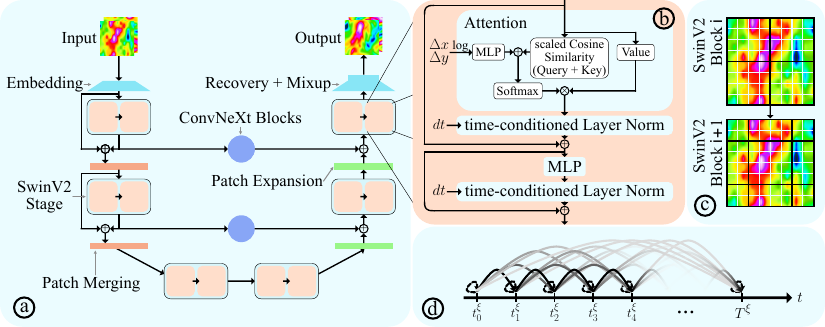}
    \caption{(a) scOT, the model underlying \textsc{Poseidon}; (b) SwinV2 Transformer block; (c) Shifting Window over patch-based tokens with window (patch) boundaries with black (white); (d) all2all Training for time-dependent PDEs.}
    \label{fig:2}
\end{figure}

\textbf{Model Architecture.} The backbone for the \textsc{Poseidon} foundation model is provided by scOT or \emph{scalable Operator Transformer}, see Figure \ref{fig:2} (a-c) for an illustrated summary. scOT is a \emph{hierarchical multiscale vision transformer with lead-time conditioning} that processes lead time $t$ and function space valued initial data input $a$ to approximate the solution operator $\sol(t,a)$ of the PDE \eqref{eq:pde}. 

For simplicity of exposition, we set $d=2$ and $D= [0,1]^2$ as the underlying domain. As in a vision transformer \cite{vit}, any underlying input is first \emph{partitioned into patches and (linearly) embedded into a latent space}. At the level of function inputs $a \in C(D;\R^{n})$, this amounts to the action of the \emph{patch partitioning and embedding} operator $\bv= \hat{{\bf E}}(a)$, with $\hat{{\bf E}}$ defined in {\bf SM} \eqref{eq:ppeo}. This operator transforms the input function into a piecewise constant function, which is constant within patches (subdivisions of the domain $D$), by taking weighted averages and then transforming these piecewise constant values into a $C$-dimensional latent space resulting in output $ \bv \in C(D;\R^C)$. In practice, a discrete version of this operator is used and is described in {\bf SM} \ref{sec:scot2}. 

As shown in Figure \ref{fig:2} (a), this patch embedded output  is then processed through a sequence of \emph{SwinV2 transformer} blocks \cite{swinv1,swinv2}, each of which has the structure of $SW_\ell: C(D;\R^C) \mapsto C(D;\R^C)$,
\begin{equation}
\label{eq:swin}
\begin{aligned}
\bv_\ell = SW_\ell(\bv_{\ell-1})&= \bv_{\ell}^\prime + LN_{\alpha^\ell_2,\beta^\ell_2} (MLP(\bv_{\ell}^\prime)), \\
\bv_{\ell}^\prime &= \bv_{\ell-1} + LN_{\alpha^\ell_1,\beta^\ell_1}(W-MSA(\bv_{\ell-1})).
\end{aligned}
\end{equation}
for layer index $\ell = 1,..., L$. The main building block of a SwinV2 transformer block \eqref{eq:swin} (see Figure \ref{fig:2} (b)) is the \emph{windowed multi-head self attention} operator defined in {\bf SM} \eqref{eq:wmsa} (see {\bf SM} \ref{sec:scot2} for its discrete version). In particular, the attention operator acts only inside each window, which is defined by another (coarser) sub-division of $D$ (see Figure \ref{fig:2} (c)), making it more computationally efficient than a standard vision transformer \cite{vit}. Moreover, the windows are shifted across layers, as depicted in Figure \ref{fig:2} (c), so that all the points in the domain can be attended to, by iteratively shifting windows across multiple layers, see {\bf SM} \ref{sec:scot2} for a detailed description of the SwinV2 block.  

The MLP in \eqref{eq:swin} is defined by {\bf SM} \eqref{eq:mlp}. We follow \cite{film} to propose a \emph{time-conditioning} strategy by introducing a \emph{lead-time conditioned} layer norm in \eqref{eq:swin},
\begin{equation}
\label{eq:LNt}
\begin{aligned}
LN_{\alpha(t),\beta(t)}(\bv)(x) &=  \alpha(t) \odot \frac{\bv(x) - \mu_{\bv}(x)}{\sigma_{\bv}(x)} + \beta(t), \\
\mu_{\bv}(x) &= \frac{1}{C}\sum\limits_{j=1}^C \bv_j(x),~\sigma^2_{\bv}(x) =   \frac{1}{C}\sum\limits_{j=1}^C (\bv_j(x) - \mu_{\bv}(x))^2,
\end{aligned}
\end{equation}
Here, $\alpha(t) = \alpha t + \overline{\alpha}$ and $\beta(t) = \beta t + \overline{\beta}$, with learnable $\alpha,\overline{\alpha},\beta,\overline{\beta}$ although more general (small) MLPs can also be considered. This choice of time embedding enables \textit{continuous-in-time evaluations}.

Finally, as depicted in Figure \ref{fig:2} (a), the SwinV2 transformer blocks \eqref{eq:swin} are arranged in a hierarchical, multiscale manner, within a U-Net style \emph{encoder-decoder} architecture \cite{swinunet}, by employing patch merging (downscaling) and patch expansion (upscaling) operations, (see {\bf SM} \ref{sec:scot2} for a detailed description). Moreover, layers at the same scale, but within the encoder and decoder stages of scOT, respectively, are connected through \emph{ConvNeXt} convolutional layers \cite{convnext}, specified in {\bf SM} \ref{sec:scot2}.

\textbf{Training and Inference.} We denote scOT by  $\sol^\ast_{\theta}:[0,T] \times \bX \mapsto  \bX$, with trainable parameters $\theta \in \Theta \subset \R^p$. For scOT to approximate the solution operator $\sol$ of \eqref{eq:pde}, the parameters $\theta$ need to be determined by minimizing the mismatch between the predictions of scOT and ground truth training data, given in the form of trajectories $\{\sol(t_k,a_i)\}$, for $0 \leq k \leq K$ and $1 \leq i \leq M$, with $a_i \sim \mu$ and $0=t_0 < t_1 <\ldots t_k <\ldots <t_K=T$, being the time points at which the data is sampled. We assume that the data is sampled at the same timepoints for each sample $a_i$ for simplicity. For training, it is natural to consider the loss function,
\begin{equation}
\label{eq:lfi}
\overline{\map}(\theta) := \frac{1}{M (K+1)}\sum\limits_{i=1}^{M} \sum\limits_{k=0}^{K} \|\sol^\ast_{\theta}(t_k,a_i) - \sol(t_k,a_i) \|^p_{L^p(D)},
\end{equation}
with the (spatial) integral in \eqref{eq:lfi} being replaced by a quadrature at some underlying sampling points and $p=1$ in our paper. Thus, we use $K+1$ samples per trajectory in order to train our model. 

Given the fact that scaling up available training data is necessary for successful foundation models \cite{kap1}, we propose a \emph{novel training strategy} that further leverages the \emph{structure} of the time-dependent PDE \eqref{eq:pde} to increase the amount of training data. To this end, we consider the modified loss function, 
\begin{equation}
\label{eq:lfij}
\widehat{\map}(\theta) := \frac{1}{M\widehat{K}}\sum\limits_{i=1}^{M} \sum\limits_{k,{\bar{k}}=0, k \leq {\bar{k}}}^{K} \|\sol(t_{\bar{k}}-t_k,u_i(t_k)) - \sol^{\ast}_{\theta}(t_{\bar{k}}-t_k,u_i(t_k))\|^p_{L^p(D)},
\end{equation}
with $u_i(t_k) = \sol(t_k,a_i)$ (approximately) solving \eqref{eq:pde} and $\widehat{K} =\frac{(K+1) (K+2)}{2}$. 
In other words, we leverage the fact that the solution operator of \eqref{eq:pde} possesses a \emph{semi-group property} and one can realize,
\begin{equation}
\label{eq:caus}
u(t^\ast) = \sol(t^\ast,a) = \sol(t^{\ast}-t,u(t)) = \sol(t^{\ast}-t,\sol(t,a)),~\forall~0\leq t \leq t^\ast \leq T, 
\end{equation}
and any initial condition $a$. We term this use of all possible data pairs $(u(t_k),u(t_{\bar{k}}))$ with $k \leq \bar{k}$, see Figure \ref{fig:2} (d) for a visual representation, within a trajectory as \emph{all2all training} and observe that it allows us to utilize \emph{quadratic} $\mathcal{O}(K^2)$ samples per trajectory, when compared to the linear $K$ samples used for training corresponding to the \emph{vanilla} loss function \eqref{eq:lfi}. In practice, we consider a relative form of Equation \ref{eq:lfij} to balance out different scales of different operator outputs, see {\bf SM} \ref{sec:mb} for details.

Once scOT has been trained with (stochastic) gradient descent to find a (local) minimum $\theta^\ast$ of the all2all loss function \eqref{eq:lfij}, the trained model, denoted as $\sol^\ast_{\theta^\ast}$ can be deployed for inference for any initial condition $a \in \bX$ and for any $t \in \R_+$ by directly applying $\sol^\ast_{\theta^\ast}(t,a)$ to provide continuous-in-time evaluation of the entire trajectory. However, it might be advantageous to infer using \emph{autoregressive rollouts} \cite{AR}. To this end, we consider a sequence 
$0=t^{\ast}_0 < t^{\ast}_1 < \ldots <t^{\ast}_\kappa=t$. Then, the rollout,
\begin{equation}
\label{eq:ar}
\sol(t,a) \approx          \sol^{\ast}_{\theta^\ast}\left(t^{\ast}_{\kappa}-t^{\ast}_{\kappa-1},\sol^{\ast}_{\theta^\ast}(\ldots\ldots\sol^{\ast}_{\theta^\ast}\left(t^{\ast}_2-t^{\ast}_1,\sol^{\ast}_{\theta^\ast}(t^{\ast}_1,a)\right)\right),
\end{equation}
of $\kappa$ successive applications of the trained scOT approximates the solution operator at any time $t$. 

\textbf{Pretraining.} The key point in the development of any foundation model is the \emph{pretraining} step, in which the model is trained on a diverse set of data distributions, rather than just on data drawn from one specific operator. To formulate pretraining and subsequent steps precisely, we introduce index sets $\Lambda,\Xi$ and let $\lambda \in \Lambda$ and $\xi \in \Xi$ correspond to indexing the PDE type and the data-distribution, respectively. To see this, we fix any $\lambda \in \Lambda,\xi \in \Xi$ and tag the differential and boundary operators $\map,\cB$ in the PDE \eqref{eq:pde} by $\map^\lambda$ and $\cB^\lambda$. Similarly the initial distribution $\mu$ in \eqref{eq:pde} is tagged by $\mu^\xi$ and the resulting solution operator for PDE \eqref{eq:pde} with $\map^\lambda,\cB^\lambda$ and initial datum $a \sim \mu^\xi$ is denoted by $\sol^{\lambda,\xi}$. In other words, $\Lambda,\Xi$ indexes the entire set of PDEs and data distributions that we consider.

Next,  we fix index sets, $\widehat{\Lambda} \subset \Lambda$ and $\widehat{\Xi} \subset \Xi$ and consider a set of PDEs \eqref{eq:pde}, indexed by $\lambda \in \widehat{\Lambda}$ and with data distributions $\mu^\xi$, indexed by $\xi \in \widehat{\Xi}$ as the \emph{pretraining dataset}, which consists of the corresponding trajectories, $\{\sol^{\lambda,\xi}(t,\cdot)\}$, for all $t$ and all $(\lambda,\xi) \in (\widehat{\Lambda}, \widehat{\Xi})$.   

Let $n^{\widehat{\Xi}}$ be the maximum number of components of the solution vectors for all the operators in the pretraining dataset. By including additional (constant $0$ over space and time) components, we augment the relevant solution operators (for which the number of components is below $n^{\widehat{\Xi}}$ ) such that for each $\lambda \in \widehat{\Lambda}, \xi \in \widehat{\Xi}$, all the input functions have the same number of $n^{\widehat{\Xi}}$ components (channels). These inputs are fed into a scOT model $\Pi^{\widehat{\Lambda},\widehat{\Xi}}_\theta:[0,\widehat{T}]\times L^p(D;\R^{n^{\widehat{\Xi}}}) \mapsto C([0,\widehat{T}];L^p(D;\R^{n^{\widehat{\Xi}}}))$, with $\widehat{T}$ being the supremum over all the final times in the pretraining dataset. The trainable parameters $\theta$ of this pretrained model are then determined by \emph{minimizing the mismatch between model predictions and ground truth over all PDEs and data distributions in the pretraining dataset} resulting in, 
\begin{equation}
\label{eq:lfpos}
\begin{aligned}
\Pi^{\widehat{\Lambda},\widehat{\Xi}}_{\ast} &= \Pi^{\widehat{\Lambda},\widehat{\Xi}}_{\theta^\ast},~{\rm with}\quad \theta_\ast = {\rm argmin}_{\theta \in \Theta} \frac{1}{|\widehat{\Lambda}||\widehat{\Xi}|}\sum\limits_{\lambda \in \widehat{\Lambda}} \sum\limits_{\xi \in \widehat{\Xi}} \widehat{\map}^{\lambda,\xi}(\theta)\ ,
\end{aligned}
\end{equation}
with $\widehat{\map}^{\lambda,\xi}$ obtained by replacing $\sol$ and $\sol^\ast_\theta$ in \eqref{eq:lfij} with $\sol^{\lambda,\xi}$ and $\Pi^{\widehat{\Lambda},\widehat{\Xi}}_\theta$, respectively. %

\textbf{Finetuning.} To \emph{finetune} the pretrained foundation model $\Pi^{\widehat{\Lambda},\widehat{\Xi}}_{\ast}$ for any downstream task, corresponding any specific solution operator $\sol^{\lambda,\xi}$ for any $\lambda\in \Lambda, \xi \in \Xi$, we decompose the vector of learnable parameters $\theta \in \Theta \subset \R^p$ as $\theta = [\widehat{\theta},\widetilde{\theta},\widetilde{\theta}^{\mathcal{N}}]$, with $\widehat{\theta} \in \R^{\hat{p}}$, $\widetilde{\theta} \in \R^{\tilde{p}}$, and $\widetilde{\theta}^{\mathcal{N}} \in \R^{\tilde{p}_{\mathcal{N}}}$ and $\hat{p} + \tilde{p} + \tilde{p}_{\mathcal{N}} = p$, with $\tilde{p}, \tilde{p}_{\mathcal{N}} \ll \hat{p}$. A gradient descent step for finetuning is then written as,
\begin{equation}
\label{eq:gdfm}
\begin{aligned}
\forall r \geq 1, \quad  &[\widehat{\theta}_{r+1},\widetilde{\theta}_{r+1}, \widetilde{\theta}^{\mathcal{N}}_{r+1}] = [\widehat{\theta}_{r},\widetilde{\theta}_{r},\widetilde{\theta}^{\mathcal{N}}_{r}] - [\widehat{\eta}_r,\widetilde{\eta}_r,\widetilde{\eta}^{\mathcal{N}}_{r}] \nabla_{\theta} \widehat{\map}^{\lambda,\xi}(\theta_r), \\
&\widehat{\theta}_0 = \widehat{\theta}_\ast, \quad \widetilde{\theta}^{\mathcal{N}}_{0} = \widetilde{\theta}^{\mathcal{N}}_{\ast}, \quad \widetilde{\theta}_0  \sim \widetilde{P}, \quad \widetilde{P}\in {\rm Prob}(\R^{\tilde{p}}).
\end{aligned}
\end{equation}
Hence, during finetuning, a subset of parameters $\widetilde{\theta}$ of the foundation model are trained from scratch with random initializations, whereas the complementary, much larger subset of $\widehat{\theta}$ and $\widetilde{\theta}_{\mathcal{N}}$ is initialized by \emph{transferring} the corresponding parameters from the pretrained model. When $\lambda \notin \widehat{\Lambda}$, $\widetilde{\theta}$ consists of the \emph{embedding/recovery} parameters. On the other hand, if $\lambda \in \widehat{\Lambda}$, then all trainable parameters, including the patch embeddings/recovery, are initialized with the corresponding parameters of the pretrained model. However, the corresponding learning rate $\widetilde{\eta}_r \gg \widehat{\eta}_r$ in \eqref{eq:gdfm} is much higher. Similarly, the time embeddings $\widetilde{\theta}^{\mathcal{N}}$, i.e., the trainable parameters in the layer-norm operators \eqref{eq:LNt} are always initialized from the corresponding time embeddings in the pretrained model but finetuned with a higher learning rate $\tilde{\eta}^{\mathcal{N}}$. 
\section{Experiments}
\label{sec:3}
\textbf{Pretraining Dataset.} We pretrain \textsc{Poseidon} on a dataset containing 6 operators, defined on the space-time domain $[0,1]^2 \times [0,1].$ 4 of these operators (CE-RP, CE-KH, CE-CRP, CE-Gauss) pertain to the compressible Euler equations ({\bf SM} \eqref{eq:ceuler}) of gas dynamics and 2 (NS-Sines, NS-Gauss) to the incompressible Navier-Stokes equations ({\bf SM} \eqref{eq:NS}) of fluid dynamics, see {\bf SM} Table \ref{tab:SM1} for abbreviations and {\bf SM} \ref{sec:datapt} for a detailed description of these datasets. These datasets have been selected to highlight different aspects of the PDEs governing fluid flows (shocks and shear layers, global and local turbulent features, and mixing layers etc.). The pretraining dataset contains 9640 and 19640 trajectories for the Euler and Navier-Stokes operators, respectively, leading to a total of 77840 trajectories. Each trajectory is uniformly sampled at 11 time snapshots. Within the all2all training procedure (Section \ref{sec:2}), this implies a total of $66$ input-output pairs per trajectory, leading to approx 5.11M training examples in the pretraining dataset. 

\textbf{Downstream Tasks.} To evaluate \textsc{Poseidon} (and the baselines), we select a suite of 15 challenging downstream tasks, see {\bf SM} Table \ref{tab:SM2} for abbreviations and {\bf SM} \ref{sec:dataft} for detailed description. Each of these tasks is a (variant of) well-known benchmarks for PDEs in the numerical analysis and computational physics literature and corresponds to a distinct PDE solution operator. They have also been selected for their diversity in terms of the PDE types as they contain linear (4) and nonlinear (11), time-dependent (12) and time-independent (3), elliptic (2), parabolic (1), hyperbolic (4) and mixed-type (8). The tasks also cover a wide gamut of physical processes across a range of spatio-temporal scales. Moreover, we emphasize that each of the downstream tasks is \emph{out-of-distribution} with respect to the pretraining data. While 6 of them do pertain to the Euler and Navier-Stokes equations seen in the pretraining dataset but with very different data distributions, the remaining 9 involve PDEs not seen during pretraining. These include 3 (NS-Tracer-PwC, FNS-KF, GCE-RT) which add new physical processes (tracer transport, forcing, gravity) to the Navier-Stokes and Euler equations. 3 more (Wave-Gauss, Wave-Layer, ACE) involve completely new time-dependent PDEs (Wave Eqn., Allen-Cahn Eqn.) and the final 3 (SE-AF, Poisson-Gauss, Helmholtz) \emph{even consider time-independent PDEs}, which is in stark contrast to the pretraining dataset where only 2 time-dependent PDEs are covered. For these steady state PDEs, we finetune them by using the interpretation of the PDE \eqref{eq:tipde} as a \emph{long-time limit} of the time-dependent PDE \eqref{eq:pde} with a normalized lead time of $1$. Finally, the tasks have also been selected to probe the ability of the foundation model to handle different \emph{task or operator types.} To this end, we point out that all the operators in the pretraining dataset simply map the initial conditions to the solution at later times in time-dependent fluid flows on the two-dimensional unit square with \emph{periodic boundary conditions.} While some of the downstream tasks (8 out of 15) do pertain to this type of operators, the remaining (7 out of 15) tasks involve different types of operators which include operators mapping the coefficients or PDE parameters to the PDE solution (5 out of 15), forcing term to the PDE solution (2) and domain shape to the PDE solution. Moreover, many of the downstream tasks are with non-periodic boundary conditions while one of them is even on a non-Cartesian domain.  Thus, these downstream tasks deviate from the setup of the pretraining operators and provide a hierarchy of challenges for any foundation model.

\textbf{Models and Baselines.} We consider three different \textsc{Poseidon} models: i) \textsc{Poseidon}-T with $\approx 21$M parameters, ii) \textsc{Poseidon}-B with $\approx 158$M parameters, and iii) \textsc{Poseidon}-L with $\approx 629$M parameters. The detailed specifications of each of these models is provided in {\bf SM} \ref{sec:pm}. As baselines, in addition to the standalone scOT, we use trained from scratch neural operators in the form of the widely used FNO \cite{FNO} and recently proposed CNO \cite{CNO}, each augmented with time-conditioned instance normalizations. Foundation model baselines are provided by  MPP-aVIT (MPP) \cite{MPP} and we also pretrain a CNO \cite{CNO} model (see details in {\bf SM} \ref{sec:cnofm}) on our pretraining dataset, resulting in an additional foundation model baseline termed CNO-FM, see {\bf SM} \ref{sec:mb} for details on baselines.

\textbf{Evaluation Metrics.} All the models and baselines are evaluated on each task in terms of the relative $L^1$ error at the underlying final time. This choice is motivated by the fact that errors tend to grow over time, making final time prediction harder than any time-averaged quantities, see {\bf SM} \ref{sec:tdep}. This also corresponds well to the interpretation of time-independent PDEs as long-time limits of \eqref{eq:pde}.  Following \cite{kap1} that advocates this approach for LLMs, we evaluate all models in terms of \emph{scaling} curves which plot the test error for each task vs. the number of task-specific training examples, see {\bf SM} \ref{sec:perf_downstream_scalings}. To extract further information from scaling plots, we introduce two evaluation metrics,
\begin{equation}
\label{eq:met}
{\rm {\bf AG}_S(model)} := \frac{\bE_S({\rm FNO})}{\bE_S({\rm model})}, \quad {\rm {\bf EG}_S(model)} := \frac{S}{s}, ~ {\rm where}~ \bE_s({\rm model}) = \bE_S({\rm FNO}),
\end{equation}
with $\bE_S({\rm model})$ being the relative error (at final time) for the model with $S$ trajectories. Thus, \emph{Accuracy Gain} ${\bf AG}_S$ measures how accurate the model is w.r.t. FNO for a given number ($S$) of samples while \emph{Efficiency Gain} ${\bf EG}_S$ measures how much fewer (greater) number of samples the model needs to attain the same error level as FNO trained on $S$ samples. {\bf AG} is the relevant metric for the \emph{limited compute} regime whereas {\bf EG} is relevant for the \emph{limited data} regime.

\textbf{\textsc{Poseidon} performs very well on all downstream tasks.} From the scaling plots {\bf SM} Figures \ref{fig:scaling_ns_pwc} to \ref{fig:scaling_helmholtz}, we observe that \textsc{Poseidon} readily outperforms FNO on \emph{all the 15 downstream tasks}. This point is further buttressed by Table \ref{tab:1}, where the {\bf EG} and {\bf AG} \eqref{eq:met} metrics are presented (see also {\bf SM} Table \ref{tab:full_res} for these metrics for the \textsc{Poseidon}-B and -T models). We observe from this table that \textsc{Poseidon} requires \emph{far fewer} task specific samples to attain the same error level as FNO does with $S=1024$ samples for time-dependent PDEs ($S=4096$ for time-independent PDEs). In fact, there are 4 tasks for which a mere 3 task-specific samples suffice for \textsc{Poseidon} to attain the same error as FNO with 1024 samples. From {\bf SM} Table \ref{tab:stat}, we observe that, on an average (median), only 20 samples are needed for \textsc{Poseidon}-L to reach the errors of FNO with 1024 samples and in 13 (of the 15) tasks, \textsc{Poseidon}-L needs an order of magnitude fewer samples than FNO. Similarly from Table \ref{tab:1} and {\bf SM} Table \ref{tab:stat}, we see that for the same number ($S=128$ for time-dependent, and $S=512$ for time-independent PDEs) of samples, \textsc{Poseidon}-L has significantly lower error than FNO, with gains ranging from anywhere between $10\%$ to a factor of 25, with the mean gain of accuracy being an \emph{entire order of magnitude}. 
\begin{table}[htbp]
    \centering
    \small
        \caption{Efficiency gain EG (\eqref{eq:met} with $S=1024$ for time-dependent and $S=4096$ for time-independent PDEs) and Accuracy Gain (\textit{AG}) (\eqref{eq:met} with $S=128$ for time-dependent and $S=512$ for time-independent PDEs) for all models and downstream tasks.}
    \begin{tabular}{r c c c c c c c c c c c c c c}
        \toprule
        & \multicolumn{6}{c}{Pretrained Models} & \multicolumn{6}{c}{Models trained from Scratch}\\
         \cmidrule(r){2-7}  \cmidrule(r){8-13}

         & \multicolumn{2}{c}{\textsc{Poseidon}-L} 
         & \multicolumn{2}{c}{CNO-FM}
         & \multicolumn{2}{c}{MPP-B}
         & \multicolumn{2}{c}{CNO}
         & \multicolumn{2}{c}{scOT}
         & \multicolumn{2}{c}{FNO}\\

         \cmidrule(r){2-3} \cmidrule(r){4-5} \cmidrule(r){6-7} \cmidrule(r){8-9} \cmidrule(r){10-11} \cmidrule(r){12-13}
        
         & EG & \textit{AG}  & EG & \textit{AG}  & EG & \textit{AG}  & EG & \textit{AG}  & EG & \textit{AG}  & EG & \textit{AG} 
        \\
        \midrule\midrule
        
NS-PwC & \bf890.6 & \bf\textit{24.7} & 16.6 & \textit{3.3} & 7.4 & \textit{2.3} & 3.7 & \textit{1.5} & 5.4 & \textit{2.0} & 1 & \textit{1} \\\midrule
NS-SVS & \bf502.9 & \bf\textit{7.3} & 59.6 & \textit{3.1} & 34.8 & \textit{2.2} & 73.2 & \textit{3.4} & 10.2 & \textit{1.2} & 1 & \textit{1} \\\midrule
NS-BB & \bf552.5 & \bf\textit{29.3} & 10.6 & \textit{3.9} & 4.6 & \textit{2.6} & 2.7 & \textit{1.7} & 3.4 & \textit{2.1} & 1 & \textit{1} \\\midrule
NS-SL & \bf21.9 & \bf\textit{5.5} & 0.4 & \textit{0.8} & 0.3 & \textit{0.8} & 0.8 & \textit{1.2} & 0.3 & \textit{0.8} & 1 & \textit{1} \\\midrule
NS-Tracer-PwC & \bf49.8 & \bf\textit{8.7} & 17.8 & \textit{3.6} & 8.5 & \textit{2.7} & 4.6 & \textit{1.9} & 4.6 & \textit{1.9} & 1 & \textit{1} \\\midrule
FNS-KF & \bf62.5 & \bf\textit{7.4} & 13.2 & \textit{2.7} & 2.0 & \textit{1.6} & 3.1 & \textit{1.5} & 3.3 & \textit{0.9} & 1 & \textit{1} \\\midrule
CE-RPUI & \bf352.2 & \bf\textit{6.5} & 33.2 & \textit{2.3} & 0.0 & \textit{1.2} & 12.5 & \textit{1.8} & 15.6 & \textit{2.1} & 1 & \textit{1} \\\midrule
CE-RM & \bf4.6 & \bf\textit{1.2} & 0.6 & \textit{1.0} & 0.0 & \textit{0.2} & 1.7 & \textit{1.1} & 0.4 & \textit{1.0} & 1 & \textit{1} \\\midrule
SE-AF & 3.4 & \textit{1.2} & 4.8 & \textit{1.3} & 2.2 & \textit{1.1} & \bf5.5 & \bf\textit{1.5} & 1.2 & \textit{1.0} & 1 & \textit{1} \\\midrule
GCE-RT & \bf5.3 & \bf\textit{2.0} & 1.2 & \textit{1.0} & 0.0 & \textit{0.3} & 1.2 & \textit{1.4} & 1.1 & \textit{1.1} & 1 & \textit{1} \\\midrule
Wave-Layer & \bf46.5 & \bf\textit{6.1} & 5.6 & \textit{2.2} & 0.0 & \textit{0.9} & 11.4 & \textit{3.0} & 13.0 & \textit{2.9} & 1 & \textit{1} \\\midrule
Wave-Gauss & \bf62.1 & \bf\textit{5.6} & 6.0 & \textit{1.8} & 0.0 & \textit{0.8} & 14.0 & \textit{2.6} & 9.2 & \textit{2.1} & 1 & \textit{1} \\\midrule
ACE & \bf17.0 & \bf\textit{11.6} & 1.7 & \textit{2.0} & 0.0 & \textit{0.3} & 4.5 & \textit{4.6} & 6.5 & \textit{5.2} & 1 & \textit{1} \\\midrule
Poisson-Gauss & \bf42.5 & \bf\textit{20.5} & 25.0 & \textit{9.2} & 17.0 & \textit{7.3} & 21.1 & \textit{7.0} & 9.8 & \textit{5.3} & 1 & \textit{1} \\\midrule
Helmholtz & \bf78.3 & \textit{6.1} & 54.0 & \textit{5.1} & 22.4 & \textit{3.0} & 68.9 & \textit{7.3} & 60.4 & \bf\textit{9.0} & 1 & \textit{1} \\
        \bottomrule
    \end{tabular}
    \label{tab:1}
\end{table}
\par Among the trained-from-scratch neural operator baselines, CNO and scOT are comparable in performance to each other, while both outperform FNO significantly on almost all tasks (see Table \ref{tab:1} and {\bf SM} Table \ref{tab:stat}). However, \textsc{Poseidon} is much superior to both of them, in terms of gains in sample efficiency (median gain of an order of magnitude) as well as accuracy (average gain of a factor of $4$).   

\textbf{\textsc{Poseidon} generalizes well to unseen physics.} This impressive performance of \textsc{Poseidon} is particularly noteworthy as all the downstream tasks are \emph{out-of-distribution} with respect to the pretraining dataset. This performance is also consistent across the 9 tasks which involve PDEs not seen during pretraining. \textsc{Poseidon} is the best performing model on 8 of these tasks, including all the time-dependent PDEs. It is only for 1 of the time-indepedent PDEs, which constitute the hardest generalization challenge, that \textsc{Poseidon} is outperformed by CNO, but only marginally. These results underscore the ability of \textsc{Poseidon} to learn completely new physical processes and contexts from a few downstream task-specific samples.   

\textbf{Architecture of the foundation model matters.} We observe from {\bf SM} \ref{sec:perf_downstream_scalings} and Table \ref{tab:1} (see also {\bf SM} Table \ref{tab:stat}) that \textsc{Poseidon} outperforms CNO-FM clearly on 14 out of 15 downstream tasks. On average (median over all tasks), CNO-FM requires approximately 100 task-specific examples to attain the error levels of FNO with 1024 samples, whereas \textsc{Poseidon} only requires approximately 20. As CNO-FM and \textsc{Poseidon} have been pretrained on exactly the same dataset, this difference in performance can be largely attributed to architectural differences as CNO-FM is based on multiscale CNNs, in contrast to the multiscale vision transformer which is the backbone of \textsc{Poseidon}.

The second baseline foundation model, MPP-B of \cite{MPP}, is based on a transformer with axial attention and is pretrained on the PDEBench dataset \cite{pdebench}. However, it has been trained to predict the next time step, given a context window of $\tau$ previous time steps, with $\tau=16$ as the default. We emphasize that this next step prediction, given a context window, \emph{does not solve the underlying operator learning task} {\bf OLT} directly as {\bf OLT} requires that the entire trajectory needs to be generated, given the initial data. Hence, we had to finetune the pretrained MPP model with varying context windows (starting with window size of 1), see {\bf SM} \ref{sec:mpp} for details. We see from Table \ref{tab:1} and {\bf SM} Table \ref{tab:stat} that the finetuned MPP modestly outperformed FNO on some (8 out of 15) of the downstream tasks but it failed on the rest of them, where MPP simply could not attain the error levels of FNO, as it did not converge or even blew up with increasing number of downstream samples (see scaling plots in {\bf SM} \ref{sec:perf_downstream_scalings}).

In this context, it can be argued that the \textsc{Poseidon}-L model is larger in size than both CNO-FM and MPP-B and perhaps, it is this size difference which explains the differential in performance. However, this is far from the case. As shown in all the scaling plots of {\bf SM} \ref{sec:perf_downstream_scalings} and {\bf SM} Tables \ref{tab:full_res} and \ref{tab:stat}, both CNO-FM and MPP-B are significantly inferior to the \textsc{Poseidon}-B model, which is comparable in size. In fact, we can see from these tables that even the \textsc{Poseidon}-T model, which is an order of magnitude smaller in size, outperforms CNO-FM and MPP-B handily. It also readily outperforms all the trained-from-scratch neural operators (CNO, FNO and scOT) which are of comparable size to it, leading us to conclude that it is the combination of the pretraining dataset as well as the underlying architecture, rather than just model size, that underpins the superior performance of \textsc{Poseidon}.  

\textbf{\textsc{Poseidon} scales with model size.} Nevertheless, the model size of \textsc{Poseidon} does matter. As seen from {\bf SM} Figure \ref{fig:model_scaling}, both the training as well as evaluation (validation) errors on the pretraining dataset clearly decrease with increasing model size of \textsc{Poseidon}. However, does this scaling with model size lead to any impact on the performance of these models, when finetuned on downstream tasks? We see from the scaling plots in {\bf SM} \ref{sec:perf_downstream_scalings} that \textsc{Poseidon}-L consistently outperforms the smaller \textsc{Poseidon}-B on most downstream tasks. This trend is reinforced by {\bf SM} Tables \ref{tab:full_res} and \ref{tab:stat}, where we find that, on an average, increasing model size correlates with a consistent decrease in test error as well as an increase in sample efficiency of the pretrained model on downstream tasks. 
 
\textbf{\textsc{Poseidon} scales with dataset size.}  In {\bf SM} Figure \ref{fig:sm4}, we show how by increasing the size of the pretraining dataset, in terms of the number of trajectories, the training and validation losses for the pretrained \textsc{Poseidon}-B model decrease. Moreover, from {\bf SM} Figures \ref{fig:quantity_quality_ablations_ns_pwc} to \ref{fig:quantity_quality_ablations_helmholtz}, where we plot the test error versus number of downstream task-specific samples for 2 different models, \textsc{Poseidon}-B trained on the full pretraining dataset and on one-eighth of the pretraining dataset, we find that for most (9 of the 15) of the downstream tasks, increasing the number of samples in the pretraining dataset, by an order of magnitude, \emph{does lead to significantly greater accuracy} even at the downstream task level. For the remaining tasks, the models trained with less data are either on par or marginally inferior to the model trained with the full dataset.     

\textbf{The quality/diversity of the pretraining dataset matters.} To demonstrate this point, we consider two different datasets: one in which half the trajectories of the pretraining dataset for \textsc{Poseidon}-B are randomly dropped (from every operator), and the other where less diversity of the pretraining dataset is imposed by dropping all the trajectories corresponding to 3 out of 6 operators, namely CE-CRP, CE-Gauss and NS-Sines. Thus, the total size of both datasets is the same but one is clearly less diverse than the other. The respective \textsc{Poseidon}-B models are then evaluated on all the downstream tasks. As shown {\bf SM} Figures \ref{fig:quantity_quality_ablations_ns_pwc} to \ref{fig:quantity_quality_ablations_helmholtz}, the model trained on less diverse data performs worse than its counterpart on 10 out of the 15 tasks and is on par on 4 of them. Thus, we demonstrate that in a large majority of downstream tasks, the quality/diversity of the pretraining dataset matters. 

\textbf{How does \textsc{Poseidon} generalize to unseen physics?} In order to understand the \emph{surprising} ability of \textsc{Poseidon} to generalize so well to unseen and \emph{a priori} unrelated PDEs and physical processes downstream, we present three case studies in {\bf SM} \ref{sec:cs} to uncover some of the inner workings of this foundation model. In particular, we first consider the CE-RPUI downstream task. This task pertains to the compressible Euler equations, which are included in the pretraining dataset. However, the underlying initial data distribution is not seen during pretraining, making the task \emph{out-of-distribution}. We show in {\bf SM} \ref{sec:cerpui}, how \textsc{Poseidon} leverages different features of different operators from the pretraining dataset to learn this task accurately with very few samples (see {\bf SM} Figure \ref{fig:case_studies_rpui}). In particular, the diversity of the pretraining dataset is more instrumental in ensuring better generalization to this unseen initial condition than the size of the dataset.  

In {\bf SM} \ref{sec:pg}, we study the Poisson-Gauss task to understand arguably the most surprising finding about the \textsc{Poseidon} foundation models, i.e., their ability to generalize well to PDEs that are completely unrelated to the Euler and Navier-Stokes equations of fluid dynamics. This task pertains to the Poisson equation \eqref{eq:pos} with a forcing term, which is a superposition of Gaussians. The task is very different from those seen during pretraining in multiple ways, namely the underlying PDE is not only time-independent (in contrast to the time-dependent PDEs of pretraining) but also elliptic (whereas the PDEs during pretraining are either hyperbolic or convection-dominated) and the boundary conditions are Dirichlet (instead of Periodic) leading to very different physics, that of diffusion and smoothing, being manifested for this task, when contrasted with the physics seen during pretraining which is dominated by transport, shock wave propagation and fluid mixing. Given this context, one would not expect \textsc{Poseidon} to perform well on this task. Yet, from {\bf SM} Figures \ref{fig:scaling_poisson_gauss} and \ref{fig:poisson_gaussians}, we know that \textsc{Poseidon} performs exceptionally well, learning the solution operator accurately with a few samples. As we elaborate in {\bf SM} \ref{sec:pg}, \textsc{Poseidon} does not use the first few training examples to \emph{forget} the physics that it has learned during pretraining and learn the new physics for this task after that. Rather surprisingly, as illustrated in {\bf SM} Figure \ref{fig:case_studies_poisson}, already with \emph{one} task specific training example, \textsc{Poseidon} outputs an (approximation of the) input, rather than the expected dynamic evolution of fluids with Gaussian inputs (see {\bf SM} Figures \ref{fig:ns_gaussians} and \ref{fig:ce_gaussians}) seen during pretraining. Then, with very few (16) examples, it is able to learn the rudiments of diffusion and smoothing of features ({\bf SM} Figure \ref{fig:case_studies_poisson}), which are characteristics of elliptic equations. To further test how the foundation model leverages physics representations learned during pretraining, we \emph{froze} the latent space by only finetuning the embeddings and freezing the latent space parameters by setting $\widehat{\theta}_r = \widehat{\theta}_{\ast}$ for all $r$, in \eqref{eq:gdfm} for finetuning. As shown in ({\bf SM} Figure \ref{fig:case_studies_poisson_1}), even this \emph{frozen latent} version of \textsc{Poseidon} is very effective at learning the underlying solution operator, demonstrating that very rich physical representations were learned during pretraining. 

Further results on the robustness of \textsc{Poseidon} for different factors and ablations as well as comparisons with other foundation models is provided in {\bf SM} \ref{sec:res} and details of computational resources are described in {\bf SM} \ref{sec:computational_resources}.

\section{Discussion}
\textbf{Summary.} In this paper, we have presented \textsc{Poseidon}, a family of foundation models for learning PDEs. The backbone of \textsc{Poseidon} is scOT, a multiscale vision transformer with shifted-windowed (SwinV2) attention that maps input functions (initial data, coefficients, sources) etc. to the solution (trajectory) of a PDE. Lead-time conditioning through a time-modulated layer norm allows for continuous-in-time evaluation and a novel all2all training strategy enables the scaling up of training data by leveraging the semi-group structure of solutions of time-dependent PDEs. \textsc{Poseidon} is pretrained on a diverse large-scale dataset of operators for the compressible Euler and incompressible Navier-Stokes PDEs. Its performance is evaluated on a challenging suite of 15 \emph{out-of-distribution} downstream tasks covering a wide variety of PDEs and data distributions. \textsc{Poseidon} displays excellent downstream performance and is the best performing model on 14 of the 15 tasks. In particular, it requires orders of magnitude (median of $50$) fewer task-specific samples to attain the same error as the widely used FNO. This large gain in sample efficiency as well as order of magnitude gains in accuracy also holds for PDEs that are not seen during pretraining, making us conclude that \textsc{Poseidon} generalizes well to \emph{new physics}. \textsc{Poseidon} also scales with model and dataset size, with respect to pretraining and even downstream task performance. To the best of our knowledge, this is the first time that it has been clearly demonstrated that by pretraining on a very small set of PDEs, a foundation model can generalize to a wider variety of unseen and unrelated PDEs and data distributions downstream. Thus, we provide an affirmative answer to the very fundamental question of whether foundation models for PDEs are even feasible. Moreover, we investigate possible mechanisms via which \textsc{Poseidon} can effectively leverage representations, learnt during pretraining, to accurately learn downstream tasks by finetuning on a few task-specific examples. Our case studies suggest hitherto undiscovered relationships between different PDEs that enable this transfer to materialize. Finally, all the models are made publicly available, as well as the pretraining and downstream datasets are open sourced in the \href{https://huggingface.co/collections/camlab-ethz/pdegym-665472c2b1181f7d10b40651}{\textsc{PDEgym} collection}.

\textbf{Related Work.} Foundation models for PDEs are of very recent vintage. The foundation model of \cite{mahoney} is limited to very specific elliptic Poisson and Helmholtz PDEs with a FNO backbone whereas ICON \cite{icon} considers a very small 1-D dataset. Neither of these models are comparable in scope to \textsc{Poseidon}. Universal physics transformers \cite{UPT} employs transformers but its focus is on incompressible fluid flows and the ability to generalize across Eulerian and Lagrangian data. Thus, a direct comparison with \textsc{Poseidon} is not possible. On the other hand, MPP \cite{MPP} and DPOT \cite{DPOT} are designed to be general purpose foundation models for PDEs that can be compared to \textsc{Poseidon}. We have already extensively compared MPP with \textsc{Poseidon} in Section \ref{sec:3} to demonstrate the very large superiority of \textsc{Poseidon} across various metrics. Although DPOT has a different architecture (Adaptive FNO) and was trained on more datasets than MPP, it follows a similar training and evaluation strategy of next time-step prediction, given a context window of previous time-steps. As argued before, this does not solve the operator learning task of generating the entire trajectory, given initial data. At the time of writing this paper, DPOT was not publicly available but it was released by the time this paper has been revised, enabling us to modify the fine-tuning procedure of DPOT and to perform comparisons between it and \textsc{Poseidon}. While directing the interested reader to {\bf SM} \ref{sec:dpot} for details, we summarize our findings by observing that \textsc{Poseidon} models are significantly better performing than DPOT foundation models, both in terms of accuracy and sample efficiency. 

\textbf{Limitations.} The range of PDEs and underlying data distributions is huge and \textsc{Poseidon} was only trained and evaluated on a few of them. Although the results here clearly demonstrate its ability to learn unseen physics from a few task-specific training examples, we anticipate that given that it is scaling with respect to both data quantity and quality, \textsc{Poseidon}'s performance as a general purpose PDE foundation model will significantly improve when it is pretrained with even more diverse PDE datasets in the future. In particular, pretraining with time-independent PDEs (particularly elliptic PDEs) as well as a larger range of time-scales in time-dependent PDEs will greatly enhance \textsc{Poseidon}. The focus here was on Cartesian geometries although \textsc{Poseidon} displayed the ability to generalize to non-Cartesian geometries, via masking, on the SE-AF task. We plan to add several non-Cartesian examples in the pretraining dataset to augment \textsc{Poseidon}'s performance on general geometries/boundary conditions. Moreover, given the fact that \textsc{Poseidon} serves as a fast and accurate neural PDE surrogate, its extension to qualitatively different downstream tasks such as uncertainty quantification \cite{LMR1}, inverse problems \cite{MM5} and PDE-constrained optimization \cite{LMRP1} is fairly straightforward and will be considered in future work. 

\begin{ack}
    This work was supported by a grant from the Swiss National Supercomputing Centre (CSCS) under project ID 1217.
\end{ack}

\bibliographystyle{abbrv}
\bibliography{refs.bib}

\appendix

\clearpage
\newpage
\begin{center}
{\large \bf Supplementary Material for}: \\
\textbf{\textsc{Poseidon}}: Efficient Foundation Models for PDEs. \\
\end{center}

\addcontentsline{toc}{section}{} %
\part{} %
\parttoc %

\newpage

\section{Architecture of the scalable Operator Transformer (scOT)}
\label{sec:scot}
\subsection{Operator Learning with scOT}
\label{sec:scot1}
First, we describe how scOT (Section \ref{sec:2} of Main Text and Figure \ref{fig:2}) transforms function space inputs into function outputs below.

For simplicity of exposition, we set $d=2$ and specify $D= [0,1]^2$ as the underlying domain. On this domain, an uniform \emph{computational grid}, with grid spacing $\Delta$, of $J^2$ equally spaced points $x_{j_x,j_y} = (j_x\Delta, j_y\Delta)$, with $J=1/\Delta$, is set. Let $1 < p < J$ such that $J{\rm mod} p =0$ and set $P=J/p$. We divide the domain $D = \cup_{\rho=1}^{P^2} D_{\rho}$ into a set of $P^2$ non-overlapping and equal (in measure) \emph{patches}. Any underlying input function $a \in C(D;\R^{n})$ is then \emph{partitioned} into a function, that is piecewise constant on patches and \emph{embedded} into a $C$-dimensional latent representation by applying the operator, 
\begin{equation}
\label{eq:ppeo}
\bv(x) = \hat{{\bf E}}(a)(x) = \sum\limits_{\rho=1}^{J^2} {\bf F}\left(\int\limits_{D_\rho} W(x) a(x) dx\right) {\mathbb{I}}_{D_\rho}(x), 
\end{equation}
with ${\bf F} \in \R^{C\times n}$ is a learnable matrix and the weight function $W$ is defined in terms of the underlying computational grid as, $W(x) = \sum\limits_{1 \leq j_x,j_y \leq J} W_{ij} \delta_{x_{j_xj_y}}$, with $\delta$ denoting the Dirac measure, and the shared (across all patches) learnable weights given by,
\begin{equation}
\label{eq:wf}
W_{j_xj_y} = \begin{cases}
\omega_{j_xj_y} &\quad {\rm if} \quad 1 \leq j_x,j_y \leq p \\
\omega_{j_x{\rm mod}p,j_y{\rm mod}p}, &\quad {\rm otherwise}.
\end{cases}
\end{equation}
The (patched and embedded) output function $\bv$ of \eqref{eq:ppeo} is then processed through a sequence of \emph{SwinV2 transformer} blocks \cite{swinv1,swinv2}, each of which has the structure of $SW_\ell: C(D;\R^C) \mapsto C(D;\R^C)$, %
for layer index $\ell = 1,\cdots, L$, formulated in Main Text \eqref{eq:swin}. 

The main building block of a SwinV2 transformer block \eqref{eq:swin} is the \emph{windowed multi-head self attention} operator,
\begin{equation}
\label{eq:wmsa}
W-MSA^\ell(\bv)(x) = \sum\limits_{h=1}^H{\bf W}^h_{\ell} \int\limits_{D^{\ell}_{q_x}} \frac{e^{\left(\cos\left({\bf Q}^h_\ell\bv(x),{\bf K}^h_{\ell}\bv(y)\right)+{\bf B}^h_{\ell}(x,y)\right)}}{\int\limits_{D^{\ell}_{q_x}} e^{\left(\cos\left({\bf Q}^h_\ell\bv(z),{\bf K}^h_{\ell}\bv(y)\right)+{\bf B}^h_{\ell}(z,y)\right)}dz} {\bf V}^h_{\ell} \bv (y) dy,
\end{equation}
for any $\bv \in C(D;\R^C)$. Here, $h$ denotes the $h$-th attention head, ${\bf W}^h_{\ell} \in \R^{C \times m}$ be the output matrix and ${\bf Q}^h_{\ell}, {\bf K}^h_{\ell}, {\bf V}^h_{\ell} \in \R^{m \times C}$ be the \emph{query, key} and \emph{value} matrices, respectively. For any two vectors $\alpha,\beta$, the cosine similarity is defined as $\langle \alpha,\beta \rangle = |\alpha||\beta|\cos(\alpha,\beta)$ and ${\bf B}_\ell^h:D \times D \mapsto \R$ is a general form for \emph{positional encodings}. To be more specific, we use \emph{relative log position encodings} by setting the inputs to ${\bf B}_\ell^h$ to be the logarithm of the relative positions $(k,\bar{k})$ within the window and the function ${\bf B}_\ell^h$ itself to be a small MLP. Finally, the domain of integration $D^\ell_{q_x}$ is simply the window where the point of interest $x$ lies, i.e., $1 \leq q_x \leq M^2$ such that $x \in D^{\ell}_{q_x}$. Underlying \eqref{eq:wmsa}, is the partition of the domain into windows such that $D = \cup_{q=1}^{M^2} D^{\ell}_{q}$, with $1 \leq \ell \leq L$ indexing the underlying layer within a SwinV2 transformer block and with $M^2$, denoting the number of windows. Moreover, the windows are shifted across layers, as depicted in Figure \ref{fig:2} (c), so that all the points can be attended to, by iteratively shifting windows across multiple layers/blocks. 

The MLP in Main Text \eqref{eq:swin} is of the form, $MLP: C(D;\R^C) \mapsto C(D;\R^C)$ with 
\begin{equation}
\label{eq:mlp}
MLP(\bv)(x) = \bar{W}\sigma\left(W \bv(x) + \hat{B}\right),
\end{equation}
for learnable weight matrices $W \in \R^{\bar{C}\times C}$,  $\bar{W} \in \R^{C \times \bar{C}}$, bias vector $\hat{B} \in \R^{\bar{C}}$ and nonlinear activation function $\sigma: \R \mapsto \R$. The Layer Norm $LN$ in Main Text \eqref{eq:swin} is given by Main Text \eqref{eq:LNt}. The remaining operations in scOT (see Main Text Figure \ref{fig:2}) are described in their discrete form below. 
\subsection{Computational Realization of scOT}
\label{sec:scot2}

The scalable Operator Transformer (scOT), forming the underlying model architecture for \textsc{Poseidon}, is constructed as an encoder-decoder architecture. Starting from patching and embedding, embedded tokens are inputted into multiple stages of SwinV2 transformer blocks, each followed by a patch merging. The encoder is connected at every level to the decoder through ConvNeXt \cite{convnext} blocks, whereas the bottleneck is convolution-free. Finally, through patch recovery and mixup, the output is assembled. We refer to Figure \ref{fig:2} (a) in the Main Text for an illustration of the overall architecture and concrete computational realizations by presenting discrete versions of the continuous operators described in the subsection above as well as elaborating on other operators used in scOT. %
 
\paragraph{Patch Partitioning.} The encoder consists of the patch partitioning operation, creating visual tokens from $n$ discretized (on the uniform computational grid described in Section \ref{sec:scot1}) input functions ${\bf a}_i \in \R^{J \times J}$, $i \in \{1, ..., n\}$. Each ${\bf a}_i$ is divided into non-overlapping patches of size $p \times p$ (with $p \ll J$) such that $P^2 = \left\lceil \frac{J}{p} \right\rceil^2$ patches arise. For an illustration, we refer to Figure \ref{fig:2} (c) of the Main Text where $P=8$. Patches are combined for every $a_i$ such that a sequence of ${\bf a}^p_j \in \R^{p \times p \times n}$, $j \in \{1, ..., P^2\}$ visual tokens can be fed to the embedding operation.

\paragraph{Embedding.} Each of these patches is \emph{linearly} embedded using a shared learnable weight ${\bf W}_{\mathcal{E}} \in \R^{C \times n \times p \times p}$ ($\in \Tilde{\Theta}$) and bias ${\bf b}_{\mathcal{E}} \in \R^C$ ($\in \Tilde{\Theta}$),
\begin{equation}
\label{eq:ppeodis}
    ({\bf v}_j)_i = ({\bf b}_\mathcal{E})_i + \sum_{k=1}^c \sum_{u,v=1}^p ({\bf W}_\mathcal{E})_{i,k,u,v} ({\bf a}^p_j)_{k,u,v}
\end{equation}
where $(\cdot)_i$ denotes the $i$-th component (for all $1\leq i \leq C$) and $C > n$ is the embedding dimension. It is straightforward to observe that \eqref{eq:ppeodis} is a discretization of the operator \eqref{eq:ppeo}, with an additional bias term. The resulting embedding is then passed through a (time-conditioned) layer norm (see Main Text Equation \ref{eq:ln}).

\paragraph{SwinV2 Stage.} At each level $i \in \{0, ..., L-1\}$ of the U-Net-style architecture, a SwinV2 stage $\mathcal{S}_i$ is employed consisting of $t_i$ chained SwinV2 transformer blocks $\mathcal{T}_{t_i}$,
\begin{equation}
    \mathcal{S}_i = \mathcal{T}_{t_i} \circ \mathcal{T}_{t_i - 1} \circ ... \circ \mathcal{T}_1.
\end{equation}
This is done in both encoder and decoder, and the same number $t_i$ of SwinV2 transformer blocks is used on each level.

\paragraph{SwinV2 Transformer Block.} A SwinV2 transformer block $\mathcal{T}$ is built as follows
\begin{align}
    {\bf v'}({\bf v}) &= (\mathcal{N} \circ \mathcal{A})({\bf v}) + {\bf v}\\
    \mathcal{T}({\bf v}) &= (\mathcal{N} \circ \mathcal{M})({\bf v'}({\bf v})) + {\bf v'}({\bf v})
\end{align}
where ${\bf v} \in \R^{P^2 / 4^i\times C\cdot 2^i}$ is the sequence of embedded tokens, $\mathcal{A}$ the shifted-window multi-head self-attention operation, $\mathcal{N}$ the (time-conditioned) Layer Norm, $\mathcal{M}$ a MLP. The attention mechanism $\mathcal{A}$ acts only on windows of size $M \times M$ patches/tokens that shift from block $\mathcal{T}_l$ to block $\mathcal{T}_{l+1}$ by doing a cyclic displacement of $M/2 \times M/2$ tokens (when the sequence is interpreted in 2D; see Figure \ref{fig:2} (c) of Main Text). So, with an input window ${\bf v} \in \R^{M^2 \times C \cdot 2^i}$,
\begin{equation}
    \mathcal{A}({\bf v}) = \text{Concat}[\mathcal{A}_1({\bf v}), ..., \mathcal{A}_{h_i}({\bf v})]{\bf W}_O + {\bf b}_O^\top \mathbb{I}
\end{equation}
where $\mathcal{A}_l$, $1\leq l \leq h_i$ is attention in head $l$ with the maximum number of heads depending on the stage $i$, with ${\bf W}_O \in \R^{C \cdot 2^i \times C \cdot 2^i}$, ${\bf b}_O \in \R^{C \cdot 2^i}$ being learnable parameters ($\in \hat{\Theta}$). $\mathcal{A}_l$ is then given by
\begin{equation}
\label{eq:wmsadis}
    \mathcal{A}_l({\bf v}) = \text{Softmax}\left({\bf B}_l({\bf v})+ \frac{\cos(({\bf v}{\bf W}_Q^l + {\bf 1}_{M^2}\cdot{\bf b}_Q^{l\top})^\top,({\bf v}{\bf W}_K^l)^\top)}{\tau_l}\right)\cdot \left({\bf v}{\bf W}_V^l  + {\bf 1}_{M^2}\cdot{\bf b}_V^{l\top}\right)
\end{equation}
with ${\bf W}_V^l, {\bf W}_Q^l, {\bf W}_K^l \in \R^{C \cdot 2^i \times C \cdot 2^i / h_i}$ and ${\bf b}_Q^l, {\bf b}_V^l \in \R^{C \cdot 2^i / h_i}$, $\tau_l \in \R$ (all learnable $\in \hat{\Theta}$), $\cos(\cdot, \cdot)$ the cosine similarity, ${\bf 1}_{M^2} \in \R^{M^2}$ a vector of ones, ${\bf B}_l({\bf v}) \in \R^{M^2 \times M^2}$ the relative position bias matrix generated from the (logarithmic) relative positions of each patch $[\Delta x, \Delta y]^\top$ within a window, parametrized through a shared MLP $\mathcal{P}$ for all heads: 
\begin{equation}
\mathcal{P}(\Delta x, \Delta y) = \text{ReLU}([\text{sign}(\Delta x)\log (1 + |\Delta x|), \text{sign}(\Delta y)\log (1 + |\Delta y|)]^\top{\bf W}_{B,1} + {\bf b}_{b,1}){\bf W}_{B,2}
\end{equation}
${\bf W}_{B,1} \in \R^{2\times512}$, ${\bf b}_{b,1} \in \R^{512}$, ${\bf W}_{B,2} \in \R^{512\times h_i}$ are all learnable ($\in \hat{\Theta}$). Note that \eqref{eq:wmsadis} is a discretization of the operator \eqref{eq:wmsa} by replacing the spatial integrals therein with uniform quadrature. 

The tokens, coming from the attention module are then fed to a layer norm \cite{ba2016layer} if the PDE to be learned is time-independent; if it is time-dependent (also in the case of \textsc{Poseidon}), it goes through a time-conditioned layer norm \cite{film} $\mathcal{N}$
\begin{align}
\label{eq:ln}
    \mu ({\bf v}) &= \frac{1}{C\cdot 2^i}\sum_{l=1}^{C \cdot 2^i} ({\bf v})_{l}\\
    \sigma^2({\bf v}) &= \frac{1}{C\cdot 2^i}\sum_{l=1}^{C \cdot 2^i} (({\bf v})_l - \mu({\bf v}))^2\\
    \mathcal{N}({\bf v}, t) &= \alpha(t) \odot \frac{{\bf v} - \mu ({\bf v}) \cdot {\bf 1}_{C \cdot 2^i}}{\sigma^2({\bf v})} + \beta(t)
\end{align}
with ${\bf v} \in \R^{C \cdot 2^i}$ be a token resulting from the attention module, $t \in \R_{\geq 0}$, ${\bf 1}_{C \cdot 2^i} \in \R^{C \cdot 2^i}$ a vector of ones, and $\alpha(t) = {\bf W}_\alpha t + {\bf b}_\alpha$, $\beta(t) = {\bf W}_\beta t + {\bf b}_\beta$ being learnable gain and bias (${\bf W}_\alpha, {\bf W}_\beta, {\bf b}_\alpha, {\bf b}_\beta \in \R^{C\cdot 2^i}$, all $\in \Tilde{\Theta}'$).

The last building block of the SwinV2 transformer block is a single-hidden-layer MLP with GeLU \cite{hendrycks2023gaussian} as pointwise activation function and four times the width of the latent dimension $C\cdot 2^i$
\begin{equation}
    \mathcal{M}({\bf v}) = \text{GeLU}({\bf v}{\bf W}_1 + {\bf b}_1){\bf W}_2 + {\bf b}_2
\end{equation}
with ${\bf W}_1 \in \R^{C \cdot 2^i \times 4 \cdot C \cdot 2^i}$, ${\bf b}_1 \in \R^{4 \cdot C \cdot 2^i}$, ${\bf W}_2 \in \R^{4 \cdot C \cdot 2^i \times C \cdot 2^i}$, ${\bf b}_2 \in \R^{C \cdot 2^i}$ all learnable parameters ($\in \hat{\Theta}$).

\paragraph{Patch Merging.} At each (resolution) level $i$ of the architecture, after each SwinV2 stage in the encoder, a \emph{linear} downsampling operation $\mathcal{D}_i$ is performed on the output of the stage added to its input (additional residual connection) such that the resolution halves. This amounts to a linear transformation on four non-overlapping, stacked patches/tokens at a time ${\bf v} \in \R^{4 \cdot C\cdot 2^i}$
\begin{equation}
    \mathcal{D}_i({\bf v}, t) = \mathcal{N}({\bf W}_{\mathcal{D}_i} {\bf v}, t)
\end{equation}
with learnable ${\bf W}_{D_i} \in \R^{C\cdot 2^{i+1} \times 4 \cdot C\cdot 2^i}$ such that the latent dimension doubles. Here, an additional (time-conditioned) layer norm is applied.

\paragraph{ConvNeXt Block.} Outputs from each encoder stage $\mathcal{S}_i$, $0 \leq i \leq L - 2$ are fed to $n_c$ chained (time-conditioned) ConvNeXt blocks \cite{convnext} $\mathcal{Q}_i$; for that, the token sequence is reshaped to a two-dimensional grid of tokens ${\bf v} \in \R^{P / 2^i \times P / 2^i \times C \cdot 2^i}$ and transformed by
\begin{equation}
    \mathcal{Q}_i ({\bf v}, t) = (\text{GeLU}(\mathcal{N}(\text{DwConv}({\bf v}), t){\bf W}_{\mathcal{Q},1} + {\bf b}_{\mathcal{Q},1}){\bf W}_{\mathcal{Q},2} + {\bf b}_{\mathcal{Q},2}) \odot {\bf W}_{\mathcal{Q},3} + {\bf v}
\end{equation}
${\bf W}_{\mathcal{Q},1} \in \R^{C \cdot 2^i \times 4 \cdot C \cdot 2^i}$, ${\bf b}_{\mathcal{Q},1} \in \R^{4 \cdot C \cdot 2^i}$, ${\bf W}_{\mathcal{Q},2} \in \R^{4 \cdot C \cdot 2^i \times C \cdot 2^i}$, ${\bf b}_{\mathcal{Q},2} \in \R^{C \cdot 2^i}$, ${\bf W}_{\mathcal{Q},3} \in \R^{C \cdot 2^i}$ all learnable parameters ($\in \hat{\Theta}$) and $\text{DwConv}$ is a depthwise convolution with kernel size 7 (and a padding of 3) and bias.

\paragraph{Patch Expansion.} Similar to patch merging, after a SwinV2 stage in the decoder, each output token ${\bf v} \in \R^{C \cdot 2^{i+1}}$ is \emph{linearly} upsampled through $\mathcal{U}_i$ to double the resolution and half the latent dimension,
\begin{equation}
    \mathcal{U}_i({\bf v}, t) = \mathcal{N}(\text{Reshape}({\bf W}_{\mathcal{U}_{i,1}}{\bf v}),t){\bf W}_{\mathcal{U}_{i,2}}
\end{equation}
where ${\bf W}_{\mathcal{U}_{i,1}} \in \R^{C \cdot 2^{i+2} \times C \cdot 2^{i+1}}$, ${\bf W}_{\mathcal{U}_{i,2}} \in \R^{C \cdot 2^{i} \times C \cdot 2^{i}}$ are both learnable ($\in \hat{\Theta}$), and $\text{Reshape}(\cdot)$ an operation that reshapes a vector of size $C \cdot 2^{i+2}$ into a matrix of size $4 \times C \cdot 2^{i}$.

\paragraph{Patch Recovery and Mixup.} Having passed through the last stage of the decoder, every patch/visual token ${\bf v}_j \in \R^{C}$ is \emph{linearly} transformed back from the latent space to form patches of the discretized output function ${\bf u}^p_j \in \R^{p\times p \times c_u}$,
\begin{equation}
    ({\bf u}^p_j)_i = ({\bf b}_{\mathcal{R}})_i \mathbb{I} + \sum_{k=1}^C ({\bf W}_{\mathcal{R}})_{i,k,\ast,\ast} ({\bf v}_j)_k
\end{equation}
where $(\cdot)_i$ denotes the $i$-th component (for all $1\leq i \leq c_u$) and $c_u$ is the number of components of the discretized output function. ${\bf W}_{\mathcal{R}} \in \R^{c_u \times C \times p \times p}$ and ${\bf b}_{\mathcal{R}} \in \R^{c_u}$ are shared across tokens and learnable ($\in \Tilde{\Theta}$). These outputs are assembled on a grid to form ${\bf \Tilde{u}} \in \R^{J \times J \times c_u}$ which is transformed to the final output ${\bf u}$ with a convolution with kernel size 5 (and padding 2 to keep the dimensionality), without bias, with all parameters being in $\Tilde{\Theta}$.

\paragraph{Summary of Hyperparameters.} In Table \ref{tab:hyperparams_scot}, we give an overview of the hyperparameters to instantiate a scOT. To reduce the number of hyperparameters, we fix $p=4$, $M=16$, $L=4$, $[h_1, h_2, h_3, h_4] = [3, 6, 12, 24]$, and $n_c=2$ in this work.
\begin{table}[h]
    \centering
    \caption{Hyperparameters of scOT.}
    \begin{tabular}{c l}
    \toprule
    Hyperparameter & Description \\
    \midrule\midrule
        $p$ & patch size \\\midrule
        $M$ & window size\\\midrule
        $C$ & embedding/latent dimension \\\midrule
        $L$ & number of levels ($L-1$ downsampling/upsampling operations)\\\midrule
        $t_i$ & number of SwinV2 transformer blocks in level $i$ \\\midrule
        $h_i$ & number of attention heads in level $i$\\\midrule
        $n_c$ & number of ConvNeXt blocks at each level\\
        \bottomrule
    \end{tabular}
    \label{tab:hyperparams_scot}
\end{table}

\clearpage
\section{Datasets}
\label{sec:data}

We describe the various datasets used for pretraining and for the downstream tasks below. All these datasets are publicly available with the \textsc{PDEgym} collection (\href{https://huggingface.co/collections/camlab-ethz/pdegym-665472c2b1181f7d10b40651}{https://huggingface.co/collections/camlab-ethz/pdegym-665472c2b1181f7d10b40651}).

\subsection{Pretraining Datasets}
\label{sec:datapt}
\begin{table}[htbp]
\centering
\caption{Abbreviations/Summary for all the pretraining datasets. IC refers to initial conditions.}
\begin{tabular}{r c c c}
    \toprule
    Abbreviation & PDE & Defining Feature & Visualization  \\
    \midrule\midrule

    NS-Sines & Navier-Stokes \eqref{eq:NS} & Sine IC & Fig. \ref{fig:ns_sines}\\\midrule
    NS-Gaussians & Navier-Stokes \eqref{eq:NS} & Gaussians (in Vorticity) IC  &  Fig. \ref{fig:ns_gaussians}\\\midrule
    CE-RP & Euler \eqref{eq:ceuler} &  4-Quadrant Riemann Problem IC & Fig. \ref{fig:ce_rp}\\\midrule
    CE-CRP & Euler \eqref{eq:ceuler} & Multiple Curved Riemann Problems & Fig. \ref{fig:ce_crp} \\ \midrule
    CE-KH & Euler \eqref{eq:ceuler} & Shear IC & Fig. \ref{fig:ce_kh} \\\midrule
    CE-Gauss & Euler \eqref{eq:ceuler} &  Gaussians (in Vorticity) IC & Fig. \ref{fig:ce_gaussians} \\
    \bottomrule
\end{tabular}
\label{tab:SM1}
\end{table}

We pretrain \textsc{Poseidon} models and CNO-FM  on a dataset containing 6 operators, defined on the space-time domain $[0,1]^2 \times [0,1]$. We include $2$ operators governed by the Navier-Stokes equations (NS-Sines, NS-Gauss) and $4$ operators governed by the Compressible Euler equations. The pretraining datasets encompass problems that exhibit a wide range of scales and complex, nonlinear dynamics.

The \textit{Incompressible Navier-Stokes equations} of fluid dynamics are given by
\begin{equation}
    \label{eq:NS}
    u_t +(u\cdot \nabla)u + \nabla p =\nu \Delta u, \quad {\rm div}~u =0,
\end{equation}
in the domain $D=[0,1]^2$ with suitable boundary conditions. Here, $u:[0,T] \times D \mapsto \R^2$ is the velocity field and $p:[0,T] \times D \mapsto \R_+$ is the pressure. In this work, a small viscosity $\nu = 4\times 10^{-4}$ is only applied to high-enough Fourier modes to approximate the inviscid limit.  

To generate the pretraining data, all the benchmarks for the Navier-Stokes equations are simulated until the time $T=1$. Furthermore, we store \emph{21 snapshots} of the numerically simulated velocity field $u$, uniformly spaced in time. Each snapshot has a spatial resolution of $128\times128$. The initial conditions are drawn from various distributions, which we will describe later. The distribution of these initial conditions is crucial for determining the complexity of the samples and the overall dynamics. 

All the Navier-Stokes experiments, both for the pretraining dataset and the downstream tasks, are simulated with the following \textit{spectral method}. Fix a mesh width $\Delta = \frac{1}{N}$ for some $N \in \mathbb{N}$. We consider the following discretization of the Navier-Stokes equations in the Fourier domain
\begin{equation} \label{eq:shv_disc}
    \begin{cases}
        \partial_t u^{\Delta} + \mathcal{P}_N(u^{\Delta}\cdot\nabla u^{\Delta}) + \nabla p^{\Delta} &= \varepsilon_N\Delta(Q_N*u^{\Delta}) \\
        \nabla\cdot u^{\Delta} &= 0 \\
        u^{\Delta}|_{t=0} &= \mathcal{P}_Nu_0
    \end{cases}
\end{equation}
where $\mathcal{P}_N$ is the spatial Fourier projection operator mapping a function $f(x,t)$ to its first $N$ Fourier modes: $\mathcal{P}_N = \sum_{|k|_{\infty}\leq N} \hat{f}_k(t)e^{ik\cdot x}$.
The artificial viscosity term we use for the stabilization of the solver consists of a resolution-dependent viscosity $\varepsilon_N$ and a Fourier multiplier $Q_N$ controlling the strength at which different Fourier modes are dampened. This allows us to not dampen the low frequency modes, while applying some diffusion to the problematic higher frequencies. The Fourier multiplier $Q_N$ is of the form
\begin{equation}
    Q_N(\mathbf{x}) = \sum_{\mathbf{k}\in\mathbb{Z}^d, |\mathbf{k}|\leq N} \hat{Q}_{\mathbf{k}} e^{i\mathbf{k}\cdot\mathbf{x}}.
\end{equation}
In order for the solver to converge, the Fourier coefficients of $Q_N$ need to fulfill \cite{Tadmor1989,Tadmor2004,LMP1}
\begin{equation}
    \hat{Q}_k = 0 \mbox{ for } |k|\leq m_N, 1-\left(\frac{m_N}{|k|}\right)^{\frac{1}{\theta}} \leq \hat{Q}_k \leq 1
\end{equation}
where we have introduced an additional parameter $\theta > 0$. The quantities $m_N$ and $\varepsilon_N$ are required to scale as
\begin{equation}
    m_N \sim N^{\theta}, \varepsilon_N \sim \frac{1}{N}, 0 < \theta < \frac{1}{2}.
\end{equation}
For the experiment described here, we choose $m_N = \sqrt{N}$, $\varepsilon_N = \frac{0.05}{N}$, and $N = 128$. This gives rise to the viscosity $\nu \approx 4 \cdot 10^{-4}$ mentioned above. The Fourier multipliers $\hat{Q}_N$ are chosen according to \cite{Lanthaler2020} and are equal to
\begin{equation}
    \hat{Q}_{\mathbf{k}}^{\text{Smooth}} = 1 - \exp\left(-\left(\frac{|\mathbf{k}|}{k_0}\right)^{\alpha}\right).
\end{equation}
The Navier-Stokes simulations were performed with the \emph{AZEBAN} spectral hyperviscosity code \cite{azeban}.

\textit{The Compressible Euler equations of gas dynamics} are given by
\begin{equation}
    \label{eq:ceuler}
    u_t + {\rm div}~F(u) =0,~u=[\rho,\rho v, E]^\perp,~ F=[\rho v, \rho v\otimes v+p{\bf I},(E+p)]v]^\perp,
\end{equation}
in the domain $[0,1]^2$ with suitable boundary conditions, with density $\rho$, velocity $v=[v_x,v_y]$, pressure $p$ and total Energy $E$ related by the ideal gas equation of state:
\begin{equation}
    E = \frac{1}{2}\rho |u|^2 + \frac{p}{\gamma - 1},
\end{equation}
where $\gamma=1.4$.
All the trajectories are simulated until time $T = 1$. The simulations for the compressible Euler equations were performed with the \emph{ALSVINN} \cite{alsvinn} code, which is based on a high-resolution finite volume scheme with piecewise quadratic WENO reconstructions and HLLC Riemann solvers. 

During pretraining, our goal is to predict four variables: $[\rho, v_x, v_y, p]$, where $\rho$ represents density, $v_x$ is the horizontal velocity, $v_y$ is the vertical velocity, and $p$ is the pressure. As in the Navier-Stokes benchmarks, all the trajectories for compressible Euler are simulated until time $T=1$. Furthermore, we store \emph{21 snapshots} of the numerically simulated solution, uniformly spaced in time. Each snapshot has a spatial resolution of $128\times128$, though being generated on $512\times 512$ and downsampled.

Next we describe each constituent of the pretraining dataset (summarized in Table \ref{tab:SM1})

\subsubsection{NS-Sines} This dataset considers the incompressible Navier-Stokes equations \eqref{eq:NS} with the following initial conditions, 
\begin{equation}
    \begin{aligned}
        u^0_x(x, y) &= \sum_{i,j=1}^p \frac{\alpha_{i,j}}{\sqrt{2\pi(i+j)}} \sin(2\pi ix+\beta_{i,j}) \sin(2\pi jy+\gamma_{i,j}) \\
        u^0_y(x, y) &= \sum_{i,j=1}^p \frac{\alpha_{i,j}}{\sqrt{2\pi(i+j)}} \cos(2\pi ix+\beta_{i,j}) \cos(2\pi jy+\gamma_{i,j})
    \end{aligned}
\end{equation}
where the random variables are chosen as $\alpha_{i,j} \sim \mathcal{U}_{[-1,1]}$, $\beta_{i,j} \sim \mathcal{U}_{[0,2\pi]}$, and $\gamma_{i,j} \sim \mathcal{U}_{[0,2\pi]}$. The number of modes $p$ is chosen to be $p = 10$. Thus, the initial conditions amount to a linear combination of sines and cosines. 

The underlying solution operator $\sol(t,\cdot)$ is given by $\sol(t,u^0_{x,y}) = u_{x,y}(t)$, with $u_x,u_y$ solving the Navier-Stokes equations \eqref{eq:NS} with periodic boundary conditions. 

We generated 20000 NS-Sines trajectories of which the first 19640 belong to the training set, the next 120 to the validation set, and the last 240 to the test set. Note that we included 11 time steps in the pretraining dataset, with every other time step selected, starting from step 0 up to step 21. A visualization of a random sample and the predictions made by \textsc{Poseidon}-B (trained on $128$ training trajectories) is shown in Figure \ref{fig:ns_sines}.

\subsubsection{NS-Gauss} Given a two-dimensional velocity field $u = (u_x,u_y)$, its \emph{vorticity} is given by the scalar $\omega = {\rm curl}~u = \partial_x u_y - \partial_y u_x$. Note that, for any time $t$, the velocity can be recovered from the vorticity using the so-called \emph{stream function} \cite{MB1}. 

For this dataset, we specify the initial conditions for the Navier-Stokes equations in terms of the vorticity given by,  
\begin{equation}
    \omega_0(x,y) = \sum_{i=1}^p\frac{\alpha_i}{\sigma_i} \exp\left( -\frac{(x-x_i)^2+(y-y_i)^2}{2\sigma_i^2} \right)
\end{equation}
where we chose $p = 100$ Gaussians with $\alpha_i \sim \mathcal{U}_{[-1, 1]}$, $\sigma_i \sim \mathcal{U}_{[0.01, 0.1]}$, $x_i \sim \mathcal{U}_{[0, 1]}$, and $y_i \sim \mathcal{U}_{[0, 1]}$. Thus, the initial vorticity field is a superposition of a large number of Gaussians. The initial velocity field is then recovered from the vorticity. 

The underlying solution operator $\sol(t,\cdot)$ is given by $\sol(t,u^0_{x,y}) = u_{x,y}(t)$, with $u_x,u_y$ solving the Navier-Stokes equations \eqref{eq:NS} with periodic boundary conditions.  

We generated 20000 NS-Gauss trajectories with the same train/validation/test split and time-stepping as for NS-Sines. A visualization of a random sample and the predictions made by \textsc{Poseidon} ($128$ training trajectories) are shown in Figure \ref{fig:ns_gaussians}.

\subsubsection{CE-RP} The well-known four-quadrant Riemann problem is the generalization of the standard Sod shock tube to two-space dimensions \cite{HEST1}. It is defined by dividing the domain $D=[0,1]^2$ into a grid of $p \times p$ square subdomains 
\begin{equation*}
D_{i,j} = \left\{(x, y) \in \mathbb{T}^2 \mid \frac{i-1}{p} \leq x < \frac{i}{p}, \frac{j-1}{p} \leq y < \frac{j}{p} \right\},
\end{equation*}
where $\mathbb{T}^2 $ is the 2d torus. We fix $p=2$ for this problem.

The initial data on each of these subdomains is constant and given by,
\begin{equation*}
    (\rho_0, v^0_x, v^0_y, p_0) = (\rho_{i,j}, (v_x)_{i,j}, (v_y)_{i,j}, p_{i,j}).
\end{equation*}
By sampling $\rho_{i,j} \sim \mathcal{U}_{[0.1, 1]}$, $(v_x)_{i,j} \sim \mathcal{U}_{[-1, 1]}$, $(v_y)_{i,j} \sim \mathcal{U}_{[-1, 1]}$, and $p_{i,j} \sim \mathcal{U}_{[0.1, 1]}$, we obtain a stochastic version of the four-quadrant Riemann problem, which also generalizes the stochastic shock tubes of \cite{schwabsid1} to two-space dimensions. 

The underlying solution operator $\sol(t,\cdot)$ is given by $\sol(t,\rho_0,v^0_{x,y},p_0) =[\rho(t),v_{x,y}(t),p(t)]$ solving the compressible Euler equations \eqref{eq:ceuler} with periodic boundary conditions. 

We generated 10000 CE-RP trajectories where the first 9640 trajectories belong to the training set, the following 120 to the validation set, and the last 240 trajectories to the test set. The time-stepping is the same as for NS-Sines and NS-Gauss. A visualization of a random sample and the predictions made by \textsc{Poseidon} ($128$ finetuning trajectories) are shown in Figure \ref{fig:ce_rp}.

\subsubsection{CE-CRP} This dataset corresponds to a \emph{curved} and multi-partitioned version of the CE-RP dataset. To define it, we denote the fractional part of $x\in \R$ as $\{x\} := x - \lfloor|x|\rfloor\sign x$ and define the functions
\begin{equation*}
    \begin{aligned}
        \sigma_x(x, y) &= \sum_{i,j=1}^p \alpha_{x,i,j}\sin(2\pi ix + jy + \beta_{x,i,j}) \\
        \sigma_y(x, y) &= \sum_{i,j=1}^p \alpha_{y,i,j}\sin(2\pi ix + jy + \beta_{y,i,j}).
    \end{aligned}
\end{equation*}
where $\alpha_{k,i,j} \sim \mathcal{U}_{[-0.1, 0.1]}$, and $\beta_{k,i,j} \sim \mathcal{U}_{[0, 1]}$. These functions are then used to create a partition of the domain into curved subdomains,
\begin{equation*}
    D_{i,j} = \{ (x, y) \in \mathbb{T}^2 \mid x_{\text{min}} \leq \{x+\sigma_x(x,y)+1\} < x_{\text{max}}, y_{\text{min}} \leq \{y+\sigma_y(x,y)+1\} < y_{\text{max}} \}.
\end{equation*}
with $x_{\text{min}} = \frac{i}{p+1}$, $x_{\text{max}} = \frac{i+1}{p+1}$, $y_{\text{min}} = \frac{j}{p+1}$, and $y_{\text{max}} = \frac{j+1}{p+1}$. Finally, the initial conditions are given by
\begin{equation*}
    (\rho, v_x, v_y, p)|_{t=0} = (\rho_{i,j}, u_{i,j}, v_{i,j}, p_{i,j}) \mbox{ in } D_{i,j}
\end{equation*}
where $\rho_{i,j} \sim \mathcal{U}_{[0.1, 1]}$, $(v_x)_{i,j} \sim \mathcal{U}_{[-1, 1]}$, $(v_y)_{i,j} \sim \mathcal{U}_{[-1, 1]}$, and $p_{i,j} \sim \mathcal{U}_{[0.1, 1]}$. A visualization of a random sample of the initial conditions is shown in Figure \ref{fig:ce_crp} and illustrates how this problem is a curved, multi-partitioned version of the standard stochastic four-quadrant Riemann problem (CE-RP).

The underlying solution operator $\sol(t,\cdot)$ is given by $\sol(t,\rho_0,v^0_{x,y},p_0) =[\rho(t),v_{x,y}(t),p(t)]$ solving the compressible Euler equations \eqref{eq:ceuler} with periodic boundary conditions. 

We generated 10000 CE-CRP trajectories with the same train/validation/test split as CE-RP. The time-stepping is the same as for NS-Sines and NS-Gauss. A visualization of a random sample and the predictions made by \textsc{Poseidon} ($128$ training trajectories) are shown in Figure \ref{fig:ce_crp}.

\subsubsection{CE-KH} This is a well-known benchmark of compressible fluid dynamics that corresponds to the well-known Kelvin-Helmholtz instability \cite{LL1}. A modern version is presented, for instance, in \cite{FKMT1}. 

The underlying initial data is,
\begin{equation*}
    (\rho, v_x, v_y, p)|_{t=0} = \begin{cases}
        (1, 0.5, 0, 2.5) &\mbox{if } y < 0.25 + \sigma_0(x) \mbox{ or } y > 0.75 + \sigma_1(x) \\
        (2, -0.5, 0, 2.5) &\mbox{otherwise}.
    \end{cases}
\end{equation*}
The perturbations $\sigma_0$ and $\sigma_1$ are given by
\begin{equation*}
    \sigma_i(x) = \frac{\varepsilon}{\sum_{j=1}^p \alpha_{i,j}} \sum_{j=1}^p \alpha_{i,j}\cos(2\pi j(x+\beta_{i,j}))
\end{equation*}
where $\varepsilon = 0.05$, $\alpha_{i,j} \sim \mathcal{U}_{[0,1]}$, and $\beta_{i,j} \sim \mathcal{U}_{[0,1]}$.

The underlying solution operator $\sol(t,\cdot)$ is given by $\sol(t,\rho_0,v^0_{x,y},p_0) =[\rho(t),v_{x,y}(t),p(t)]$ solving the compressible Euler equations \eqref{eq:ceuler} with periodic boundary conditions. 

We generated 10000 CE-KH trajectories with the same train/validation/test split as CE-RP. The time-stepping is the same as for NS-Sines and NS-Gauss. A visualization of a random sample and the predictions made by \textsc{Poseidon} ($128$ training trajectories) are shown in Figure \ref{fig:ce_kh}.

\subsubsection{CE-Gauss}
As in the NS-Sines dataset, we initialize the curl $\omega$ of the initial velocity with a superposition of Gaussians,
\begin{equation*}
    \omega_0(x,y) = \sum_{i=1}^p \frac{\alpha_i}{\sigma_i} \exp\left( -\frac{(x-x_i)^2+(y-y_i)^2}{2\sigma_i^2} \right)
\end{equation*}
where we chose $p = 100$ Gaussians with $\alpha_i \sim \mathcal{U}_{[-1, 1]}$, $\sigma_i \sim \mathcal{U}_{[0.01, 0.1]}$, $x_i \sim \mathcal{U}_{[0, 1]}$, and $y_i \sim \mathcal{U}_{[0, 1]}$. Then, the initial field is generated from the vorticity by using the incompressibility condition. The underlying density and pressure are initialized with constants, $\rho = 0.1$ and $p = 2.5$, respectively.

The underlying solution operator $\sol(t,\cdot)$ is given by $\sol(t,\rho_0,v^0_{x,y},p_0) =[\rho(t),v_{x,y}(t),p(t)]$ solving the compressible Euler equations \eqref{eq:ceuler} with periodic boundary conditions. 

We generated 10000 CE-Gauss trajectories with the same train/validation/test split as CE-RP. Time-stepping is the same as for NS-Sines and NS-Gauss. A visualization of a random sample and the predictions made by \textsc{Poseidon} ($128$ training trajectories) are shown in Figure \ref{fig:ce_gaussians}.

We remark that out of the 6 operators that consitute the pretraining dataset, 2 of them (CE-KH and CE-RP) are well known in the literature where as the other four (NS-Sines, NS-Gauss, CE-Gauss, CE-CRP) are novel to the best of our knowledge. 

\subsection{Downstream Tasks}
\label{sec:dataft}
Next, we describe the suite of downstream tasks on which \textsc{Poseidon} and baselines are evaluated. The list of tasks is summarized in Table \ref{tab:SM2}. 
\begin{table}[htbp]
\centering
\caption{Abbreviations/Summary for all the downstream tasks. IC and RP stand for initial conditions and Riemann problem, respectively. Datasets where (*) is checked mark datasets where solutions are learned depending on PDE parameters/sources/coefficients.}
\small
\begin{tabular}{r c c c c}
    \toprule
    Abbreviation & PDE & (*) & Defining Feature & Visualization \\
    \midrule\midrule
    NS-PwC & Navier-Stokes \eqref{eq:NS} &  & Piecewise constant vorticity IC &  Fig. \ref{fig:ns_pwc} \\\midrule
    NS-BB & Navier-Stokes \eqref{eq:NS} & & Brownian Bridge IC & Fig. \ref{fig:ns_bb} \\\midrule
    NS-SL & Navier-Stokes \eqref{eq:NS} & & Shear Layer IC & Fig. \ref{fig:ns_sl} \\\midrule
    NS-SVS & Navier-Stokes \eqref{eq:NS} & & Sine Vortex sheet IC &  Fig. \ref{fig:ns_svs} \\\midrule
    NS-Tracer-PwC & Navier-Stokes + Transport \eqref{eq:passive_sc} & & Scalar Advection &  Fig. \ref{fig:ns_tracer_pwc} \\\midrule
    FNS-KF & Forced Navier-Stokes \eqref{eq:FNS} & \checkmark & Kolmogorov Flow &  Fig. \ref{fig:fns_kf} \\\midrule
    CE-RPUI & Euler \eqref{eq:ceuler} & & RP with uncertain interfaces &  Fig. \ref{fig:ce_rpui} \\\midrule
    CE-RM & Euler \eqref{eq:ceuler} & & Richtmeyer-Meshkov &  Fig. \ref{fig:ce_rm} \\\midrule
    GCE-RT & Euler+Gravity \eqref{eq:gce} & \checkmark & Rayleigh-Taylor& Fig. \ref{fig:gce_rt} \\\midrule
    Wave-Gauss & Wave Eqn. \eqref{eq:wave} & \checkmark & Waves in Gaussian medium &  Fig. \ref{fig:wave_gaussians}\\\midrule
    Wave-Layer & Wave Eqn \eqref{eq:wave}& \checkmark & Waves in layered medium &  Fig. \ref{fig:wave_layer} \\\midrule
    ACE & Allen-Cahn Eqn. \eqref{eq:ac} & & Reaction-Diffusion &   Fig. \ref{fig:ace} \\\midrule
    SE-AF & steady state of Euler \eqref{eq:ceuler} & \checkmark & Flow past airfoil &  Fig. \ref{fig:se_af}\\\midrule
    Poisson-Gauss & Poisson Eqn. \eqref{eq:pos} & \checkmark & Stationary diffusion &  Fig. \ref{fig:poisson_gaussians}\\\midrule
    Helmholtz & Helmholtz Eqn \eqref{eq:helmholtz} & \checkmark & Waves in frequency domain &  Fig. \ref{fig:helmholtz}\\
    \bottomrule
\end{tabular}
\label{tab:SM2}
\end{table}

\subsubsection{NS-PwC} This downstream task is based on the Navier-Stokes equations \eqref{eq:NS} on the space-time domain $[0,1]^2 \times [0,1]$ with periodic boundary conditions. The initial conditions are based on the vorticity, which is assumed to be constant along a uniform (square) partition of the underlying domain. To be more specific, the initial vorticity is given by,
\begin{equation}
    \label{eq:pwc_ic}
    \omega_0(x,y) = c_{i,j} \mbox{ in } [x_{i-1}, x_i] \times [y_{j-1}, y_j]
\end{equation}
for $x_i = y_i = \frac{i}{p}$ for $i = 0, 1, 2, ..., p$, and $c_{i,j} \sim \mathcal{U}_{[-1,1]}$. The number of squares in each direction was chosen to be $p = 10$. Thus, this problem is an analogue of multiple \emph{Riemann problems}, but on the vorticity. The underlying initial velocity field $u_x(0),u_y(0)$ is then recovered from the vorticity by using the incompressibility condition. 

The underlying solution operator $\sol(t,\cdot)$ is given by $\sol(t,u^0_{x,y}) = u_{x,y}(t)$, with $u_x,u_y$ solving the Navier-Stokes equations \eqref{eq:NS} with periodic boundary conditions.

We generated 20000 NS-PwC trajectories with the same train/validation/test split as NS-Sines. Note that we included 8 time steps in the training dataset, with every other time step selected, starting from step 0 up to step 14. The testing error is evaluated at the 14th time step (i.e. $t = 0.7$). A visualization of a random sample and the predictions made by \textsc{Poseidon}-B, CNO and FNO by ($128$ training trajectories) are shown in Figure \ref{fig:ns_pwc}.

\subsubsection{NS-BB} (Fractional) Brownian Bridges are widely used as an initial conditions for the Navier-Stokes equations to study statistical properties of turbulent flows in the computational physics literature, see for instance \cite{LMP1} and references therein.  

We generate Brownian Bridges directly in Fourier space with the following method:
\begin{equation}
    W(x) = \sum_{|\mathbf{k}|_{\infty} \leq N} \frac{1}{\left\|\mathbf{k}\right\|_2^{\frac{3}{2}}} \sum_{m,n,\ell\in\{0,1\}} \alpha_k^{(mn\ell)}\text{sc}_{m}(x)\text{sc}_{n}(x)\text{sc}_{\ell}(x)
\end{equation}
where
\begin{equation}
    \text{sc}_i(x) = \begin{cases}
        \sin(x) &\mbox{ for } i = 0 \\
        \cos(x) &\mbox{ for } i = 1
    \end{cases}
\end{equation}
and the $\alpha_k^{(mn\ell)} \sim \mathcal{U}_{[-1,1]}$. These Brownian Bridges are propagated through the discretized Navier-Stokes system (\ref{eq:shv_disc}) from time $t = -0.5$ to $t=0$. The resulting flow fields are then taken as initial conditions for this dataset. 

The underlying solution operator $\sol(t,\cdot)$ is given by $\sol(t,u^0_{x,y}) = u_{x,y}(t)$, with $u_x,u_y$ solving the Navier-Stokes equations \eqref{eq:NS} with periodic boundary conditions.

We generated 20000 NS-BB trajectories with the same train/validation/test split as NS-Sines. The same time-stepping is used as for NS-PwC. The testing error is evaluated at the 14th time step (i.e. $t = 0.7$). A visualization of a random sample and the predictions made by \textsc{Poseidon}, CNO and FNO ($128$ training trajectories)  are shown in Figure \ref{fig:ns_bb}.

\subsubsection{NS-SL} The Shear Layer (SL) is a well-known benchmark for the Navier-Stokes equations \eqref{eq:NS}, stemming all the way from \cite{BCG1}, if not earlier, see \cite{LMP1} for a modern (stochastic) version, whose variant we consider here. 

We take as initial conditions the shear layer, 
\begin{equation}
    \begin{aligned}
        u_0(x, y) &= \begin{cases}
            \tanh\left(2\pi\frac{y-0.25}{\rho}\right) &\mbox{ for } y+\sigma_{\delta}(x) \leq \frac{1}{2} \\
            \tanh\left(2\pi\frac{0.75-y}{\rho}\right) &\mbox{ otherwise}
        \end{cases} \\
        v_0(x, y) &= 0
    \end{aligned}
\end{equation}
where $\sigma_{\delta}: [0,1] \to \mathbb{R}$ is a perturbation of the initial data given by
\begin{equation}
    \sigma_{\delta}(x) = \xi + \delta \sum_{k=1}^{p} \alpha_k\sin(2\pi kx - \beta_k).
\end{equation}
The parameters are chosen to be $p \sim \mathcal{U}_{\{7, 8, \dots 12\}}$ $\alpha_k \sim \mathcal{U}_{[0, 1]}$, $\beta_k \sim \mathcal{U}_{[0, 2\pi]}$, $\delta = 0.025$, $\rho \sim \mathcal{U}_{[0.08, 0.12]}$, and $\xi \sim \mathcal{U}_{[-0.0625, 0.0625]}$.

The underlying solution operator $\sol(t,\cdot)$ is given by $\sol(t,u^0_{x,y}) = u_{x,y}(t)$, with $u_x,u_y$ solving the Navier-Stokes equations \eqref{eq:NS} with periodic boundary conditions.

We generated 40000 NS-SL trajectories of which the first 39640 are in the training split, the next 120 in the validation split, and the remaining 240 in the test split. The same time-stepping is used as for NS-PwC. The testing error is evaluated at the 14th time step (i.e. $t = 0.7$). A visualization of a random sample and the predictions made by \textsc{Poseidon}, CNO and FNO ($128$ training trajectories) are shown in Figure \ref{fig:ns_sl}.

\subsubsection{NS-SVS} The \emph{Sinusoidal Vortex Sheet} (SVS) is another classic numerical benchmark for the Navier-Stokes equations \cite{Kras} and references therein. We consider a modern (stochastic) version from \cite{LMP1} here. The initial datum for this problem is specified in terms of the vorticity, by setting, 

\begin{equation}
    \omega_0^{\rho} = \psi_{\rho} * \omega_0
\end{equation}
where
\begin{align}
    \omega_0(x) &= \delta(x - \Gamma) - \int_{\mathbb{T}^2}\! \,\mathrm{d}\Gamma \\
    \phi_{\rho}(x) &= \rho^{-2}\psi\left(\frac{\|x\|}{\rho}\right) \\
    \psi(r) &= \frac{80}{7\pi} \left[ (r+1)_+^3 - 4(r+\frac{1}{2})_+^3 + 6r_+^3 - 4(r-\frac{1}{2})_+^3 + (r-1)_+^3 \right] \\
    \Gamma &= \{ (x,y) \in \mathbb{T}^2 \mid y = \frac{1}{2}+0.2\sin(2\pi x) + \sum_{i=1}^p \alpha_i\sin(2\pi(x+\beta_i)) \}.
\end{align}
We choose $p = 10$ and the random variables $\alpha_i$ and $\beta_i$ are given by $\alpha_i \sim \mathcal{U}_{[0,0.003125]}$, $\beta_i \sim \mathcal{U}_{[0, 1]}$. The parameter $\rho$ is chosen to be $\rho = \frac{5}{128}$.

We generated 20000 NS-SL trajectories with the same training/validation/test split as NS-Sines. The same time-stepping is used as for NS-PwC. The testing error is evaluated at the 14th time step (i.e. $t = 0.7$). A visualization of a random sample and the predictions made by \textsc{Poseidon}, CNO and FNO ($128$ training trajectories)  are shown in Figure \ref{fig:ns_svs}.

\subsubsection{NS-Tracer-PwC} This downstream task is the first of our tasks, where the underlying physics has not been completely encountered in the pretraining dataset.

In this experiment, we focus on the important problem of transport of a passive tracer, for instance a pollutant in a river. This tracer is carried along by the Navier-Stokes flow field without feeding back into the velocity. Let $c = c(x,y,t)$ be the concentration of the passive scalar in the fluid. The equation that governs $c$ is given by
\begin{equation}
    \label{eq:passive_sc}
    \frac{\partial c}{\partial t} + \mathbf{u} \cdot \nabla c = 
    \kappa \Delta c,
\end{equation}
where $\mathbf{u}$ is the velocity field of the fluid, which in turn is governed by the Navier-Stokes equations \eqref{eq:NS}, and $\kappa$ is the diffusivity constant. We choose $\kappa$ to be equal to the the artificial viscosity term used in the simulation of the flow (see \ref{sec:datapt} for clarification).

The fluid velocity field $\mathbf{u}$ has the exact same initial data as in the NS-PwC experiment. The tracer concentration $c$ is initialized as a sphere centered in the center of the domain
\begin{equation}
    c_0(x, y) = \mathbbm{1}_{B_{\frac{1}{4}}(\frac{1}{2},\frac{1}{2})}(x, y).
\end{equation}
Thus, the source of stochasticity in this problem is purely the random initial condition driving the fluid flow. 

The underlying solution operator $\sol(t,\cdot)$ is now given by $\sol(t,u^0_x,u^0_y,c_0) = [u_x(t),u_y(t),c(t)]$, with $u_x,u_y$ solving the Navier-Stokes equations \eqref{eq:NS} with periodic boundary conditions and $c$ solving the transport equation \eqref{eq:passive_sc}. 

We generated 20000 NS-Tracer-PwC trajectories with the same train/validation/test split as NS-Sines. The same time-stepping is used as for NS-PwC. The testing error is evaluated for the 14th time step (i.e. $t = 0.7$). A visualization of a random sample and the predictions made by \textsc{Poseidon}, CNO and FNO ($128$ training trajectories)  are shown in Figure \ref{fig:ns_tracer_pwc}.

\subsubsection{FNS-KF} Another downstream task which introduces a physical process that has not been encountered in the pretraining dataset, a two-dimensional version of the well-known Kolmogorov Flow \cite{MB1} is modeled by Navier-Stokes equations with a forcing term, namely
\begin{equation}
    \label{eq:FNS}
    u_t +(u\cdot \nabla)u + \nabla p - \nu \Delta u = f, \quad {\rm div}~u =0,
\end{equation}
in the domain $[0,1]^2$ with periodic boundary conditions. The forcing term $f$ is chosen to be constant in time and is equal to
\begin{equation}
\label{eq:fin}
    f(x, y) = 0.1\sin(2\pi(x+y)).
\end{equation}
The fluid velocity field $u$ is initialized in the exact same way as in the NS-PwC experiment. The data is simulated with the same method as the other flows governed by Navier-Stokes equations (see \ref{sec:datapt} for clarification).

We also remark that this problem can be readily recast in the generic form \eqref{eq:pde} by considering the augmented solution vector $U = [u_x,u_y,f]$ and augmenting the PDE \eqref{eq:pde} with the trivial equation $f_t = 0$ and augmenting the initial data with \eqref{eq:fin}. The underlying solution operator $\sol(t,\cdot)$ is then given by $\sol(t,U^0_{x,y}) = [u_x(t),u_y(t),f]$, with $u_x,u_y$ solving the forced Navier-Stokes equations \eqref{eq:FNS} with periodic boundary conditions and $f$ being given by \eqref{eq:fin}. 

We generated 20000 FNS-KF trajectories with the same train/validation/test split as NS-Sines. The same time-stepping is used as for NS-PwC. The testing error is evaluated at the 14th time step (i.e. $t = 0.7$). A visualization of a random sample and the predictions made by \textsc{Poseidon}, CNO and FNO ($128$ training trajectories) are shown in Figure \ref{fig:fns_kf}.

\subsubsection{CE-RPUI} This downstream task considers the compressible Euler equations and is a variant of the uncertain interface problem considered in \cite{schwabsid1} as well as a (hard) perturbation of CE-RP, where not just the amplitude of the jumps for each Riemann problem is randomly varied, but even the location and shape of the initial interfaces is randomly perturbed. To realize this construction, we denote the fractional part any $x \in \R$ as $\{x\} := x - \lfloor|x|\rfloor\sign x$ and define the functions
\begin{equation*}
    \begin{aligned}
        \sigma_x(x, y) &= \sum_{i,j=1}^p \alpha_{x,i,j}\sin(2\pi (i+2p^2)x + (j+2p^2)y + \beta_{x,i,j}) \\
        \sigma_y(x, y) &= \sum_{i,j=1}^p \alpha_{y,i,j}\sin(2\pi (i+2p^2)x + (j+2p^2)y + \beta_{y,i,j}).
    \end{aligned}
\end{equation*}
where $\alpha_{k,i,j} \sim \mathcal{U}_{[-0.01, 0.01]}$, and $\beta_{k,i,j} \sim \mathcal{U}_{[0, 1]}$. These functions are then used to create a partitioning of the domain into subdomains
\begin{equation*}
    D_{i,j} = \{ (x, y) \in \mathbb{T}^2 \mid x_{\text{min}} \leq \{x+\sigma_x(x,y)+1\} < x_{\text{max}}, y_{\text{min}} \leq \{y+\sigma_y(x,y)+1\} < y_{\text{max}} \}.
\end{equation*}
with $x_{\text{min}} = \frac{i}{p+1}$, $x_{\text{max}} = \frac{i+1}{p+1}$, $y_{\text{min}} = \frac{j}{p+1}$, and $y_{\text{max}} = \frac{j+1}{p+1}$. Finally, the initial conditions are given by
\begin{equation*}
    (\rho, v_x, v_y, p)|_{t=0} = (\rho_{i,j}, v^x_{i,j}, v^x_{i,j}, p_{i,j}) \mbox{ in } D_{i,j}
\end{equation*}
where $\rho_{i,j} \sim \mathcal{U}_{[1, 3]}$, $v^x_{i,j} \sim \mathcal{U}_{[-10, 10]}$, $v^y_{i,j} \sim \mathcal{U}_{[-10, 10]}$, and $p_{i,j} \sim \mathcal{U}_{[5, 7]}$.

The underlying solution operator $\sol(t,\cdot)$ is given by $\sol(t,\rho_0,v^0_{x,y},p_0) =[\rho(t),v_{x,y}(t),p(t)]$ solving the compressible Euler equations \eqref{eq:ceuler} with periodic boundary conditions. 

We generated 10000 CE-RPUI trajectories with the train/validation/test split being the same as for CE-RP. The same time-stepping is used as for NS-PwC. The testing error is evaluated at the 14th time step (i.e. $t = 0.7$). A visualization of a random sample and the predictions made by \textsc{Poseidon}, CNO and FNO ($128$ training trajectories) are shown in Figure \ref{fig:ce_rpui}.

\subsubsection{CE-RM} Another well-known benchmark for the compressible Euler equations \eqref{eq:ceuler} is the Richtmeyer-Meshkov problem, see \cite{LL1}. A modern (stochastic) version is provided in \cite{FKMT1}. The compressible Euler equations are considered with the initial data given by, 

\begin{equation}\label{eq:rm}
p_0(x,y)=\begin{cases}
20 & \text{if } \sqrt{x^2 + y^2} < 0.1\\
1 & \text{otherwise.}
\end{cases}\qquad \rho_0(x,y) = \begin{cases} 2 & \text{if }   |x|< I(x,y,\omega)\\
1 & \text{otherwise}\end{cases} \quad v^x_0=w^y_0=0
\end{equation}
We assign periodic boundary conditions on $D=[0,1]^2$. The interface between the two states is given as
\begin{equation}
I(x, y,\omega) = 0.25+\epsilon \sum_{j=1}^K a_j(\omega)\sin(2\pi((x,y)+b_j(\omega))),
\end{equation}
where $K=10$, $\epsilon>0$, and $a_j$ and $b_j$ (for $j=1,\dots,K$) are uniform random variables on the interval $[0,1]$. We normalize the $a_j$ such that $\sum_j a_j=1$. We simulate up to $T=2$.

The underlying solution operator $\sol(t,\cdot)$ is given by $\sol(t,\rho_0,v^0_{x,y},p_0) =[\rho(t),v_{x,y}(t),p(t)]$ solving the compressible Euler equations \eqref{eq:ceuler} with periodic boundary conditions. 

We generated 1260 CE-RM trajectories with a train/validation/test split of 1030/100/130. The approximate solutions where generated with the FISH hydrodynamic code, see \cite{FKMT1} and references therein which implements high-resolution finite volume schemes. We save 21 snapshots, evenly spaced in time. The testing error is evaluated at the 14th time step (i.e. $t = 1.4$). A visualization of a random sample and the predictions made by \textsc{Poseidon}, CNO and FNO ($128$ training trajectories)  are shown in Figure \ref{fig:ce_rm}.

\subsubsection{GCE-RT} The compressible Euler equations with gravitation (GCE) are given by, 
\begin{equation}
    \label{eq:gce}
    \begin{aligned}
    u_t + {\rm div}~F(u) &=S,~u=[\rho,\rho v, E]^\perp,~ F=[\rho v, \rho v\otimes v+p{\bf I},
    (E+p)]v]^\perp, \\
    S &= [0,-\rho,0,-\rho v_x]\frac{\partial \varphi}{\partial x} + [0,0,-\rho,-\rho v_y]\frac{\partial \varphi}{\partial y}, 
    \end{aligned}
\end{equation}
with $\rho,v_{x,y},p$ be as defined in \eqref{eq:ceuler} and $\varphi$ being the \emph{gravitational potential}. 

For this experiment, we follow a well-known benchmark in astrophysics, namely the Rayleigh-Taylor (RT) instability on a model \emph{neutron star}, realized as a $\gamma = 2$ polytrope in gravitational equilibrium. Our benchmark is a two-dimensional stochastic variant of the setup of \cite{kapsid1}, Section 3.2.4, with the only variation being provided by the random fields used to generate the initial conditions.
The domain is $D = [-1/2, +1/2]^2$ and the pressure and gravitational potential are given by
\begin{equation}
p(r) = K_{0} \left( \rho_{0} \frac{\sin(\alpha r)}{\alpha r} \right)^2
,
\quad
\varphi(r) = - 2 K_{0} \rho_{0} \frac{\sin(\alpha r)}{\alpha r}
,
\end{equation}
where $r = \sqrt{x^{2} + y^{2}}$ is the radius, $K_{0} = p_{0} / \rho_{0}^{2}$ is the polytropic constant,
\begin{equation}
\alpha = \sqrt{\frac{4 \pi G}{2 K_{0}}}
\end{equation}
and $G = 1$ is the gravitational constant.
The initial velocity is set to vanish.
The density profile is set as
\begin{equation}
\rho(r) = \sqrt{\frac{K_{0}}{\tilde{K}(r)}} \rho_{0} \frac{\sin(\alpha r)}{\alpha r}
,
\end{equation}
where
\begin{equation}
  \tilde{K}(r)
= \begin{cases}
    K_{0} , & r < r_{\mathrm{RT}} \\
    \left(\frac{1 - A}{1 + A}\right)^{2} K_{0}, & r \geq r_{\mathrm{RT}}.
  \end{cases}
\end{equation}
Here, $A$ is the Atwood number which parameterizes the density jump between the heavier and lighter fluid, characterizing the Rayleigh-Taylor instability.
The interface between the fluids is given as
\begin{equation}
r_{\mathrm{RT}} = 0.25(1+ a \cos\left(\operatorname{atan2}(y, x)+b\right)),
\end{equation}
where the amplitude $a$ and phase $b$ are uniform random variables on $[-1,1]$ and $[-\pi,\pi]$, respectively.
Similarly, we perturb the central density $\rho_{0}$, pressure $p_{0}$ and Atwood number as
\begin{equation}
\rho_{0} = 1 + 0.2 c, \quad p_{0} = 1 + 0.2 d, A = 0.05 (1 + 0.2 e)
,
\end{equation}
where $c, d, e$ are uniform random variables on $[-1,1]$. We evolve the initial state up to a final time of $T=5$ and save 11 snapshots, evenly spaced in time.

The new physical phenomena that we add in this case is \emph{gravitational forcing} and the underlying solution operator $\sol(t,\cdot)$ is given by $\sol(t,\rho_0,v^0_{x,y},p_0,\varphi) =[\rho(t),v_{x,y}(t),p(t),\varphi]$ solving the gravitational Euler equations \eqref{eq:gce} with periodic boundary conditions.

We generated 1260 GCE-RT trajectories (with the same train/validation/test split as CE-RM) with a well-balanced second-order finite volume method, as described in \cite{kapsid1}, at $256^2$ resolution, then downsampled to $128^2$. The testing error is evaluated at the 7th time step, and we take every snapshot up to and including the 7th as training data. A visualization of a random sample and the predictions made by \textsc{Poseidon}, CNO and FNO ($128$ training trajectories)  are shown in Figure \ref{fig:gce_rt}.

\subsubsection{Wave-Gauss}

We consider the wave equation with a spatially varying propagation speed, i.e.
\begin{equation}
\label{eq:wave}
u_{tt}- (c(x))^2 \Delta u = 0, ~ {\rm in}~D\times(0,T),
\end{equation}
in order to model the propagation of acoustic waves in a spatially varying medium. 

The initial condition $u_0$ is given by a sum of several Gaussians whose parameters are drawn uniformly at random. First, we draw a random integer $n$ from the set $\{2,3,4,5,6\}$. Then, for $1 \leq i \leq n$, we draw two locations, $x_{c,i},y_{c,i} \sim \mathcal{U}_{[1/6, 5/6]}$. We fix the amplitude of the $i$th Gaussian to $1.0$ and draw the $i$th standard deviation $s_i \sim \mathcal{U}_{[0.039, 0.156]}$ Note that we restrict any two centers of the Gaussians to be closer than $2$ standard deviations from each other. If this happens, we draw a new point and discard the old one. The $i$th Gaussian is defined as
\begin{equation*}
g_i(x,y) = \exp \left( -\frac{(x_{c,i} - x)^2 + (y_{c,i} -  y)^2}{2 
s_i^2} \right), \quad x,y \in (0,1).
\end{equation*}

Finally, the initial condition $u_0$ is defined by
\begin{equation}
\label{eq:wave_ic}
u_0(x,y) = \sum_{i = 1}^{n} g_i(x,y), \quad x,y \in (0,1).
\end{equation}

We use absorbing boundary conditions. The propagation speed $c$ is spatially dependent and is generated as a sum of Gaussians in several steps. First, a random \textit{base} speed $c_0$ is generated such that $c_0 \sim \mathcal{U}_{[1500, 2500]}$. Then, we select $4$ points in the domain, namely, $(x_1, y_1) = (0.25, 0.25)$, $(x_2, y_2) = (0.25, 0.75)$, $(x_3, y_3) = (0.75, 0.25)$ and $(x_4, y_4)= (0.75, 0.75)$. For each point $i$, we define a random vector $(dx_i, dy_i)$, where $dx_i, dy_i \sim \mathcal{U}_{[-0.3125, 0.3125]}$.  We also draw an amplitude $v_i \sim \mathcal{U}_{[1000, 2500]}$ of a Gaussian that corresponds to the $i$-th point, as well as its standard deviation $\sigma_i \sim \mathcal{U}_{[1/12, 1/6]}$. The $i$th Gaussian is defined by
\begin{equation*}
f_i(x,y) = v_i \cdot \exp \left( -\frac{(x_i + dx_i - x)^2 + (y_i + dy_i -y)^2}{2 
\sigma_i^2} \right), \quad x,y \in (0,1).
\end{equation*}

Finally, the propagation speed is defined by
\begin{equation*}
c(x,y) = c_0 + \sum_{i = 1}^{4} f_i(x,y), \quad x,y \in (0,1).
\end{equation*}

Trajectories are generated with a finite-difference method, similar to the DeVITO code \cite{devito} at $128^2$ resolution. The final time of all the simulations is $T = 1$. We save 15 snapshots, evenly spaced in time.

Thus, this benchmark models the propagation of acoustic waves, generated by seismic sources, which propagate in a smoothly varying medium. The wave equation \eqref{eq:wave} can be readily recast into the generic form \eqref{eq:pde} by adding the time-derivative $v=u_t$ and the coefficient $c$ into the solution vector $U = [u(x,t),v(x,t),c(x)]$. Thus, the differential operator in \eqref{eq:pde} can be rewritten as, 
\begin{equation}
\label{eq:wavesys}
u_t = v, \quad v_t = c^2 \Delta u, \quad c_t = 0,
\end{equation}
and the resulting solution operator is $\sol(t,U_0) = [u(t),v(t),c]$. 

We generated $10512$ Wave-Gauss trajectories with a train/validation/test split of 10212/60/240. The same time-stepping is used as for NS-PwC. The testing error is evaluated at the 14th time step (i.e. $t = 0.7$). A visualization of a random sample and the predictions made by \textsc{Poseidon}, CNO and FNO are shown in Figure \ref{fig:wave_gaussians}.

\subsubsection{Wave-Layer}

In the Wave-Layer experiment, we also consider the wave equation with spatially dependent propagation speed \eqref{eq:wave}, initial conditions given by \eqref{eq:wave_ic}. We use absorbing boundary conditions.

The propagation speed $c$ varies spatially and is generated as a (vertically) layered medium, with each layer having a constant propagation speed drawn uniformly at random. To generate one instance of $c$, we first draw a random integer $n$ from $\{3,4,5,6\}$, where $n$ represents a number of layers in $c$. Then, for each $2\leq i \leq n$, we generate a $x-$dependent \textit{frontier}, defined by
\begin{equation*}
a_i(x) = \frac{i}{n} + c_0 +  \sum_{i = 1}^{10} \frac{a_i}{i} \sin(2\pi i x),
\end{equation*}

where, first, $a_i$ values are drawn uniformly at random from $(0,1)$ and then $c_0$ is drawn uniformly at random from $(0,1)$ and it is rescaled by a constant that depends on $i$ so that the adjacent frontiers are impossible to intersect. Finally, a point $(x,y) \in (0,1)^2$ is in $i$-th frontier if and only if $a_i(x) \leq y\leq a_{i+1}(x)$, with $a_1 = 0$ and $a_{n + 1} = 0$. Each layer $i$ has a constant speed of propagation $c_i \sim \mathcal{U}_{[2000, 5000]}$. Trajectories are generated by a finite-difference method at $128^2$ resolution. The final time of all the simulations is $T = 1$. We save 21 snapshots, evenly spaced in time.

As in the Wave-Gauss benchmark, the resulting solution operator is $\sol(t,U_0) = [u(t),v(t),c]$, with $v=u_t$ and coefficient $c$. The wave-layer experiment models the propagation of acoustic waves, generated by seismic sources, inside a layered subsurface medium. 

We generated 10512 Wave-Layer trajectories with the same train/validation/test split as Wave-Gauss. The same time-stepping is used as for NS-PwC. The testing error is evaluated at the 14th time step (i.e. $t = 0.7$). A visualization of a random sample and the predictions made by \textsc{Poseidon}, CNO and FNO are shown in Figure \ref{fig:wave_layer}.

We remark that both the Wave-Gauss and Wave-Layer tasks are very different from the pretraining dataset as the wave equation is a linear second-order (in time and space) equation that is different from the compressible Euler and incompressible Navier-Stokes equations that form the pretraining dataset. 

\subsubsection{ACE}
The Allen-Cahn equation for modeling phase transitions in material science is given by
\begin{equation}
    \label{eq:ac}
    u_t= \Delta u - \epsilon^2 u(u^2-1), 
\end{equation}
with a reaction rate of $\epsilon=220$. We consider this equation with periodic boundary conditions and initial conditions given by
\begin{equation*}
u_0(x,y) = \frac{1}{K^2} \sum_{i,j=1}^{K} a_{ij}\cdot (i^2 + j^2)^{-r} \sin(\pi i x) \sin(\pi j y),\quad \forall x, y \in (0, 1),
\end{equation*}

where $K$ is a random integer drawn uniformly at random from $[16, 32]$, $r\sim \mathcal{U}_{[0.7, 1.0]}$ and $a_{ij} \sim \mathcal{U}_{[-1, 1]}$.

Trajectories are generated by a finite-difference method at $128^2$ resolution. The final time of all the simulations is $T = 0.0002$. We save 20 snapshots, evenly spaced in time.

The corresponding solution operator is $\sol(t,u_0) = u(t)$ and maps the initial concentration to the concentration at time $t$.

We generated 15000 ACE trajectories with a train/validation/test split of 14700/60/240. The same time-stepping is used as for NS-PwC. The testing error is evaluated at the 14th time step. A visualization of a random sample and the predictions made by \textsc{Poseidon}, CNO and FNO ($128$ training trajectories)  are shown in Figure \ref{fig:ace}.

Again, it is essential to emphasize that the Allen-Cahn equation is a nonlinear parabolic reaction-diffusion equation that is very different from the PDEs used in constructing the pretraining dataset. 

\subsubsection{SE-AF}
This dataset contains the samples that describe the classical computational physics benchmark of flow past airfoils, modeled by the compressible Euler equations \eqref{eq:ceuler}. The samples are \textit{not} time-dependent, as we are interested in the \emph{steady-state} solution.

We follow standard practice in aerodynamic shape optimization and consider a reference airfoil shape with upper and lower surface of the airfoil are located at $(x, y^\text{U}_{\text{ref}}(x/c))$ and $(x, y^\text{L}_{\text{ref}}(x/c))$ where $c$ is the chord length and $y^\text{U}_{\text{ref}}$ and $y^\text{L}_{\text{ref}}$ corresponding to the well-known RAE2822 airfoil \cite{LMRP1}. The reference shape is then perturbed by \emph{Hicks-Henne Bump functions} \cite{HH1} :
\begin{gather*}
    y^\text{L}(\xi) = y^\text{L}_{\text{ref}}(\xi) + \sum_{i=1}^{15} a_i^\text{L}B_i(\xi), \ \ \
    y^\text{U}(\xi) = y^\text{U}_{\text{ref}}(\xi) + \sum_{i=1}^{15} a_i^\text{U}B_i(\xi), 
    \\
    B_i(\xi) = \text{sin}^3(\pi\xi^{q_i}),\ \ \ q_i = \frac{\text{ln}2}{\text{ln}14 - \text{ln}i}, \ \ \ \xi = \frac{x}{c},
    \\
    a_i^\text{L} = 2(\psi_i - 0.5)(i+1)\times 10^{-3},\ \ \ 
    a_i^\text{U} = 2(\psi_{i+10} - 0.5)(11-i)\times 10^{-3}, \quad i=1,...,15
\end{gather*}
with $ \psi\in [0,1]^d$.
We can now formally define the airfoil shape as $\mathcal{S}=\{(x,y)\in D : x\in[0,c], y^L\leq y\leq y^U\}$ and accordingly the shape function $f=\chi_{[\mathcal{S}]}(x,y)$, with $\chi$ being the \textit{characteristic function}.

The underlying operator of interest maps the shape function  $f$ into the density of the flow $\rho$ at steady state of the compressible Euler equations.

\begin{figure}
    \centering
    \includegraphics[width=0.45\textwidth]{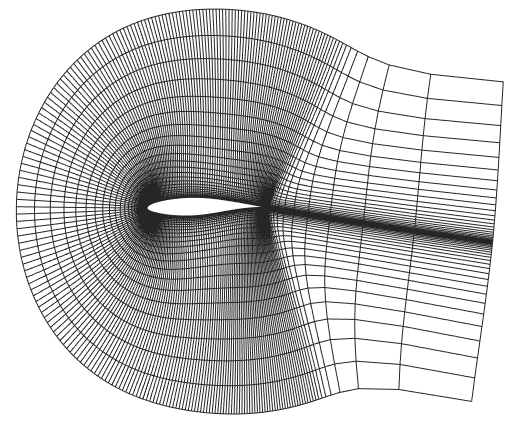}
    \caption{Elliptic mesh for the airfoil problem}
    \label{fig:airfoil_mesh}
\end{figure}

The equations are solved with the solver NUWTUN, see \cite{LMR1} and references therein, on $243 \times 43$ elliptic mesh (see Figure \ref{fig:airfoil_mesh}) given the following  free-stream boundary conditions,
\begin{equation*}
    T^\infty = 1, \ \ \ M^\infty = 0.729, \ \ \ p^\infty = 1, \ \ \ \alpha = 2.31^\circ.
\end{equation*}
The data is ultimately interpolated onto a Cartesian grid of dimensions $128 \times 128$ on the underlying domain $D=[-0.75, 1.75]^2$, and unit values are assigned to the density $\rho(x,y)$ for all $(x,y)$ in the set $\mathcal{S}$. The shapes of the training data samples correspond to $30$ bump functions, with coefficients $\psi$ sampled uniformly from $[0,1]^{30}$. During the training and evaluation processes, the difference between the learned solution and the ground truth is exclusively calculated for the points $(x,y)$ that do not belong to the airfoil shape $\mathcal{S}$.

We generated 10869 SE-AF solutions with a train/validation/test split of 10509/120/240. A visualization of a random sample and the predictions made by \textsc{Poseidon}, CNO and FNO ($128$ training samples)  are shown in Figure \ref{fig:se_af}.

We note here that this SE-AF benchmark differs from what has been seen during pretraining in many aspects, namely i) the problem is time-independent, in contrast to the time-dependent PDEs for pretraining, ii) the solution operator is very different as it maps a (shape) coefficient into the steady state solution, and iii) the geometry of the underlying domain is non-Cartesian and the boundary conditions are very different from what was encountered during pretraining. 

\subsubsection{Poisson-Gauss}
We consider the Poisson equation,
\begin{equation}
\label{eq:pos}
-\Delta u = f, ~ {\rm in}~(0,1)^2,
\end{equation}
with homogeneous Dirichlet boundary conditions. The solution operator maps the source term $f$ to the solution $u$. The source term $f$ consists of a superposition of a random number of Gaussians 
\begin{equation*}
f(x, y) = \sum_{i=1}^N \exp\left(-\frac{(x - \mu_{x, i})^2 + (y - \mu_{y, i})^2}{2 \sigma_i^2}\right)
\end{equation*}
with $N$ being an integer drawn from a geometric distribution $\text{Geom}(0.4)$, $\mu_{x, i}, \mu_{y, i} \sim \mathcal{U}_{[0, 1]}$ and $\sigma_i \sim \mathcal{U}_{[0.025, 0.1]}$. Thus, this experiment models the diffusion of an input (source) which is a superposition of Gaussians. 

We generated 20000 Poisson-Gauss solutions (with a train/validation/test split of 19640/120/240) with a finite element method based on FENICS \cite{fenics}. A visualization of a random sample and the predictions made by \textsc{Poseidon}, CNO and FNO ($128$ training samples)  are shown in Figure \ref{fig:poisson_gaussians}.

We note here that both the Poisson-Gauss and Helmholtz benchmarks differ from what has been seen during pretraining in many aspects, namely i) the problems are time-independent, in contrast to the time-dependent PDEs for pretraining, ii) the solution operator is very different as it maps coefficients into the steady-state solution, and iii) the boundary conditions are very different from the periodic boundary conditions, seen during pretraining.

\subsubsection{Helmholtz}
The Helmholtz equation models wave propagation in the frequency domain. We consider a variant of this equation given by
\begin{equation}
\label{eq:helmholtz}
-\Delta u - \omega^2 a(x,y) u =  0, \quad x,y \in D,
\end{equation}
and Dirichlet boundary conditions
\begin{equation*}
u(x,y) = b, \quad x,y \in \partial D,
\end{equation*}
where $\omega =  5\pi/2$ is the frequency, $D = (0,1)^2$ is the domain, $a$ is the spatial dependent function that defines properties of the medium of the wave propagation and $b$ is the fixed value of the solution $u$ at the boundary $\partial D$. 

The boundary value $b$ follows uniform ditribution, namely $b \sim \mathcal{U}_{[0.25, 0.5]}$. The function $a$ is defined as a sum of random number of Gaussians and is generated in several steps. First, we draw an integer $n$ from $[2,7]$ uniformly at random. This number represents the number of Gaussians that will be randomly generated. For $1 \leq i \leq n$, it holds that $A_i \sim \mathcal{U}_{[0.5, 10.0]}$ and $\sigma_i \sim \mathcal{U}_{[0.05, 0.1]}$. Additionally, two numbers $x_i, y_i$ that represent x and y coordinates of the Gaussians are generated such that $x_i, y_i \sim \mathcal{U}_{[0.2, 0.8]}$. The unnormalized function $\bar{a}$ is obtained by
\begin{equation*}
\bar{a}(x,y) = - \sum_{i =1} ^n A_i \exp \left( -\frac{(x_i - x)^2 + (y_i -  y)^2}{2 
\sigma_i^2} \right), \quad x,y \in (0,1).
\end{equation*}
Function $a$ is obtained by normalizing $\bar{a}$, i.e.
\begin{equation*}
a(x,y) = \frac{\bar{a} - \min(\bar{a})}{\max(\bar{a}) - \min(\bar{a})}.
\end{equation*}
The solution operator maps the tuple $(a, b)$ to the solution $u$. This problem is a \emph{steady-state} problem. Trajectories are generated by a finite-difference method at $128^2$ resolution, similar to DeVito \cite{devito}. 

We generated 19675 Helmholtz solutions with a train/validation/test split of 19035/128/512. A visualization of a random sample and the predictions made by \textsc{Poseidon}, CNO and FNO ($128$ training trajectories) are shown in Figure \ref{fig:helmholtz}.

\clearpage
\section{Models and Baselines}
\label{sec:mb}

We compare multiple versions of \textsc{Poseidon} with foundation model baselines, namely MPP \cite{MPP} and a CNO \cite{CNO} foundation model that is trained in a similar manner as \textsc{Poseidon}. In addition, we compare against state-of-the-art neural operators trained from scratch, namely CNO and FNO \cite{FNO}, as well as scOT trained from scratch. All these models and their training strategies are described in the following. Their approximate model sizes can be read off from Table \ref{tab:model_sizes}.
\begin{table}[htb]
    \centering
    \caption{Approximate model sizes of all the models considered in this paper.}
    \begin{tabular}{r c}
    \toprule
    Model & Number of parameters \\
    \midrule\midrule
        \textsc{Poseidon}-L & 629M \\\midrule
        \textsc{Poseidon}-B & 158M \\\midrule
        \textsc{Poseidon}-T & 21M\\\midrule
        CNO-FM & 109M\\\midrule
        MPP-B & 116M\\\midrule
        CNO & 39M\\\midrule
        scOT & 40M\\\midrule
        FNO & 37M\\
        \bottomrule
    \end{tabular}
    \label{tab:model_sizes}
\end{table}

We train all models on a realization of Equation \ref{eq:lfij} in the Main Text; in particular, we set $p=1$. For each gradient step, we draw from the set of $N$ available trajectory snapshots $\{{\bf u}^l_{t_k} | {\bf u}^l_{t_k} \in \R^{c \times J \times J}\}_{l=1}^N$ where $c$ is the number of input/output functions, $J$ is the size of the computational grid, and $t_k$ is the (lead) time from the initial condition to the $k$-th snapshot in the trajectory, i.e. we get a batch of size $B$ of pairs $\{({\bf u}^m_{t_i}, {\bf u}^m_{t_j})_l\}_{l=1}^B$ where $i \leq j$. The loss is then computed as
\begin{equation}
    \mathcal{L}(\{({\bf u}^m_{t_i}, {\bf u}^m_{t_j})_l\}_{l=1}^B) = \frac{1}{c} \sum_{s=1}^c \frac{\sum_{l=1}^B\sum_{u,v=1}^J\abs{({\bf u}^m_{t_j})^l_{s,u,v} - \sol_\theta((t_j)^l - (t_i)^l, ({\bf u}^m_{t_i})^l_{s,u,v})}}{\sum_{l=1}^B\sum_{u,v=1}^J\abs{({\bf u}^m_{t_j})^l_{s,u,v}} + \epsilon}
\end{equation}
where $\epsilon = 10^{-10}$ for numerical stability, $\sol_\theta$ is the model, $(\cdot)^l$ is the $l$-th sample from the batch and $(\cdot)_{s,u,v}$ denotes the value at indices $(s,u,v)$.

During training, we create a checkpoint after every epoch, but only keep the checkpoint corresponding to the lowest validation loss (evaluated at the end of the epoch) which is then also used for testing.

\subsection{\textsc{Poseidon} Models}
\label{sec:pm}
In the following, we give thorough details for all \textsc{Poseidon} models that we pretrained, as well as details on finetuning these pretrained models. All models are pretrained on the datasets introduced in Section \ref{sec:datapt}, i.e. they expect four dimensional inputs and outputs, density $\rho$, velocities $u$ and $v$, and pressure $p$. During pretraining, we set $\rho=1$ and mask $p$ for all pretraining datasets corresponding to incompressible flow. Further, we use the full set of 77840 pretraining trajectories, unless otherwise specified.

To finetune the pretrained model on tasks whose input/output functions are not in the set of pretraining input/outputs ($\rho$, $u$, $v$, $p$) or where there are additional inputs/outputs -- this corresponds to the tasks NS-Tracer-PwC, FNS-KF, SE-AF, GCE-RT, Wave-Layer, Wave-Gauss, ACE, Poisson-Gauss, and Helmholtz -- we transfer all parameters from the pretrained model, except
\begin{itemize}
    \item the embedding weight ${\bf W}_{\mathcal{E}}$,
    \item the patch recovery weight ${\bf W}_{\mathcal{R}}$ and bias ${\bf b}_{\mathcal{R}}$, and
    \item the mixup convolutional kernel.
\end{itemize}
We refer to Section \ref{sec:scot2} for notation. This means that solely embedding/recovery is trained from scratch and just the parameters whose dimensions would not match, i.e. \emph{a minimal set of parameters} is trained from random initialization. All other parameters are kept trainable and \emph{no parameter is frozen}.

In general, we do not apply any weight decay on (time-conditioned) layer norm parameters. We finetune all parameters using the same optimizer with different learning rates, i.e. we build two or three parameter groups, depending on the finetuning task. In case the embedding and recovery do not have to be replaced, we finetune all parameters, except parameters of the (time-conditioned) layer norm, with learning rate $\widehat{\eta}$, and parameters of the layer norm with learning rate $\widetilde{\eta}_{\mathcal{N}}$. If the embedding/recovery is to be replaced and trained from scratch, we finetune all embedding/recovery parameters (including the embedding bias ${\bf b}_{\mathcal{E}}$) with learning rate $\widetilde{\eta}$, layer norm parameters with learning rate $\widetilde{\eta}_{\mathcal{N}}$, and all other parameters with learning rate $\widehat{\eta}$.

\subsubsection{\textsc{Poseidon}-T}
\textsc{Poseidon}-T is the smallest pretrained model, an instantiated scOT with the following hyperparameters:
\begin{itemize}
    \item \textbf{Embedding/latent dimension} $C$: 48
    \item \textbf{Number of SwinV2 transformer blocks at each level ($\forall i$)} $t_i$: 4
\end{itemize}

This results in a model with 21M parameters (for \textsc{Poseidon} models, we exclude embedding and recovery parameters in this count).

\paragraph{Pretraining} The model is pretrained on 8 NVIDIA RTX 4090 GPUs using the following (data-parallel) training protocol:
\begin{itemize}
 \item  \textbf{Optimizer:} AdamW \cite{loshchilov2019decoupled}
 
   \item  \textbf{Scheduler:} Cosine Decay with linear warmup of 2 epochs

   \item  \textbf{Maximum learning rate:} $10^{-3}$

    \item  \textbf{Weight decay:} $0.1$

   \item  \textbf{Effective batch size:} $640$, resulting in a per-device batch size of $80$
   
   \item  \textbf{Number of epochs:} $40$

   \item \textbf{Early stopping:} No

   \item \textbf{Gradient clipping (maximal norm):} 5
\end{itemize}

\paragraph{Finetuning} The pretrained model is finetuned on every task on a single GPU following this finetuning protocol ($\widetilde{\eta}$ is only applicable to certain downstream tasks):
\begin{itemize}
 \item  \textbf{Optimizer:} AdamW \cite{loshchilov2019decoupled}
 
   \item  \textbf{Scheduler:} Cosine Decay

   \item  \textbf{Initial learning rate} $\widehat{\eta}$: $5 \cdot 10^{-5}$

   \item  \textbf{Initial learning rate} $\widetilde{\eta}$: $5 \cdot 10^{-4}$

   \item  \textbf{Initial learning rate} $\widetilde{\eta}_{\mathcal{N}}$: $5 \cdot 10^{-4}$

    \item  \textbf{Weight decay:} $10^{-6}$

   \item  \textbf{Batch size:} 40
   
   \item  \textbf{Number of epochs:} $200$

   \item \textbf{Early stopping:} No

      \item \textbf{Gradient clipping (maximal norm):} 5
\end{itemize}

\subsubsection{\textsc{Poseidon}-B}
\textsc{Poseidon}-B is the base model, an instantiated scOT with the following hyperparameters:
\begin{itemize}
    \item \textbf{Embedding/latent dimension} $C$: 96
    \item \textbf{Number of SwinV2 transformer blocks at each level ($\forall i$)} $t_i$: 8
\end{itemize}

This results in a model with 158M parameters.

\paragraph{Pretraining} The model is pretrained on 8 NVIDIA RTX 4090 GPUs using the following (data-parallel) training protocol:
\begin{itemize}
 \item  \textbf{Optimizer:} AdamW \cite{loshchilov2019decoupled}
 
   \item  \textbf{Scheduler:} Cosine Decay with linear warmup of 2 epochs

   \item  \textbf{Maximum learning rate:} $5 \cdot 10^{-4}$

    \item  \textbf{Weight decay:} $0.1$

   \item  \textbf{Effective batch size:} $320$, resulting in a per-device batch size of $40$
   
   \item  \textbf{Number of epochs:} $39$ ($40$ were initially planned)

   \item \textbf{Early stopping:} No

   \item \textbf{Gradient clipping (maximal norm):} 5
\end{itemize}

\paragraph{Finetuning} The pretrained model is finetuned on every task on a single GPU following this finetuning protocol ($\widetilde{\eta}$ is only applicable to certain downstream tasks):
\begin{itemize}
 \item  \textbf{Optimizer:} AdamW \cite{loshchilov2019decoupled}
 
   \item  \textbf{Scheduler:} Cosine Decay

   \item  \textbf{Initial learning rate} $\widehat{\eta}$: $5 \cdot 10^{-5}$

   \item  \textbf{Initial learning rate} $\widetilde{\eta}$: $5 \cdot 10^{-4}$

   \item  \textbf{Initial learning rate} $\widetilde{\eta}_{\mathcal{N}}$: $5 \cdot 10^{-4}$

    \item  \textbf{Weight decay:} $10^{-6}$

   \item  \textbf{Batch size:} 40
   
   \item  \textbf{Number of epochs:} $200$

   \item \textbf{Early stopping:} No

      \item \textbf{Gradient clipping (maximal norm):} 5
\end{itemize}

\subsubsection{\textsc{Poseidon}-L}
\textsc{Poseidon}-L is the largest model we trained, an instantiated scOT with the following hyperparameters:
\begin{itemize}
    \item \textbf{Embedding/latent dimension} $C$: 192
    \item \textbf{Number of SwinV2 transformer blocks at each level ($\forall i$)} $t_i$: 8
\end{itemize}

This results in a model with 629M parameters.

\paragraph{Pretraining} The model is pretrained on 8 NVIDIA RTX 4090 GPUs using the following (data-parallel) training protocol:
\begin{itemize}
 \item  \textbf{Optimizer:} AdamW \cite{loshchilov2019decoupled}
 
   \item  \textbf{Scheduler:} Cosine Decay with linear warmup of 1 epoch

   \item  \textbf{Maximum learning rate:} $2 \cdot 10^{-4}$

    \item  \textbf{Weight decay:} $0.1$

   \item  \textbf{Effective batch size:} $128$, resulting in a per-device batch size of $16$
   
   \item  \textbf{Number of epochs:} $20$

   \item \textbf{Early stopping:} No

   \item \textbf{Gradient clipping (maximal norm):} 5
\end{itemize}

\paragraph{Finetuning} The pretrained model is finetuned on every task on a single GPU following this finetuning protocol ($\widetilde{\eta}$ is only applicable to certain downstream tasks):
\begin{itemize}
 \item  \textbf{Optimizer:} AdamW \cite{loshchilov2019decoupled}
 
   \item  \textbf{Scheduler:} Cosine Decay

   \item  \textbf{Initial learning rate} $\widehat{\eta}$: $5 \cdot 10^{-5}$

   \item  \textbf{Initial learning rate} $\widetilde{\eta}$: $5 \cdot 10^{-4}$

   \item  \textbf{Initial learning rate} $\widetilde{\eta}_{\mathcal{N}}$: $5 \cdot 10^{-4}$

    \item  \textbf{Weight decay:} $10^{-6}$

   \item  \textbf{Batch size:} 16
   
   \item  \textbf{Number of epochs:} $200$

   \item \textbf{Early stopping:} No

      \item \textbf{Gradient clipping (maximal norm):} 5
\end{itemize}

\subsubsection{Models for Dataset Ablations (see Section \ref{sec:perf_downstream_data})}
For models used in the pretraining dataset ablations, we utilize the same pretraining and finetuning strategies as for \textsc{Poseidon}-B. For the model trained on half of the pretraining dataset, we only train on the first half of each subset (NS-Sines, NS-Gaussians, CE-RP, CE-CRP, CE-KH, CE-Gauss); the same logic applies to the model trained on an eighth of the pretraining dataset. The model trained on a less diverse pretraining dataset is not trained on NS-Sines, CE-CRP, and CE-Gauss, such that the pretraining dataset size is directly comparable to the model trained on half of the pretraining dataset.

\subsection{scOT}
We additionally train a scOT from scratch on every downstream task, to compare its performance to \textsc{Poseidon} and other baselines. Its hyperparameters are as follows:
\begin{itemize}
    \item \textbf{Embedding/latent dimension} $C$: 48
    \item \textbf{Number of SwinV2 transformer blocks at each level ($\forall i$)} $t_i$: 8
\end{itemize}

This results in a model with 40M parameters. It is trained on one or multiple GPUs (depending on the dataset size) with the following parameters:
\begin{itemize}
 \item  \textbf{Optimizer:} AdamW \cite{loshchilov2019decoupled}
 
   \item  \textbf{Scheduler:} Cosine Decay with linear warmup of 20 epochs

   \item  \textbf{Maximum learning rate} $\widehat{\eta}$: $5 \cdot 10^{-4}$

    \item  \textbf{Weight decay:} $10^{-6}$

   \item  \textbf{Batch size:} 40 (on a single GPU, else the effective batch size is larger)
   
   \item  \textbf{Number of epochs:} $400$

   \item \textbf{Early stopping:} If the validation loss does not improve for 40 epochs

      \item \textbf{Gradient clipping (maximal norm):} 5
\end{itemize}

\subsection{CNO}
\label{sec:CNO}

A \textit{Convolutional Neural Operator} ({CNO}) is a model that (approximately) maps bandlimited functions to bandlimited functions \cite{CNO}. Let $\mathcal{B}_{w}$ be the space of bandlimited functions with the bandlimit $w$. A CNO is compositional mapping between function spaces $\mathcal{G}:\mathcal{B}_{w}(D) \to \mathcal{B}_{w}(D)$ and is defined as
\begin{equation}
\label{eq:CNO}
\mathcal{G}: u \mapsto P(u) = v_0 \mapsto v_1  \mapsto  \ldots v_{L} \mapsto Q(v_L) = \bar{u}, 
\end{equation}
where
\begin{equation}
\label{eq:udb}
v_{l+1} = \mathcal{P}_l \circ \Sigma_l \circ \mathcal{K}_l(v_l), \quad 1 \leq \ell \leq L-1,
\end{equation}
where $L$ is the number of CNO blocks and $D = (0,1)^2$ is the domain.

First, the input function $u \in \mathcal{B}_{w}(D)$ is lifted to the latent space of bandlimited functions through a \emph{lifting layer}:
\begin{equation*}
P : \left\{ u \in \mathcal{B}_w(D, \mathbb{R}^{d_\mathcal{X}}) \right\} \to \left\{ v_0 \in \mathcal{B}_w(D, \mathbb{R}^{d_{0}}) \right\}.
\end{equation*}

Here, $d_{0}\geq d_{\mathcal{X}}$ is the number of channels in the lifted, latent space. The lifting operation is performed by a convolution operator and activation operator which will be defined below. 

Then, the lifted function is processed through the composition of a series of mappings between functions (layers), with each layer consisting of three elementary mappings, i.e., $\mathcal{P}_l$ is either the \emph{upsampling} or \emph{downsampling} operator, $\mathcal{K}_l$ is the convolution operator and $\Sigma_l$ is the activation operator.

Finally, the last output function in the iterative procedure $v_L$ is projected to the output space with a \emph{projection operator} $Q$, defined as
\begin{equation*}
Q : \left\{ v_L \in \mathcal{B}_w(D, \mathbb{R}^{d_L}) \right\} \to \left\{ \overline{u} \in \mathcal{B}_w(D, \mathbb{R}^{d_\mathcal{Y}}) \right\}.
\end{equation*}

The projection operation is also performed by a convolution operator and activation operator.

\textit{Upsampling and Downsampling Operators. }
For some $\overline{w} > w$, we can \emph{upsample} a function $f \in \mathcal{B}_w$ to the \emph{higher band} $\mathcal{B}_{\overline{w}}$ by simply setting,
\begin{equation*}
\mathcal{U}_{w, \overline{w}} : \mathcal{B}_{w}(D) \to \mathcal{B}_{\overline{w}}(D), \quad \mathcal{U}_{w, \overline{w}}f(x) = f(x),\quad \forall x\in D.
\end{equation*}

On the other hand, for some $\underline{w} < w$, we can \emph{downsample} a function $f \in \mathcal{B}_w$ to the \emph{lower band} $\mathcal{B}_{\underline{w}}$ by setting $\mathcal{D}_{w, \underline{w}} : \mathcal{B}_w(D) \to \mathcal{B}_{\underline{w}}(D)$, defined by 
\begin{equation*}
\mathcal{D}_{w, \underline{w}} f(x) = \left(\frac{\underline{w}}{w}\right)^2 (h_{\underline{w}} \star f)(x) = \left(\frac{\underline{w}}{w}\right)^2\int_{D} h_{\underline{w}}(x - y) f(y) dy,\quad \forall x\in D,
\end{equation*}

where $\star$ is the convolution operation on functions defined above and $h_{\underline{w}}$ is the so-called \emph{interpolation sinc filter}:
\begin{equation}\label{eq:sinc-filter}
h_w (x_0, x_1) = \textnormal{sinc}(2wx_0)\cdot\textnormal{sinc}(2wx_1),\quad (x_0, x_1) \in \mathbb{R}^2.
\end{equation}

\textit{Activation Operator. }
First, the input function $f \in \mathcal{B}_w$ is upsampled to a higher bandlimit $\overline{w} > w$, then the activation function is applied and finally the result is downsampled back to the original bandlimit $w$. Implicitly assuming that $\overline{w}$ is large enough such that $\sigma\left(\mathcal{B}_{w}\right) \subset \mathcal{B}_{\overline{w}}$, we define the activation operator in \eqref{eq:CNO} as,
\begin{equation}
    \label{eq:act}
\Sigma_{w,\overline{w}} : \mathcal{B}_{w}(D) \to \mathcal{B}_w(D), \quad 
\Sigma_{w,\overline{w}} f(x) = \mathcal{D}_{\overline{w},w} (\sigma \circ \mathcal{U}_{w, \Tilde{w}} f) (x),\quad \forall x\in D.
\end{equation}

\begin{figure}[htbp]
  \centering
  \includegraphics[width=1.0\linewidth]{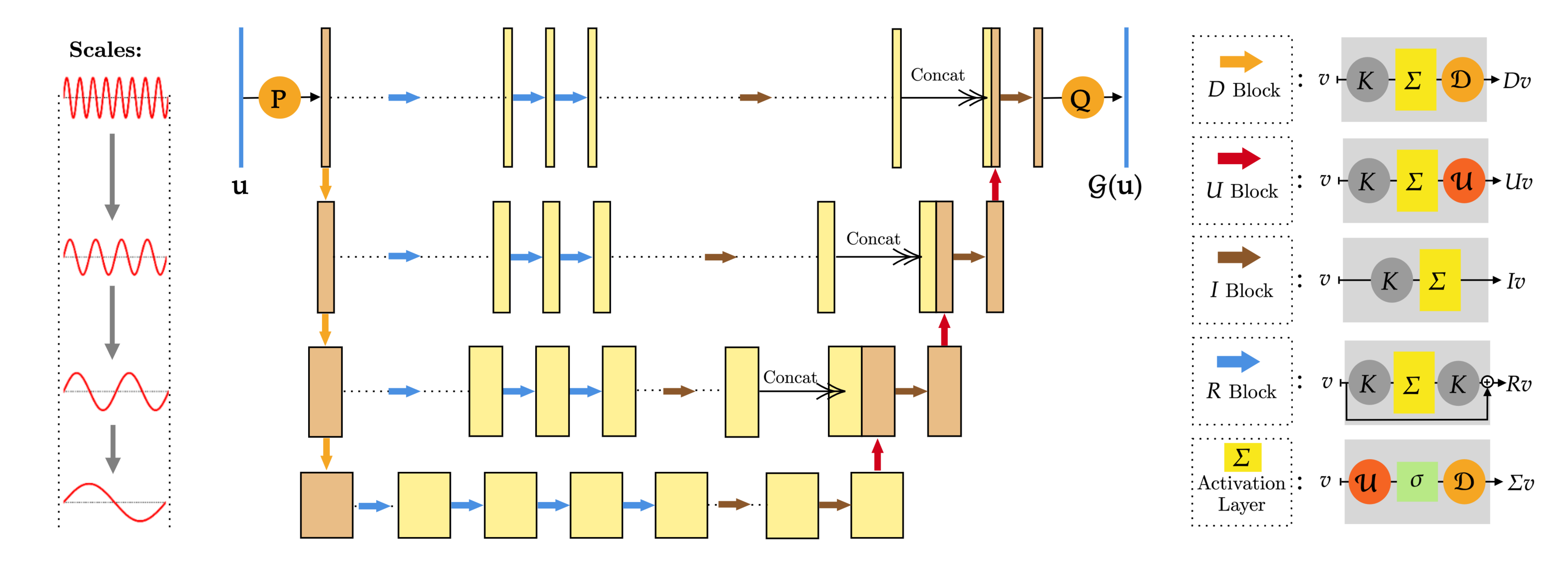}
\caption{Schematic representation of CNO \eqref{eq:CNO} as a modified U-Net with a sequence of layers mapping between bandlimited functions.}
\label{fig:cno_architecture}
\end{figure}

The above ingredients are assembled together in the form of an Operator U-Net architecture that has bandlimited functions as inputs and outputs. In addition to the blocks that have been defined above, one also needs additional ingredients, namely incorporate \emph{skip connections} through \emph{ResNet} blocks of the form,
$\mathcal{R}_{w,\overline{w}} : \mathcal{B}_{w}(D, \mathbb{R}^d) \to \mathcal{B}_{w}(D, \mathbb{R}^d)$ such that
\begin{equation}
\mathcal{R}_{w,\overline{w}}(v) = v + \mathcal{K}_{w}\circ \Sigma_{w,\overline{w}} \circ \mathcal{K}_{w} (v),\quad \forall v\in \mathcal{B}_{w}(D, \mathbb{R}^d).  
\label{eq_res}
\end{equation}
Additionally, the so-called \emph{Invariant blocks} of the form,
$\mathcal{I}_{w,\overline{w}} : \mathcal{B}_{w}(D, \mathbb{R}^d) \to \mathcal{B}_{w}(D, \mathbb{R}^d)$ is defined such that
\begin{equation}
\label{eq:inv}
\mathcal{I}_{w,\overline{w}}(v) = \Sigma_{w,\overline{w}} \circ \mathcal{K}_{w} (v),\quad \forall v\in \mathcal{B}_{w}(D, \mathbb{R}^d).
\end{equation}
Finally, all these ingredients are assembled together in a modified Operator U-Net architecture which is graphically depicted in Figure \ref{fig:cno_architecture}. Note that instead of a \emph{lead-time conditioned} layer normalization \ref{eq:LNt}, we incorporate a \emph{lead-time conditioned instance normalization} into CNO. A lead-time conditioned instance normalization is applied to an input $\mathbf{v}$ by
\begin{equation}
\label{eq:conditioned_IN}
IN_{\alpha(t),\beta(t)}(\bv)(x) =  \alpha(t) \odot IN(\bv)(x) + \beta(t)
\end{equation}
where $IN(\bv)$ is a regular instance normalization. In the case of CNO, we use (small) MLPs to parametrize $\alpha(t)$ and $\beta(t)$. This choice of conditional layer is similar to the FILM layer introduced in \cite{film}, applied on top of the instance normalization. Additionally, we observed that \textit{including time} $t$ as an additional, constant input channel of the CNO slightly enhances its performance.

The specifications of the CNO model that we used and trained from scratch in all the experiments, as well as the training details are summarized in the following list:
\begin{itemize}

   \item  \textbf{Lifting dimension:} $54$
   
   \item  \textbf{Number of up/downsampling layers:} $4$

   \item  \textbf{Number of residual blocks in the bottleneck:} $6$

   \item  \textbf{Number of residual blocks in the middle layers:} $6$
    
   \item  \textbf{Trainable parameters: }
   $39.1$M

   \item  \textbf{Optimizer: }
   AdamW \cite{loshchilov2019decoupled}
   
   \item  \textbf{Scheduler: }
   Linear with decreasing factor of $0.9$ every $10$ epochs

   \item  \textbf{Initial learning rate: }
   $5\cdot 10^{-4}$

    \item  \textbf{Weight decay: }
   $10^{-6}$

   \item  \textbf{Number of epochs: }
   $400$

   \item \textbf{Batch size:} 32

   \item \textbf{Early stopping:} If the validation loss does not improve for 40 epochs
\end{itemize}

Source code for CNO is available at \href{https://github.com/camlab-ethz/ConvolutionalNeuralOperator}{https://github.com/camlab-ethz/ConvolutionalNeuralOperator}.

\subsection{FNO}

A \textit{Fourier neural operator} ({FNO}) $\mathcal{G}$ \cite{FNO} is a composition
\begin{equation}
\label{eq:fno}
\mathcal{G}: \mathcal{X} \to \mathcal{Y}:\quad 
\mathcal{G} = Q \circ \mathcal{L}_T \circ \dots \circ \mathcal{L}_1 \circ R.
\end{equation}
It has a ``lifting operator'' ${u}(x) \mapsto R({u}(x),x)$, where $R$ is represented by a linear function $R: \R^{d_u} \to \R^{d_v}$ where $d_u$ is the number of components of the input function and $d_v$ is the ``lifting dimension''. The operator $Q$ is a non-linear projection, instantiated by a shallow neural network with a single hidden layer and leaky ReLU activation function, such that $v^{L+1}(x) \mapsto \mathcal{G}({u})(x) = Q \left( v^{L+1}(x)\right)$. 

Each \emph{hidden layer} $\mathcal{L}_\ell: v^\ell(x) \mapsto v^{\ell+1}(x)$ is of the form 
\[
v^{\ell+1}(x)  = (\sigma \circ IN)\left(
W_\ell \cdot v^{\ell}(x) + \left( K_\ell v^{\ell} \right) (x) 
\right),
\]
with $W_\ell \in \R^{d_v\times d_v}$ a trainable weight matrix (residual connection), $\sigma$ an activation function, corresponding to leaky ReLU, $IN$ standard instance normalization or time-conditioned instance normalization (see Equation \ref{eq:conditioned_IN}) and the \emph{non-local Fourier layer},
\[
K_\ell v^\ell = \cF_N^{-1} \left(P_\ell(k) \cdot \cF_N v^\ell(k) \right),
\]
where $\mathcal{F}_N v^\ell (k)$ denotes the (truncated)-Fourier coefficients of the discrete Fourier transform (DFT) of $v^\ell(x)$, computed based on the given $J$ grid values in each direction. Here, $P_\ell(k) \in \C^{d_v \times d_v}$ is a complex Fourier multiplication matrix indexed by $k\in \Z^d$, and $\cF_N^{-1}$ denotes the inverse DFT. As with CNO (Section \ref{sec:CNO}), we include time as an additional channel -- in addition to the time-conditioned instance normalization layers -- for all time-dependent problems.

We used the following hyperparameters and training details to train the FNO models:
\begin{itemize}
   \item \textbf{Lifting dimension:} 96 
   \item \textbf{Number of Fourier layers:} 5
   \item \textbf{Number of Fourier modes:} 20
   \item  \textbf{Trainable parameters:} 37.0M
   \item  \textbf{Optimizer:} AdamW \cite{loshchilov2019decoupled}
   \item  \textbf{Scheduler:} Cosine Decay
   \item  \textbf{Initial learning rate:} $5 \cdot 10^{-4}$
    \item  \textbf{Weight decay:} $10^{-6}$
   \item  \textbf{Number of epochs:} 400
   \item \textbf{Batch size:} 40
   \item \textbf{Early stopping:} If the validation loss does not improve for 40 epochs
\end{itemize}

\subsection{CNO-FM}
\label{sec:cnofm}
In addition to the \textsc{Poseidon} models, we also pretrain a CNO foundation model baseline. We use the same pretraining datasets as the \textsc{Poseidon} models (see \ref{sec:datapt}), i.e. NS-Sines, NS-Gauss, CE-RP, CE-KH, CE-CRP and CE-Gauss datasets. The inputs and the outputs of the model have $4$ channels, i.e. $\rho$, $u$, $v$ and $p$. For the NS-Sines and NS-Gauss datasets, we mask out the pressure predictions during training, while predicting a constant value $\rho = 1$ for density.

The specifications of the CNO-FM model that we pretrained, as well as the training details are summarized in the following list:
\begin{itemize}

   \item  \textbf{Lifting dimension:} $82$
   
   \item  \textbf{Number of up/downsampling layers:} $4$

   \item  \textbf{Number of residual blocks in the bottleneck:} $8$

   \item  \textbf{Number of residual blocks in the middle layers:} $8$
    
   \item  \textbf{Trainable parameters: }
   $109$M

   \item  \textbf{Optimizer: }
   AdamW
 
   \item  \textbf{Scheduler: }
   Linear with decreasing factor of $0.9$ every epoch

   \item  \textbf{Initial learning rate: }
   $5\cdot 10^{-4}$

    \item  \textbf{Weight decay: }
   $10^{-6}$

   \item  \textbf{Effective batch size: }
   $256$, resulting in a per-device batch size of $32$
   
   \item  \textbf{Number of epochs: }
   $40$

   \item \textbf{Early stopping:} No
\end{itemize}

To finetune the CNO-FM, we differentiate between two scenarios: one where the input and output share the same context as the pretrained models (comprising the variables $\rho$, $u$, $v$, and $p$, either masked or unmasked), and another where the downstream task is out-of-context (i.e. when the input and target variables differ from those used during pretraining). 

To explain the finetuning technique, let us denote the pretrained CNO model by $\mathcal{G}_{FM}$ and decompose it to
\begin{equation*}
\mathcal{G}_{FM} = Q \circ \mathcal{G}_{FM, b} \circ P,
\end{equation*}
where $P$ is the lifting layer, $Q$ is the projection layer and $\mathcal{G}_{FM, b}$ is the base part of the CNO-FM.

When the context of variables is retained in the downstream task, we introduce an additional linear layer $\mathcal{L}$ that is applied prior to the lifting layer $P$. All other parameters from $\mathcal{G}_{FM}$ are transferred over to the downstream task model. Hence, the model that is finetuned is
\begin{equation}
\label{eq: cno_ft_ic}
    \mathcal{G}_{FT} =  Q \circ \mathcal{G}_{FM, b} \circ P \circ \mathcal{L}.
\end{equation}

\begin{figure}[htbp]
  \centering
  \includegraphics[width=0.6\linewidth]{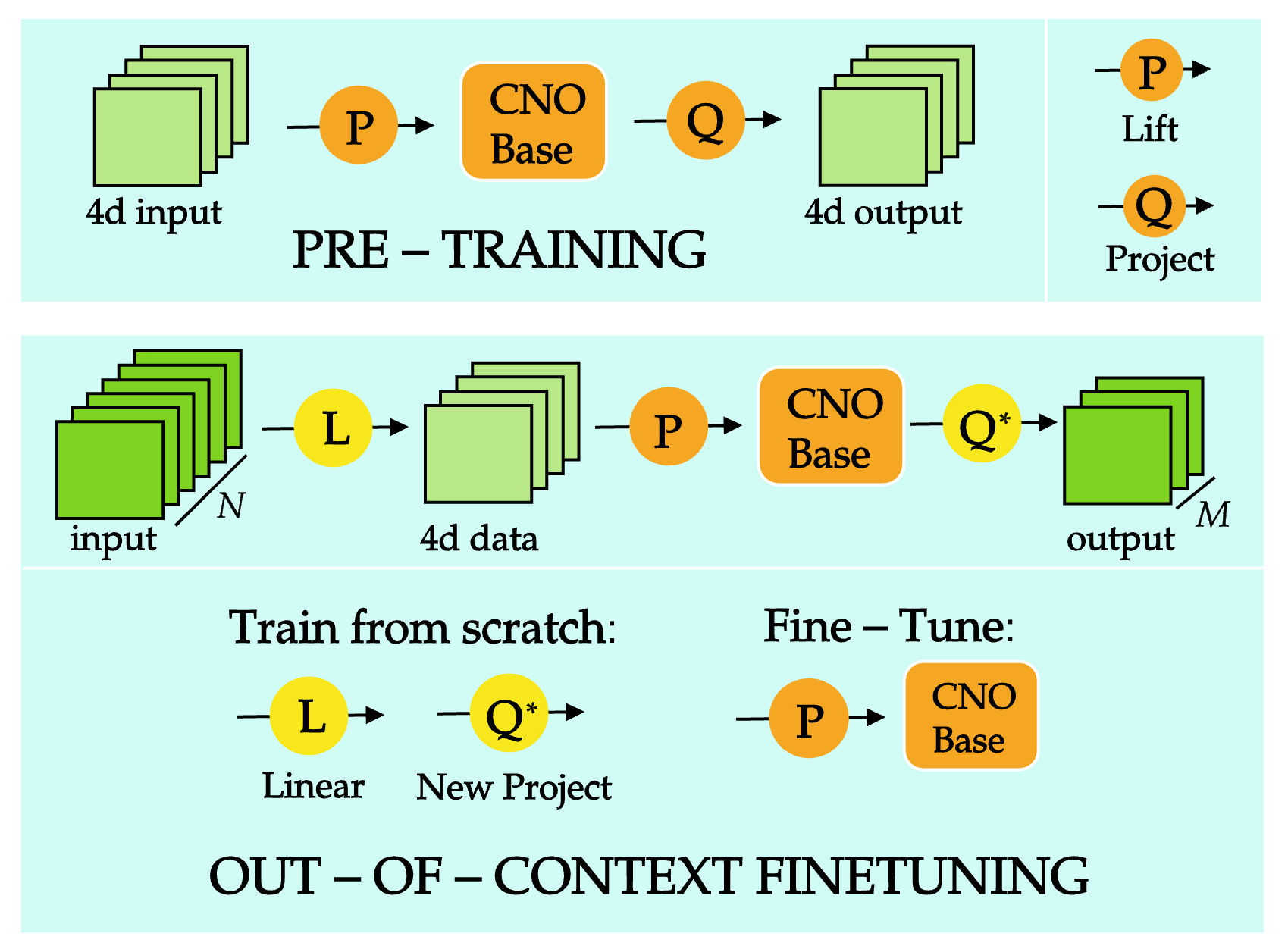}
\caption{Schematic representation of the finetuning procedure of CNO-FM.}
\label{fig:cno_fm}
\end{figure}

A schematic representation of the CNO-FM finetuning procedure is shown in Figure \ref{fig:cno_fm}. When the downstream task is out-of-context, in addition to the linear layer $\mathcal{L}$ that is applied before $P$, the projection layer is replaced by a new, randomly initialized projection layer $Q^\star$. Other parameters are transferred over to the downstream task model. The model that is finetuned is
\begin{equation}
\label{eq: cno_ft}
    \mathcal{G}_{FT} =  Q^\star \circ \mathcal{G}_{FM, b} \circ P \circ \mathcal{L}.
\end{equation}

We set the number of epochs for the downstream tasks to $200$. Since the loss converges significantly faster than when training from scratch, even $50 - 100$ epochs were sufficient to effectively finetune the CNO-FM. Parameters of $\mathcal{G}_{FT}$ are divided into three distinct groups
\begin{itemize}
    \item \textbf{Group 1}: Projection $Q$ (or $Q^\star$), Lifting $P$ and Linear layer $\mathcal{L}$
    \item \textbf{Group 2}: All the conditional instance normalization layers $IN_{\alpha(t),\beta(t)}$
    \item \textbf{Group 3}: Other parameters in $\mathcal{G}_{FM, b}$
\end{itemize}

We experimented with learning rates for each group of parameters, as well as schedulers. An efficient way to finetune CNO-FM for in-context downstream tasks was to set the initial learning rates of the parameter groups to $lr_{1} = 2.5\cdot 10^{-4}$, $lr_{2} = 5\cdot 10^{-4}$ and $lr_{3} = 10^{-4}$. For out-of-context tasks, the learning rates that we used are
$lr_{1} = 7.5\cdot 10^{-4}$, $lr_{2} = 5\cdot 10^{-4}$ and $lr_{3} = 10^{-4}$. In both cases, the learning rate scheduler is linear with with decreasing factor of $0.9$ every $5$ epochs.

The CNO codes are available at \href{https://github.com/camlab-ethz/ConvolutionalNeuralOperator}{https://github.com/camlab-ethz/ConvolutionalNeuralOperator}.

\subsection{MPP}
\label{sec:mpp}
\textit{Multiple physics pretraining} (MPP) is a pretraining approach for autoregressive
physical surrogate modeling \cite{MPP}. MPP uses a \textit{scalable axial attention} transformer backbone to reduce the quadratic complexity of the usual attention mechanism. Multiple input fields of  MPP are projected onto a single, shared embedding space. MPP also uses spatial and temporal attention blocks to capture spatial and temporal dependencies in the data. To train or finetune MPP models, one uses the normalized MSE loss. We will finetune the \textbf{MPP-AVIT-B} foundation model for all our downstream tasks. The \textbf{MPP-AVIT-B} model has $116$M trainable parameters. 

MPP models are autoregressive models with fixed context size of $T^S$. They predict the solution at a time step $N$ of a PDE of interest given the previous $T^S$ time steps. Thus, they rely on the \textit{history of the solution}, encompassing multiple time steps, to forecast future time steps accurately. This differs from the the task that we are interested in (i.e. \textbf{OLT} defined in the Main Text), which aims to generate the entire solution trajectory given only the initial datum and boundary conditions.

\begin{figure}[htbp]
  \centering
  \includegraphics[width=1.0\linewidth]{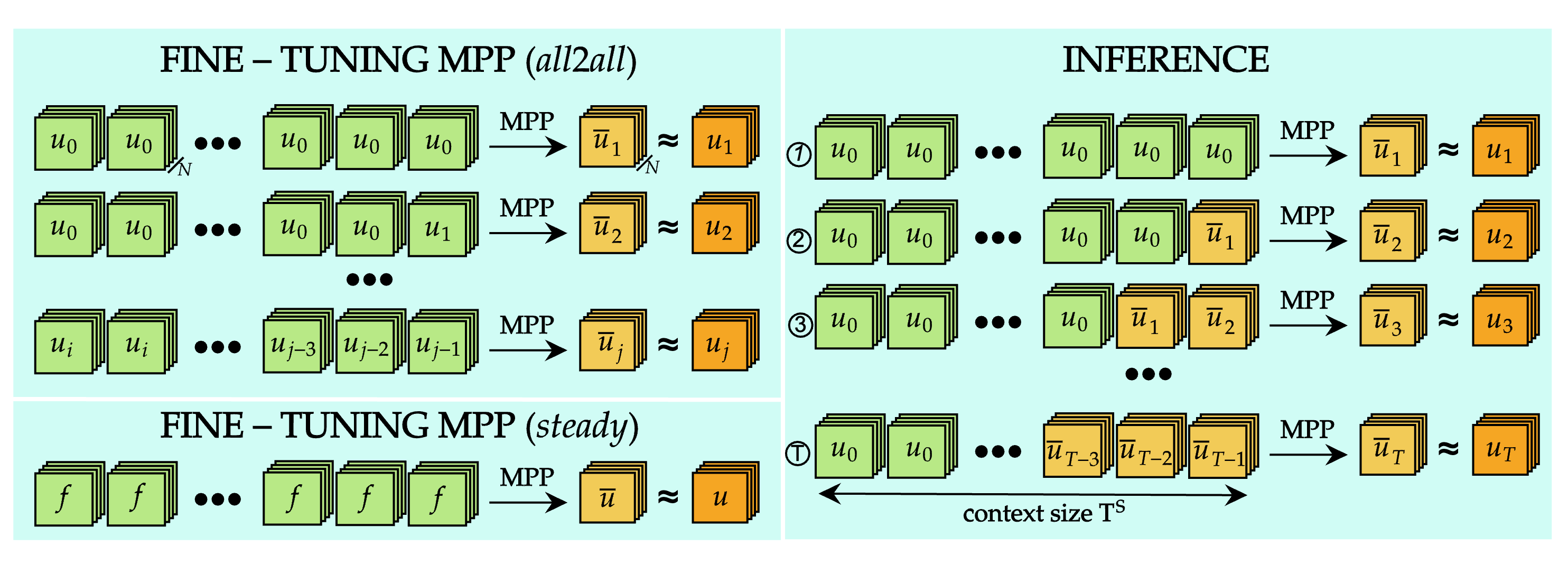}
\caption{Schematic representation of MPP \textit{all2all} and \textit{steady} finetuning and inference strategies.}
\label{fig:mpp}
\end{figure}

Therefore, we need to adjust the MPP finetuning strategy. We adapt the \textit{all2all} strategy. Let $U = (u_0, u_1, \dots, u_T)$ be a solution trajectory of length $T + 1$. Let $(i,j)$ be two integers such that $j > i$. We rely on the fact that MPP predicts one snapshot at a time and finetune MPP to predict $u_{j}$ based on the history $u_{j-1}, u_{j-2}, \dots u_i$. Since there are not always $T^S$ past time steps in the training samples, we fill the remaining time steps with copies of $u_i$ (see Figure \ref{fig:mpp}). We generate $T(T+1)/2$ samples out of the trajectory $U$. For steady-state operators of the form $(f_1, f_2, \dots, f_L) \to u$, the $L$ channels are copied $T^S$ times, and MPP is finetuned using these samples (see Figure \ref{fig:mpp}). The inference strategy for the time-dependent problems is straightforward. Given the initial snapshot $u_0$, one autoregressively applies the finetuned model $T$ times to predict $u_T$ (see Figure \ref{fig:mpp}). During the finetuning of MPP, we do not predict dummy variables like the speed function in the Wave equation or the forcing term in Kolmogorov flow, as the model had difficulties in predicting them, so the errors accumulated fast. This contrasts with other models that predict the dummy variables alongside the solution. Final testing errors for all the models are not calculated for these dummy variables.

For each downstream task, we finetuned \textbf{MPP-AVIT-B} model for $100$ epochs. We did not use more than $100$ epochs as the training usually converged after $10$ to $50$ epochs. We used the Adam \cite{kingma2014adam} optimizer with a cosine annealing scheduler and linear warmup. 

\clearpage
\section{Results}
\label{sec:res}
\subsection{Performance on Downstream Tasks}
\label{sec:perf_downstream_scalings}

We evaluate all models on the \emph{median relative $L^1$ error} at a certain snapshot in time on each solution function of interest. For vectorized functions such as the velocity in fluid flow, we evaluate it over the entire vector. Since all our downstream tasks have different solution functions of interest, we provide an overview over the actual functions of interest in Table \ref{tab:performance_metrics}; should there be a list, we compute the mean over all metrics. It also depicts, how the rollout in time was done for each task i.e., either the solution at final time is computed directly with the final time as lead time or autoregressively (AR) as presented in Main Text \eqref{eq:ar}. Within AR, \textsc{Poseidon} models were always evaluated with uniform (homogeneous) rollout whereas in the case of CNO and CNO-FM, autoregressive (AR) rollout is heterogeneous; for MPP, it is always uniformly autoregressive.

\begin{figure}
    \centering
    \includegraphics[width=\textwidth]{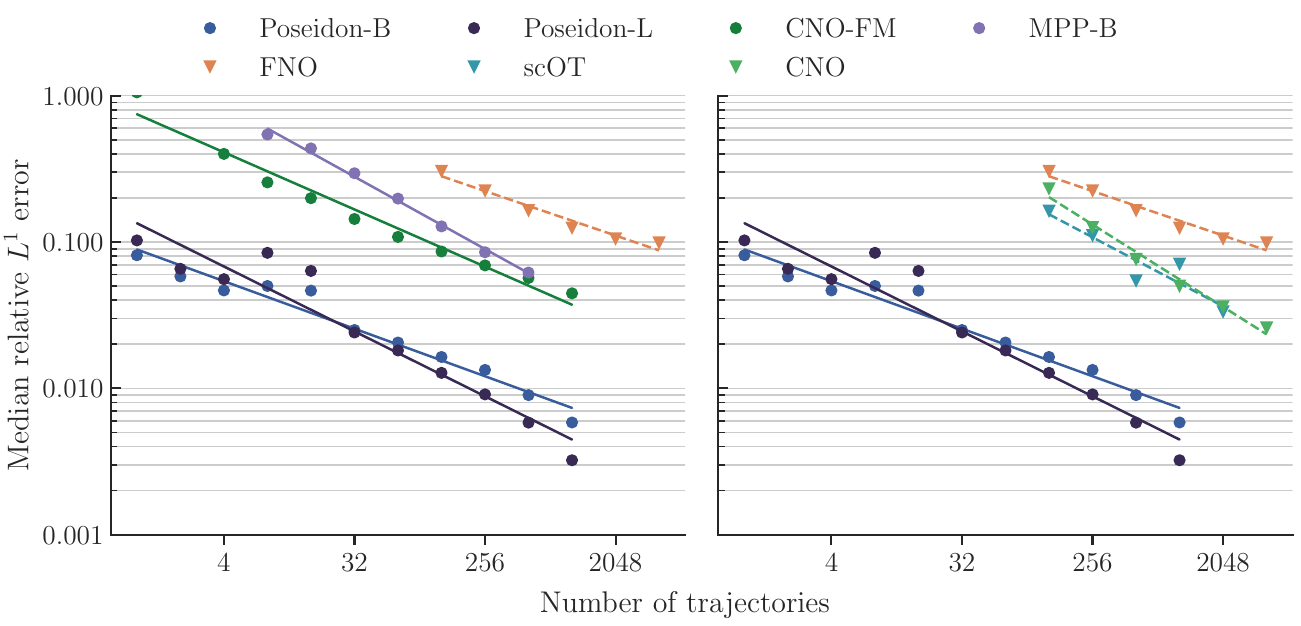}
    \caption{NS-PwC. Number of trajectories vs. median relative $L^1$ error on the test set.}
    \label{fig:scaling_ns_pwc}
\end{figure}

\begin{figure}
    \centering
    \includegraphics[width=\textwidth]{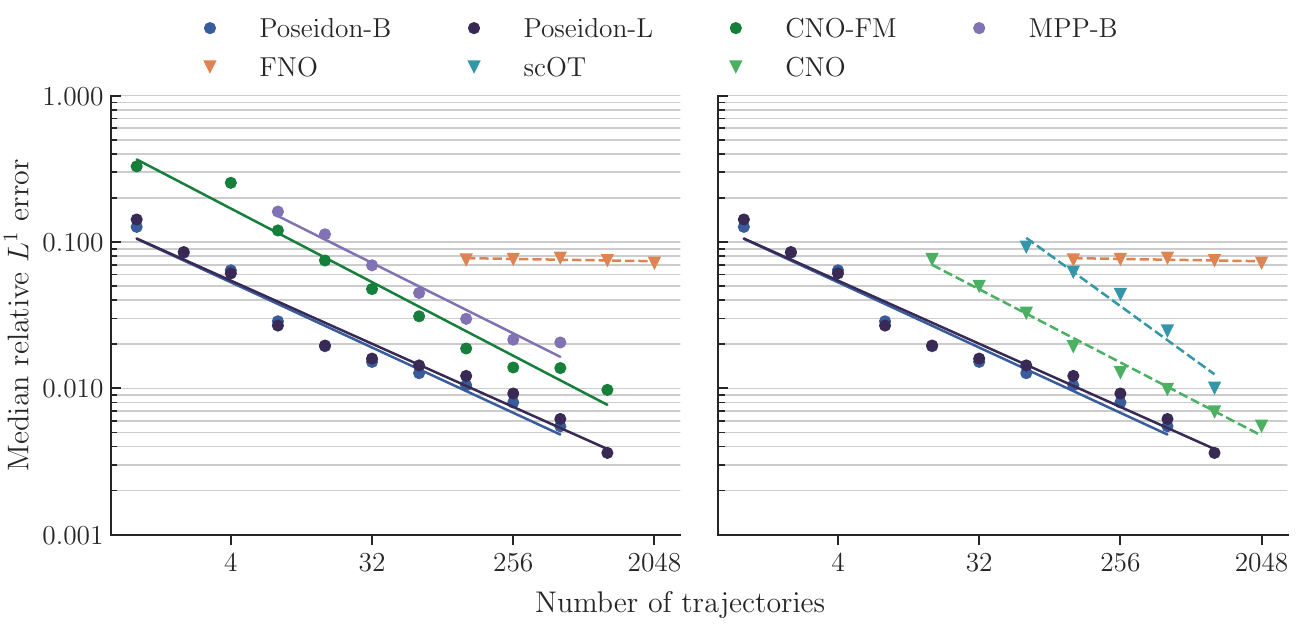}
    \caption{NS-SVS. Number of trajectories vs. median relative $L^1$ error on the test set.}
    \label{fig:scaling_ns_svs}
\end{figure}

\begin{figure}
    \centering
    \includegraphics[width=\textwidth]{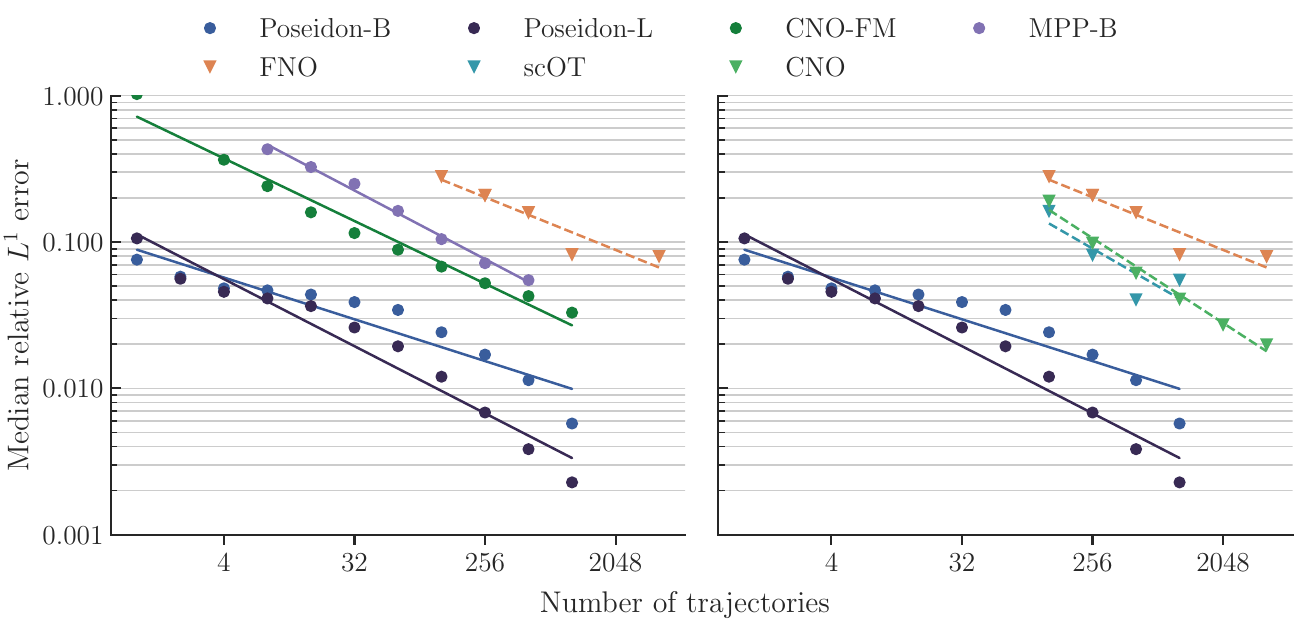}
    \caption{NS-BB. Number of trajectories vs. median relative $L^1$ error on the test set.}
    \label{fig:scaling_ns_bb}
\end{figure}

\begin{figure}
    \centering
    \includegraphics[width=\textwidth]{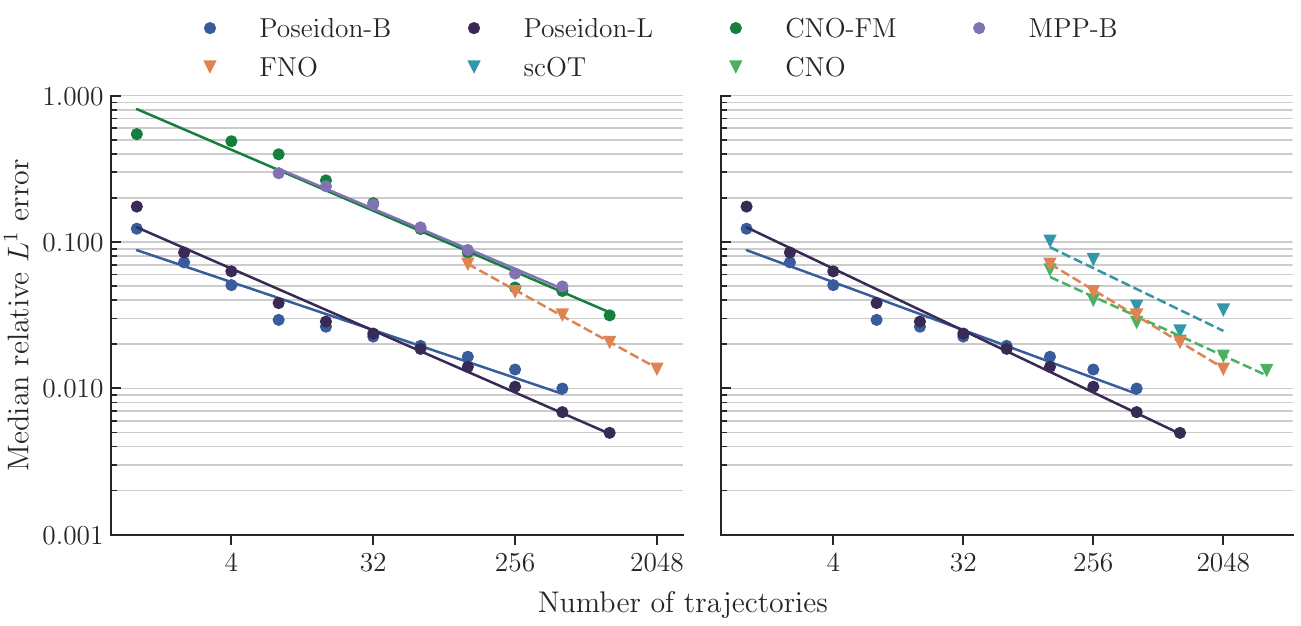}
    \caption{NS-SL. Number of trajectories vs. median relative $L^1$ error on the test set.}
    \label{fig:scaling_ns_sl}
\end{figure}

\begin{figure}
    \centering
    \includegraphics[width=\textwidth]{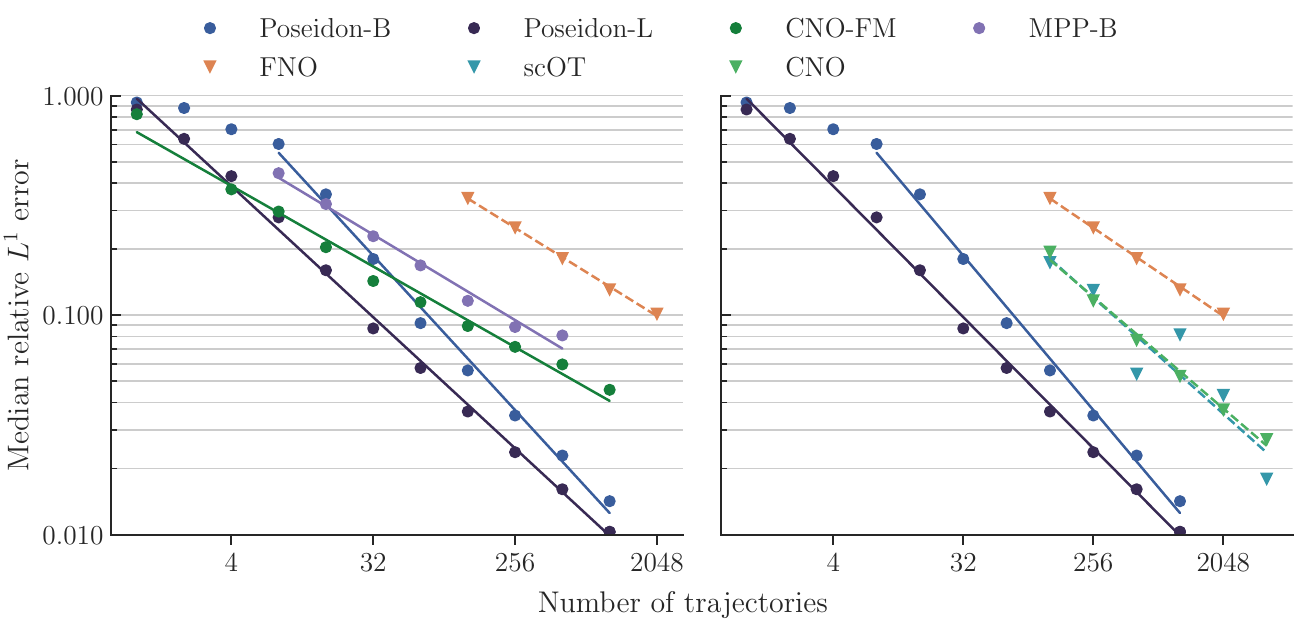}
    \caption{NS-Tracer-PwC. Number of trajectories vs. median relative $L^1$ error on the test set.}
    \label{fig:scaling_ns_pwc_tracer}
\end{figure}

\begin{figure}
    \centering
    \includegraphics[width=\textwidth]{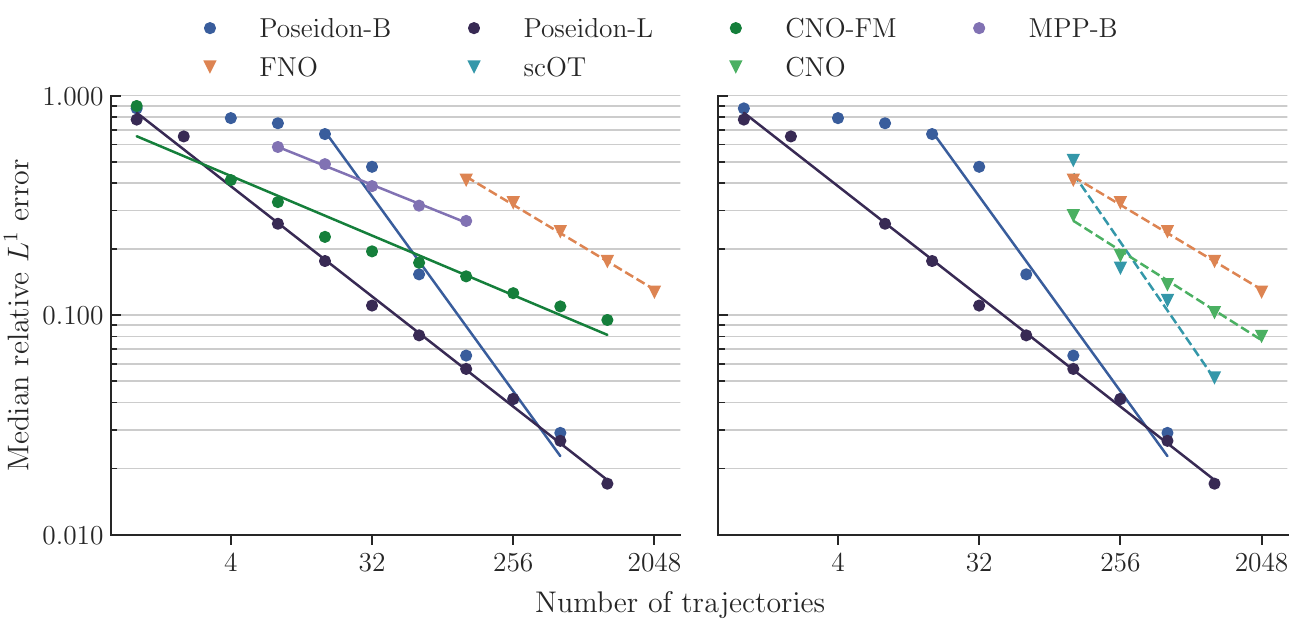}
    \caption{FNS-KF. Number of trajectories vs. median relative $L^1$ error on the test set.}
    \label{fig:scaling_fns_kf}
\end{figure}

\begin{figure}
    \centering
    \includegraphics[width=\textwidth]{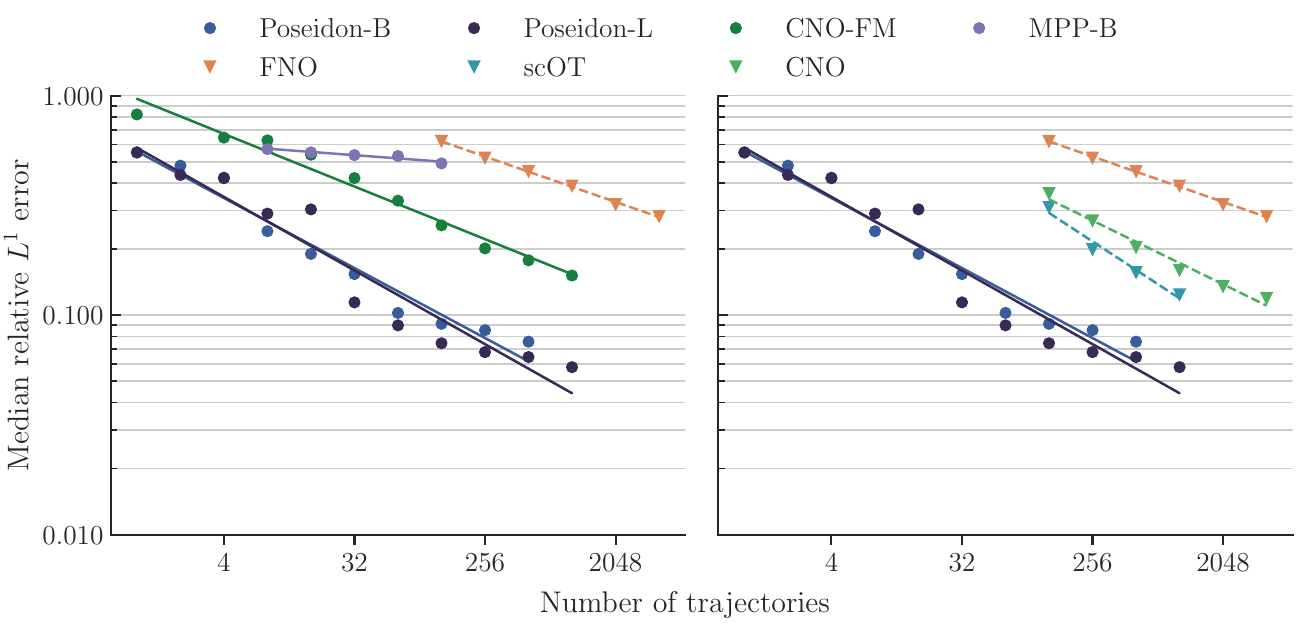}
    \caption{CE-RPUI. Number of trajectories vs. median relative $L^1$ error on the test set.}
    \label{fig:scaling_ce_rpui}
\end{figure}

\begin{figure}
    \centering
    \includegraphics[width=\textwidth]{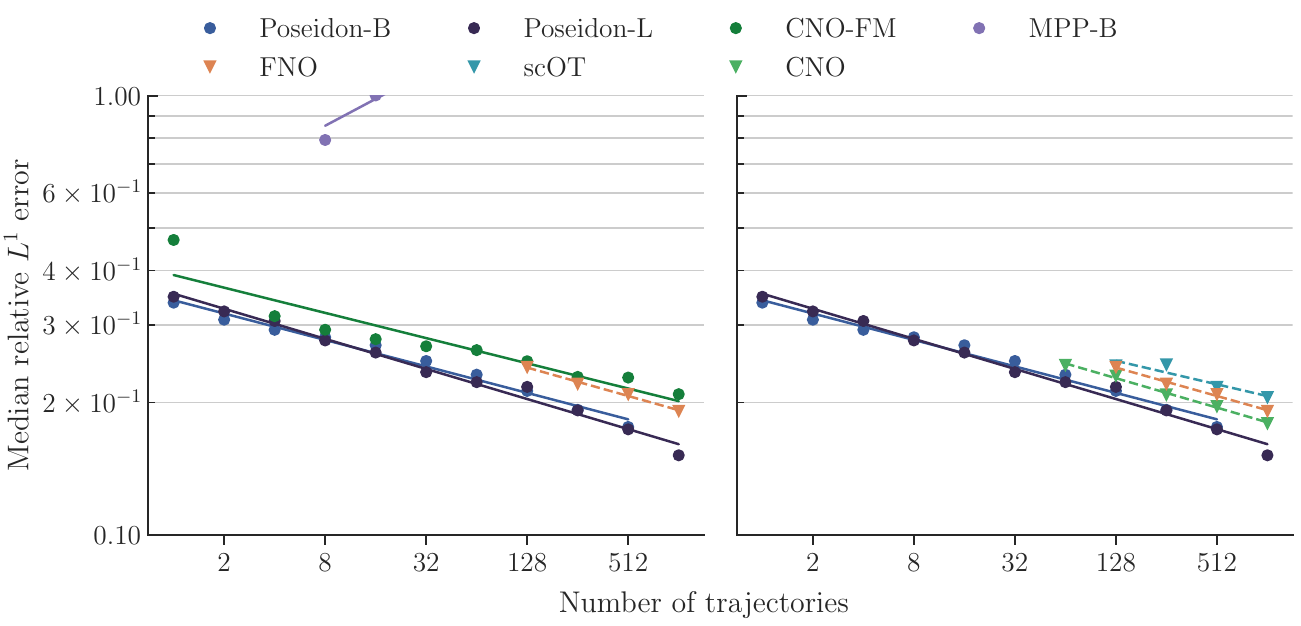}
    \caption{CE-RM. Number of trajectories vs. median relative $L^1$ error on the test set.}
    \label{fig:scaling_ce_rm}
\end{figure}

\begin{figure}
    \centering
    \includegraphics[width=\textwidth]{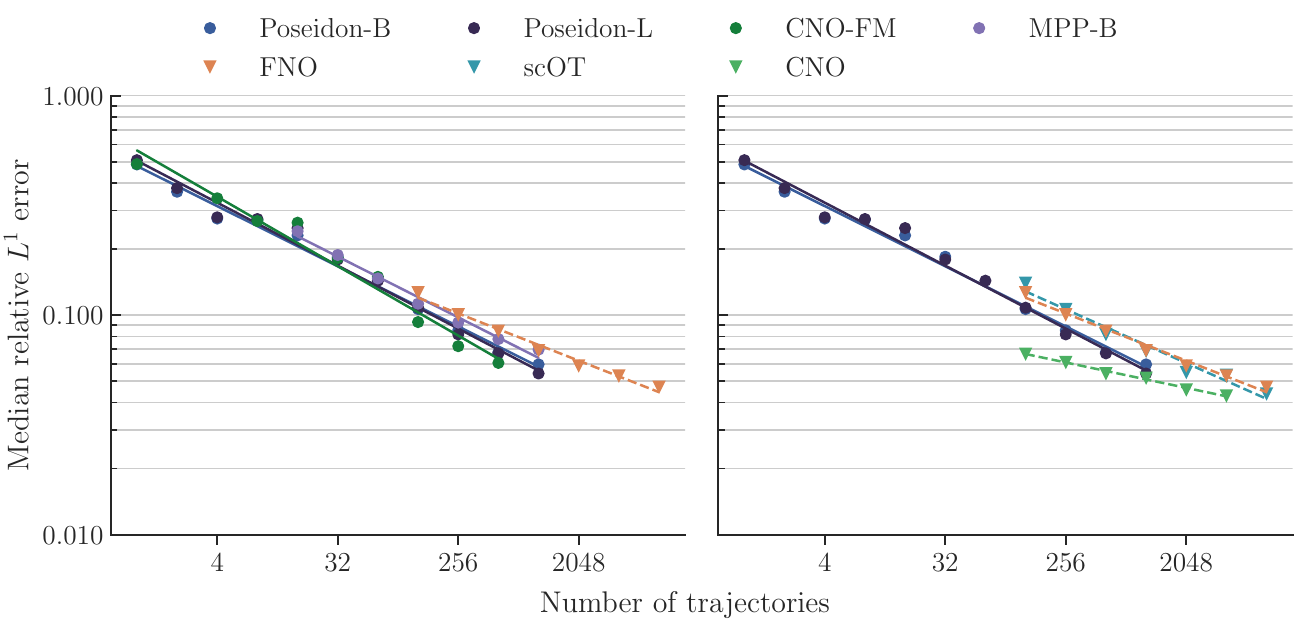}
    \caption{SE-AF. Number of samples vs. median relative $L^1$ error on the test set.}
    \label{fig:scaling_se_af}
\end{figure}

\begin{figure}
    \centering
    \includegraphics[width=\textwidth]{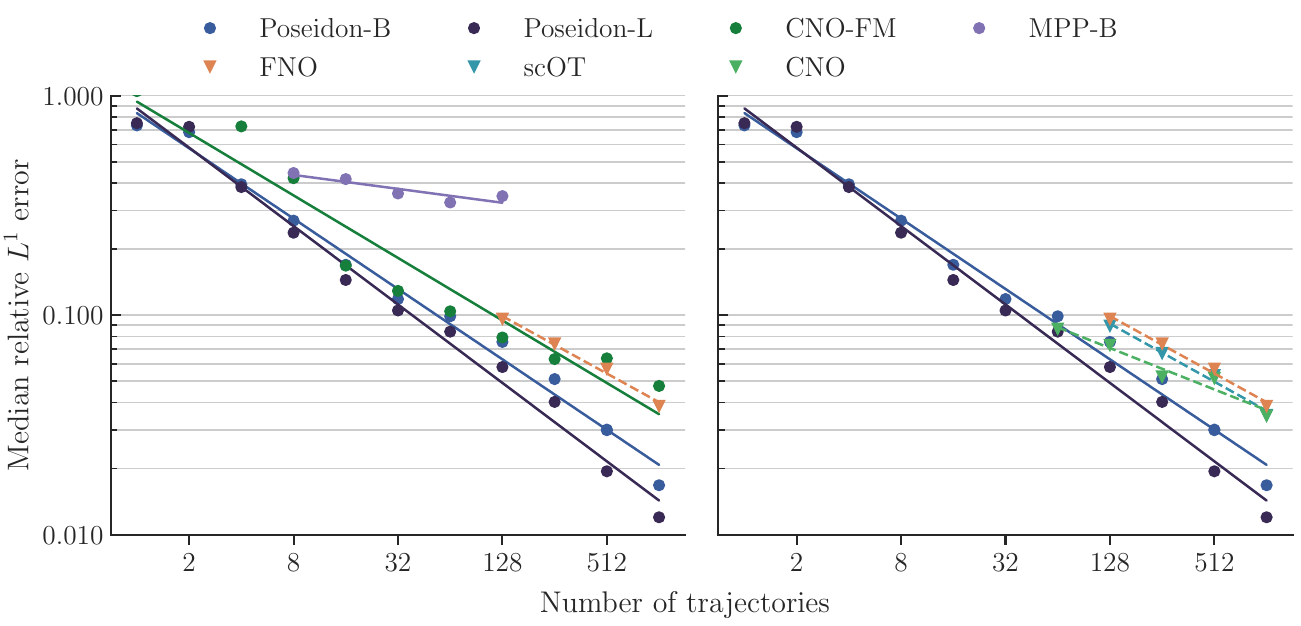}
    \caption{GCE-RT. Number of trajectories vs. median relative $L^1$ error on the test set.}
    \label{fig:scaling_gce_rt}
\end{figure}

\begin{figure}
    \centering
    \includegraphics[width=\textwidth]{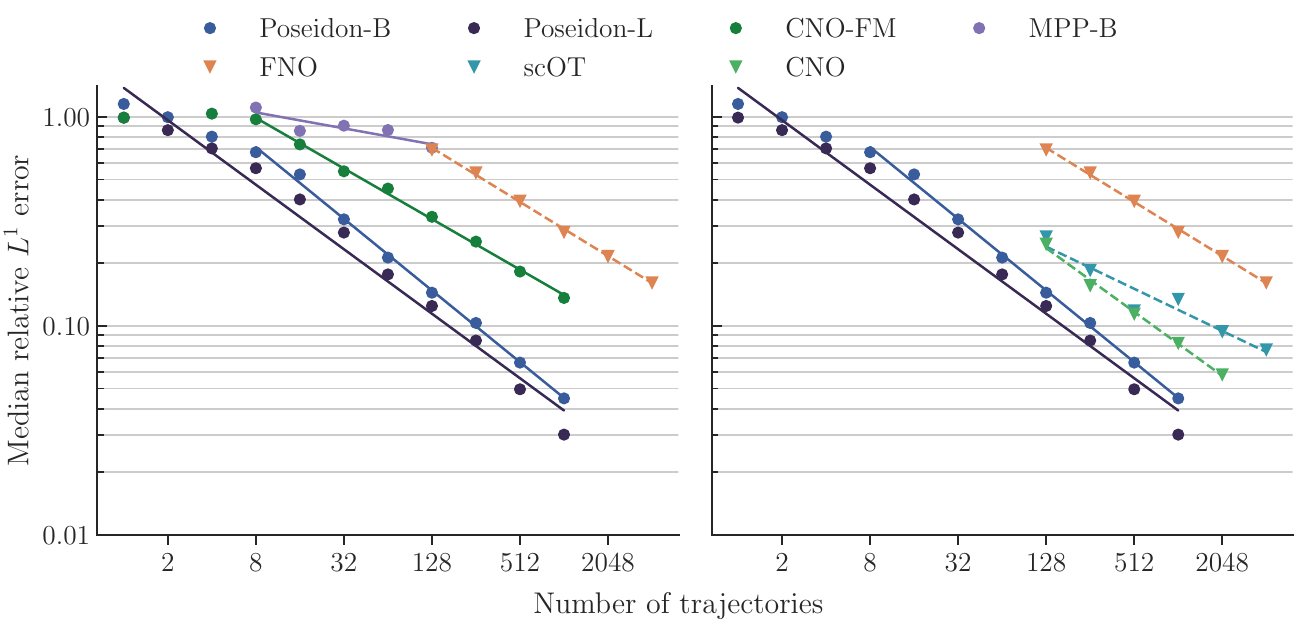}
    \caption{Wave-Layer. Number of trajectories vs. median relative $L^1$ error on the test set.}
    \label{fig:scaling_wave_layer}
\end{figure}

\begin{figure}
    \centering
    \includegraphics[width=\textwidth]{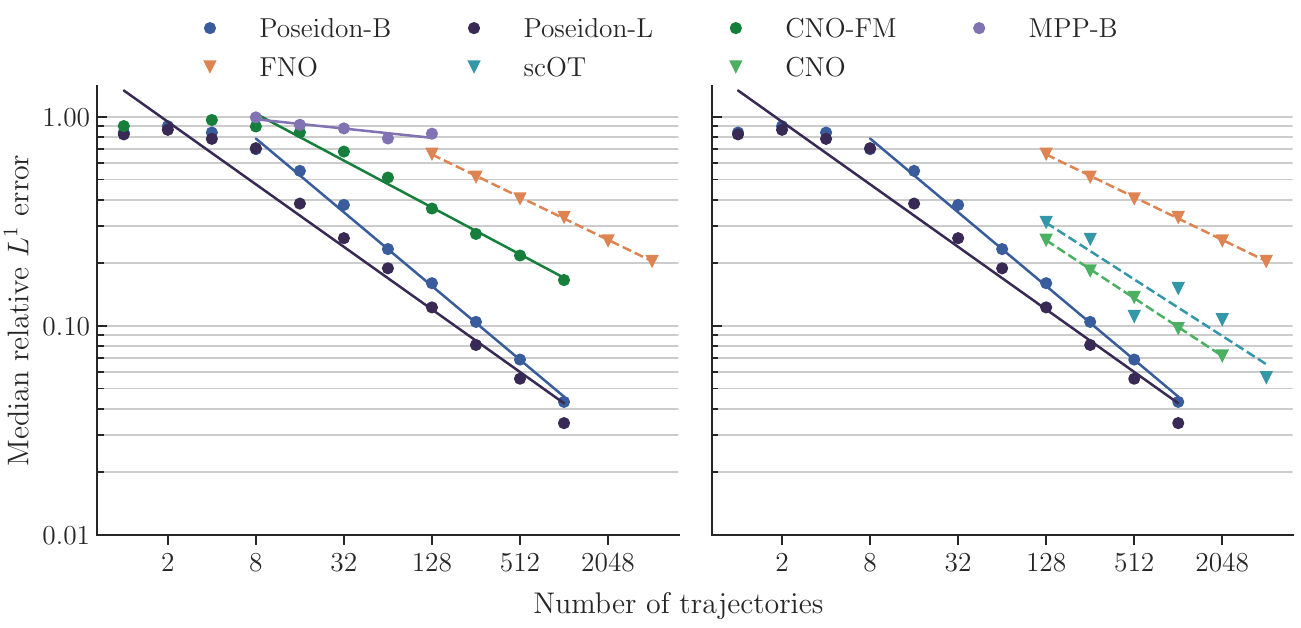}
    \caption{Wave-Gauss. Number of trajectories vs. median relative $L^1$ error on the test set.}
    \label{fig:scaling_wave_gaussians}
\end{figure}

\begin{figure}
    \centering
    \includegraphics[width=\textwidth]{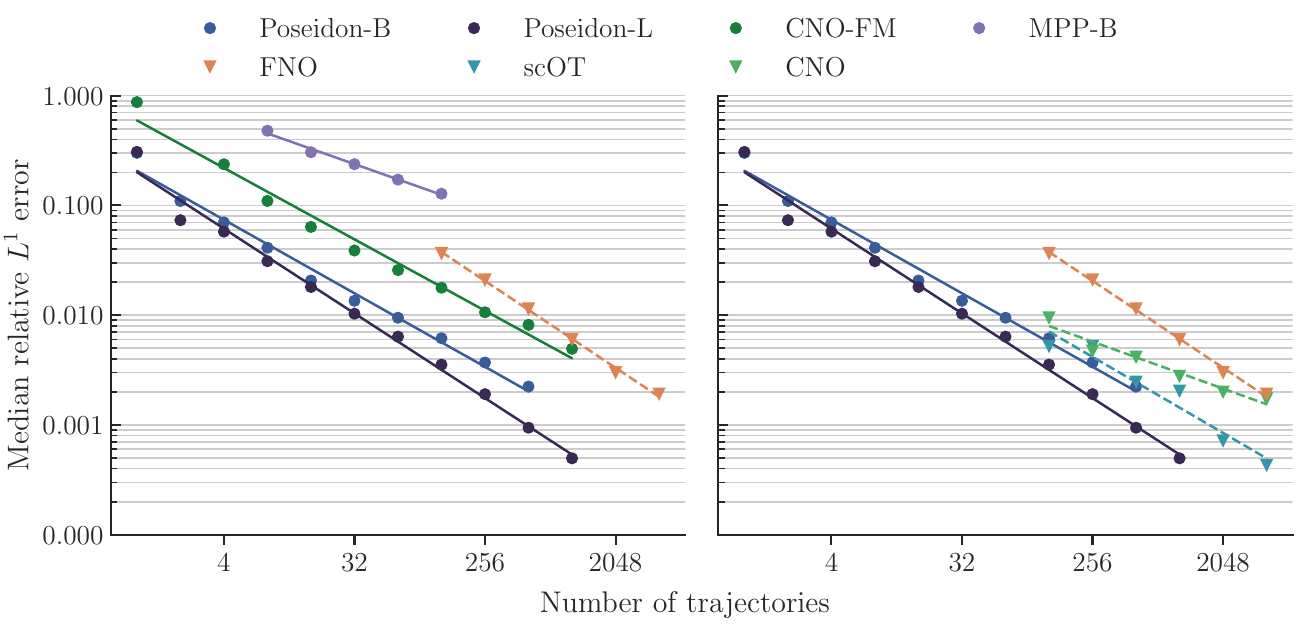}
    \caption{ACE. Number of trajectories vs. median relative $L^1$ error on the test set.}
    \label{fig:scaling_ace}
\end{figure}

\begin{figure}
    \centering
    \includegraphics[width=\textwidth]{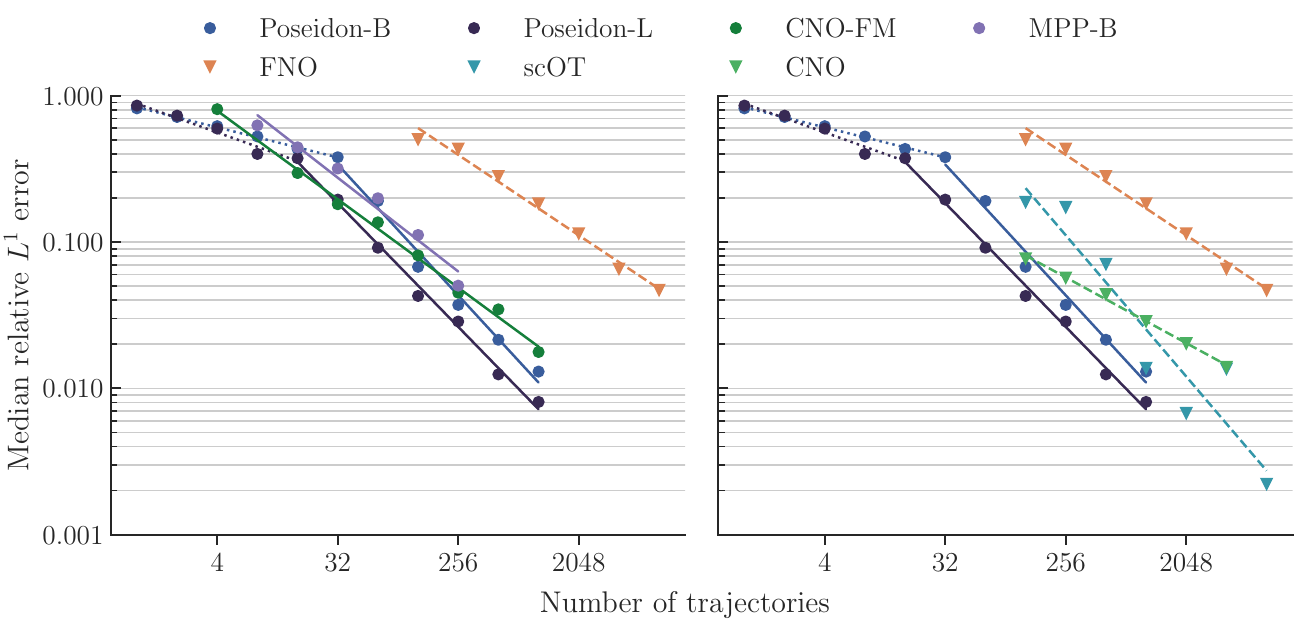}
    \caption{Poisson-Gauss. Number of samples vs. median relative $L^1$ error on the test set.}
    \label{fig:scaling_poisson_gauss}
\end{figure}

\begin{figure}
    \centering
    \includegraphics[width=\textwidth]{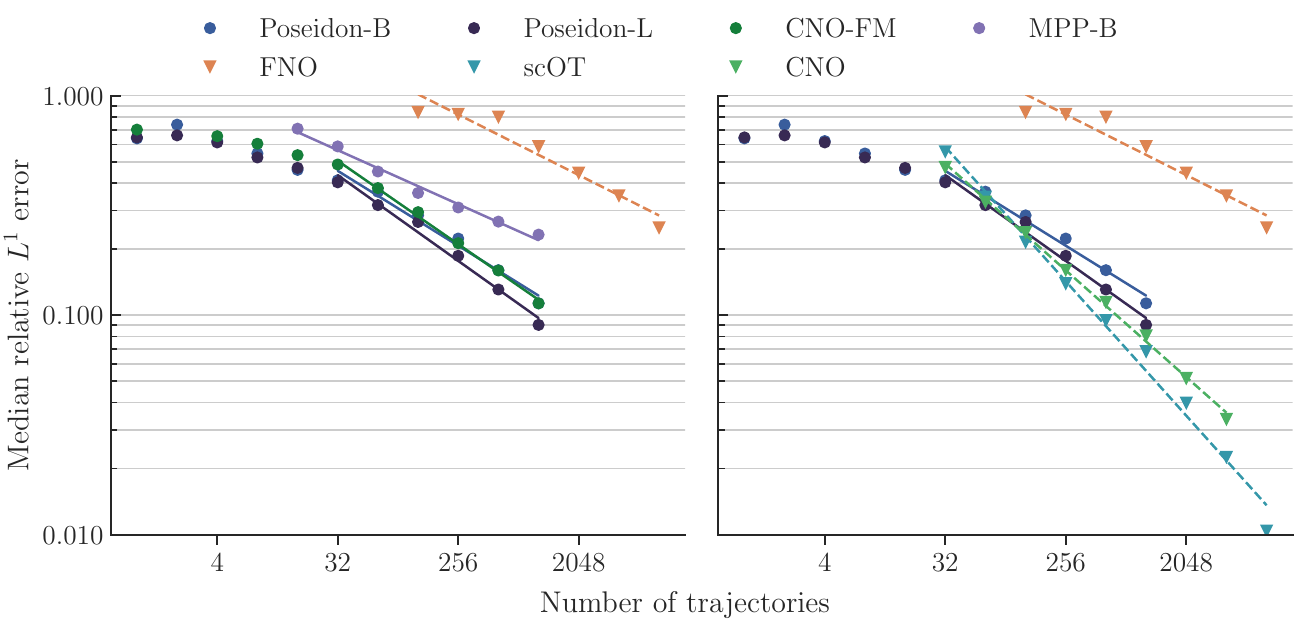}
    \caption{Helmholtz. Number of samples vs. median relative $L^1$ error on the test set.}
    \label{fig:scaling_helmholtz}
\end{figure}

\textbf{Evaluation of Downstream Tasks with Scaling Plots.} In Figures \ref{fig:scaling_ns_pwc} to \ref{fig:scaling_helmholtz}, we present the test errors (y-axis) for all the models (\textsc{Poseidon}-L, \textsc{Poseidon}-B, CNO-FM, MPP, FNO on the left sub-figure and  \textsc{Poseidon}-L, \textsc{Poseidon}-B, CNO, scOT and FNO on the right sub-figure of each figure) vs. the total number of trajectories (time-dependent PDEs) or total number of samples (for time-dependent PDEs) on the x-axis. As mentioned in the Main Text, the \textsc{Poseidon} models clearly outperform all the baselines on most of the tasks as the corresponding test errors are significantly lower than the baselines for the same number of samples. Note that the metrics {\bf EG} and {\bf AG} \eqref{eq:met} were computed based on these plots. We do not include the \textsc{Poseidon}-T results as they would further clutter the scaling plots. However, the {\bf EG} and {\bf AG} metrics for \textsc{Poseidon}-T are presented in Table \ref{tab:full_res}.  

\textbf{On Scaling Laws.} If we denote the number of trajectories (samples) by $M$, we can fit power laws of the form,
\begin{equation}
\label{eq:plaw}
\er_{\rm model}(M) \approx C_{\rm model} M^{-\alpha_{\rm model}},
\end{equation}
to the scaling plots in Figure \ref{fig:scaling_ns_pwc} to \ref{fig:scaling_helmholtz}. Here, $C_{\rm model}$ denotes the model-specific scaling factor and the scaling exponent is $\alpha_{\rm model}$. The scaling exponents, resulting from these fits are presented in Table \ref{tab:scaling_exp}. We observe from this table, that all the models that we consider obey \emph{scaling laws} of the form \eqref{eq:plaw}, with different scaling exponents for different problems (MPP-B does not converge in some cases). These include the \textsc{Poseidon} foundation models which show consistent scaling laws. For instance, \textsc{Poseidon}-L has a scaling exponent of approximately $0.5$ or higher in all the cases except for CE-RM (where all models converge very slowly). Nevertheless, we would like to emphasize that the scaling exponent alone does not govern the final error, except in the asymptotic infinite data limit. Rather, the scaling factor $C_{\rm model}$ in \eqref{eq:plaw} plays a decisive role in determining errors in the pre-asymptotic limited data regime that all downstream tasks correspond to.

Moreover, a closer analysis of the scaling plots reveals a more nuanced picture for the \textsc{Poseidon} models. In some of the downstream tasks, for instance the Poisson-Gauss benchmark, we see from the scaling plot Figure \ref{fig:scaling_poisson_gauss} that both \textsc{Poseidon}-L and \textsc{Poseidon}-B display a \emph{biphasic} behavior, with a scaling law of the form, 
\begin{equation}
\label{eq:biplaw}
\er_{\rm model}(M) \approx \begin{cases} 
                           C^w_{\rm model} M^{-\alpha^w_{\rm model}}, &{\rm if} \quad M \leq M^{pt}_{\rm model} , \\
                            C^\ell_{\rm model} M^{-\alpha^\ell_{\rm model}}, &{\rm if} \quad M \geq M^{pt}_{\rm model},
                            \end{cases}
\end{equation}
with $\alpha^w < \alpha^\ell$. Thus, the scaling behavior is characterized by two phases, with different exponents. For instance, for the Poisson-Gauss benchmark (Figure \ref{fig:scaling_poisson_gauss}), we find that $M^{pt}=32$ for both models. Moreover, for \textsc{Poseidon}-B, $\alpha_w = 0.23$ and $\alpha_\ell=0.99$ and for \textsc{Poseidon}-L, $\alpha_w = 0.33$ and $\alpha_\ell=0.94$.  We speculate that these \emph{phase transitions} separate two phases, a \emph{warmup} phase where \textsc{Poseidon} is slowly learning about an operator that is very different from those encountered in the pretraining dataset (as is the case with Poisson-Gauss) and a \emph{learning} phase, where fast learning takes place and the model is able to quickly learn the specifics of the downstream task.  

\paragraph{Summarizing Downstream Task Performance.} In Table \ref{tab:stat}, we provide a statistical summary of the performance of all models on all downstream tasks by presenting the (median) {\bf EG} and the (mean) {\bf AG} over all tasks. These statistics provide an (average) account of model performance over all downstream tasks and clearly quantify how the \textsc{Poseidon} family of foundation models significantly outperforms all the baselines.

\begin{table}[htb]
    \centering
    \caption{The evaluation metrics are computed for each downstream task on different functions of interest, and rollout is done differently.}
    \begin{tabular}{r c c}
    \toprule
    Downstream Task & Functions of Interest & Rollout\\
    \midrule\midrule
        NS-PwC & $(u_x,u_y)$ & AR\\\midrule
        NS-SVS & $(u_x,u_y)$ & AR\\\midrule
        NS-BB & $(u_x,u_y)$ & AR\\\midrule
        NS-SL & $(u_x,u_y)$ & AR\\\midrule
        NS-Tracer-PwC & $(u_x,u_y)$, $c$ & AR\\\midrule
        FNS-KF & $(u_x,u_y)$ & direct\\\midrule
        CE-RPUI & $\rho$, $(v_x,y)$, $p$ & AR\\\midrule
        CE-RM & $\rho$, $(v_x,v_y)$, $p$ & direct\\\midrule
        SE-AF & $\rho$ & direct\\\midrule
        GCE-RT & $\rho$, $(v_x,v_y)$, $p$, $\phi$ & direct\\\midrule
        Wave-Layer & $u$ & direct \\\midrule
        Wave-Gauss & $u$ & direct \\\midrule
        ACE & $u$ & direct \\\midrule
        Poisson-Gauss & $u$ & direct \\\midrule
        Helmholtz & $u$ & direct\\
        \bottomrule
    \end{tabular}
    \label{tab:performance_metrics}
\end{table}

\begin{table}[]
    \centering
    \caption{Scaling exponents with a power law fit \eqref{eq:plaw}}
    \begin{tabular}{r c c c c c c c}
         \toprule
Dataset & \textsc{Poseidon}-B & \textsc{Poseidon}-L & scOT & CNO-FM & CNO & MPP-B & FNO\\\midrule\midrule
NS-PwC & 0.36 & 0.49 & 0.52 & 0.43 & 0.62 & 0.55 & 0.34\\\midrule
NS-SVS & 0.49 & 0.48 & 0.77 & 0.56 & 0.55 & 0.53 & 0.02\\\midrule
NS-BB & 0.32 & 0.51 & 0.57 & 0.47 & 0.64 & 0.52 & 0.40\\\midrule
NS-SL & 0.36 & 0.47 & 0.48 & 0.46 & 0.45 & 0.45 & 0.59\\\midrule
NS-Tracer-PwC & 0.78 & 0.66 & 0.59 & 0.41 & 0.56 & 0.43 & 0.44\\\midrule
FNS-KF & 0.98 & 0.56 & 1.04 & 0.30 & 0.45 & 0.29 & 0.43\\\midrule
CE-RPUI & 0.35 & 0.37 & 0.43 & 0.27 & 0.32 & 0.05 & 0.23\\\midrule
CE-RM & 0.10 & 0.11 & 0.09 & 0.10 & 0.11 & -0.20 & 0.11\\\midrule
SE-AF & 0.30 & 0.32 & 0.27 & 0.35 & 0.13 & 0.31 & 0.24\\\midrule
GCE-RT & 0.53 & 0.59 & 0.44 & 0.47 & 0.31 & 0.11 & 0.43\\\midrule
Wave-Layer & 0.57 & 0.51 & 0.33 & 0.40 & 0.51 & 0.13 & 0.43\\\midrule
Wave-Gauss & 0.59 & 0.50 & 0.45 & 0.37 & 0.46 & 0.07 & 0.34\\\midrule
ACE & 0.74 & 0.85 & 0.77 & 0.72 & 0.47 & 0.46 & 0.88\\\midrule
Poisson-Gauss & 0.99 & 0.94 & 1.07 & 0.67 & 0.50 & 0.71 & 0.61\\\midrule
Helmholtz & 0.38 & 0.43 & 0.68 & 0.42 & 0.54 & 0.27 & 0.31\\
         \bottomrule
    \end{tabular}
    \label{tab:scaling_exp}
\end{table}

\begin{table}[htbp]
    \caption{Efficiency gain (EG) and Accuracy Gain (\textit{AG}) for the \textsc{Poseidon} models on all downstream tasks.}
    \label{tab:full_res}
    \centering
    \small
    \begin{tabular}{r c c c c c c c c}
        \toprule
        & \multicolumn{6}{c}{Pretrained Models} & \multicolumn{2}{c}{Scratch}\\
         \cmidrule(r){2-7}  \cmidrule(r){8-9}

        & \multicolumn{2}{c}{\textsc{Poseidon}-L}
         & \multicolumn{2}{c}{\textsc{Poseidon}-B}
          & \multicolumn{2}{c}{\textsc{Poseidon}-T}
         & \multicolumn{2}{c}{FNO}\\

         \cmidrule(r){2-3} \cmidrule(r){4-5} \cmidrule(r){6-7} \cmidrule(r){8-9}
        
         & EG & \textit{AG} & EG & \textit{AG} & EG & \textit{AG} & EG & \textit{AG} 
        \\
        \midrule\midrule
NS-PwC & 890.6 & \textit{24.7} & 1024.0 & \textit{19.7} & 1024.0 & \textit{19.8} & 1 & \textit{1} \\\midrule
NS-SVS & 502.9 & \textit{7.3} & 518.9 & \textit{7.9}  & 212.0 & \textit{6.1}  & 1 & \textit{1} \\\midrule
NS-BB & 552.5 & \textit{29.3} & 816.0 & \textit{14.7}  & 365.0 & \textit{19.4}  & 1 & \textit{1} \\\midrule
NS-SL & 21.9 & \textit{5.5} & 19.1 & \textit{4.7}  & 9.7 & \textit{3.7}  & 1 & \textit{1} \\\midrule
NS-Tracer-PwC & 49.8 & \textit{8.7} & 20.4 & \textit{5.4}  & 35.1 & \textit{6.2}  & 1 & \textit{1} \\\midrule
FNS-KF & 62.5 & \textit{7.4} & 16.1 & \textit{4.7}  & 77.9 & \textit{5.9}  & 1 & \textit{1} \\\midrule
CE-RPUI & 352.2 & \textit{6.5} & 370.8 & \textit{6.2}  & 909.7 & \textit{5.8}  & 1 & \textit{1} \\\midrule
CE-RM & 4.6 & \textit{1.2} & 3.1 & \textit{1.1}  & 2.8 & \textit{1.1}  & 1 & \textit{1} \\\midrule
SE-AF & 3.4 & \textit{1.2} & 2.9 & \textit{1.2}  & 2.4 & \textit{1.1}  & 1 & \textit{1} \\\midrule
GCE-RT & 5.3 & \textit{2.0} & 3.2 & \textit{1.5}  & 1.7 & \textit{1.2}  & 1 & \textit{1} \\\midrule
Wave-Layer & 46.5 & \textit{6.1} & 24.9 & \textit{4.7}  & 14.5 & \textit{3.4}  & 1 & \textit{1} \\\midrule
Wave-Gauss & 62.1 & \textit{5.6} & 29.3 & \textit{4.3}  & 19.5 & \textit{3.1}  & 1 & \textit{1} \\\midrule
ACE & 17.0 & \textit{11.6} & 8.7 & \textit{6.5}  & 9.8 & \textit{7.2}  & 1 & \textit{1} \\\midrule
Poisson-Gauss & 42.5 & \textit{20.5} & 24.4 & \textit{13.0}  & 18.2 & \textit{8.4}  & 1 & \textit{1} \\\midrule
Helmholtz & 78.3 & \textit{6.1} & 64.7 & \textit{5.0}  & 64.7 & \textit{4.9}  & 1 & \textit{1} \\

        \bottomrule
    \end{tabular}
\end{table}

\begin{table}[htbp]
    \caption{(Median) Efficiency gain (EG) and (Mean) Accuracy Gain (\textit{AG}) over all downstream tasks for all models. We also present $\bN$(EG) as the number of tasks for which the EG of the model is greater than 10 and $\bN$(AG) as the number of tasks where the AG of the model is greater than 2.}
    \label{tab:stat}
    \centering
    \small
    \begin{tabular}{r c c c c}
        \toprule
            & Median EG & Mean AG & $\bN$(EG) & $\bN$(AG) \\
            \midrule \midrule
            \textsc{Poseidon}-L & {\bf 49.8}  & {\bf 9.58} & {\bf 12} &  {\bf 13} \\
            \midrule
             \textsc{Poseidon}-B & 24.4  & 6.71 & 11 & 12 \\
              \midrule
              \textsc{Poseidon}-T & 19.5  & 6.49 & 10 & 12 \\
               \midrule
               CNO-FM & 10.6   & 2.91 & 8 &  10 \\
                \midrule
            MPP-B & 2.0  & 1.82 & 3 & 6 \\
             \midrule
             CNO & 4.6  & 2.61 & 5 & 6 \\
              \midrule
              scOT & 5.4  & 2.57 & 4 & 8 \\

         \bottomrule
    \end{tabular}
\end{table}

\clearpage

\clearpage

\subsection{Scaling with respect to Model Size}
\begin{figure}[htbp]
    \centering
    \includegraphics[width=\textwidth]{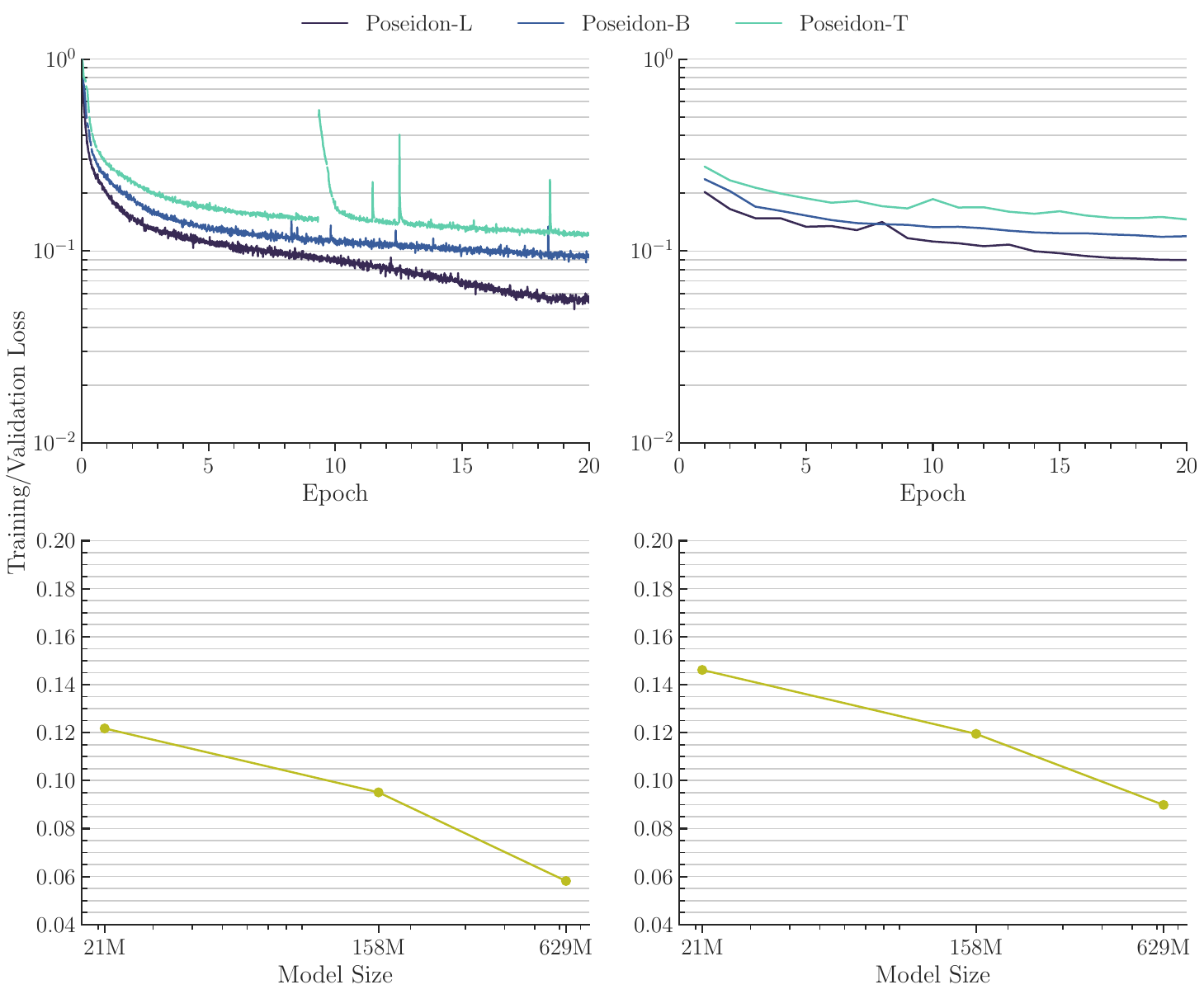}
    \caption{(Top) Training (left) and evaluation (right) losses up to epoch 20 for different model sizes. (Bottom) Scaling at epoch 20 for training loss (left) and evaluation loss (right).}
    \label{fig:model_scaling}
\end{figure}

In Figure \ref{fig:model_scaling}, we plot how the training loss and evaluation (validation) loss during pretraining changes with model size for the \textsc{Poseidon} models. We observe from this figure (bottom row) that there is a consistent decay in losses with increasing model size. The role of model size vis a vis downstream tasks has already been shown in the scaling plots of the previous subsection where we compared \textsc{Poseidon}-L with the smaller \textsc{Poseidon}-B. The corresponding metrics {\bf EG} and {\bf AG} are shown in Table \ref{tab:full_res}. We also see from the statistical summary Table \ref{tab:stat} that there is a gain, on average, in downstream performance with increasing model size for the \textsc{Poseidon} family of models.

\clearpage
\subsection{Scaling with respect to Pretraining Dataset Size and Quality}
\label{sec:perf_downstream_data}
\begin{figure}[htbp]
\centering
\includegraphics[width=\textwidth]{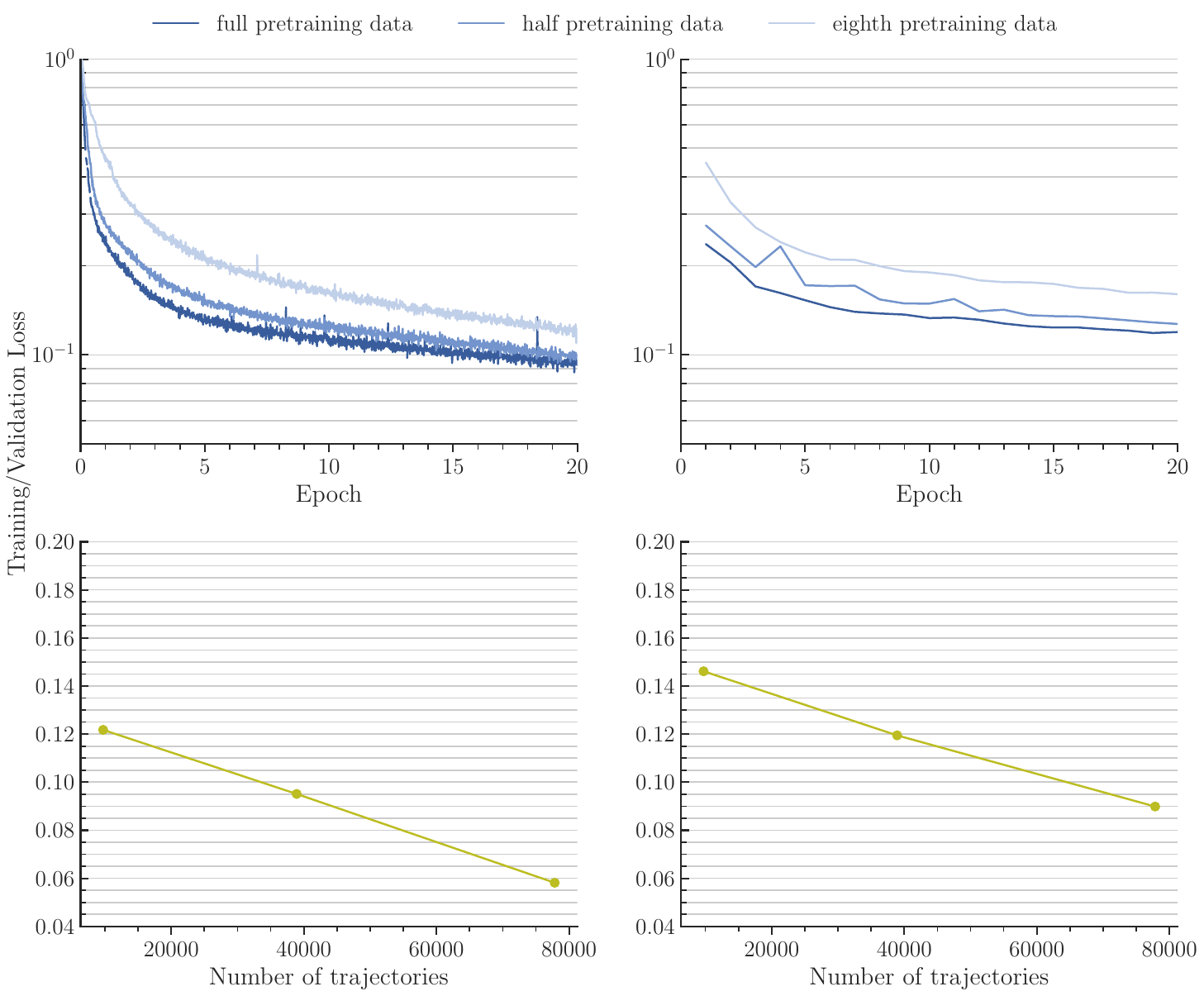}
\caption{(Top) Training (left) and evaluation (right) losses up to epoch 20 for different pretraining dataset sizes. (Bottom) Scaling at epoch 20 for training loss (left) and evaluation loss (right).}
\label{fig:sm4}
\end{figure}

\begin{figure}
    \centering
    \includegraphics[width=\textwidth]{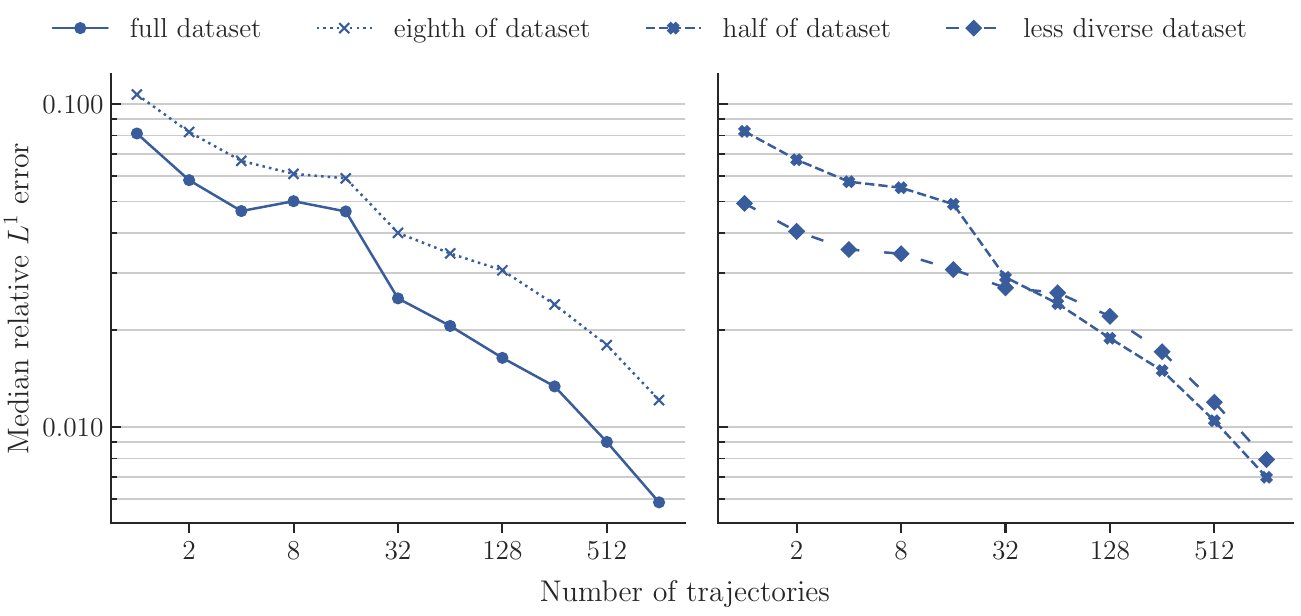}
    \caption{NS-PwC. Number of trajectories vs. median relative $L^1$ error on the test set.}
    \label{fig:quantity_quality_ablations_ns_pwc}
\end{figure}

\begin{figure}
    \centering
    \includegraphics[width=\textwidth]{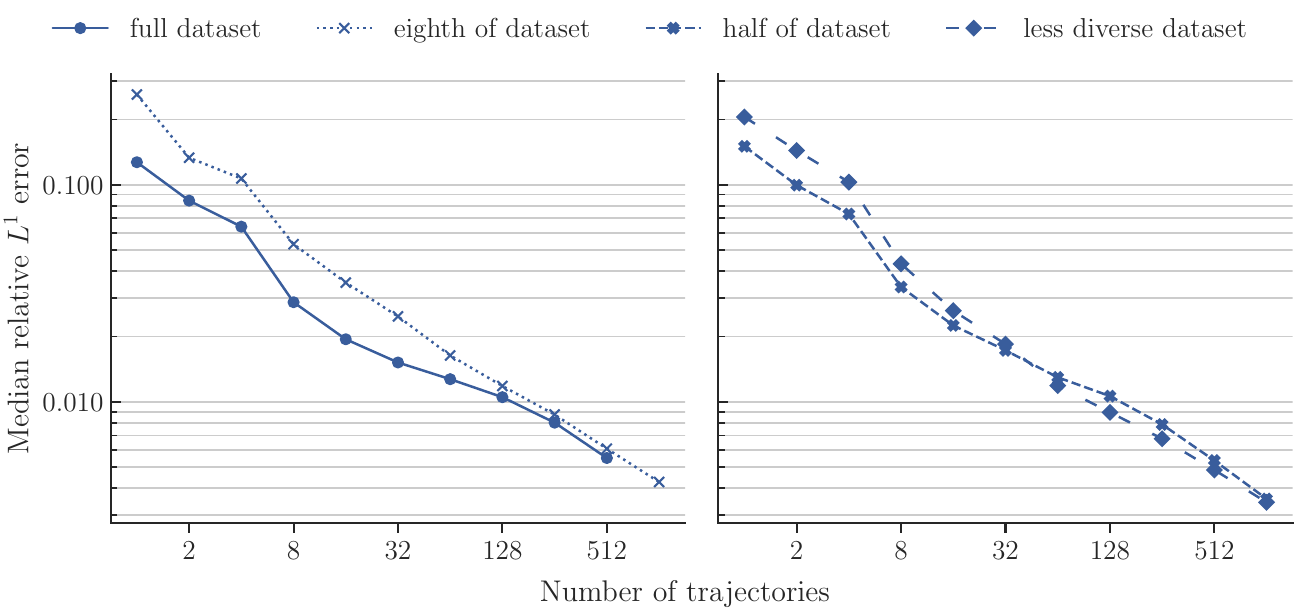}
    \caption{NS-SVS. Number of trajectories vs. median relative $L^1$ error on the test set.}
    \label{fig:quantity_quality_ablations_ns_svs}
\end{figure}

\begin{figure}
    \centering
    \includegraphics[width=\textwidth]{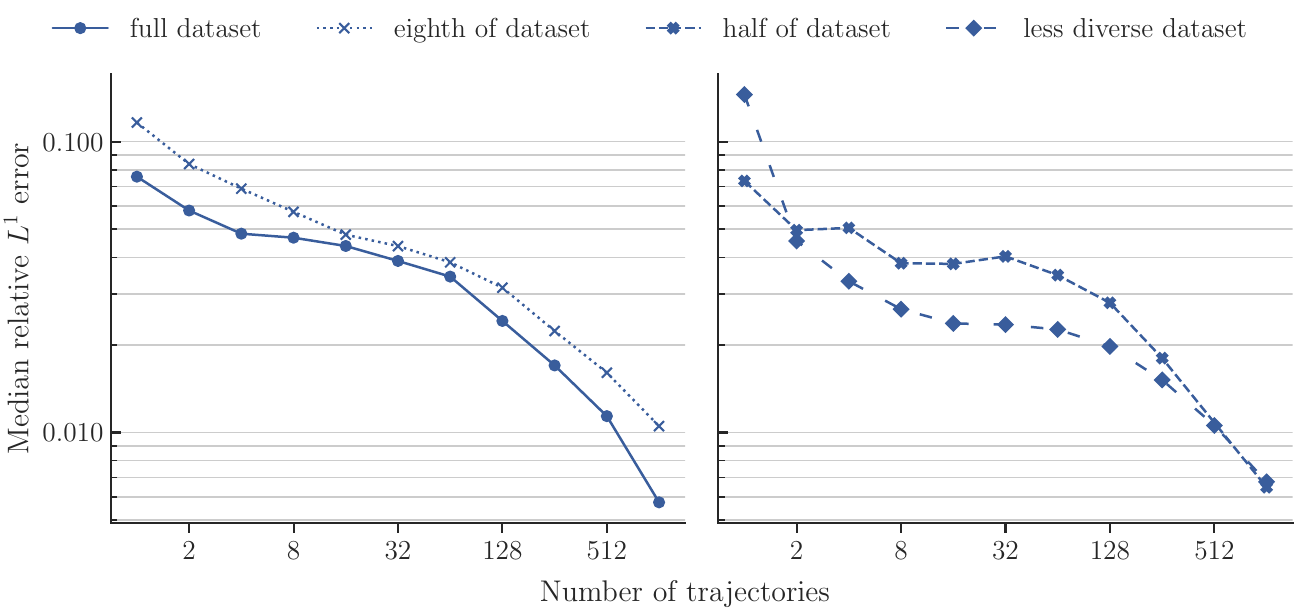}
    \caption{NS-BB. Number of trajectories vs. median relative $L^1$ error on the test set.}
    \label{fig:quantity_quality_ablations_ns_bb}
\end{figure}

\begin{figure}
    \centering
    \includegraphics[width=\textwidth]{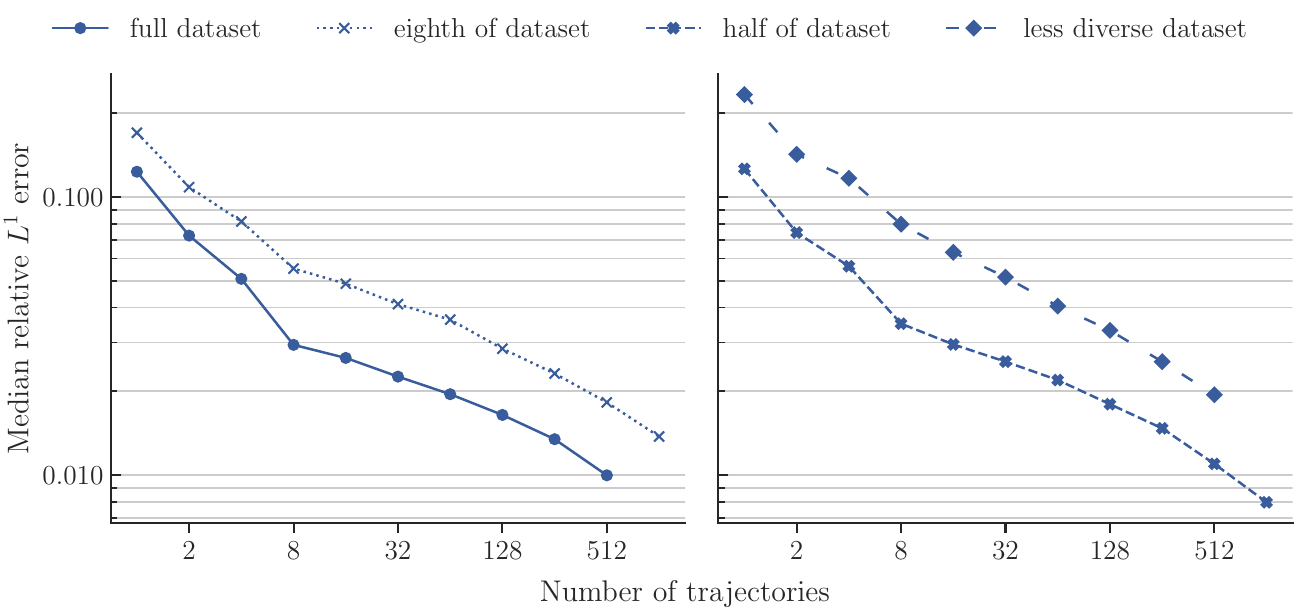}
    \caption{NS-SL. Number of trajectories vs. median relative $L^1$ error on the test set.}
    \label{fig:quantity_quality_ablations_ns_sl}
\end{figure}

\begin{figure}
    \centering
    \includegraphics[width=\textwidth]{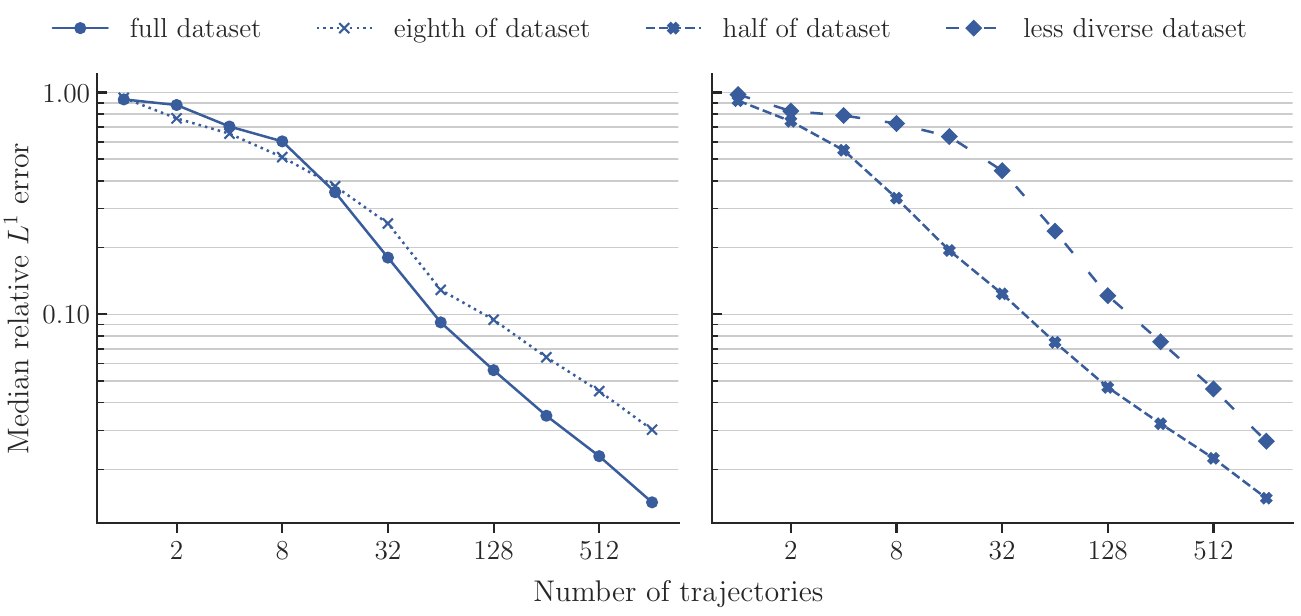}
    \caption{NS-Tracer-PwC. Number of trajectories vs. median relative $L^1$ error on the test set.}
    \label{fig:quantity_quality_ablations_ns_pwc_tracer}
\end{figure}

\begin{figure}
    \centering
    \includegraphics[width=\textwidth]{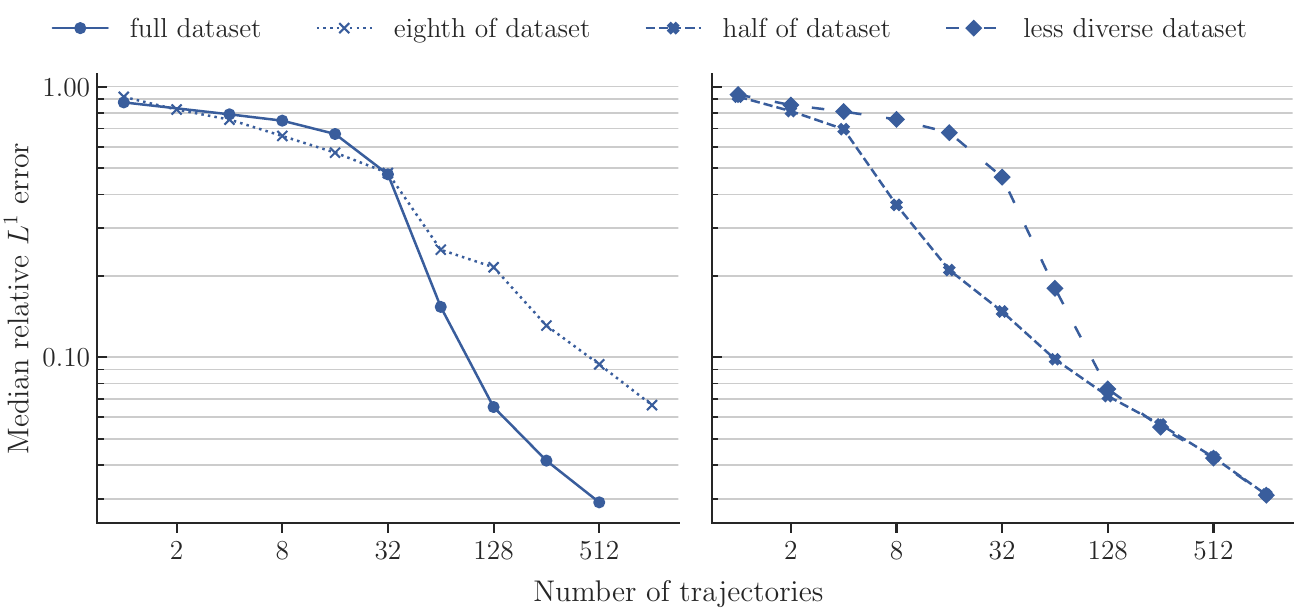}
    \caption{FNS-KF. Number of trajectories vs. median relative $L^1$ error on the test set.}
    \label{fig:quantity_quality_ablations_fns_kf}
\end{figure}

\begin{figure}
    \centering
    \includegraphics[width=\textwidth]{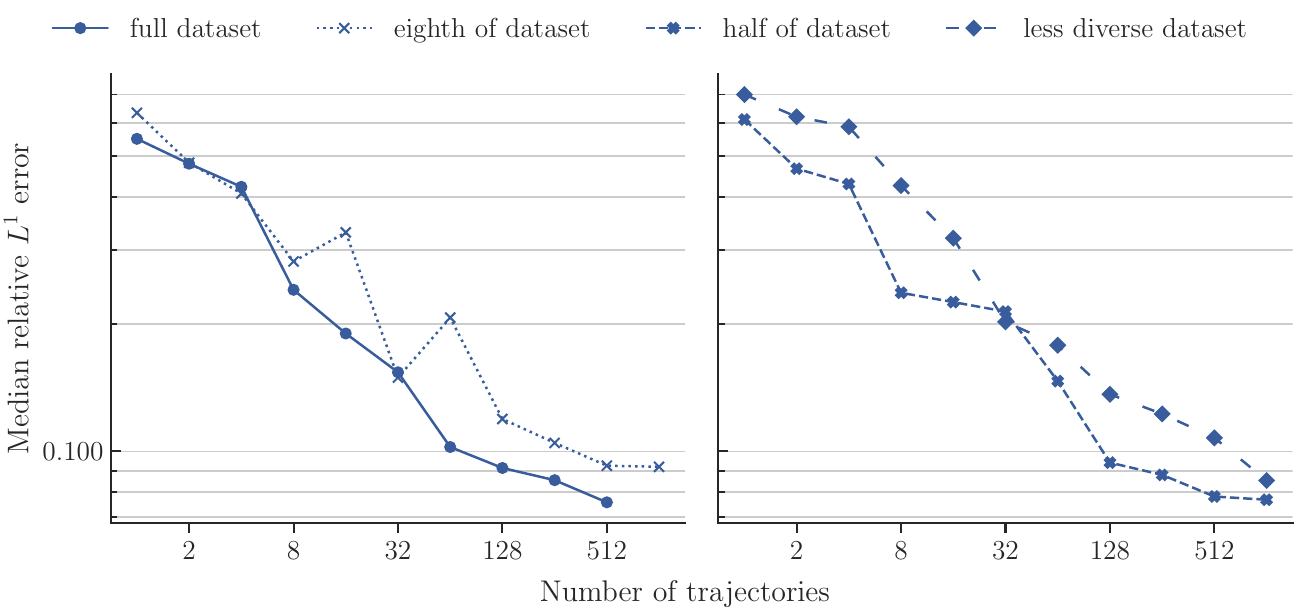}
    \caption{CE-RPUI. Number of trajectories vs. median relative $L^1$ error on the test set.}
    \label{fig:quantity_quality_ablations_ce_rpui}
\end{figure}

\begin{figure}
    \centering
    \includegraphics[width=\textwidth]{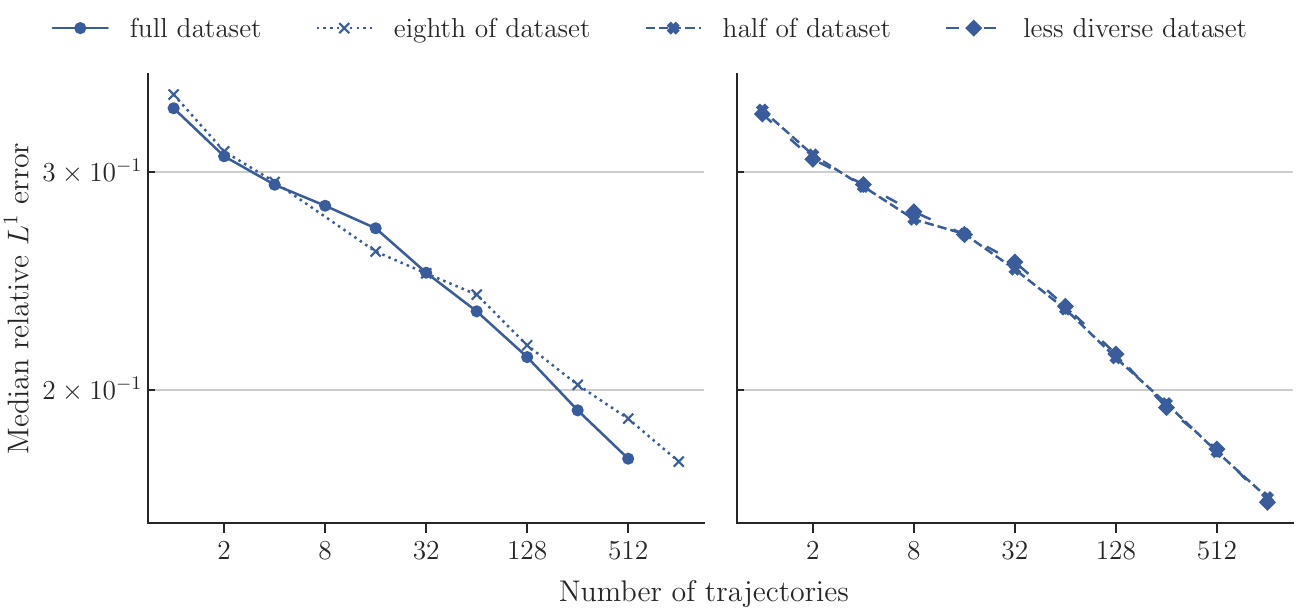}
    \caption{CE-RM. Number of trajectories vs. median relative $L^1$ error on the test set.}
    \label{fig:quantity_quality_ablations_ce_rm}
\end{figure}

\begin{figure}
    \centering
    \includegraphics[width=\textwidth]{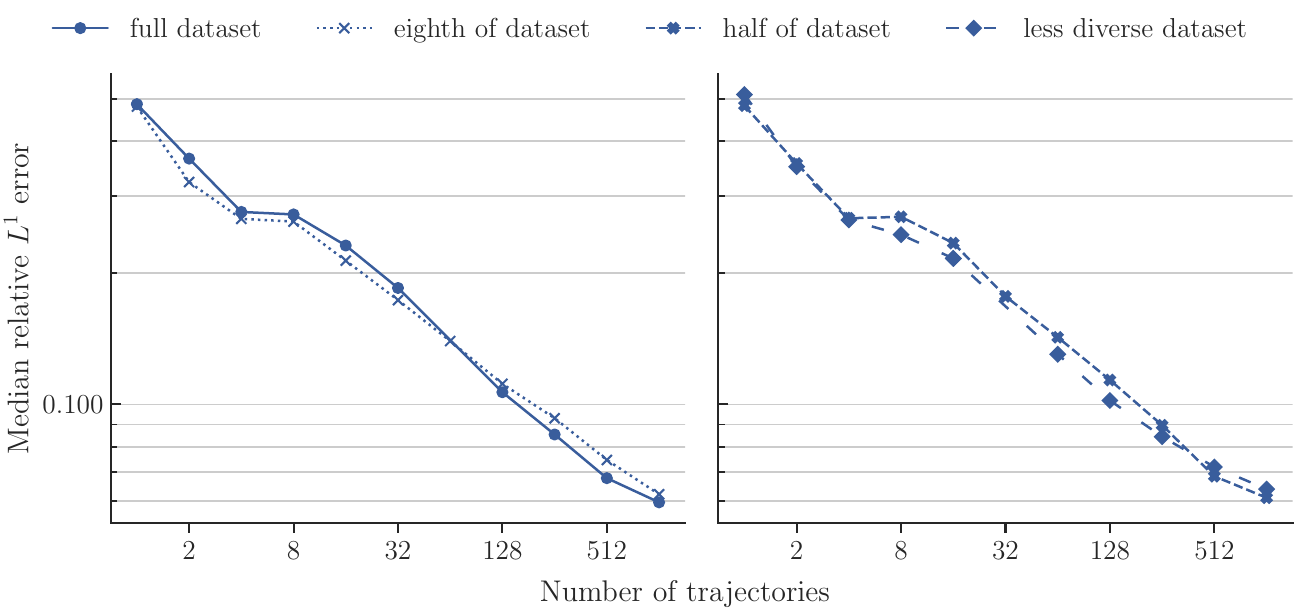}
    \caption{SE-AF. Number of samples vs. median relative $L^1$ error on the test set.}
    \label{fig:quantity_quality_ablations_se_af}
\end{figure}

\begin{figure}
    \centering
    \includegraphics[width=\textwidth]{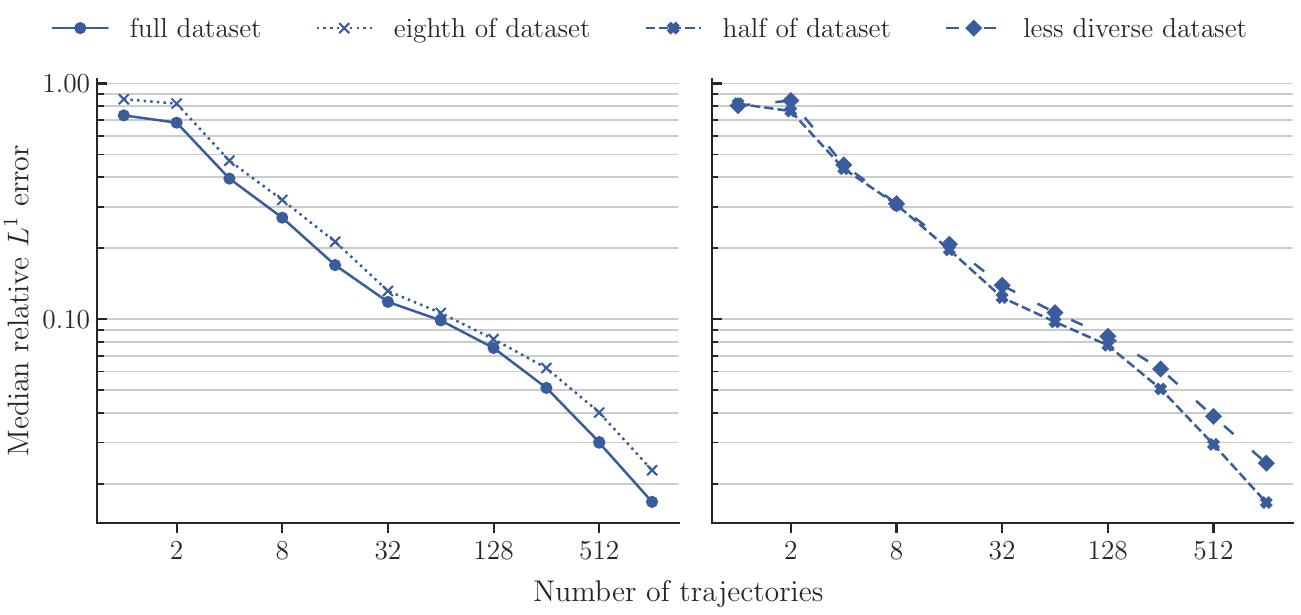}
    \caption{GCE-RT. Number of trajectories vs. median relative $L^1$ error on the test set.}
    \label{fig:quantity_quality_ablations_gce_rt}
\end{figure}

\begin{figure}
    \centering
    \includegraphics[width=\textwidth]{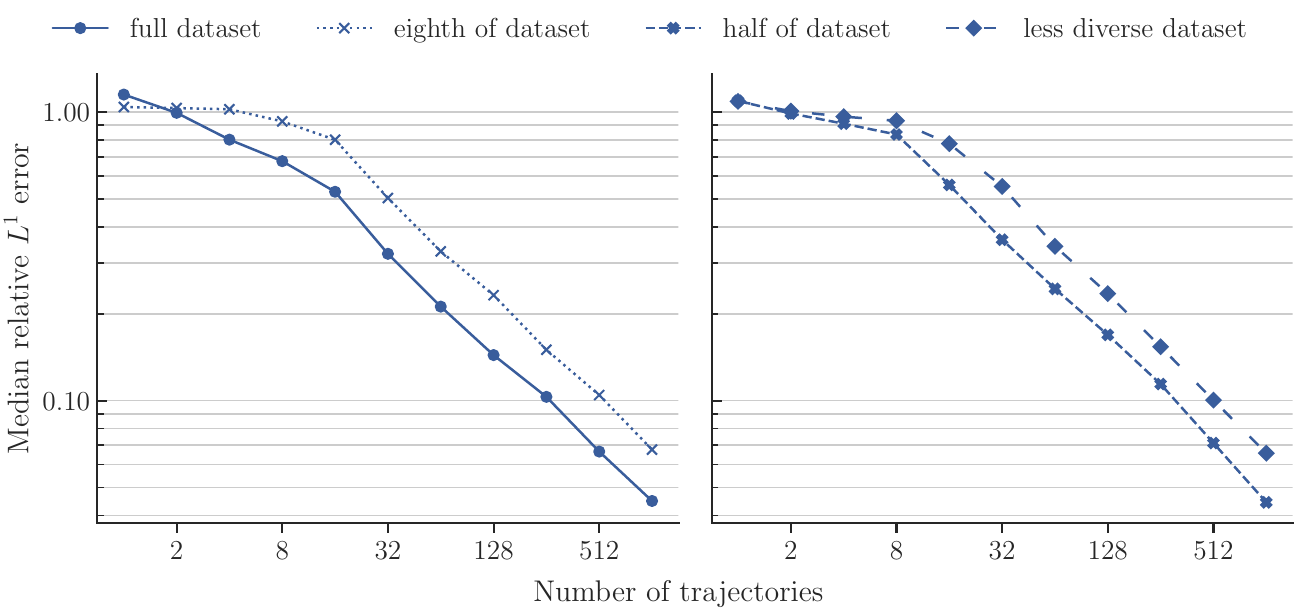}
    \caption{Wave-Layer. Number of trajectories vs. median relative $L^1$ error on the test set.}
    \label{fig:quantity_quality_ablations_wave_layer}
\end{figure}

\begin{figure}
    \centering
    \includegraphics[width=\textwidth]{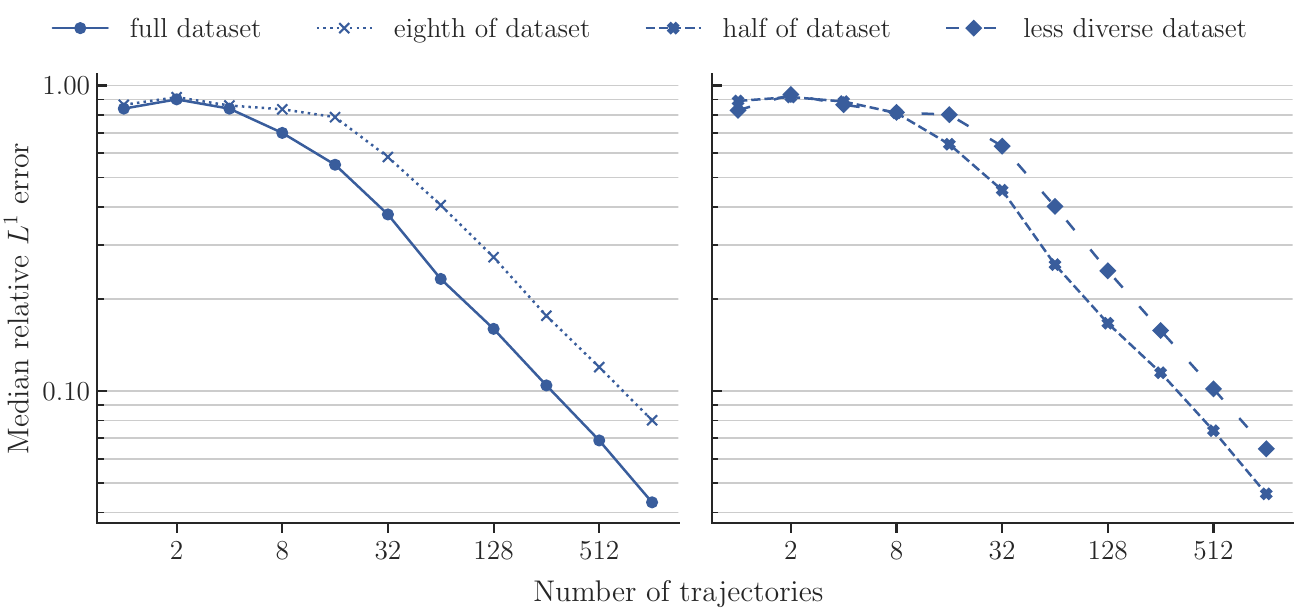}
    \caption{Wave-Gauss. Number of trajectories vs. median relative $L^1$ error on the test set.}
    \label{fig:quantity_quality_ablations_wave_gaussians}
\end{figure}

\begin{figure}
    \centering
    \includegraphics[width=\textwidth]{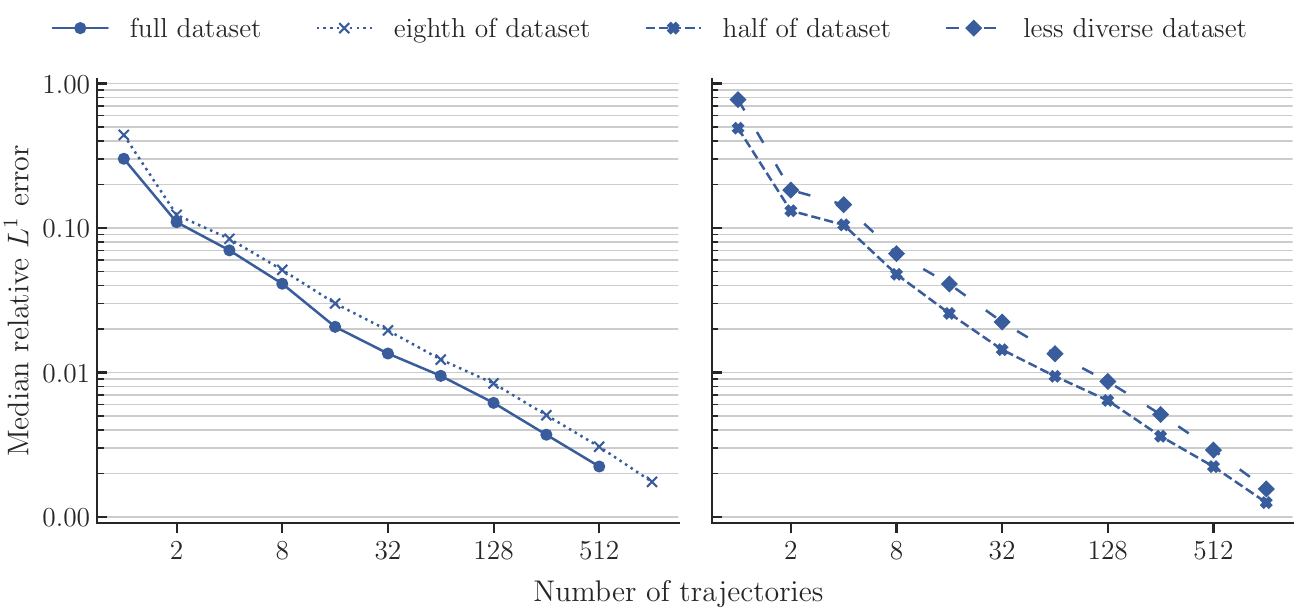}
    \caption{ACE. Number of trajectories vs. median relative $L^1$ error on the test set.}
    \label{fig:quantity_quality_ablations_ace}
\end{figure}

\begin{figure}
    \centering
    \includegraphics[width=\textwidth]{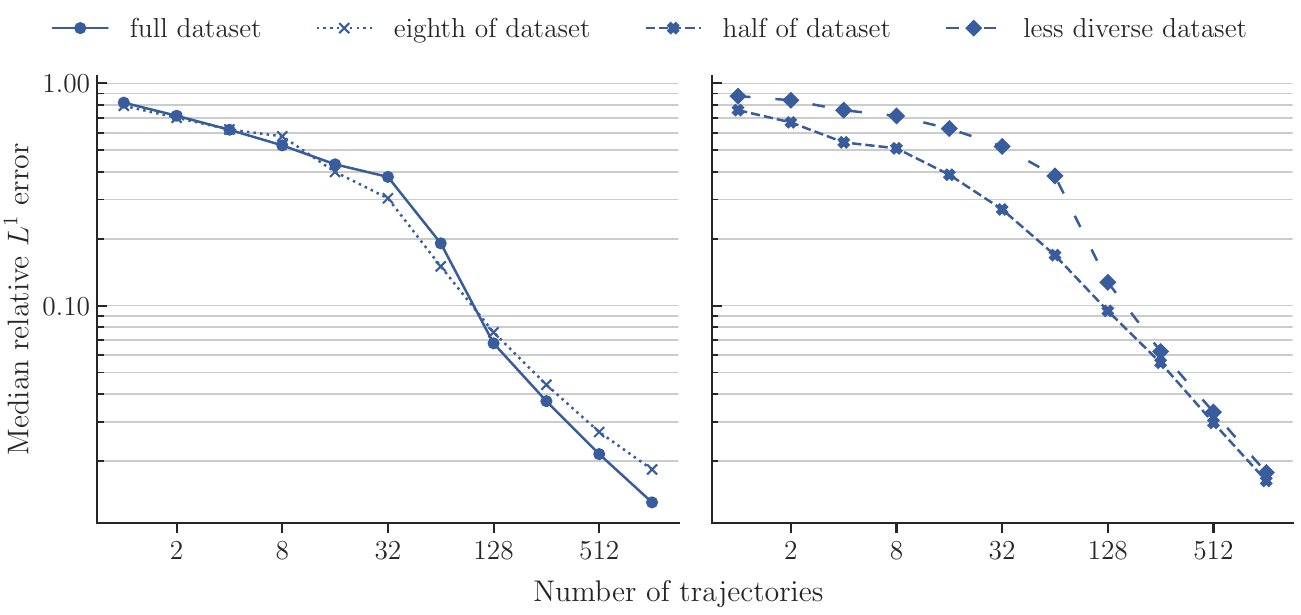}
    \caption{Poisson-Gauss. Number of samples vs. median relative $L^1$ error on the test set.}
    \label{fig:quantity_quality_ablations_poisson_gauss}
\end{figure}

\begin{figure}
    \centering
    \includegraphics[width=\textwidth]{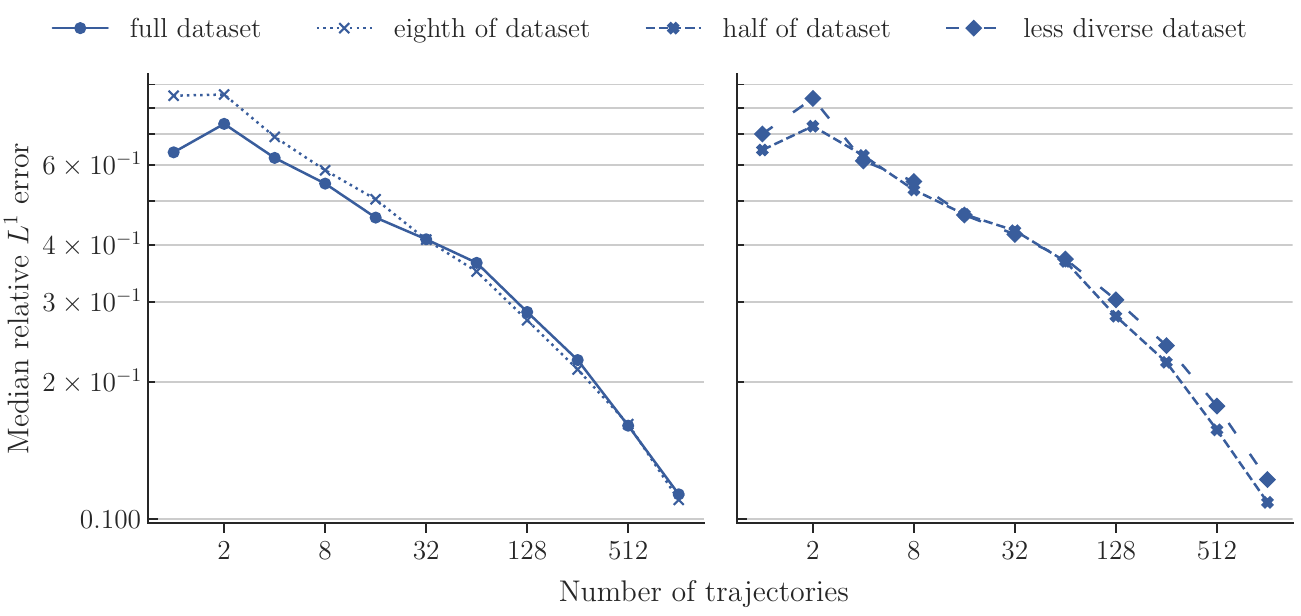}
    \caption{Helmholtz. Number of samples vs. median relative $L^1$ error on the test set.}
    \label{fig:quantity_quality_ablations_helmholtz}
\end{figure}

As mentioned in the Main Text, scaling of the \textsc{Poseidon} models with dataset size is of great interest. To that end, in Figure \ref{fig:sm4}, we plot the training and evaluation losses, during pretraining, for the \textsc{Poseidon}-B model, trained with one-eighth, one-half and full size of the pretraining dataset (the details of the corresponding setups are given in Section \ref{sec:mb}). We see from this figure that \textsc{Poseidon}-B scales with dataset size on the pretraining dataset. Moreover, in Figures \ref{fig:quantity_quality_ablations_ns_pwc} to \ref{fig:quantity_quality_ablations_helmholtz} (left subfigure of each figure), we compare the performance of \textsc{Poseidon}-B, trained on the full pretraining dataset with the same model trained on one-eighth of it. 

Moreover, in Figures \ref{fig:quantity_quality_ablations_ns_pwc} to \ref{fig:quantity_quality_ablations_helmholtz} (right subfigure of each figure), we compare the performance of \textsc{Poseidon}-B, trained on the half the pretraining dataset with the same model pretrained on a less diverse dataset (see Main Text and Section \ref{sec:mb} for details of the setup).  

\clearpage
\subsection{Case Studies}
\label{sec:cs}
Given the excellent performance of \textsc{Poseidon} models across the board, including on tasks that involve PDEs (physical processes) not encountered during pretraining, it is important to understand what underpins this performance. To this end, we will present three case studies in order to explain \textsc{Poseidon}'s robust performance.

\subsubsection{CE-RPUI}
\label{sec:cerpui}
First we consider the CE-RPUI downstream task. Clearly, \textsc{Poseidon} models perform very well on this task, as shown in Figure \ref{fig:scaling_ce_rpui} as well as Tables \ref{tab:1} and \ref{tab:full_res}. Also, as seen from Figure \ref{fig:ce_rpui}, where we visualize a single random sample for all the variables at time $T=0.7$, \textsc{Poseidon}-B is much more accurate, when finetuned on $128$ trajectories, than CNO and FNO, which are trained from scratch with the same number of trajectories. Note that the underlying solution is very complex, with a mixture of shocks and roll-up vortices. While \textsc{Poseidon} captures these shocks and vortices very sharply, CNO and (especially) FNO fail to do so. What explains this impressive performance of \textsc{Poseidon} on this difficult downstream task? 

\begin{figure}
    \centering
    \includegraphics[width=\textwidth]{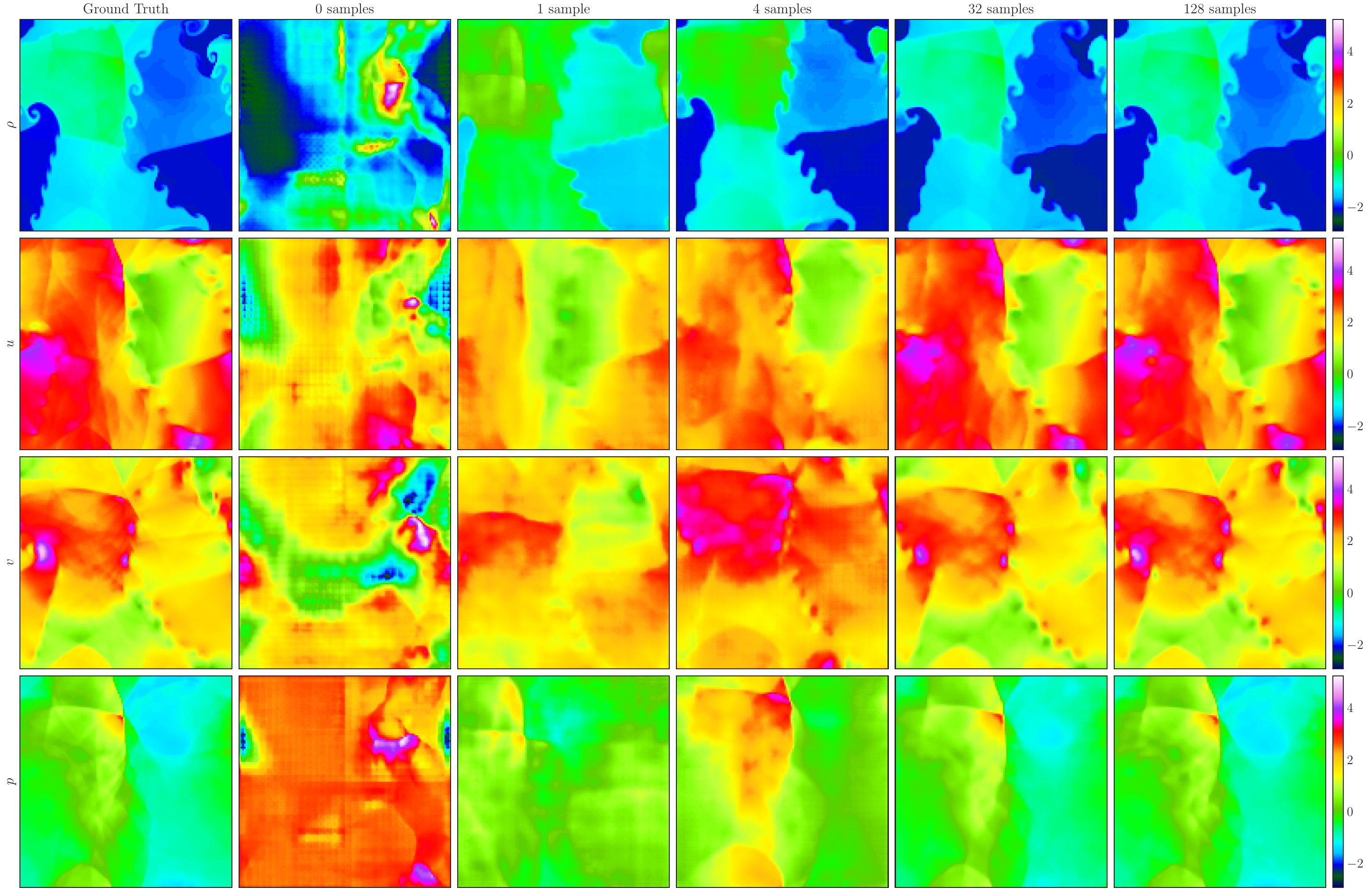}
    \caption{How \textsc{Poseidon}-B approximates a random sample for the CE-RPUI task when trained with different numbers of task-specific trajectories.}
    \label{fig:case_studies_rpui}
\end{figure}

We start by observing that, like all other downstream tasks, this task is out-of-distribution (o.o.d.) with respect to the pretraining dataset. Although the underlying PDE (compressible Euler equations \eqref{eq:ceuler}) is present in the pretraining dataset, this data distribution has not been seen during pretraining. This \emph{o.o.d.} nature of the task is clearly seen from Figure \ref{fig:case_studies_rpui}, where we plot how the same random sample (visualized in Figure \ref{fig:ce_rpui})) is inferred with a \textsc{Poseidon}-B model \emph{zero-shot}. We see from the figure (second column from the left) that the zero-shot results are rather poor. However, even with 1 task-specific example, we see from Figure \ref{fig:case_studies_rpui} (third column from left) that at least, some large scale features (such shock locations) are approximated reasonably accurately. With just 4 downstream trajectories, the quality of approximation improves dramatically and even the vortex roll-ups are captured accurately. The quality of approximation continues to improve with 32 and 128 downstream trajectories, as shown in the right-most columns of Figure \ref{fig:case_studies_rpui}. Thus, from this figure we conclude that a few task-specific samples suffice to accurately approximate the underlying solution operator. This is also evidenced in the scaling plot Figure \ref{fig:scaling_ce_rpui}.  

How does \textsc{Poseidon} succeed in learning this complex solution operator with so few samples? We know that the pretraining dataset contains the CE-RP operator, where the initial condition (see Figure \ref{fig:ce_rp} for a sample) has a similar four-quadrant Riemann problem structure as the intial conditions in the CE-RPUI benchmark, the main difference being that the interfaces, across which the initial data is discontinuous, are now perturbed sinusoidally, instead of being axis-aligned. However, it is precisely these perturbations that are responsible for the roll-up of small-scale vortices that are absent in the CE-RP operator. Thus, the model potentially needs to learn how to approximate small-scale vortices accurately from some other operator in the pretraining dataset, while learning how to propagate large-scale shocks from the CE-RP operator. 

\begin{figure}
    \centering
    \includegraphics[width=0.7\textwidth]{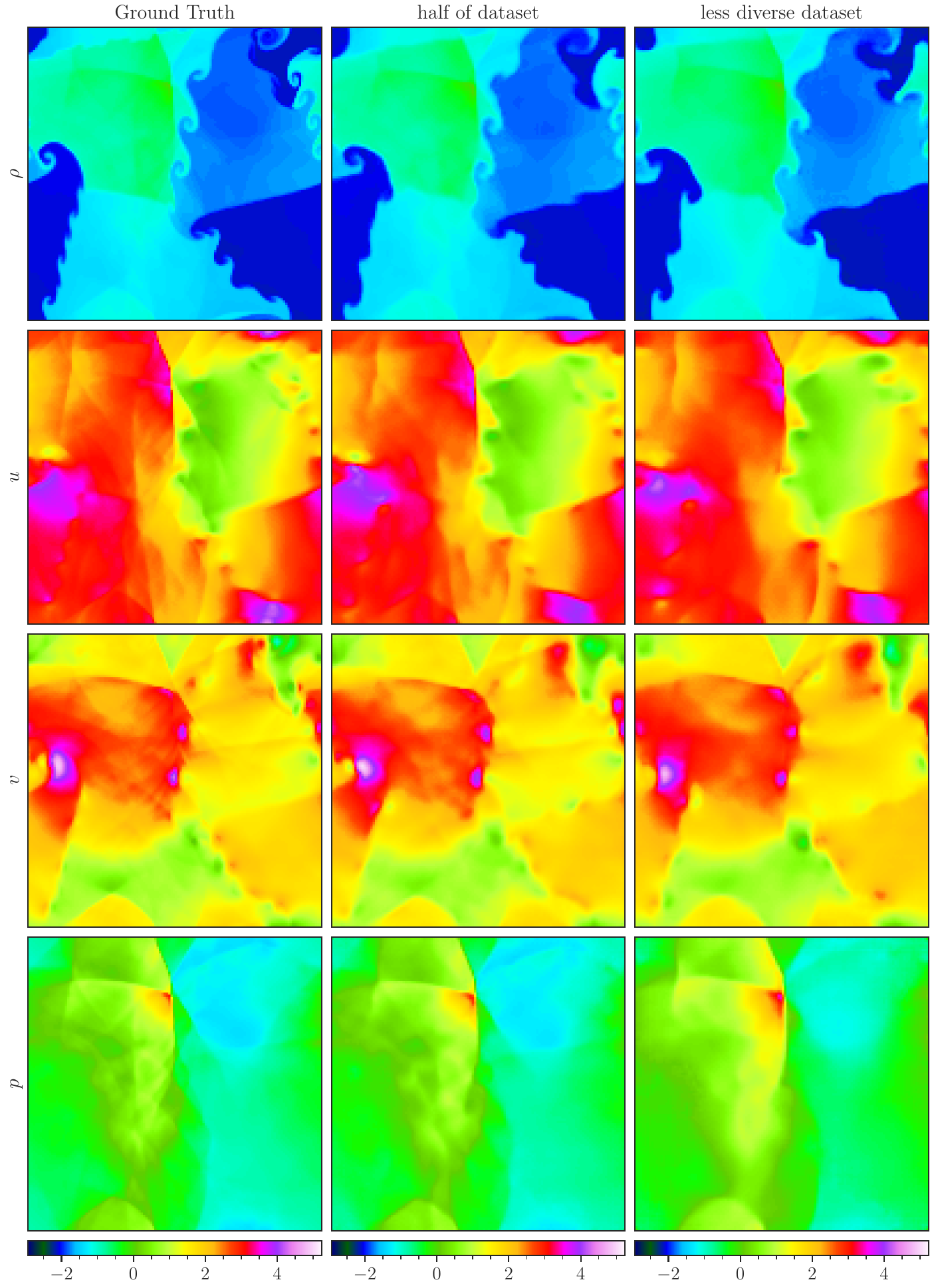}
    \caption{A sample of CE-RPUI when \textsc{Poseidon}-B is pretrained on half of the pretraining dataset vs. a less diverse pretraining dataset.}
    \label{fig:half_ldd_ce-rpui}
\end{figure}

One would think that the CE-KH operator in the pretraining dataset provides the information about vortex roll-ups, see Figure \ref{fig:ce_kh} for visualizing a sample. However, the underlying vortices are much larger. So, where does this missing information come from? One possible source could be the CE-CRP operator (see Figure \ref{fig:ce_crp}) where vortices of many different scales are being formed. Perhaps, the model leverages shock propagation from CE-RP, large vortex roll-ups from CE-KH and small-scale vortex dynamics, as well as curved shock propagation, from CE-CRP in order to provide very good approximation with a few training examples on CE-RPUI. A partial test of this hypothesis is to check if the model, pretrained with the less-diverse dataset that excludes CE-CRP performs worse than the model pretrained with the full dataset. This is already shown in Figure \ref{fig:quantity_quality_ablations_ce_rpui} (Right) where the performance of the model, pretrained with the less-diverse dataset is worse than the model trained with the similarly sized but fully diverse dataset. This behavior is further reinforced from Figure \ref{fig:half_ldd_ce-rpui}, where the approximation of the same sample, considered in Figure \ref{fig:case_studies_rpui}, with these ablated models is shown. As predicted, the model pretrained on the less diverse dataset is clearly less accurate at resolving small-scale vortices than the competing one trained on the more-diverse dataset. It misses the input from the CE-CRP operator regarding small-scale vortex dynamics. 

This qualitative analysis illustrates how the \textsc{Poseidon} model leverages different operators from its pretraining dataset to amortize different aspects in order to construct accurate approximations during finetuning with a few task-specific examples and throws some light into how a foundation model for PDEs can learn effective representations from its pretraining phase. 

\subsubsection{ACE}
Next, we consider the ACE downstream task, where the underlying PDE is the nonlinear parabolic Allen-Cahn equation \eqref{eq:ac}, which is clearly not included in the pretraining dataset for \textsc{Poseidon}. More importantly, the type of physics that the Allen-Cahn Equation models is that of \emph{reaction-diffusion}. On the other hand, the PDEs included in the pretraining dataset, Compressible Euler and Incompressible Navier-Stokes at very high Reynolds number, are convection-dominated. Hence, one does not expect that the pretrained model has learned effective representations about reaction-diffusion. Yet, we see from Figure \ref{fig:scaling_ace} that \textsc{Poseidon} is very effective at learning this solution operator from a few training examples. This point is also reinforced from Figure \ref{fig:ace}, where we show how a single randomly chosen sample is well-approximated by \textsc{Poseidon}. How does \textsc{Poseidon} learn these \emph{new physics}? 

\begin{figure}
    \centering
    \includegraphics[width=\textwidth]{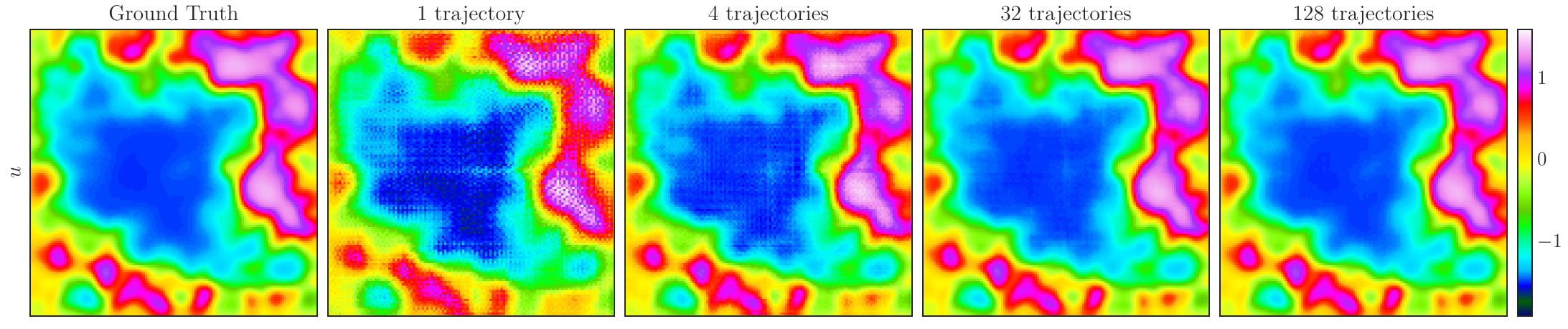}
    \caption{How \textsc{Poseidon}-B approximates a random sample for the ACE task when trained with different numbers of task-specific trajectories.}
    \label{fig:case_studies_ace}
\end{figure}

To understand the factors behind \textsc{Poseidon}'s performance, we plot how the same random sample, visualized in Figure \ref{fig:ace}, is approximated by \textsc{Poseidon}-B, when fine-tuned with different number of task-specific examples, ranging from 1 to 128, in Figure \ref{fig:case_studies_ace}. We observe from this figure that already with just 1 task-specific trajectory, \textsc{Poseidon} is able to learn the large-scale features of the solution of Allen-Cahn approximately. In particular, it has learnt both front propagation (potentially from all propagating shock waves seen during pretraining) as well as diffusion (spreading) of localized features. With more downstream trajectories, it is able to adjust local features quite well to further approximate the diffuse fronts. This case study shows how \textsc{Poseidon} can learn new features from a few task-specific training examples. 

\begin{figure}
    \centering
    \includegraphics[width=\textwidth]{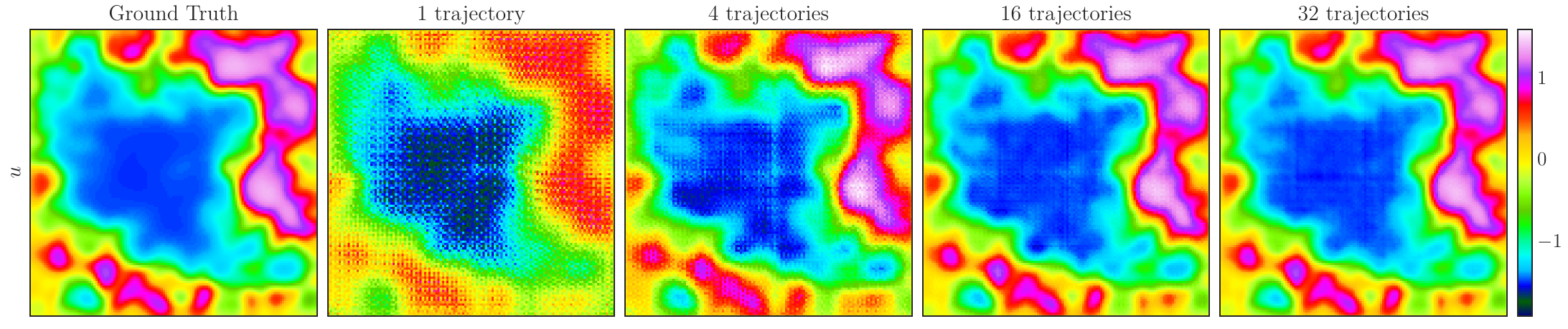}
    \caption{How \textsc{Poseidon}-B with a \emph{Frozen Latent Representation} approximates the same random sample as in Figure \ref{fig:case_studies_ace} for the ACE task when trained with different numbers of task-specific trajectories.}
    \label{fig:case_studies_ace_1}
\end{figure}

Given these encouraging results on the ability of \textsc{Poseidon} to generalize to the unseen physics underlying the Allen-Cahn equation, we investigate this ability further by \emph{freezing the latent space} of \textsc{Poseidon} during finetuning by setting $\widehat{\theta}_r = \widehat{\theta}_\ast$, for all $r$, in the gradient descent procedure \eqref{eq:gdfm} for finetuning. Thus, only the \emph{embedding and recovery} parameters are learned and the rest frozen. This results in an \emph{extremely lightweight} model for training as less than $0.5\%$ of the total parameters in \textsc{Poseidon} are being retrained. Nevertheless, as shown in Figure \ref{fig:case_studies_ace_1}, even this very parsimonious form of \textsc{Poseidon} has already learned the solution of the Allen-Cahn equation qualitatively with only one training trajectory, although there is a quantitative mismatch. This mismatch is corrected when further samples are shown to the model. In particular, with $32$ trajectories, the error with this model ($0.031$) is actually lower than FNO with $128$ trajectories ($0.037$) although it is higher than the corresponding model where all the parameters of \textsc{Poseidon}-B are finetuned ($0.014$). This experiment demonstrates that the latent representations learned from the equations of fluid dynamics during pretraining are very rich and can unexpectedly contain information about reaction-diffusion equations, which are then leveraged by the \emph{frozen-latent} model to learn the underlying solution operator.

\subsubsection{Poisson-Gauss}
\label{sec:pg}
In our final case study, we consider the Poisson-Gauss task. The underlying PDE is the Poisson equation \eqref{eq:pos} and the solution operator maps the coefficient, which is a superposition of Gaussians, into the solution. A visualization of the solution operator for a single random sample is shown in Figure \ref{fig:poisson_gaussians} and shows how the source is \emph{diffused and smoothed out}. 

\begin{figure}
    \centering
    \includegraphics[width=\textwidth]{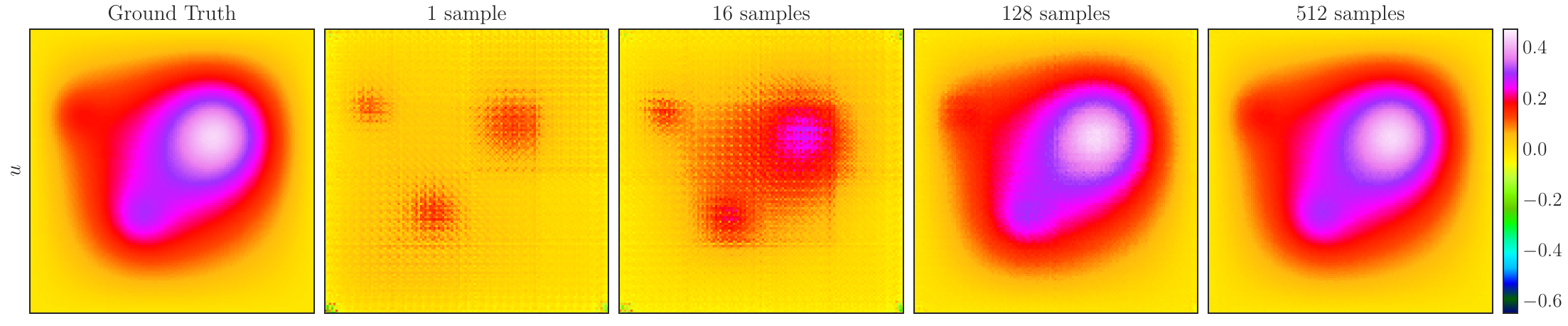}
    \caption{How \textsc{Poseidon}-B approximates a random sample for the Poisson-Gauss task when trained with different numbers of task-specific samples.}
    \label{fig:case_studies_poisson}
\end{figure}

We remark that this task is very different from the pretraining dataset in various ways. First, the underlying PDE is time-independent in contrast to the two time-dependent PDEs seen during pretraining. Second, the underlying physics of \emph{diffusion} of features and their \emph{smoothing out} is patently different from the physics seen in the pretraining dataset. Finally, the Dirichlet boundary conditions considered here are also different from the periodic boundary conditions of the pretraining dataset. Nevertheless, we see from the scaling plot Figure \ref{fig:scaling_poisson_gauss} and Table \ref{tab:1} and \ref{tab:full_res} that \textsc{Poseidon} models perform very well in this case. This is also observed from Figure \ref{fig:poisson_gaussians}, where we observe that \textsc{Poseidon}-B learns this particular random sample far better than CNO and FNO, with the same number (512) of training samples. To understand the reasons behind \textsc{Poseidon}'s performance, in Figure \ref{fig:case_studies_poisson}, we again plot how this foundation model approximates this particular random sample, when trained with an increasing number of task-specific samples. We see from this figure that for 1 sample, the approximation is very poor, indicating how much \emph{out-of-distribution} this task is, with reference to the pretraining dataset. In fact, the model simply learns to approximate the input. However, within a few samples (16), it has learnt that the input needs to be spread (diffused) out. It takes about 128 samples for the model to realize that the input needs to be both spread out as well as smoothened and by 512 samples, the local adjustments needed to further smoothen the output have been made. 

A few remarks are in order to explain this qualitative picture. First, \textsc{Poseidon} could have used the first few samples in training to \emph{forget} the information from the pretraining phase. Yet, it does not do that and uses the very first sample to already just output the identity operator. Then, there appears to be a \emph{warmup} phase where the model slowly learns the underlying physics, for instance diffusion and smoothening and then a fast learning phase where the operator can be better approximated. This qualitative picture is also consistent with the observed \emph{biphasic} power scaling, see the subsection on scaling laws in section \ref{sec:perf_downstream_scalings}, and the fact that there is a \emph{phase transition} between the warmup and fast learning phases in the power law \eqref{eq:biplaw}. This case study sheds further light into how \textsc{Poseidon} can learn \emph{unseen physics} from a few task-specific training examples. 

\begin{figure}
    \centering
    \includegraphics[width=\textwidth]{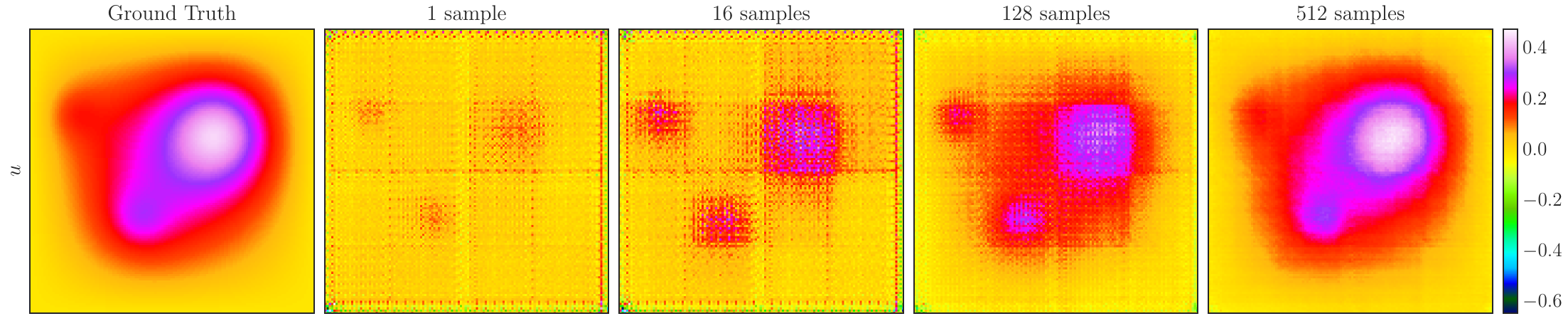}
    \caption{How \textsc{Poseidon}-B, with a \emph{Frozen Latent} Representation, approximates a random sample for the Poisson-Gauss task when trained with different numbers of task-specific samples.}
    \label{fig:case_studies_poisson_1}
\end{figure}

As with the Allen-Cahn equations of the previous section, we further study the factors underpinning the ability of \textsc{Poseidon} to generalize to this PDE by \emph{freezing the latent space} of \textsc{Poseidon} during finetuning by setting $\widehat{\theta}_r = \widehat{\theta}_\ast$, for all $r$, in the gradient descent procedure \eqref{eq:gdfm} for finetuning. Thus, only the \emph{embedding and recovery} parameters are learned and the rest frozen. As shown in Figure \ref{fig:case_studies_poisson_1}, even this frozen-latent form of \textsc{Poseidon} has already learned the basic features of the underlying solution operator, i.e., Diffusion and Smoothing, qualitatively with only a few training samples, although there is a quantitative mismatch. This mismatch is corrected when further samples are shown to the model. In particular, with $512$ trajectories, the error with this model ($0.11$) is significantly lower than FNO ($0.282$) although it is higher than the corresponding model where all the parameters of \textsc{Poseidon}-B are finetuned ($0.022$). This experiment further demonstrates that the latent representations learned from the equations of fluid dynamics during pretraining are rich enough to even contain information about the a priori unrelated physics of steady state diffusion, which are then leveraged by the \emph{frozen-latent} model to learn the underlying solution operator. 

\clearpage
\subsection{Results with DPOT}
\label{sec:dpot}
\begin{table}[htbp]
    \centering
    \small
        \caption{Efficiency gain EG (\eqref{eq:met} with $S=1024$ for time-dependent and $S=4096$ for time-independent PDEs) and Accuracy Gain (\textit{AG}) (\eqref{eq:met} with $S=128$ for time-dependent and $S=512$ for time-independent PDEs) for DPOT and tested downstream tasks.}
    \begin{tabular}{r c c c c c c c c c c}
        \toprule
        & \multicolumn{4}{c}{Finetuned DPOT} & \multicolumn{4}{c}{DPOT from Scratch}\\
         \cmidrule(r){2-5}  \cmidrule(r){6-9}

         & \multicolumn{2}{c}{M}
         & \multicolumn{2}{c}{L}
         & \multicolumn{2}{c}{M}
         & \multicolumn{2}{c}{L}\\

         \cmidrule(r){2-3} \cmidrule(r){4-5} \cmidrule(r){6-7} \cmidrule(r){8-9}
        
         & EG & \textit{AG}  & EG & \textit{AG}  & EG & \textit{AG}  & EG & \textit{AG} 
        \\
        \midrule\midrule
        NS-PwC  & 44.8 & \textit{12.5} & 39.7 & \textit{12.0} & 17.0 & \textit{6.1} & 23.3 & \textit{10.2}\\\midrule
        NS-SL  & 4.5 & \textit{2.4} & 4.7 & \textit{2.4} & 2.1 & \textit{1.3} & 3.0 & \textit{1.6}\\\midrule
        FNS-KF  & 0.0 & \textit{1.0} & 0.0 & \textit{0.9} & 0.0 & \textit{0.8} & 0.0 & \textit{0.8}\\\midrule
        CE-RPUI & 53.5 & \textit{3.7} & 53.6 & \textit{3.6}  & 26.1 & \textit{2.5} & 31.2 & \textit{2.9}\\\midrule
        SE-AF & 3.5 & \textit{1.2} & 4.7 & \textit{1.4} & 4.4 & \textit{1.3} & 5.1 & \textit{1.4}\\\midrule
        Wave-Layer & 23.5 & \textit{5.5} & 28.9 & \textit{6.0}  & 14.1 & \textit{3.6} & 17.8 & \textit{4.2}\\\midrule
        Wave-Gauss  & 25.2 & \textit{4.4} & 27.8 & \textit{4.5} & 18.0 & \textit{3.3} & 20.5 & \textit{3.6}\\
        \bottomrule
    \end{tabular}
    \label{tab:dpot}
\end{table}
The DPOT foundation model \cite{DPOT} has been trained on operators for the compressible and incompressible Navier-Stokes equations, Reaction-Diffusion equations and Shallow-Water equations. The model has been setup to take a sequence of time steps for a time-dependent PDE and output the next time step. However, we can modify it for finetuning for our {\bf OLT} operator learning task by following exactly the same procedure as for finetuning the MPP foundation model. For steady state problems, an identical procedure as with MPP is used. This allows us to perform a fair comparison between DPOT and the \textsc{Poseidon} models proposed here. 

To this end, we consider DPOT-M (with $120$ M parameters) and DPOT-L (with $509$ M parameters) which are comparable in size to the \textsc{Poseidon-B} and \textsc{Poseidon-L} models, respectively. Given compute constraints, we focus this comparison on a representative subset of 7 downstream tasks which are listed in Table \ref{tab:dpot}. Moreover, a trained-from-scratch DPOT model, with the Adaptive FNO architecture, is also employed for each task to evaluate DPOT's model performance. For both finetuning and training models from scratch, we employed the Adam optimizer \cite{kingma2014adam} with a weight decay of $10^{-6}$, and a \textit{1cycle} learning rate policy. For finetuning DPOT models, the maximum learning rate was set to $10^{-4}$, and training was conducted for 100 epochs. When training models from scratch, we used a maximum learning rate of $10^{-3}$ and trained for 200 epochs.

The resulting EG and AG scores are presented in Table \ref{tab:dpot}. These scores should be compared with the corresponding EG and AG scores of \textsc{Poseidon}-L and scOT from Table \ref{tab:1} and \textsc{Poseidon}-B from Table \ref{tab:full_res}. Comparing these results, we make the following observations, 
\begin{itemize}
\item For all these tasks except SE-AF, \textsc{Poseidon} is significantly better, both in terms of efficiency and accuracy gains, to the corresponding DPOT model. Even for SE-AF, the models are very comparable. The superiority in performance of \textsc{Poseidon} is seen very clearly when we consider the mean AG scores over these 7 downstream tasks which amount to \textsc{Poseidon}-L ($8.14$), \textsc{Poseidon}-B ($6.5$), DPOT-M ($4.39$) and DPOT-L ($4.4$). Hence, the \textsc{Poseidon}-L model is almost \emph{twice} more accurate than both the DPOT models considered here. In fact, the DPOT models' performance lies in-between CNO-FM with an average AG score of $2.66$ and the \textsc{Poseidon} models. Similar results also hold for the efficiency gain score. 

\item Surprisingly, DPOT foundation models do not seem to scale with model size, at least on this set of 7 representative downstream tasks as seen from the mean AG scores of $4.4$ for both the DPOT-M and DPOT-L models where an increase of the number of parameters by a factor of $5$ does not lead to any noticeable increase in model performance on downstream tasks.

\item As surprisingly, the stand-alone DPOT neural operators performed well on this dataset. For instance, the average AG score of trained-from-scratch DPOT-L is $3.53$, which is only $25\%$ lower than the DPOT-L foundation model. On the other hand, \textsc{Poseidon}-L is almost $5$ times more accurate than the underlying scOT neural operator. These results indicate that DPOT foundation models do not harness latent representations as well as \textsc{Poseidon} does and they rely on the capacity of the underlying neural operator to learn downstream tasks. 

\end{itemize}

Taken together, our results indicate that (our modification of) DPOT performs better than the CNO-FM and MPP foundation models but is significantly inferior to the \textsc{Poseidon} models. Moreover, the lack of scaling with model size for DPOT on downstream tasks and questions over how it uses latent representations further point to the advantages of \textsc{Poseidon} over this competing model. Nevertheless, this comparison merits further study.

\clearpage
\subsection{Further Ablations and Results}
\label{sec:fa}
\subsubsection{On all2all training}
\begin{figure}
    \centering
    \includegraphics[width=0.6\textwidth]{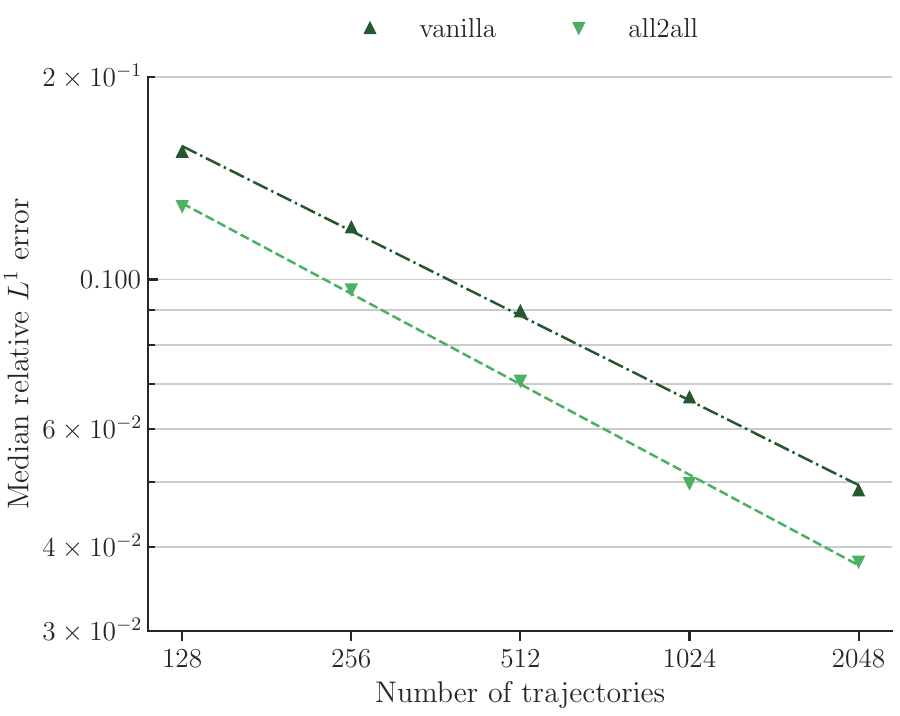}
    \caption{NS-SL. Testing errors of the CNO models trained in an \textit{all2all} and \textit{vanilla} manner. Performance improves with all2all training.}
    \label{fig:all2all_vanilla_sl}
\end{figure}

The all2all training strategy, described in the Main Text, aims to leverage the semi-group structure of the solution operator of the time-dependent PDE \eqref{eq:pde} to scale-up the training data per trajectory. As shown in Figure \ref{fig:2} (d), we use every possible pair of snapshots, per trajectory, in the learning process leading to the loss function \eqref{eq:lfij}. It is instructive to compare this strategy with the \emph{vanilla} training strategy based on the loss function \eqref{eq:lfi}. As this strategy is applicable for any (time-dependent) operator learning algorithm, we study it for the CNO model \cite{CNO} here. To this end, we consider the NS-SL task and compare the all2all and vanilla strategies and plot the results in Figure \ref{fig:all2all_vanilla_sl} to observe that the all2all training strategy significantly outperforms the vanilla training strategy for this task. 

However, there is a caveat with the all2all strategy. It lies in the computational cost of training as the number of training pairs grows \emph{quadratically} with the number of available time snapshots at which the trajectory is sampled. One option to reduce this cost is to \emph{select a subset of snapshots} from within all available snapshots per trajectory and apply all2all training to this subset, bringing down the computational cost proportionately by the relative reduction in the cardinality of the selected subset. Yet, there is the possibility that by sampling too few snapshots, the overall error will increase. 

To investigate this trade-off, we consider the NS-PwC task and the CNO model. The data for this task is available in the time-interval [0,0.7], sampled at 14 time snapshots (excluding the initial time 0). Denoting the ith-snapshot by $t_i$ with $i=0,1,\ldots,14$, We select the following subsets of time snapshots,
\begin{itemize}
\item [$\mathcal{T}_{14}$:] Snapshots at $t_{0} = 0$ and $t_{14}=0.7$. The training only considers learning the map between initial datum and solution at final time $t_{14}$. Samples corresponding to identity function are also included.

\item [$\mathcal{T}_{7}$:] Snapshots at $t_0, t_7,t_{14}$
\item [$\mathcal{T}_{2}$:] Snapshots at $t_0,t_2,t_4,t_6,t_8,t_{10},t_{12},t_{14}$
\item [$\mathcal{T}_{1}$:] Snapshots at $t_{j}$, for all $0 \leq j \leq 14$
\end{itemize}
For each of the above subsets of time snapshots, all2all training is used leading to $3,6,36$ and $120$ training pairs per trajectory for $\mathcal{T}_{14},\mathcal{T}_{7},\mathcal{T}_{2},\mathcal{T}_{1}$, respectively. 

\begin{figure}
    \centering
    \includegraphics[width=\textwidth]{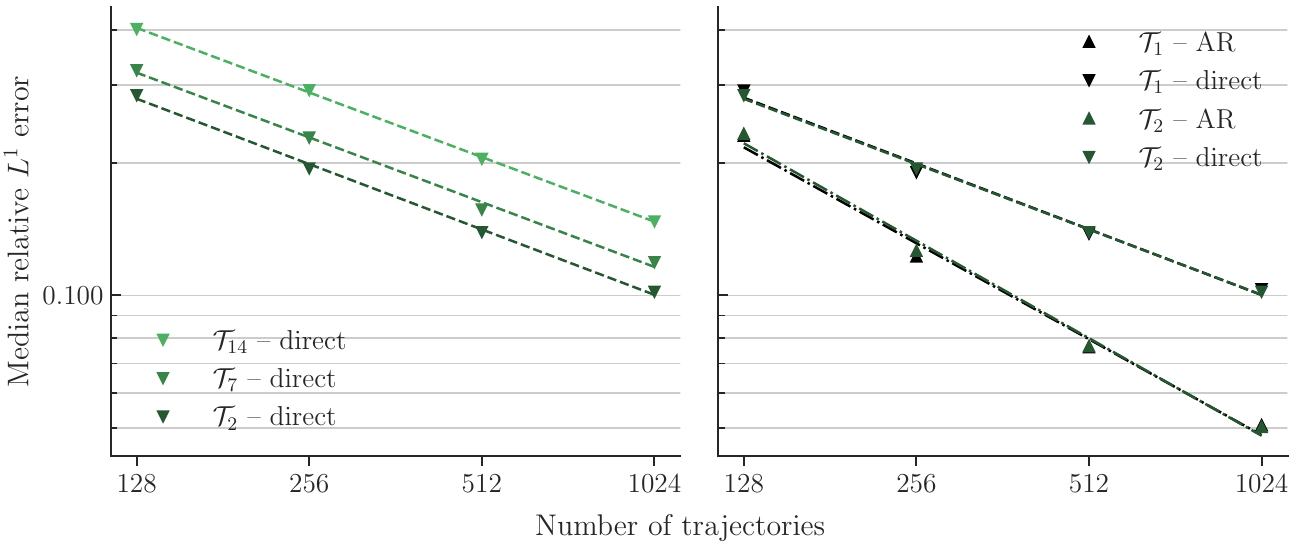}
    \caption{NS-PwC. Testing errors of the CNO models trained in an \textit{all2all} manner on different $\mathcal{T}_i$ trajectories. (Left) Errors from directly evaluating the trained models. Performance improves as denser trajectories are incorporated. (Right) Saturation effect observed. Adding denser trajectories no longer enhances performance, as the additional samples are statistically less significant.}
    \label{fig:all2all_pwc}
\end{figure}

In Figure \ref{fig:all2all_pwc}, we plot the test error vs. number of trajectories. From the left panel of this figure, we see that there is consistent gain in accuracy as a more dense sampling of the snapshots is performed. The models are monotonically more accurate as we go from $\mathcal{T}_{14}$ through $\mathcal{T}_{7}$ to $\mathcal{T}_2$. However, we also observe from Figure \ref{fig:all2all_pwc} that going beyond $\mathcal{T}_2$ to $\mathcal{T}_1$ does not yield any further decrease in test error as the difference between the newly added snapshots and the existing ones in $\mathcal{T}_2$ is not statistically significant enough to aid the training process. Moreover, by choosing $\mathcal{T}_2$ over $\mathcal{T}_1$, we reduce the computational cost of training by a factor of $3.3$. These considerations motivate us to a not too dense sampling strategy for pretraining (and finetuning) our foundation models.

\subsubsection{Direct. vs. Autoregressive Inference}
As mentioned in the Main Text, our time-conditioned models can either be directly evaluated at the time of interest, or an autoregressive rollout can be performed (see Equation \ref{eq:ar} of the Main Text). This can per se be of any form that the user wants, i.e. with homogeneous step-sizes in time, or with heterogeneous step-sizes in time. For simplicity, we only consider homogeneous autoregressive rollouts for \textsc{Poseidon}, scOT and FNO models, for the CNO models we find a slight performance boost with a heterogeneous rollout strategy.

\begin{figure}[htb]
    \centering
    \includegraphics[width=\textwidth]{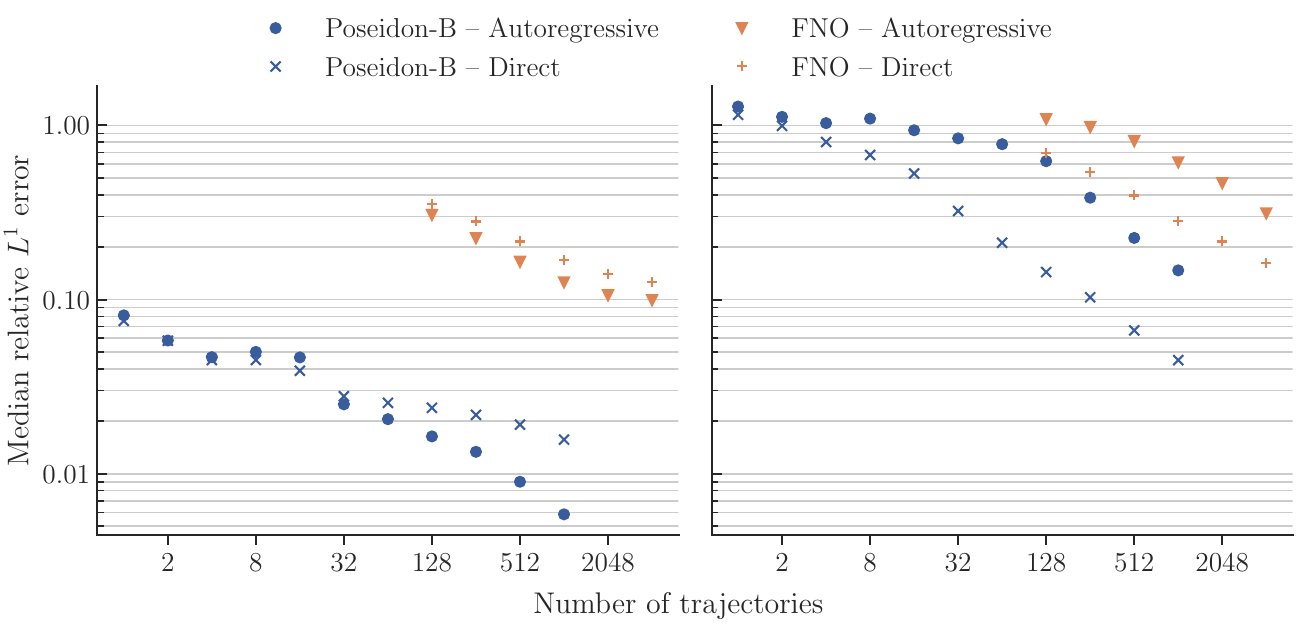}
    \caption{Homogenenous autoregressive rollout vs. direct lead-time input on NS-PwC (left) and Wave-Layer (right).}
    \label{fig:direct_vs_ar}
\end{figure}

Figure \ref{fig:direct_vs_ar} shows for the NS-PwC and the Wave-Layer downstream task how the error behaves when using direct or (homogeneous) autoregressive rollouts. We can directly see that it depends very much on the task at hand, as autoregressive rollout works better for the NS-PwC task, whereas direct lead-time input works better for Wave-Layer; this seems to be very dataset- and dynamics-dependent. We therefore choose the best strategy for each task which is listed in Table \ref{tab:performance_metrics}.

\subsubsection{Error Growth over Time for \textsc{Poseidon}-B}
\label{sec:tdep}
\begin{figure}[htb]
    \centering
    \includegraphics[width=0.6\textwidth]{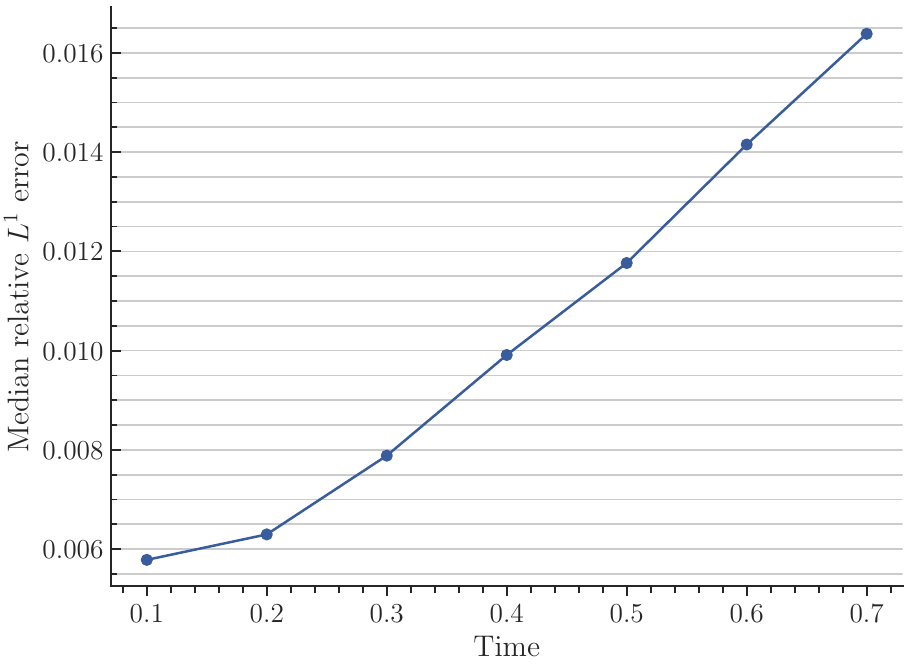}
    \caption{Error accumulation for autoregressive rollout of \textsc{Poseidon}-B finetuned on 128 trajectories of the NS-PwC dataset.}
    \label{fig:error_accumulation}
\end{figure}

Autoregressive inference can only work better than direct lead-time input when the error that accumulates at every step is smaller than the error obtained by direct lead time input. In Figure \ref{fig:error_accumulation}, we can directly see that the error scales \emph{sub-linearly} for the NS-PwC experiment and this is true in general for our downstream tasks. This leads to two observations. First, there is no blow-up (for instance exponential growth) of error in time with these models. Second, the fact that the error grows in time proves that it is harder to predict the solution at final time from initial data than predicting time-averaged quantities. In other words, the $L^\infty$-error in time will be greater than the $L^1$-error. This justifies our choice of evaluating different models at the final lead time of the underlying task. 

\begin{figure}
    \centering
    \begin{subfigure}[t]{0.49\textwidth}
        \centering
        \includegraphics[width=\textwidth]{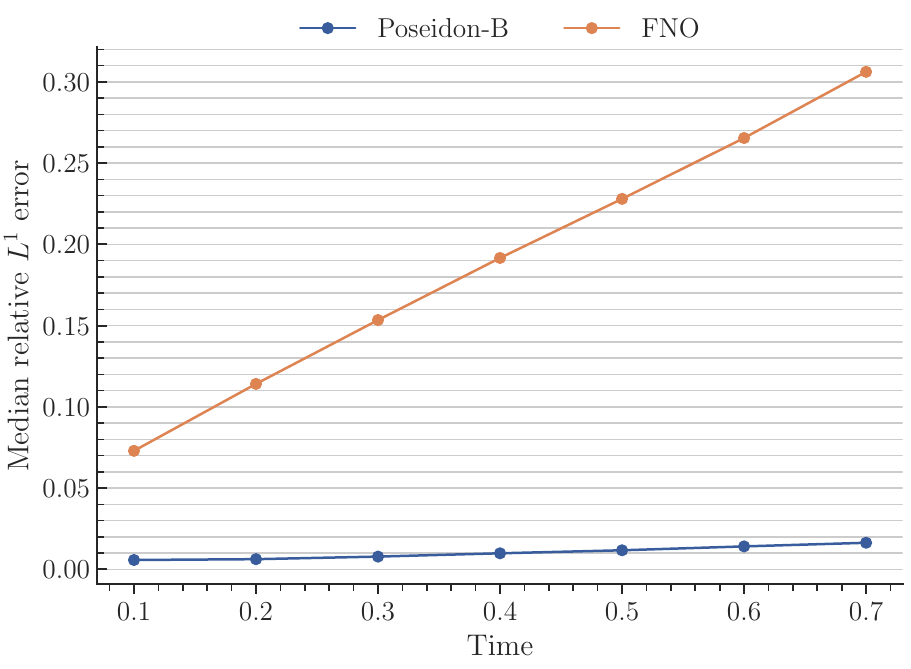}
    \end{subfigure}
    \hfill
    \begin{subfigure}[t]{0.49\textwidth}
        \centering
        \includegraphics[width=\textwidth]{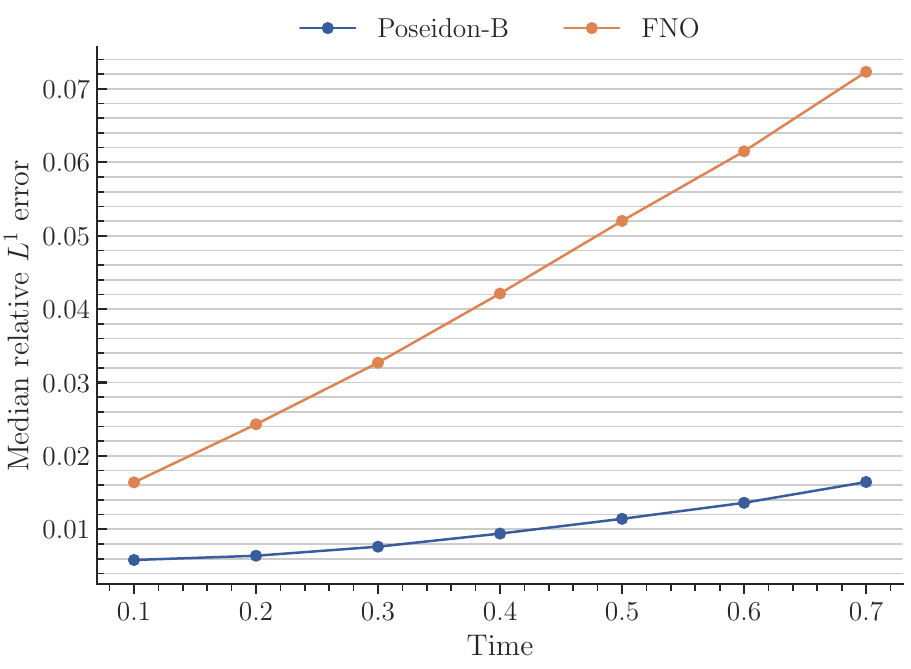}
    \end{subfigure}
    \caption{Error accumulation for the finetuned \textsc{Poseidon}-B and FNO for 128 training trajectories on NS-PwC (left) and NS-SL (right).}
    \label{fig:timeerr}
\end{figure}

To further demonstrate how \textsc{Poseidon} compares with FNO over time, we plot errors for the NS-PwC and NS-SL experiments as a function of time with both models in Figure \ref{fig:timeerr}. We observe from this figure that the difference in error between FNO and \textsc{Poseidon}-B actually grows over time and is the highest at the final time as FNO has much larger rate of error growth over time than \textsc{Poseidon}, justifying our decision to evaluate models with respect to error at the final time.

\begin{figure}[htb]
    \centering
    \includegraphics[width=0.75\textwidth]{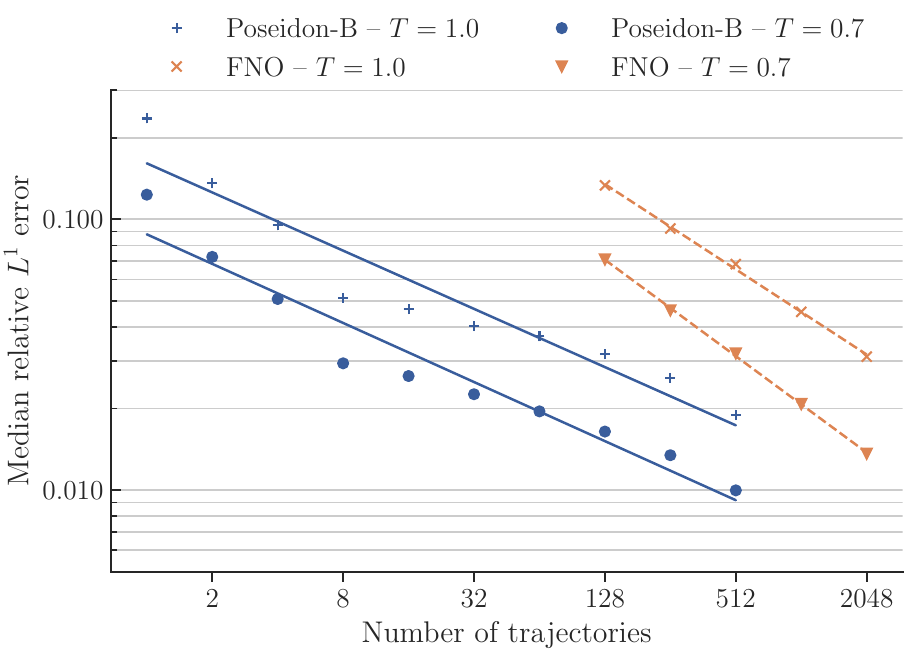}
    \caption{Out-of-distribution extrapolation in time for \textsc{Poseidon}-B and FNO on NS-SL up to $T=1$.}
    \label{fig:ood_time}
\end{figure}

\subsubsection{Out-of-distribution Time Extrapolation}
Here, we consider the NS-SL downstream task. As mentioned before, FNO (and other neural operators) were trained from scratch as well as \textsc{Poseidon} (and other foundation models) were finetuned to learn the solution up to a final lead time of $T=0.7$. We want to investigate how the \textsc{Poseidon} foundation model and the neural operator baseline (relatively) perform when we consider an \emph{out-of-distribution time extrapolation} at the downstream task level. To this end, in Figure \ref{fig:ood_time}, we plot the test errors, with respect to increasing number of task-specific trajectories, for both FNO and \textsc{Poseidon}-B, but evaluated at final times of $T=0.7$ and the extrapolated final time of $T=1.0$. A homogeneous autoregressive rollout is used in all cases. We observe from this figure that both \textsc{Poseidon}-B and FNO are worse at extrapolating in time than they are at predicting within the time-period that they have been trained on. In addition to significantly outperforming FNO at both time $T=0.7$ and at the extrapolated time of $T=1.0$, \textsc{Poseidon}-B in fact performs relatively better at out-of-distribution than FNO. It is best seen from the {\bf EG} metric \eqref{eq:met}, where \textsc{Poseidon}'s {\bf EG} $\approx 20$ for time $T=0.7$ is improved to  {\bf EG} $\approx 30$ for time $T=1.0$. This gain can be attributed to the fact that during pretraining, \textsc{Poseidon} models have been trained for a longer time horizon. 

\subsubsection{Generalization of \textsc{Poseidon} with respect to Changing PDE Parameters}
\begin{figure}
    \centering
        \includegraphics[width=0.75\textwidth]{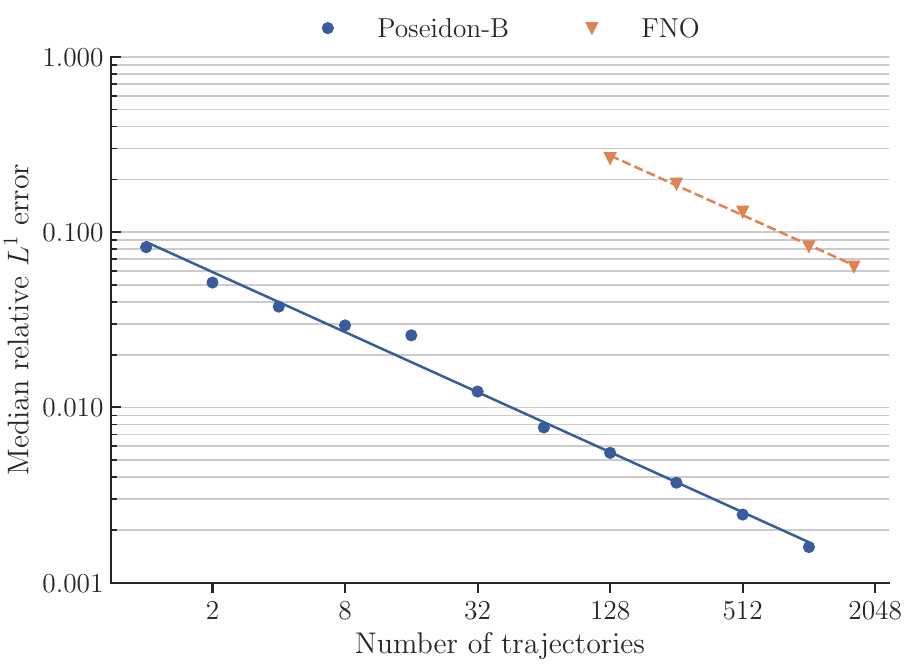}
    \caption{Error for the NS-PwC downstream task, but with viscosity $\nu = 4 \times 10^{-3}$ (on all modes) instead of $4 \times 10^{-4}$ applied only on high-enough Fourier modes to simulate the inviscid limit}
    \label{fig:diffvisc}
\end{figure}

Several of our downstream tasks such as GCE-RT, Wave-Layer, Wave-Gauss and Helmholtz involve operators that map the coefficient in the PDE to its solution. This setup is very different from the pretraining dataset where the underlying solution operators only map the initial data to solutions at later times and there is no PDE coefficient that is encountered. Nevertheless, from Tables \ref{tab:1} and \ref{tab:full_res}, we observe that the \textsc{Poseidon} models generalize very well to these very different setups for the operators for downstream tasks. To further test the ability of \textsc{Poseidon} to generalize for different PDE parameters, we consider the Navier-Stokes Equations (\eqref{eq:NS}) with a viscosity coefficient $\nu=4\times 10^{-3}$. The ground truth data is generated using the Azeban spectral hyper viscosity solver \cite{azeban}. This new viscosity coefficient is very different from the setup of the pretraining data and downstream tasks considered so far as in all of them, only a hyperviscosity of $4 \times 10^{-4}$ was applied to high-enough Fourier modes in order to model the incompressible Euler equations with zero viscosity. In this \emph{new} task, the initial conditions are identical to the NS-PwC downstream task. We see from Figure \ref{fig:diffvisc} that Poseidon-B generalizes very well to this new viscosity coefficient and outperforms FNO readily, in terms of both sample efficiency and accuracy. In particular, the AG and EG scores of Poseidon-B are $EG=925.5$ and $AG=47.5$, which are completely comparable to (even better than) the scores of $EG=1024$ and $AG=19.7$ (see Table \ref{tab:full_res} for the original NS-PwC task). Taken together with other downstream tasks involving different PDE coefficients, this experiment clearly demonstrates the ability of \textsc{Poseidon} to generalize to different PDE parameters via finetuning.

\subsubsection{\textsc{Poseidon} Evaluated on Different Grids}
\begin{figure}[htb]
    \centering
    \includegraphics[width=0.75\textwidth]{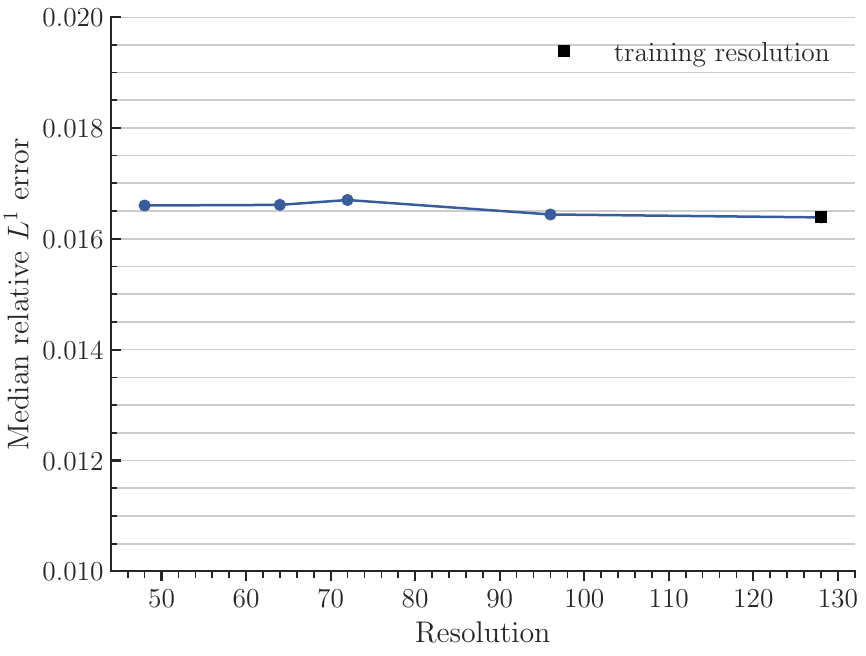}
    \caption{Test performance of \textsc{Poseidon}-B finetuned on 128 trajectories of the NS-PwC dataset for multiple resolutions.}
    \label{fig:resolutions}
\end{figure}

As \textsc{Poseidon} is based on an operator transformer (scOT), it can be evaluted on grid resolutions, different from the underlying computational grid. Following \cite{Reno}, we can simply downsample (upsample) the input function from the given grid to the computational grid, process the input with \textsc{Poseidon} and upsample (downsample) the output from the computational grid to the given grid resolution. We perform this evaluation of \textsc{Poseidon}-B on multiple grid resolutions for the NS-PwC task and present the result in Figure \ref{fig:resolutions} to observe that the test error is (approximately) invariant to the grid resolution. 

\subsubsection{Robustness of Poseidon with respect to Noise}
\begin{figure}
    \centering
    \includegraphics[width=0.75\textwidth]{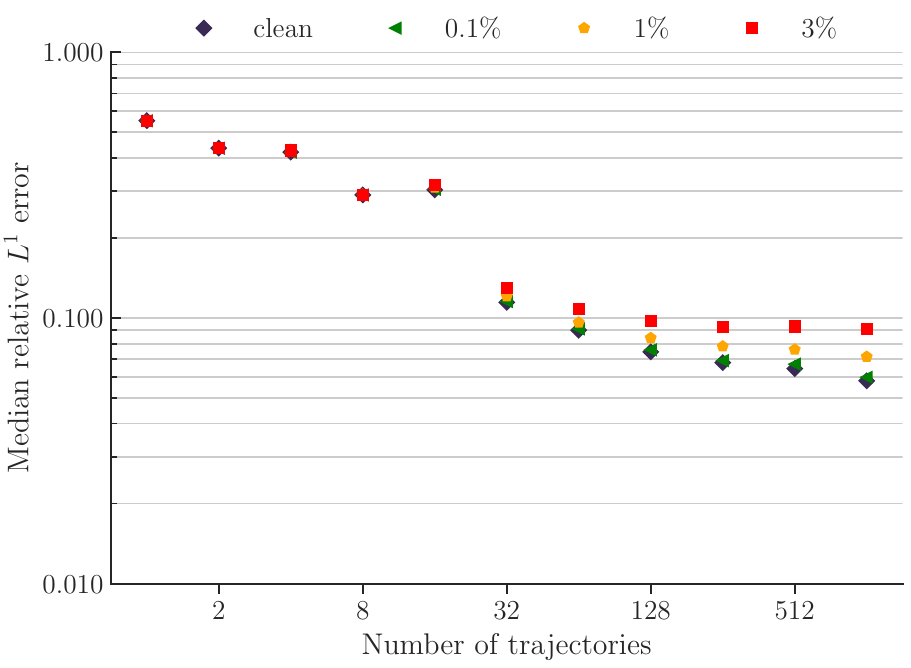}
    \caption{Effect of injecting Gaussian noise in the initial condition on CE-RPUI (before normalizing the data; normalization constants are as before) with \textsc{Poseidon}-L.}
    \label{fig:noiserobust}
\end{figure}
To study how robust \textsc{Poseidon} is to noise, we consider the downstream CE-RPUI task and at inference time, we add Gaussian noise to the inputs (initial conditions) at different noise-to-signal ratios (NSRs) of $0.1\%$, $1\%$ and $3\%$ respectively. The resulting errors, computed with respect to a Ground Truth where the outputs are not noisy, for varying numbers of training trajectories, are shown in Figure \ref{fig:noiserobust}. The errors in the zero noise (clean) case are also shown in this Figure. We observe from this figure that \textsc{Poseidon}-L's performance is robust to input noise and the error does not grow significantly even when the noise level is an appreciable $3\%$, demonstrating the robustness of this foundation model with respect to noise.

\subsubsection{Histograms of Errors for Different Tasks}
\label{sec:histogram}
\begin{figure}[p]
    \centering
    \includegraphics[width=\textwidth]{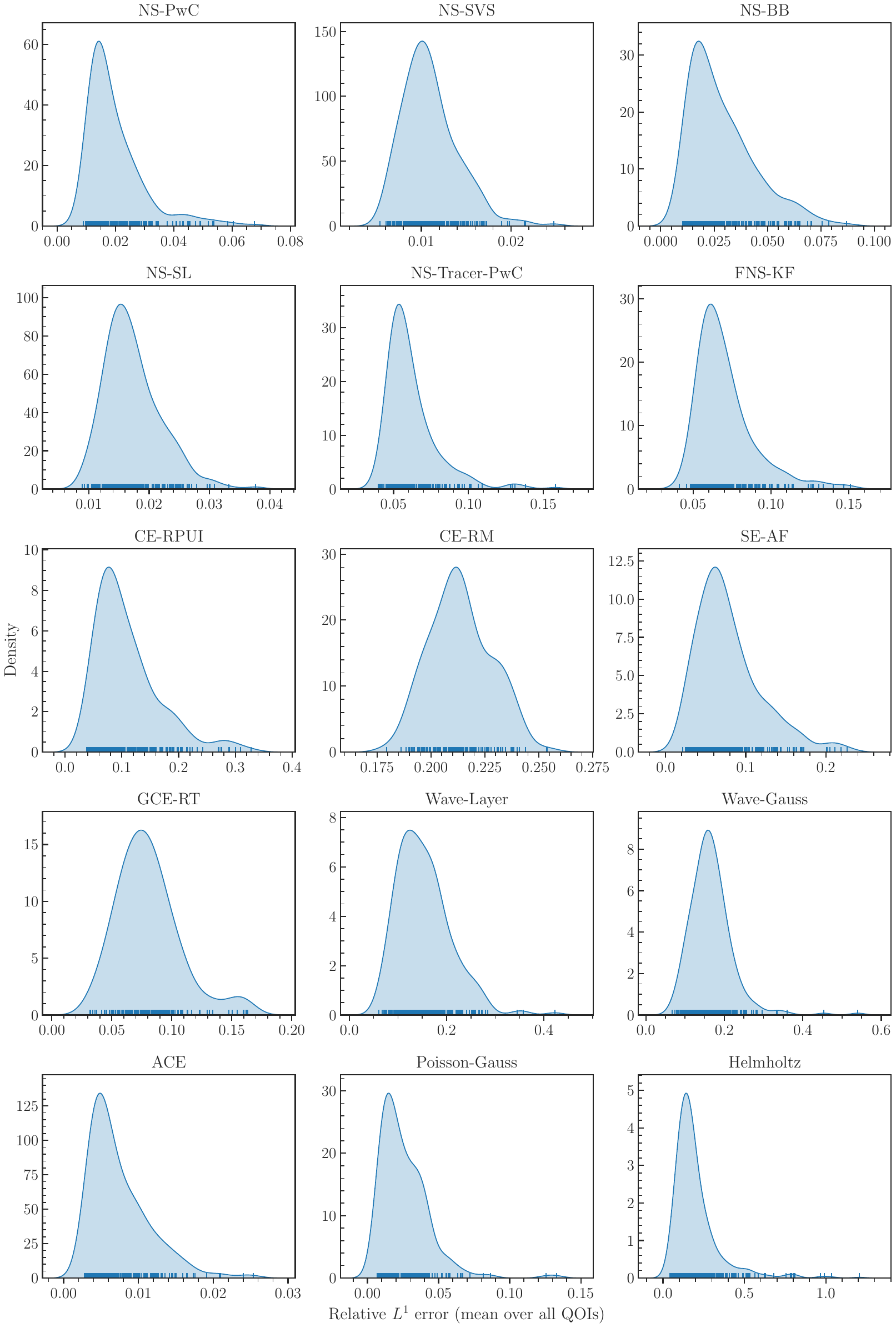}
    \caption{Error distribution of \textsc{Poseidon}-B finetuned on all downstream tasks (for 128 trajectories in the time-dependent, and 512 in the time-independent case). The kernel density estimate is done over the mean of all functions/quantities of interest.}
    \label{fig:error_histograms}
\end{figure}

\begin{figure}[h!]
    \centering
    \begin{subfigure}[t]{\textwidth}
        \centering
        \includegraphics[width=0.7\textwidth]{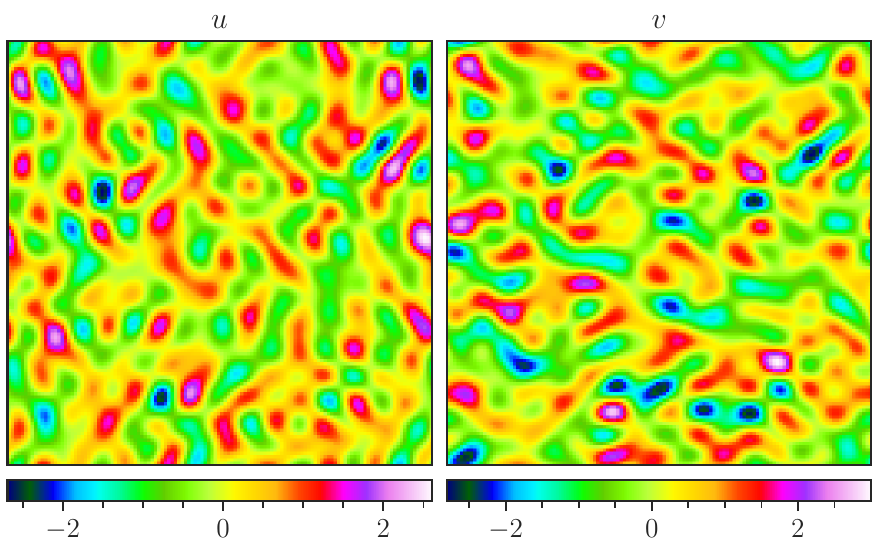}
        \caption{Inputs: horizontal velocity $u$ and vertical velocity $v$.}
    \end{subfigure}
    \hfill
    \begin{subfigure}[b]{\textwidth}
        \centering
        \includegraphics[width=0.7\textwidth]{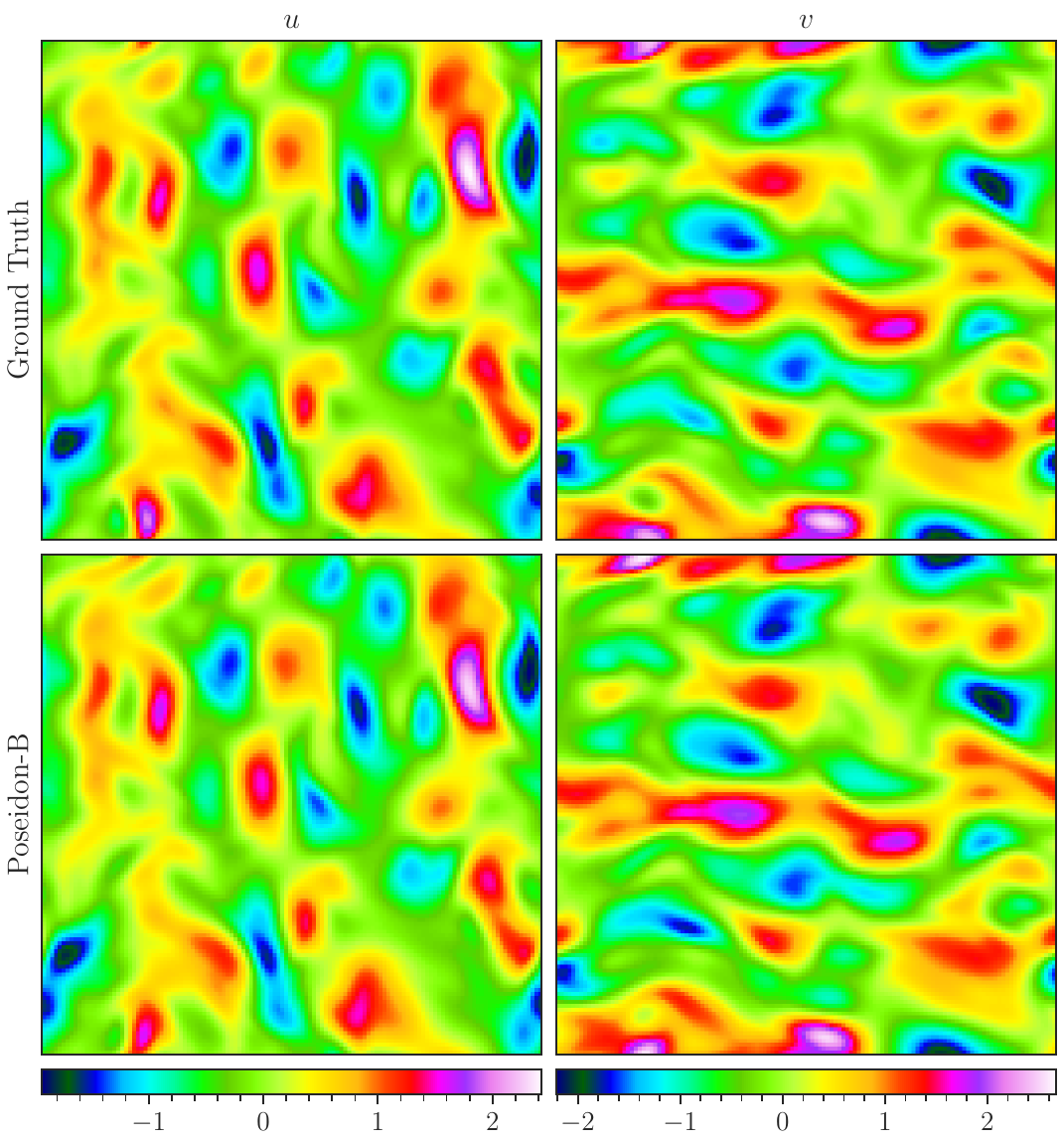}
        \caption{(Top) Ground truth. (Bottom) Samples predicted by \textsc{Poseidon}-B at $T = 1$.}
    \end{subfigure}
    \caption{NS-Sines. Visualization of a random sample.}
    \label{fig:ns_sines}
\end{figure}

\begin{figure}[p]
    \centering
    \begin{subfigure}[t]{\textwidth}
        \centering
        \includegraphics[width=0.7\textwidth]{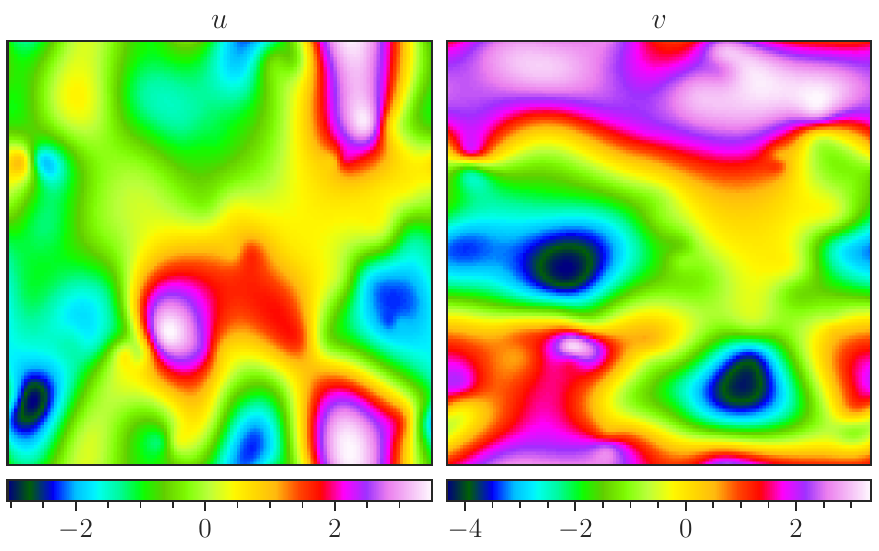}
        \caption{Inputs: horizontal velocity $u$ and vertical velocity $v$.}
    \end{subfigure}
    \hfill
    \begin{subfigure}[b]{\textwidth}
        \centering
        \includegraphics[width=0.7\textwidth]{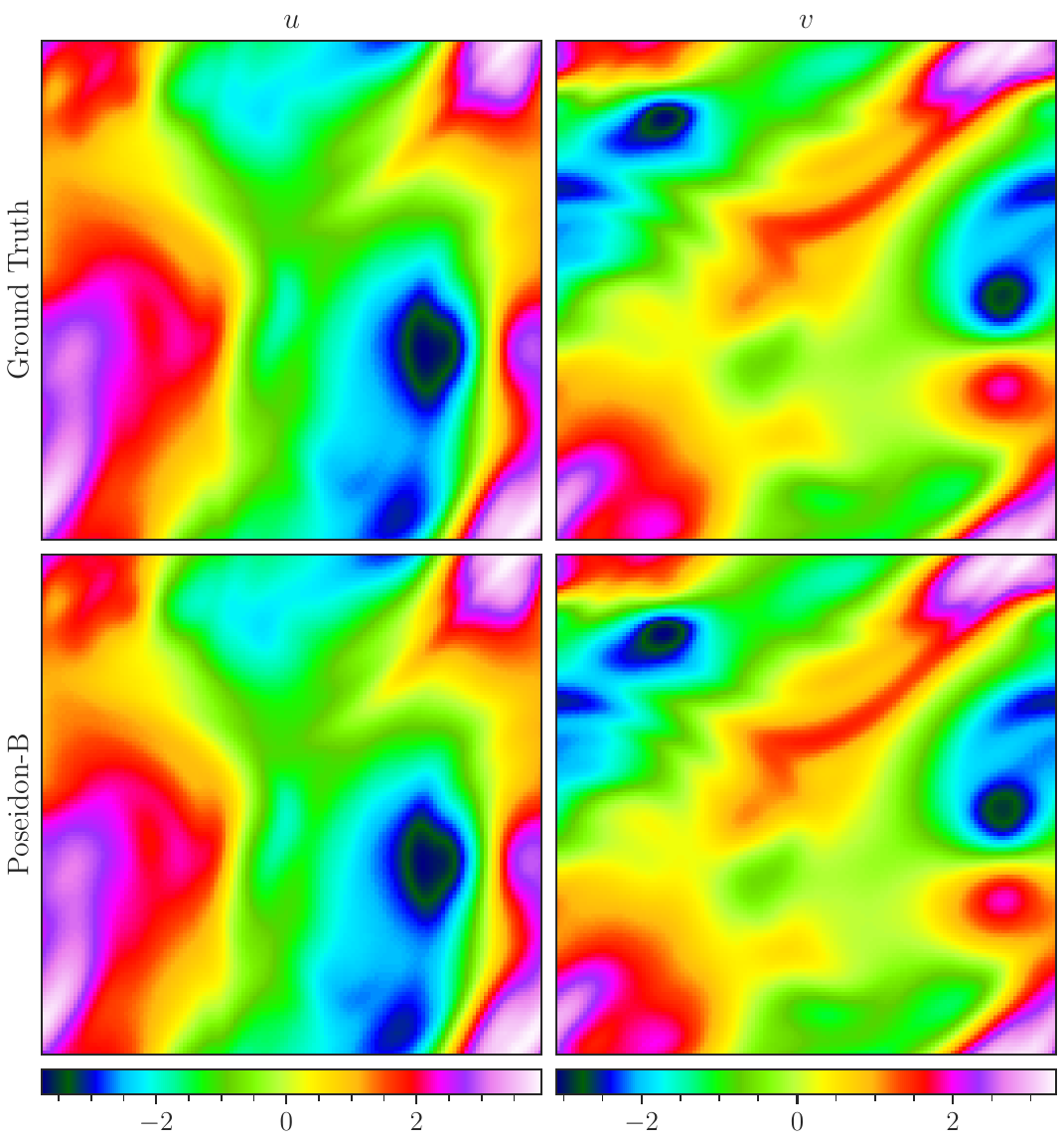}
        \caption{(Top) Ground truth. (Bottom) Samples predicted by \textsc{Poseidon}-B at $T = 1$.}
    \end{subfigure}
    \caption{NS-Gauss. Visualization of a random sample.}
    \label{fig:ns_gaussians}
\end{figure}

\begin{figure}[p]
    \centering
    \begin{subfigure}[t]{\textwidth}
        \centering
        \includegraphics[width=\textwidth]{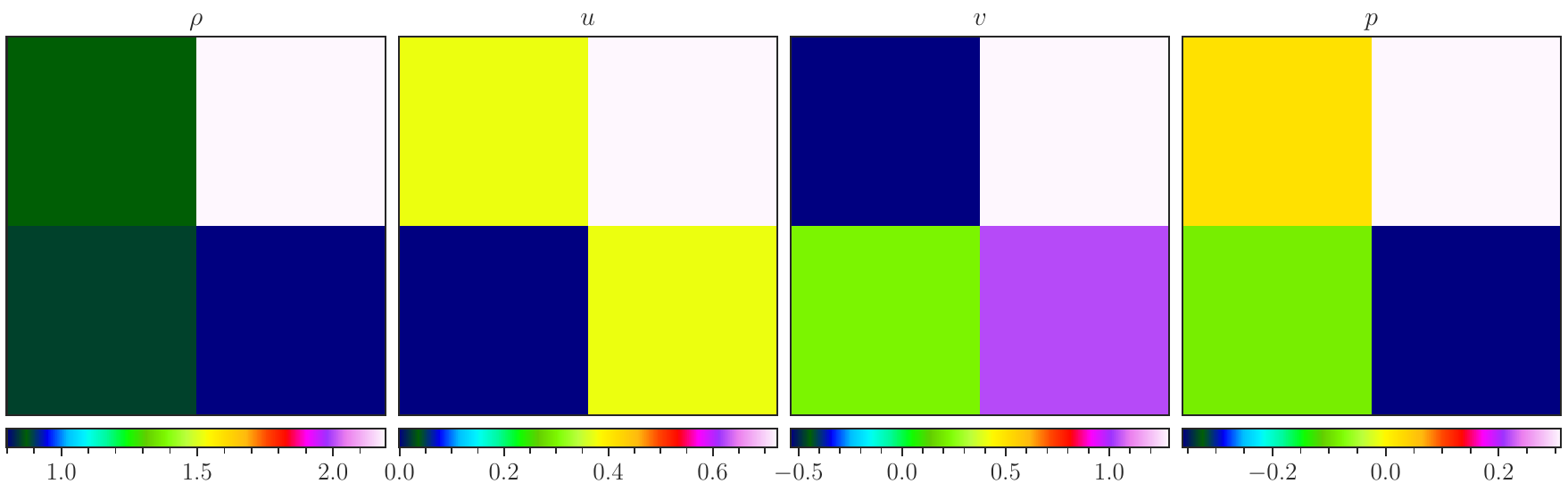}
        \caption{Inputs: density $\rho$, horizontal velocity $u$, vertical velocity $v$ and pressure $p$.}
    \end{subfigure}
    \hfill
    \begin{subfigure}[b]{\textwidth}
        \centering
        \includegraphics[width=\textwidth]{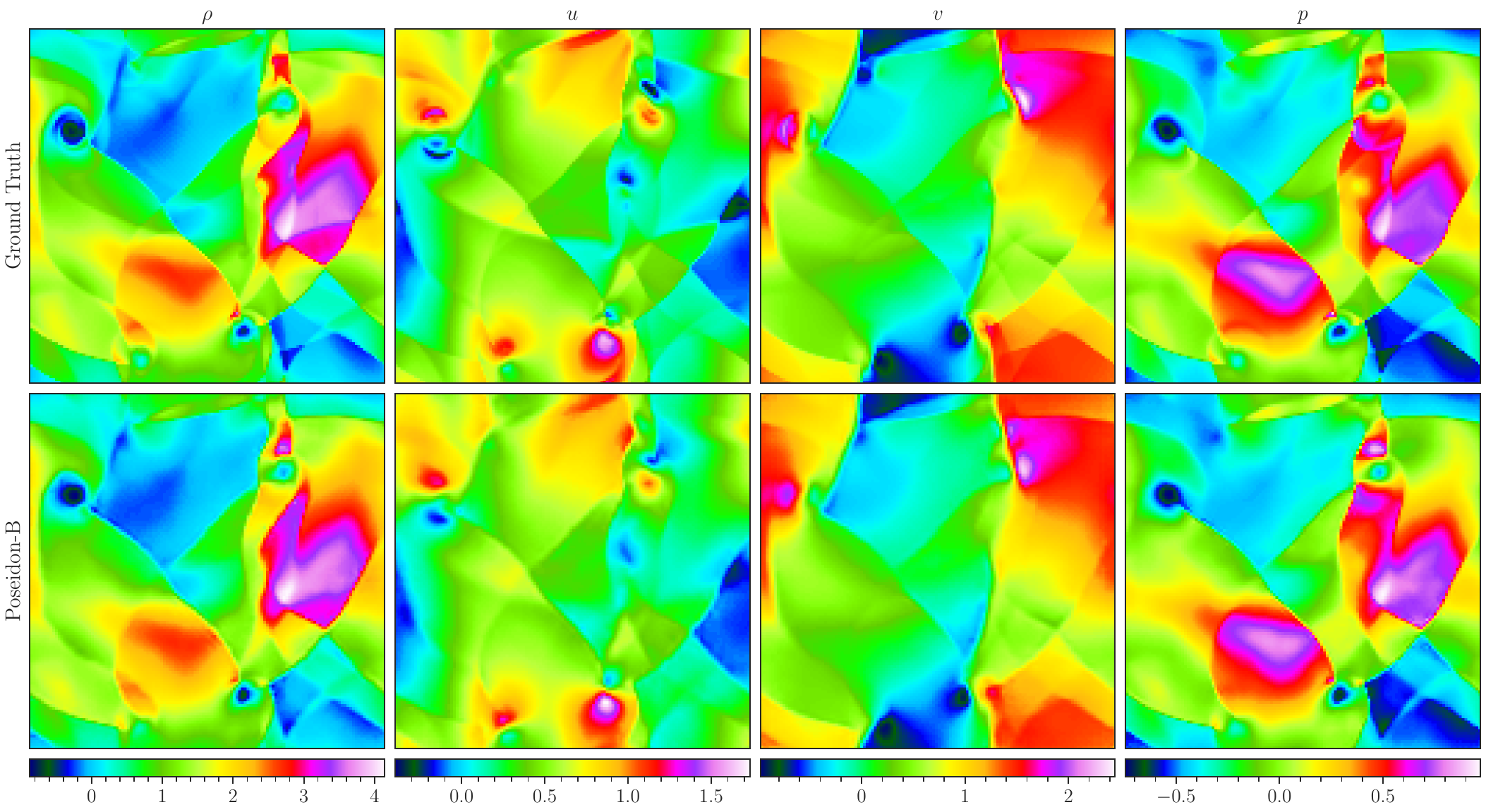}
        \caption{(Top) Ground truth. (Bottom) Samples predicted by \textsc{Poseidon}-B at $T = 1$.}
    \end{subfigure}
    \caption{CE-RP. Visualization of a random sample.}
    \label{fig:ce_rp}
\end{figure}

\begin{figure}[p]
    \centering
    \begin{subfigure}[t]{\textwidth}
        \centering
        \includegraphics[width=\textwidth]{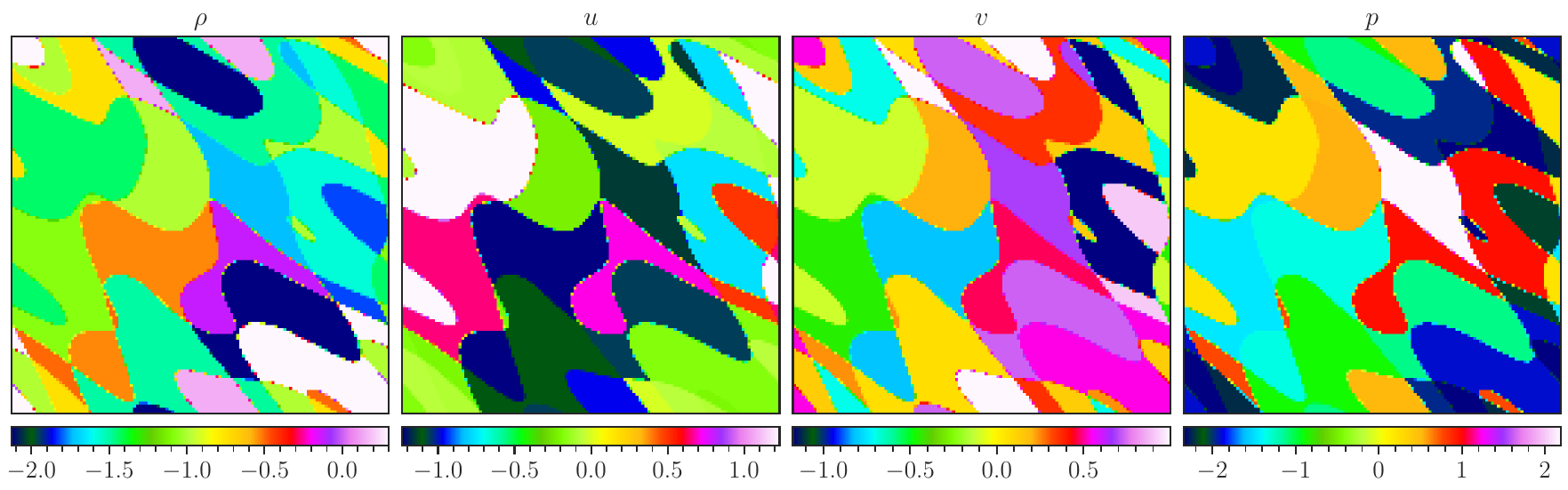}
        \caption{Inputs: density $\rho$, horizontal velocity $u$, vertical velocity $v$ and pressure $p$.}
    \end{subfigure}
    \hfill
    \begin{subfigure}[b]{\textwidth}
        \centering
        \includegraphics[width=\textwidth]{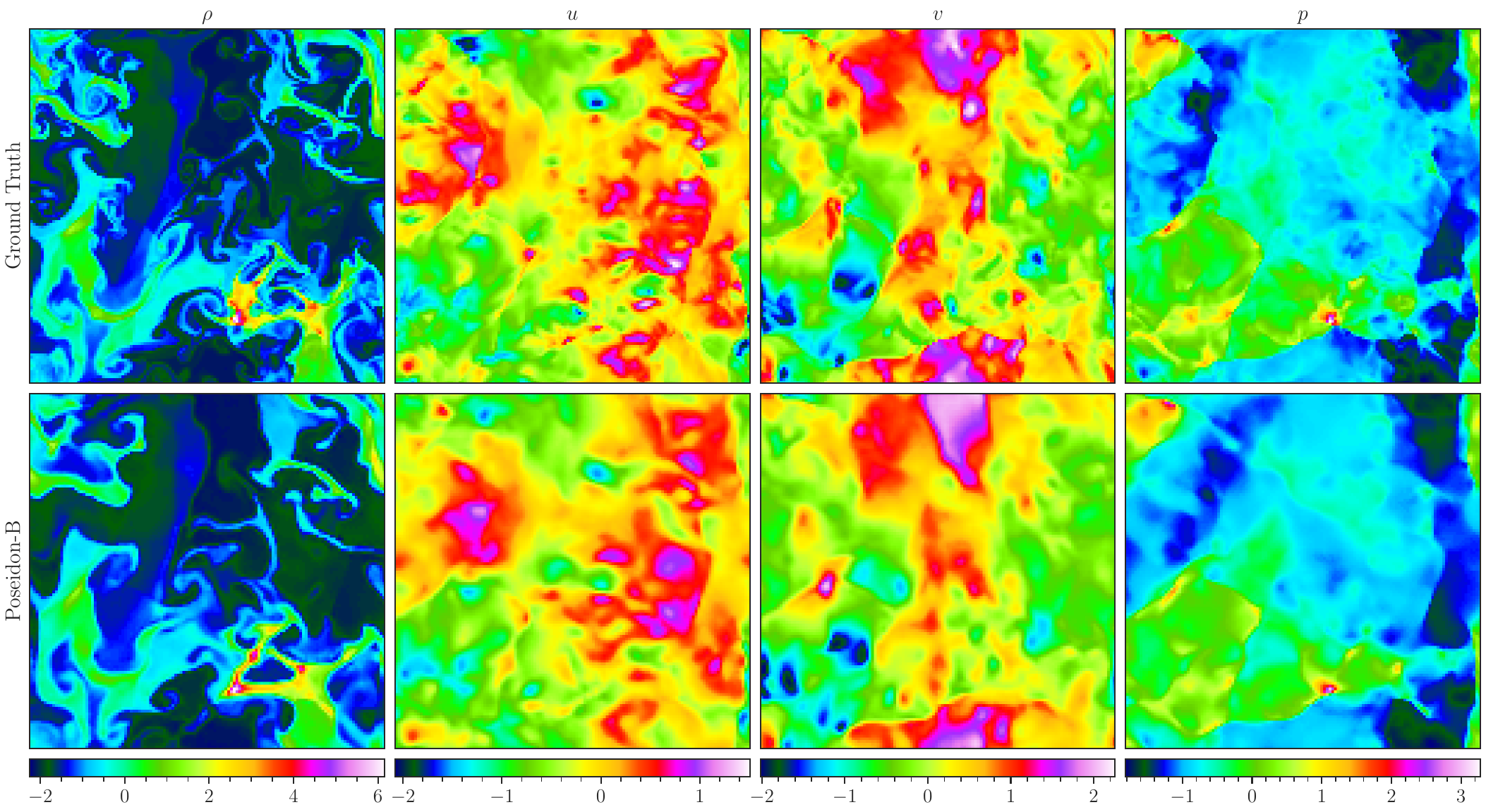}
        \caption{(Top) Ground truth. (Bottom) Samples predicted by \textsc{Poseidon}-B at $T = 1$.}
    \end{subfigure}
    \caption{CE-CRP. Visualization of a random sample.}
    \label{fig:ce_crp}
\end{figure}

\begin{figure}[p]
    \centering
    \begin{subfigure}[t]{\textwidth}
        \centering
        \includegraphics[width=\textwidth]{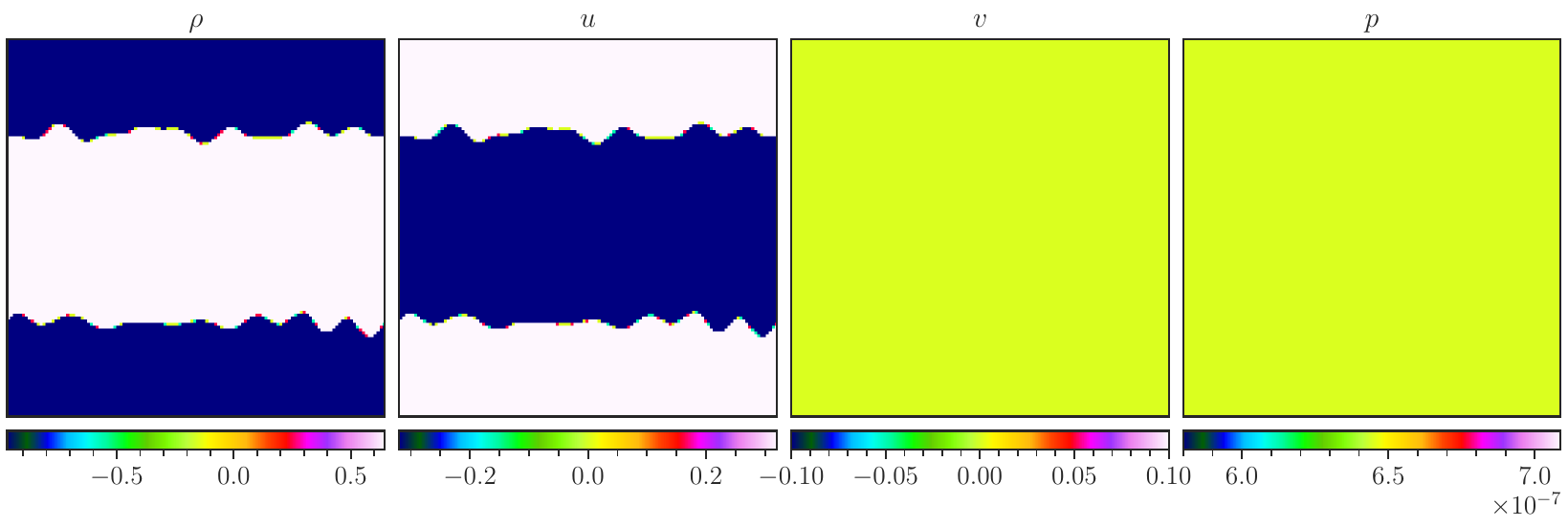}
        \caption{Inputs: density $\rho$, horizontal velocity $u$, vertical velocity $v$ and pressure $p$.}
    \end{subfigure}
    \hfill
    \begin{subfigure}[b]{\textwidth}
        \centering
        \includegraphics[width=\textwidth]{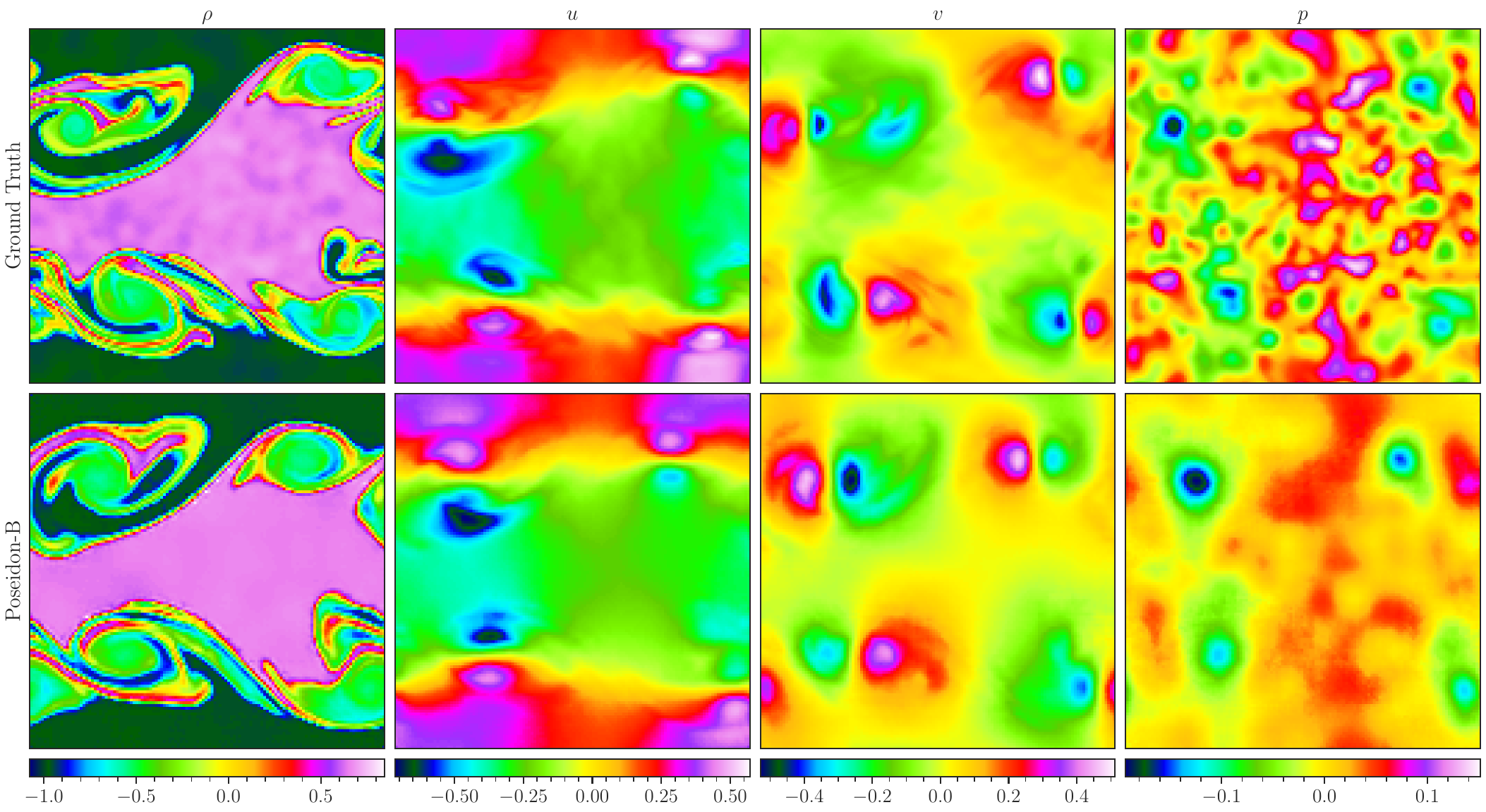}
        \caption{(Top) Ground truth. (Bottom) Samples predicted by \textsc{Poseidon}-B at $T = 1$.}
    \end{subfigure}
    \caption{CE-KH. Visualization of a random sample.}
    \label{fig:ce_kh}
\end{figure}

\begin{figure}[p]
    \centering
    \begin{subfigure}[t]{\textwidth}
        \centering
        \includegraphics[width=\textwidth]{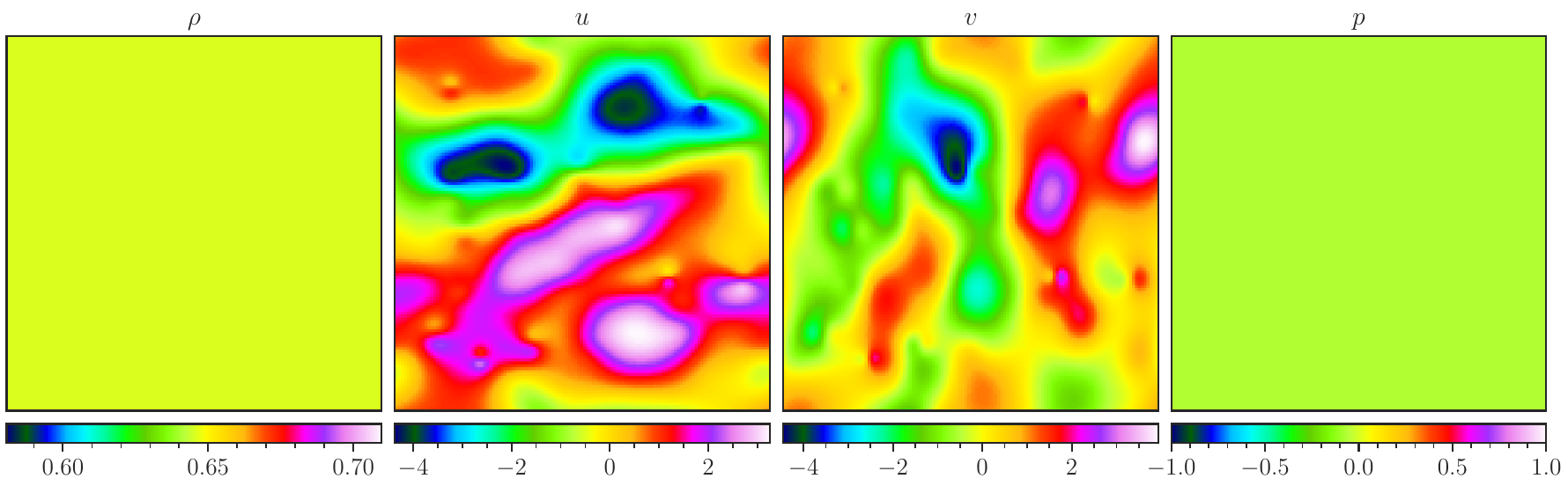}
        \caption{Inputs: density $\rho$, horizontal velocity $u$, vertical velocity $v$ and pressure $p$.}
    \end{subfigure}
    \hfill
    \begin{subfigure}[b]{\textwidth}
        \centering
        \includegraphics[width=\textwidth]{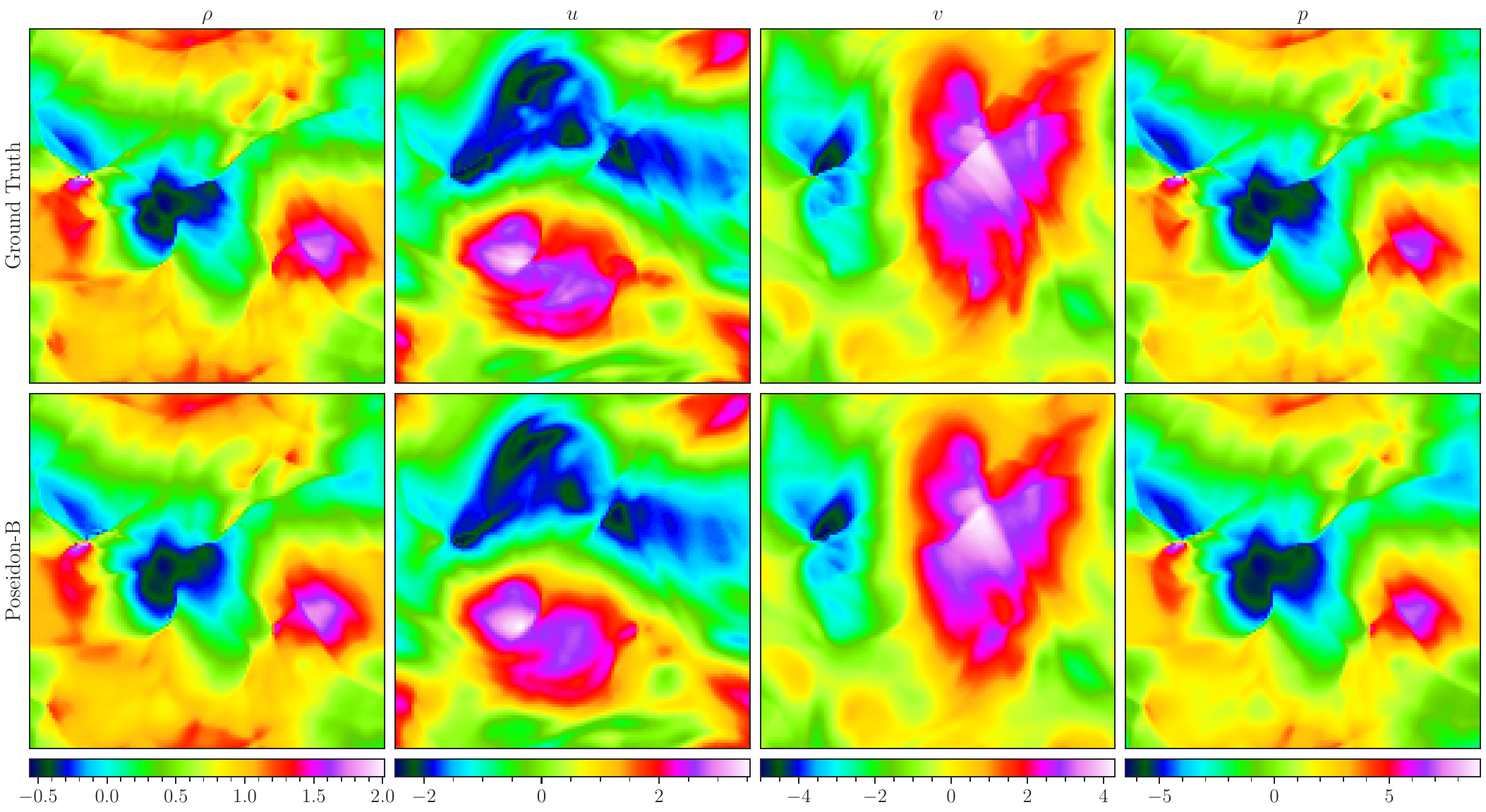}
        \caption{(Top) Ground truth. (Bottom) Samples predicted by \textsc{Poseidon}-B at $T = 1$.}
    \end{subfigure}
    \caption{CE-Gauss. Visualization of a random sample.}
    \label{fig:ce_gaussians}
\end{figure}

\begin{figure}[p]
    \centering
    \begin{subfigure}[t]{\textwidth}
        \centering
        \includegraphics[width=0.55\textwidth]{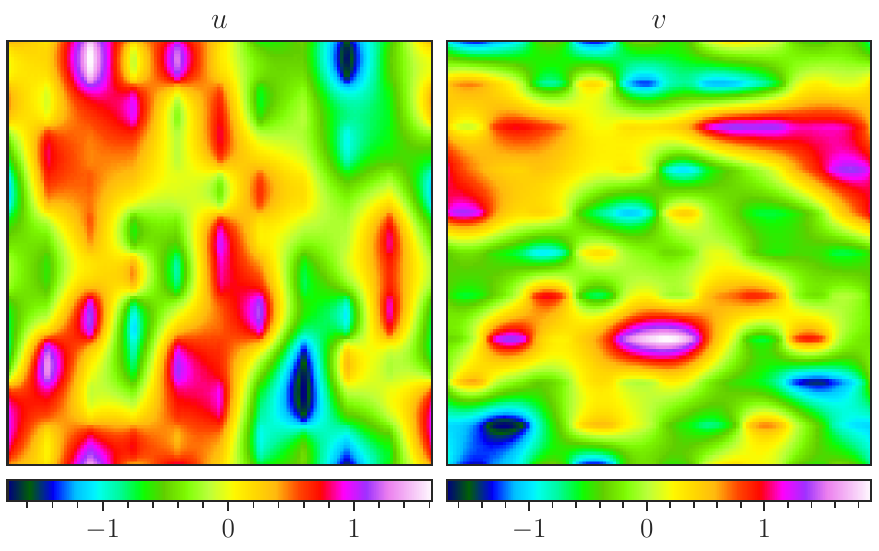}
        \caption{Inputs: horizontal velocity $u$ and vertical velocity $v$.}
    \end{subfigure}
    \hfill
    \begin{subfigure}[b]{\textwidth}
        \centering
        \includegraphics[width=0.55\textwidth]{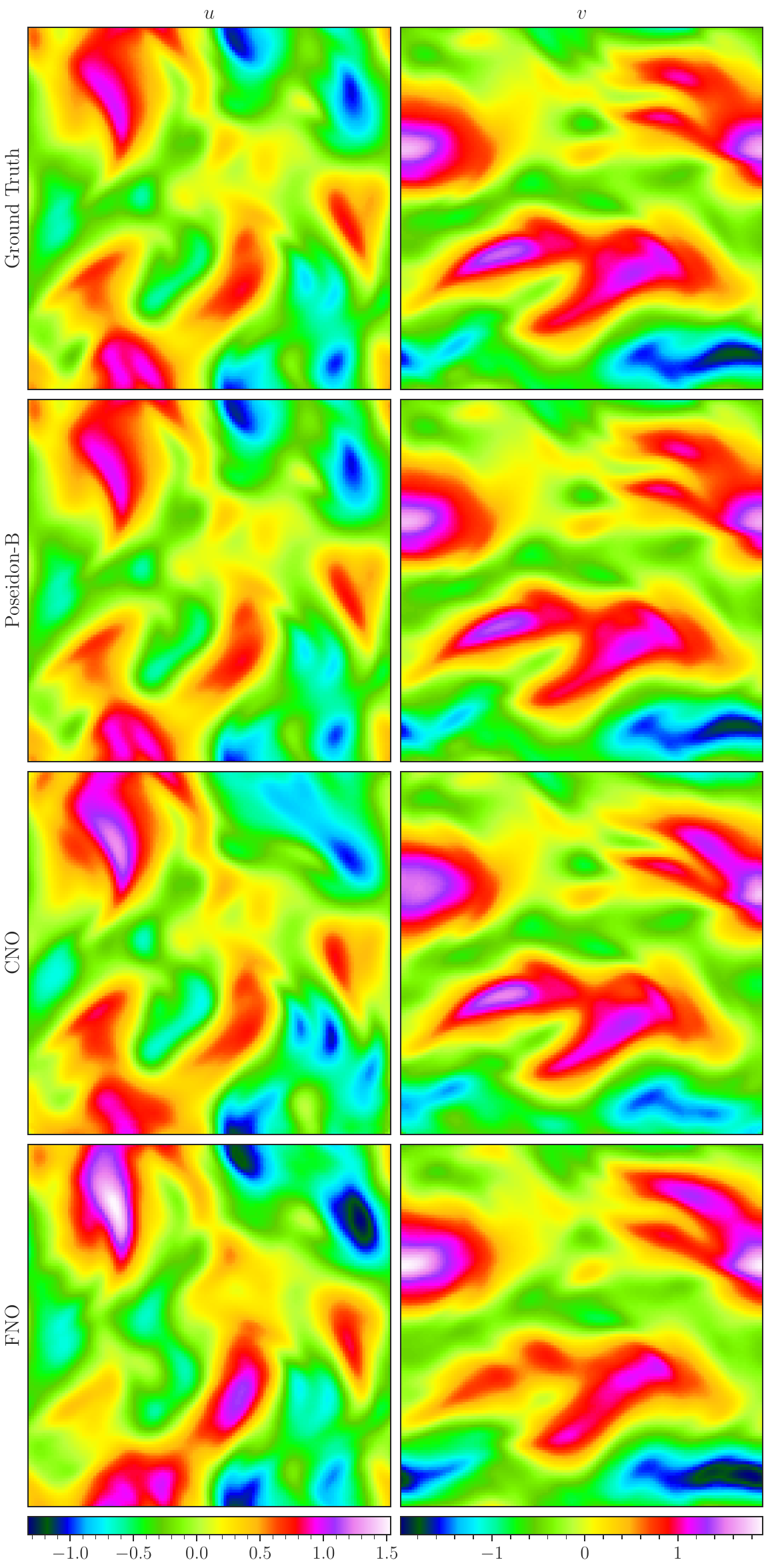}
        \caption{(Top) Ground truth. (From second row onwards) Samples predicted by the finetuned \textsc{Poseidon}-B, CNO, and FNO at $T = 0.7$.}
    \end{subfigure}
    \caption{NS-PwC. Visualization of a random sample.}
    \label{fig:ns_pwc}
\end{figure}

\begin{figure}[p]
    \centering
    \begin{subfigure}[t]{\textwidth}
        \centering
        \includegraphics[width=0.55\textwidth]{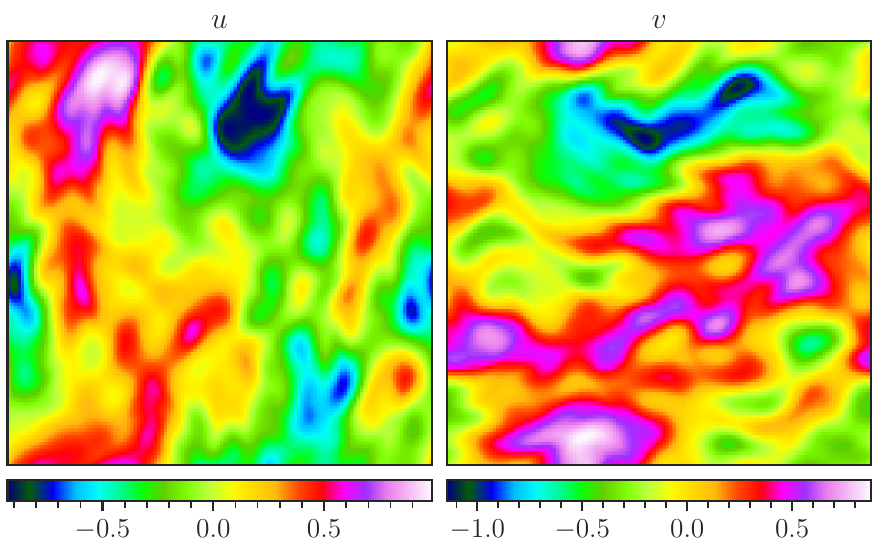}
        \caption{Inputs: horizontal velocity $u$ and vertical velocity $v$.}
    \end{subfigure}
    \hfill
    \begin{subfigure}[b]{\textwidth}
        \centering
        \includegraphics[width=0.55\textwidth]{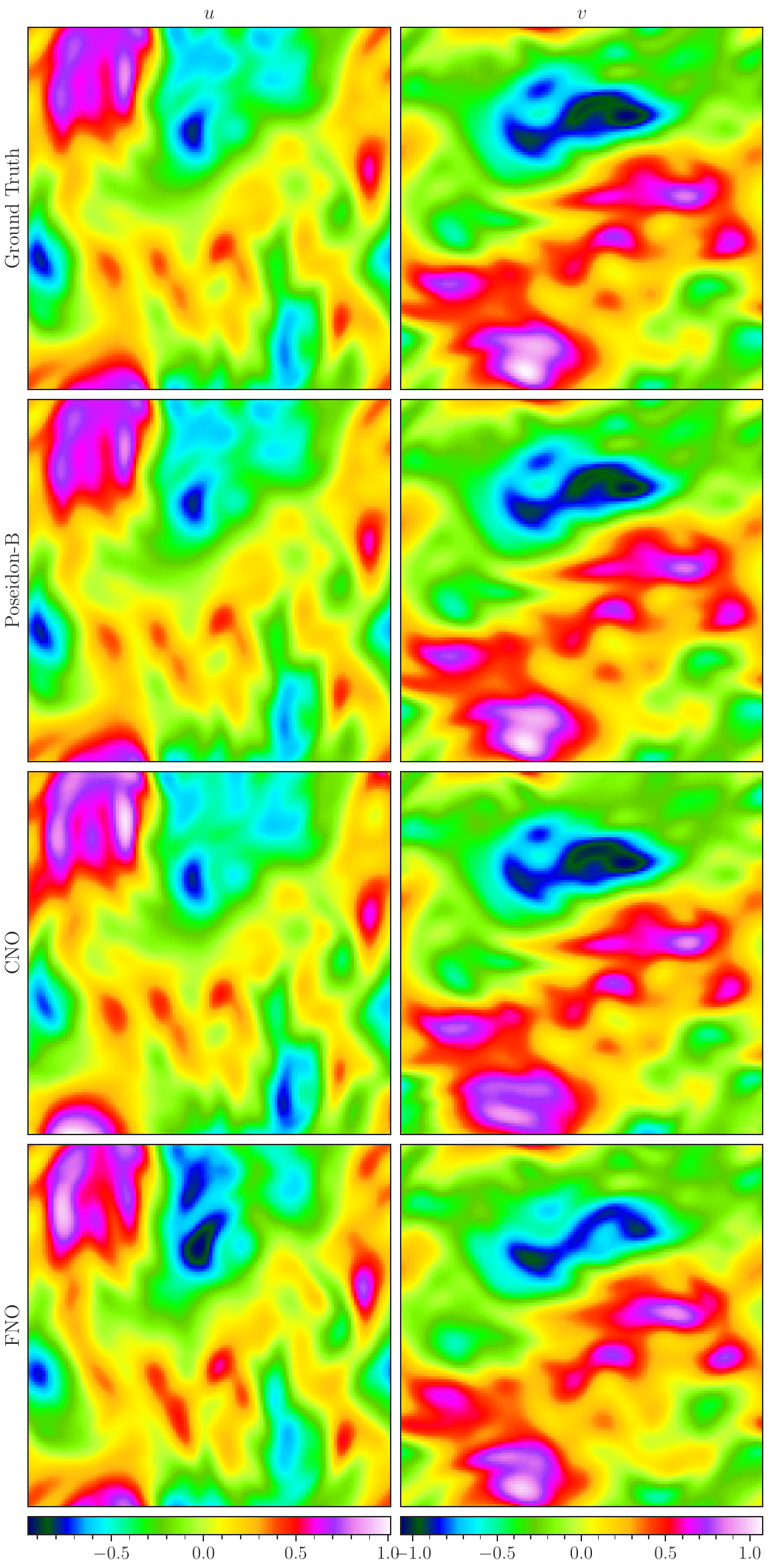}
        \caption{(Top) Ground truth. (From second row onwards) Samples predicted by the finetuned \textsc{Poseidon}-B, CNO, and FNO at $T = 0.7$.}
    \end{subfigure}
    \caption{NS-BB. Visualization of a random sample.}
    \label{fig:ns_bb}
\end{figure}

\begin{figure}[p]
    \centering
    \begin{subfigure}[t]{\textwidth}
        \centering
        \includegraphics[width=0.55\textwidth]{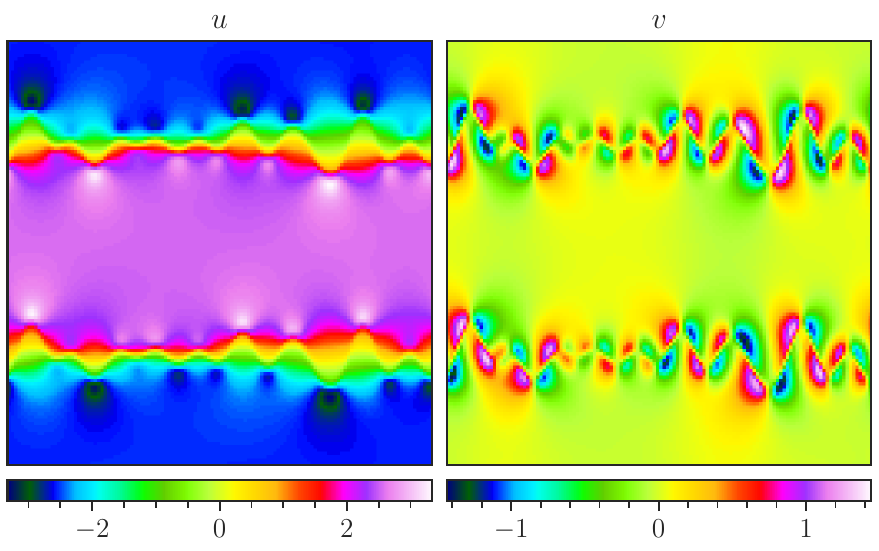}
        \caption{Inputs: horizontal velocity $u$ and vertical velocity $v$.}
    \end{subfigure}
    \hfill
    \begin{subfigure}[b]{\textwidth}
        \centering
        \includegraphics[width=0.55\textwidth]{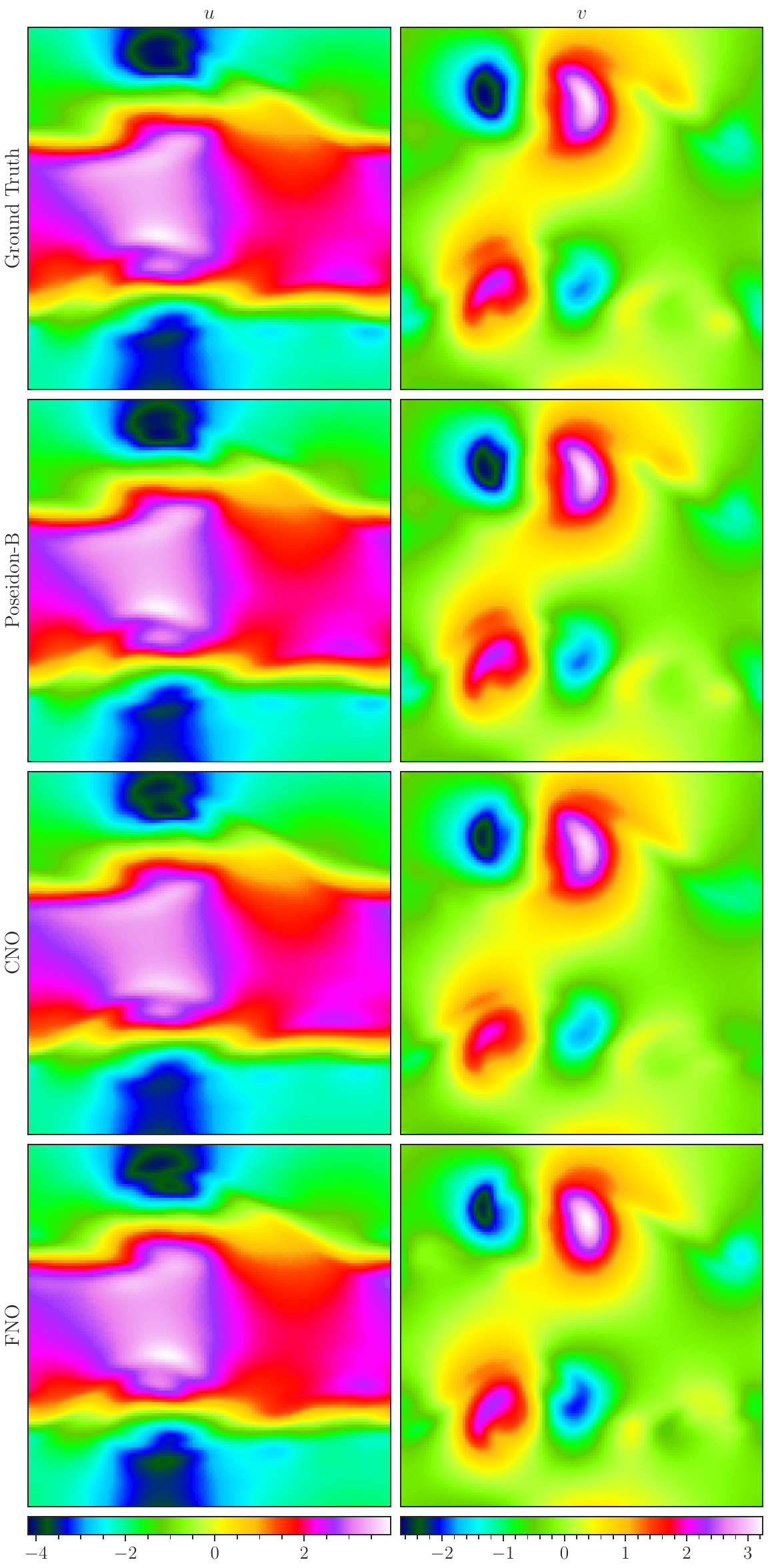}
        \caption{(Top) Ground truth. (From second row onwards) Samples predicted by the finetuned \textsc{Poseidon}-B, CNO, and FNO at $T = 0.7$.}
    \end{subfigure}
    \caption{NS-SL. Visualization of a random sample.}
    \label{fig:ns_sl}
\end{figure}

\begin{figure}[p]
    \centering
    \begin{subfigure}[t]{\textwidth}
        \centering
        \includegraphics[width=0.55\textwidth]{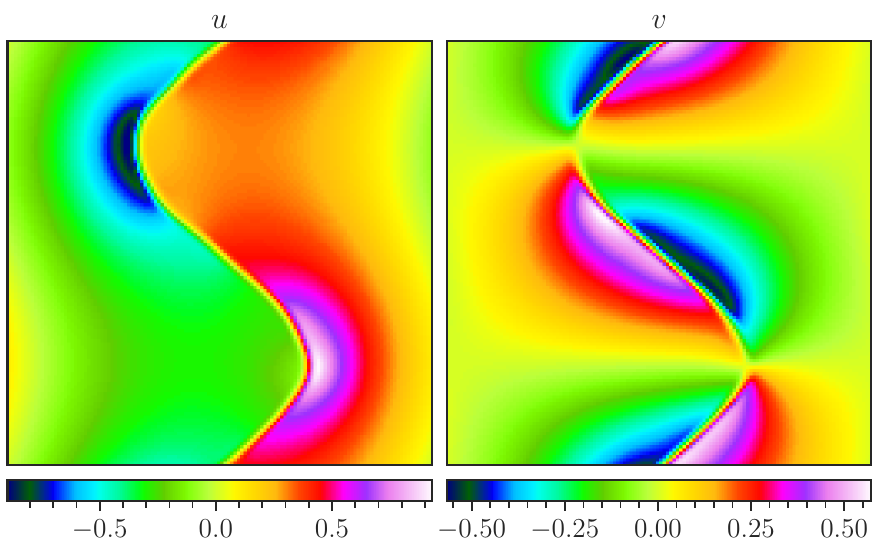}
        \caption{Inputs: horizontal velocity $u$ and vertical velocity $v$.}
    \end{subfigure}
    \hfill
    \begin{subfigure}[b]{\textwidth}
        \centering
        \includegraphics[width=0.55\textwidth]{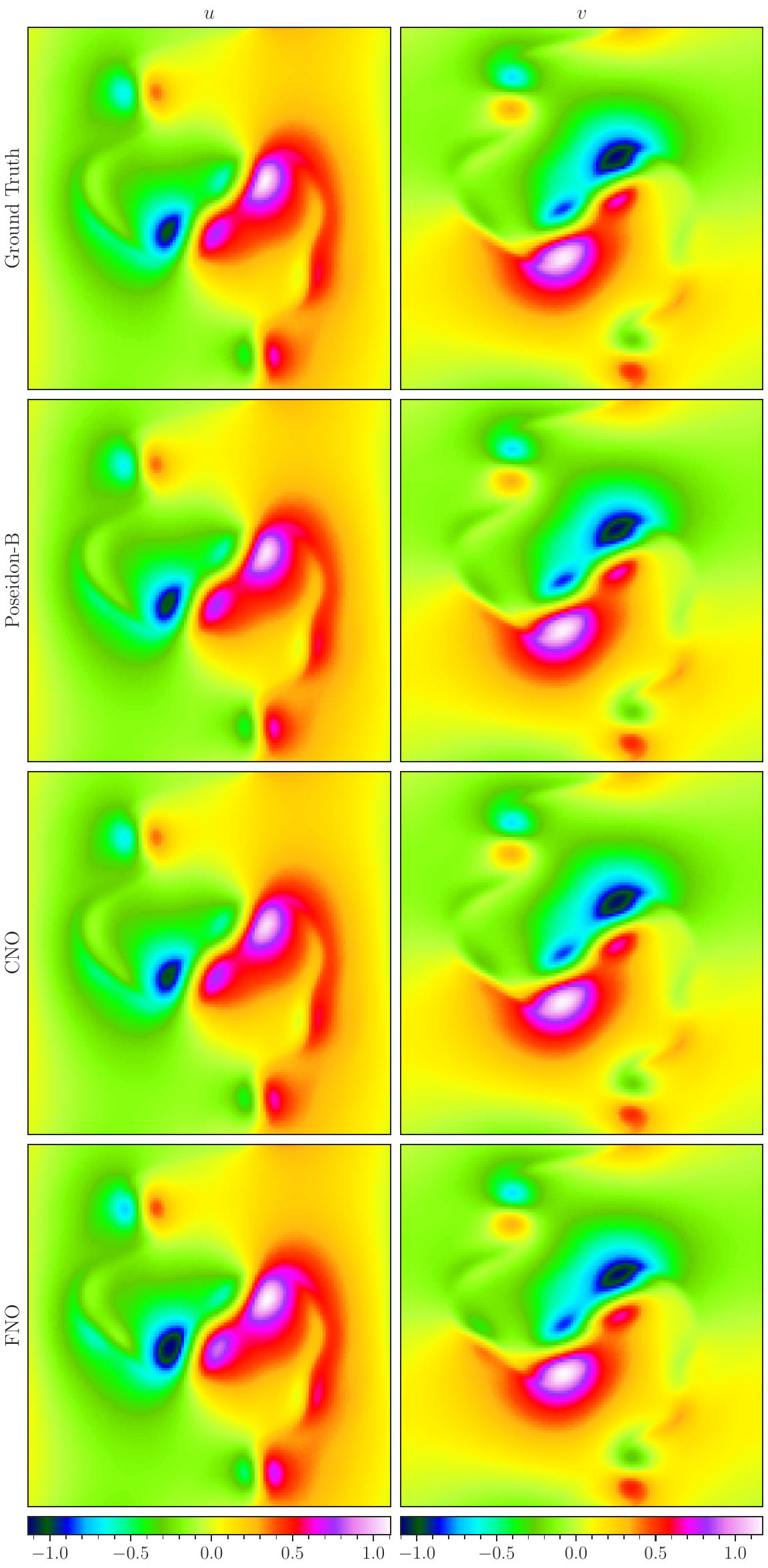}
        \caption{(Top) Ground truth. (From second row onwards) Samples predicted by the finetuned \textsc{Poseidon}-B, CNO, and FNO at $T = 0.7$.}
    \end{subfigure}
    \caption{NS-SVS. Visualization of a random sample.}
    \label{fig:ns_svs}
\end{figure}

\begin{figure}[p]
    \centering
    \begin{subfigure}[t]{\textwidth}
        \centering
        \includegraphics[width=0.85\textwidth]{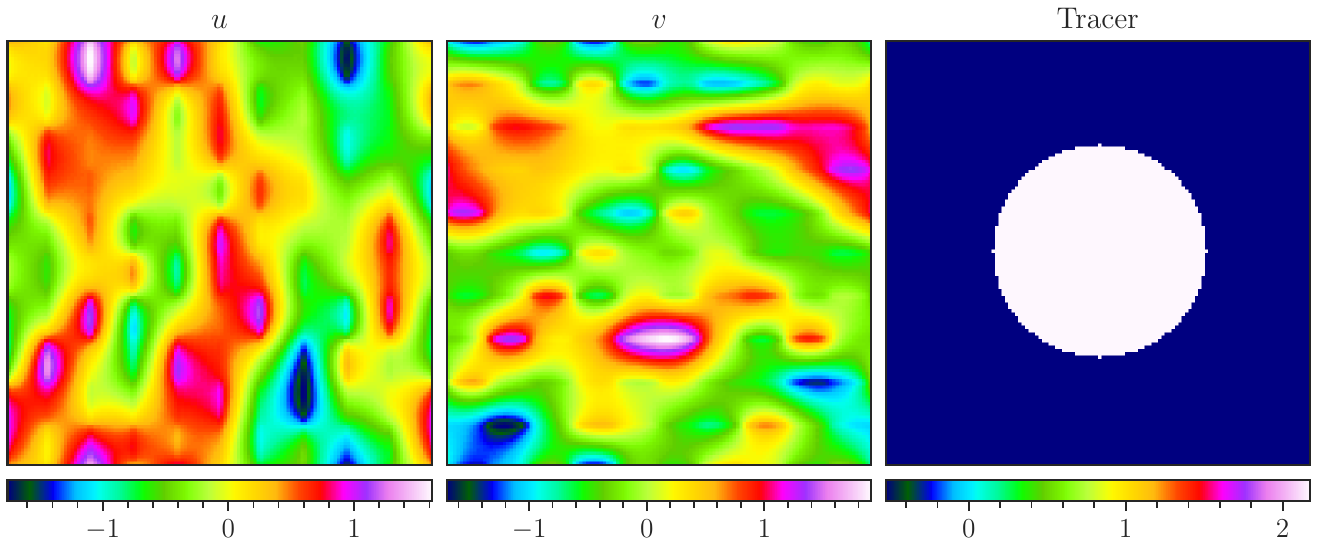}
        \caption{Inputs: horizontal velocity $u$, vertical velocity $v$ and tracer concentration $c$.}
    \end{subfigure}
    \hfill
    \begin{subfigure}[b]{\textwidth}
        \centering
        \includegraphics[width=0.85\textwidth]{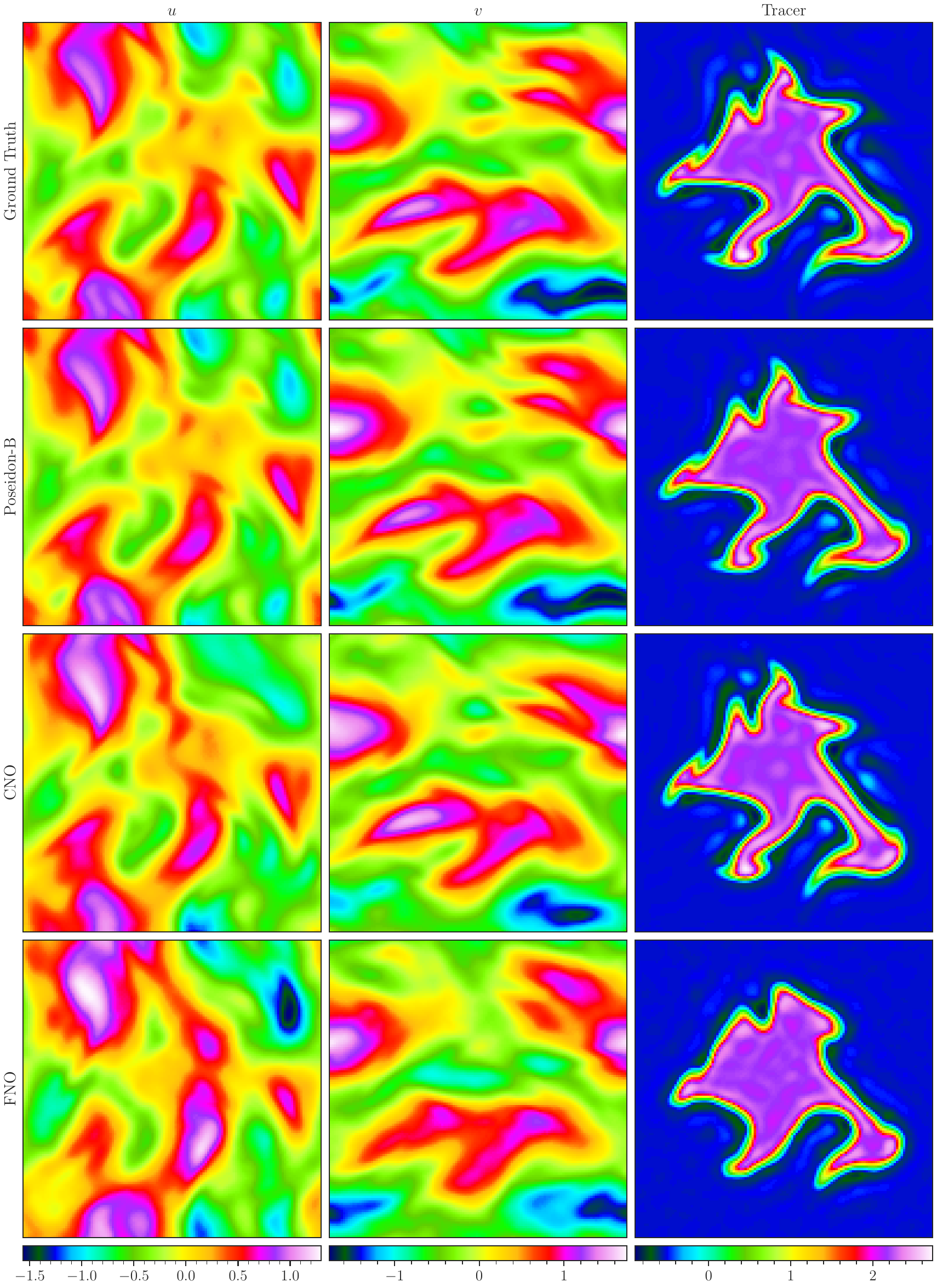}
        \caption{(Top) Ground truth. (From second row onwards) Samples predicted by the finetuned \textsc{Poseidon}-B, CNO, and FNO at $T = 0.7$.}
    \end{subfigure}
    \caption{NS-Tracer-PwC. Visualization of a random sample.}
    \label{fig:ns_tracer_pwc}
\end{figure}

\begin{figure}[p]
    \centering
    \begin{subfigure}[t]{\textwidth}
        \centering
        \includegraphics[width=0.82\textwidth]{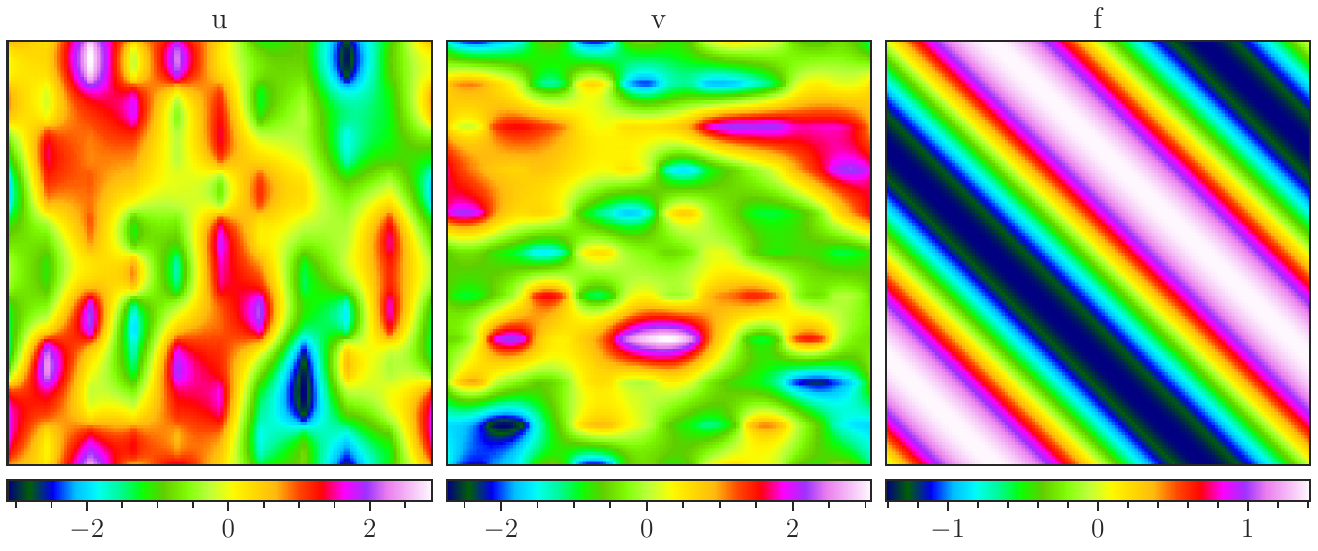}
        \caption{Inputs: horizontal velocity $u$, vertical velocity $v$ and forcing term $f$.}
    \end{subfigure}
    \hfill
    \begin{subfigure}[b]{\textwidth}
        \centering
        \includegraphics[width=0.82\textwidth]{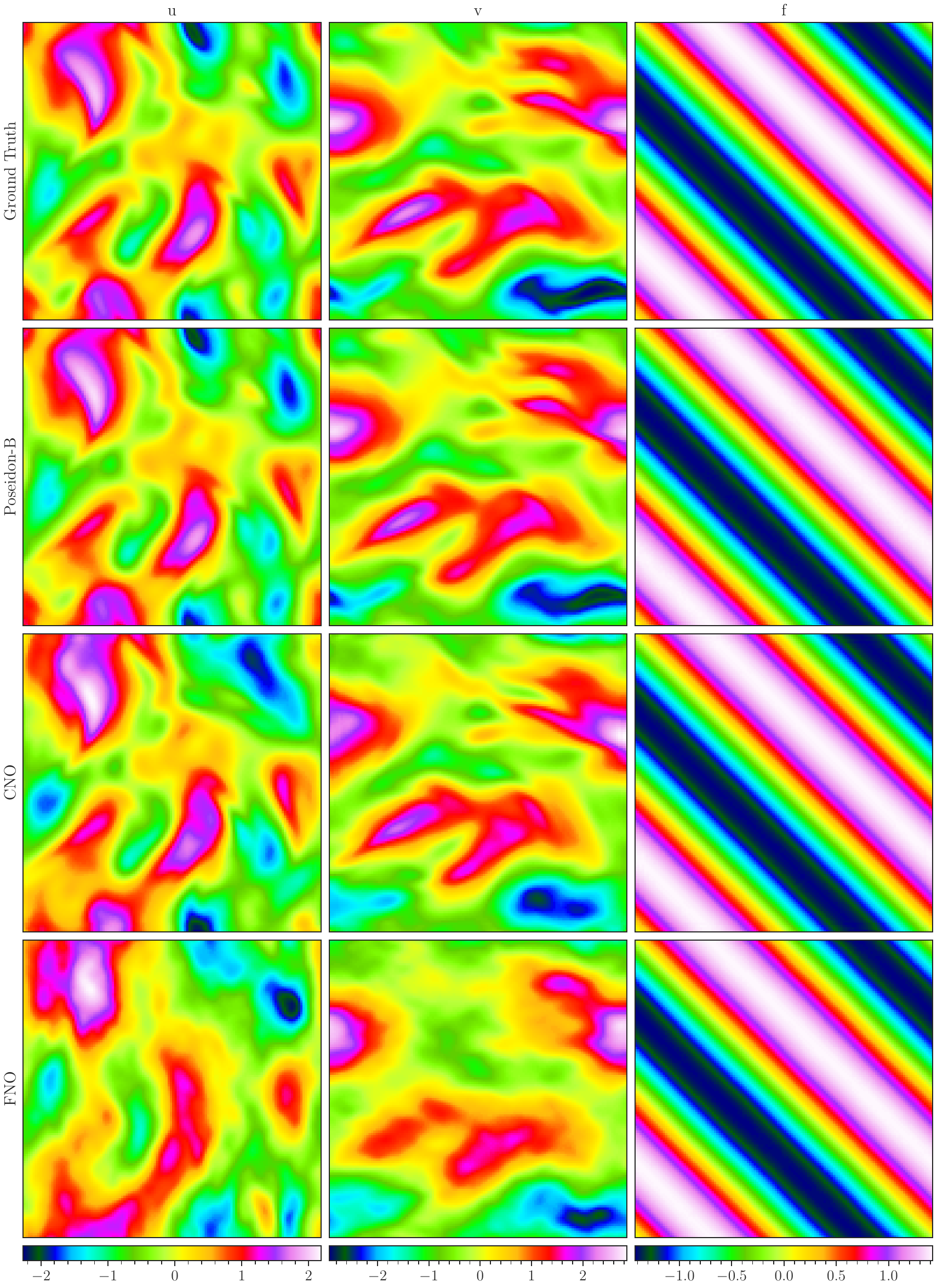}
        \caption{(Top) Ground truth. (From second row onwards) Samples predicted by the finetuned \textsc{Poseidon}-B, CNO, and FNO at $T = 0.7$.}
    \end{subfigure}
    \caption{FNS-KF. Visualization of a random sample.}
    \label{fig:fns_kf}
\end{figure}

\begin{figure}[p]
    \centering
    \begin{subfigure}[t]{\textwidth}
        \centering
        \includegraphics[width=\textwidth]{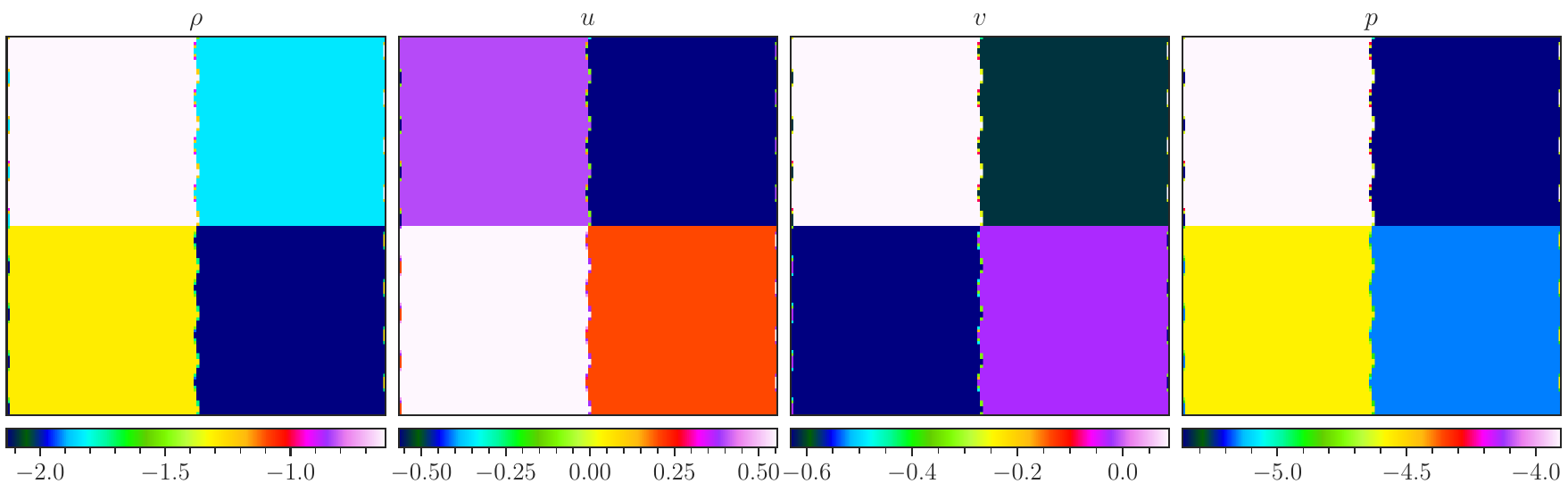}
        \caption{Inputs: density $\rho$, horizontal velocity $u$, vertical velocity $v$ and pressure $p$.}
    \end{subfigure}
    \hfill
    \begin{subfigure}[b]{\textwidth}
        \centering
        \includegraphics[width=\textwidth]{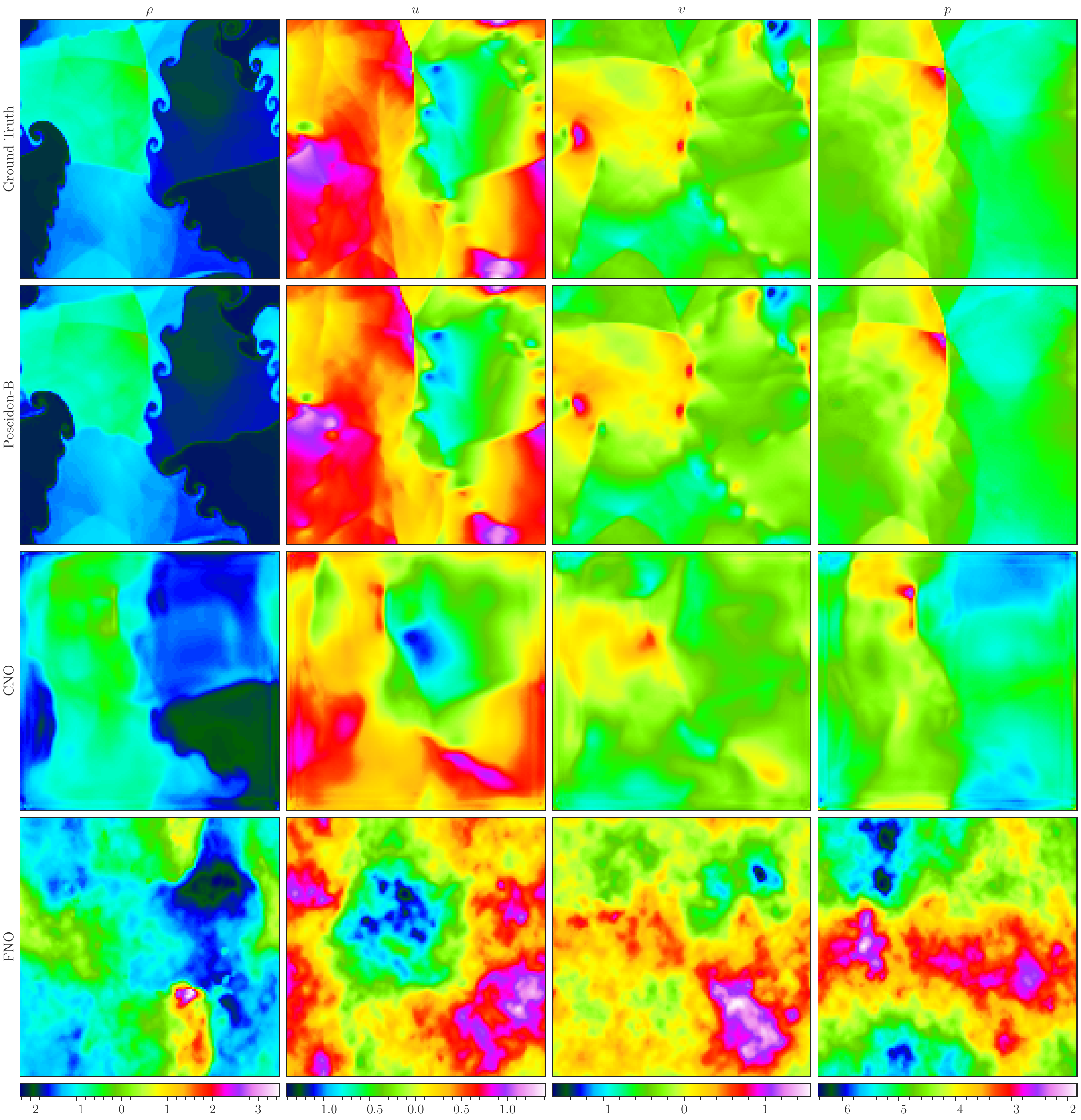}
        \caption{(Top) Ground truth. (From second row onwards) Samples predicted by the finetuned \textsc{Poseidon}-B, CNO, and FNO at $T = 0.7$.}
    \end{subfigure}
    \caption{CE-RPUI. Visualization of a random sample.}
    \label{fig:ce_rpui}
\end{figure}

\begin{figure}[p]
    \centering
    \begin{subfigure}[t]{\textwidth}
        \centering
        \includegraphics[width=\textwidth]{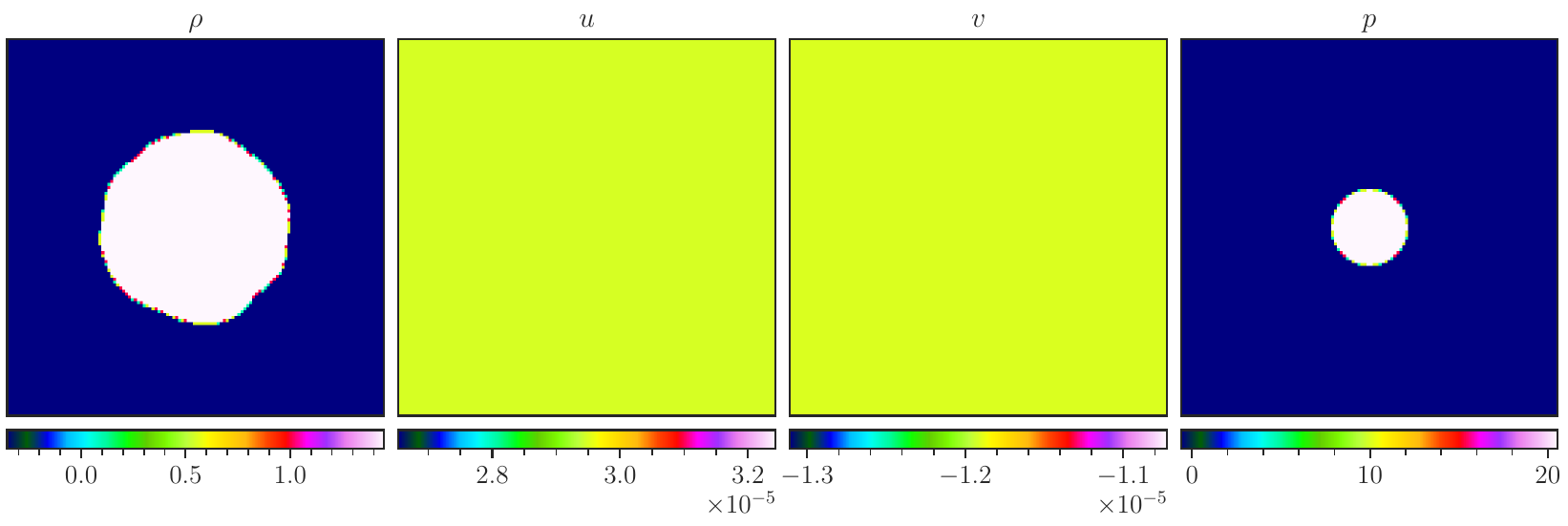}
        \caption{Inputs: density $\rho$, horizontal velocity $u$, vertical velocity $v$ and pressure $p$.}
    \end{subfigure}
    \hfill
    \begin{subfigure}[b]{\textwidth}
        \centering
        \includegraphics[width=\textwidth]{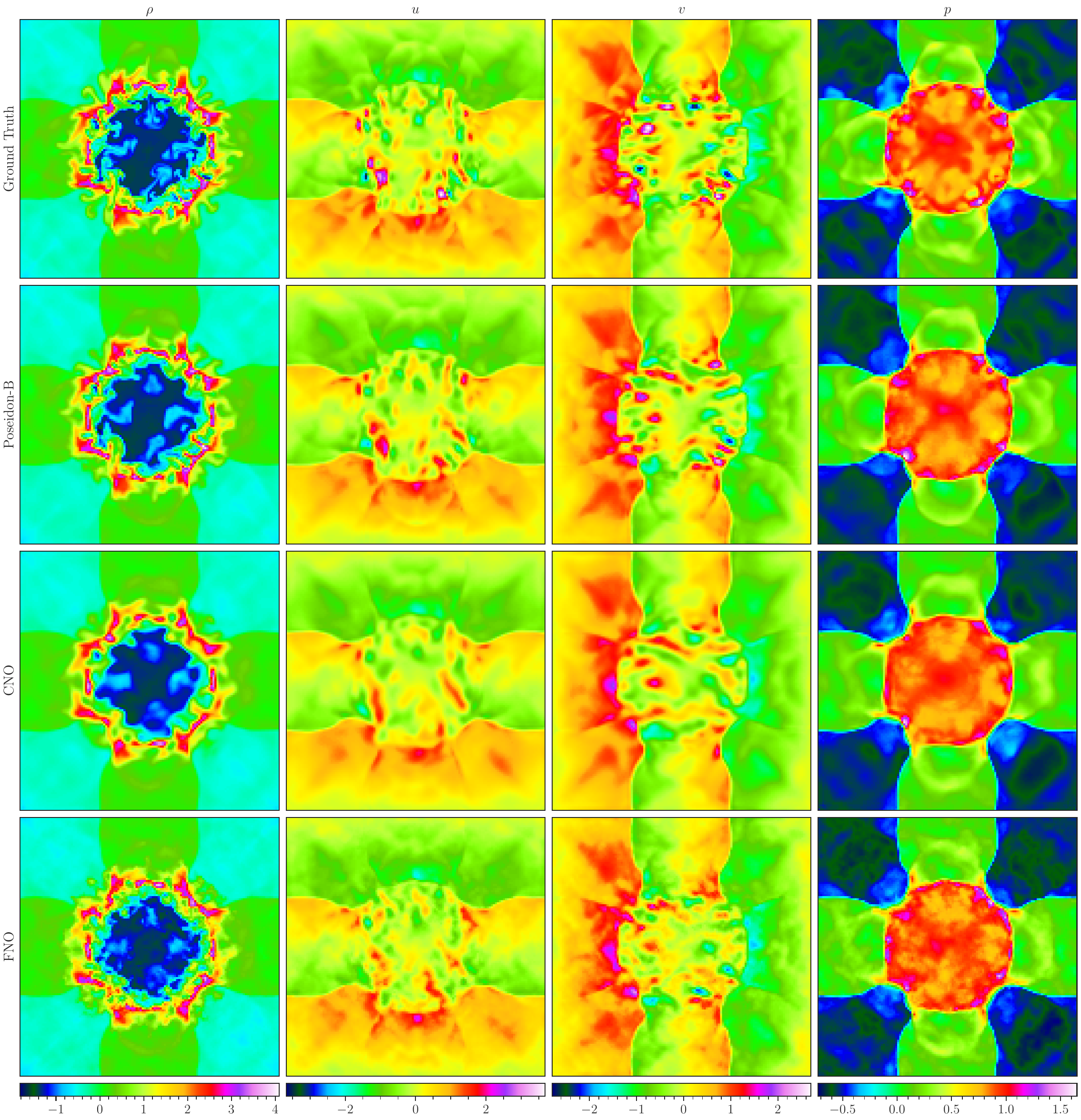}
        \caption{(Top) Ground truth. (From second row onwards) Samples predicted by the finetuned \textsc{Poseidon}-B, CNO, and FNO at $T = 1.4$.}
    \end{subfigure}
    \caption{CE-RM. Visualization of a random sample.}
    \label{fig:ce_rm}
\end{figure}

\begin{figure}[p]
    \centering
    \begin{subfigure}[t]{\textwidth}
        \centering
        \includegraphics[width=\textwidth]{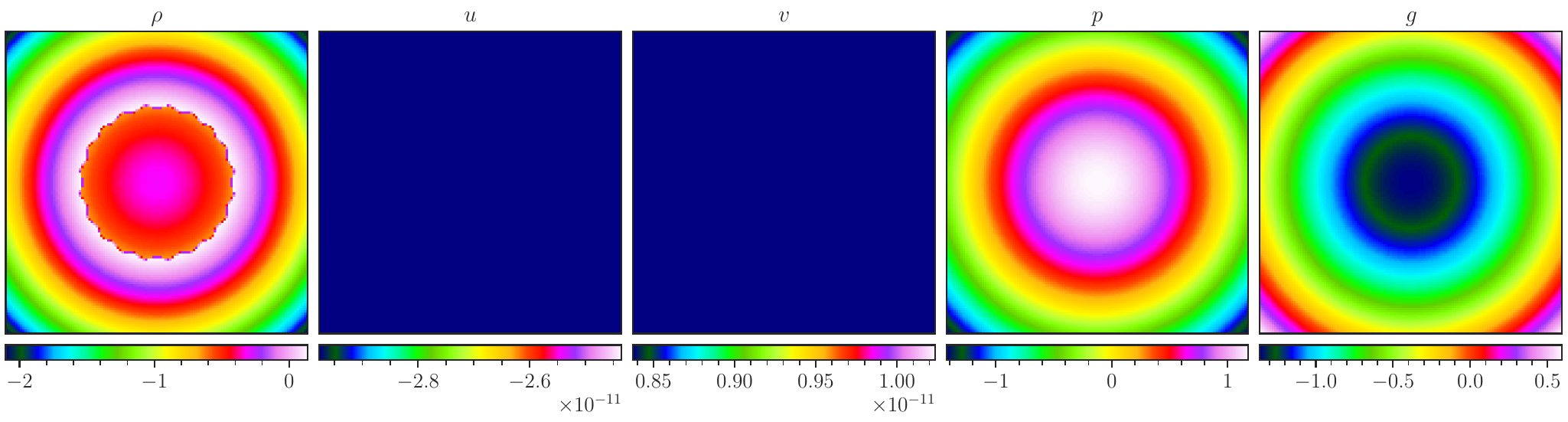}
        \caption{Inputs: density $\rho$, horizontal velocity $u$, vertical velocity $v$, pressure $p$, and gravitational potential $g$.}
    \end{subfigure}
    \hfill
    \begin{subfigure}[b]{\textwidth}
        \centering
        \includegraphics[width=\textwidth]{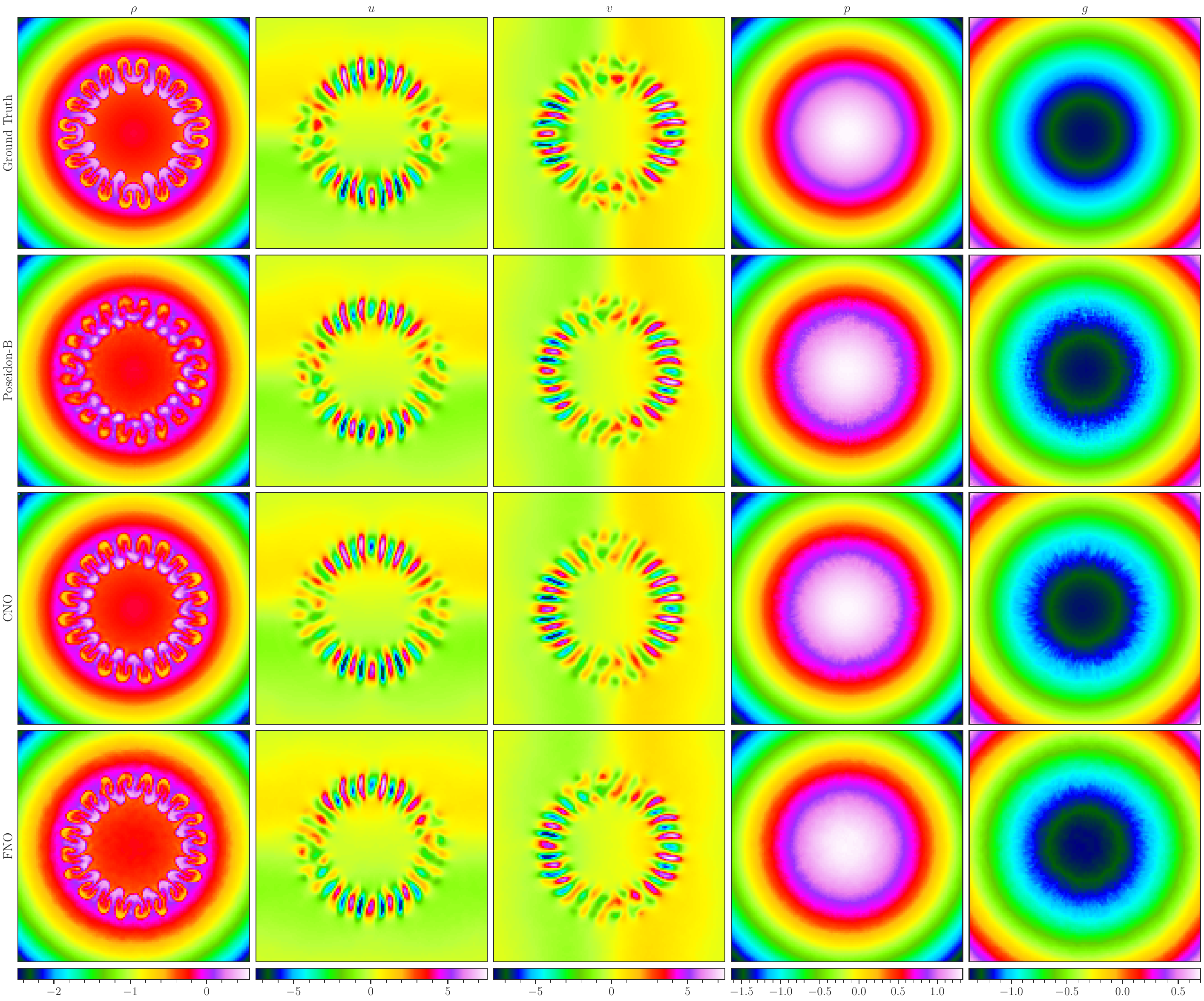}
        \caption{(Top) Ground truth. (From second row onwards) Samples predicted by the finetuned \textsc{Poseidon}-B, CNO, and FNO at the seventh time step.}
    \end{subfigure}
    \caption{CE-RM. Visualization of a random sample.}
    \label{fig:gce_rt}
\end{figure}

\begin{figure}[p]
    \centering
    \begin{subfigure}[t]{\textwidth}
        \centering
        \includegraphics[width=0.55\textwidth]{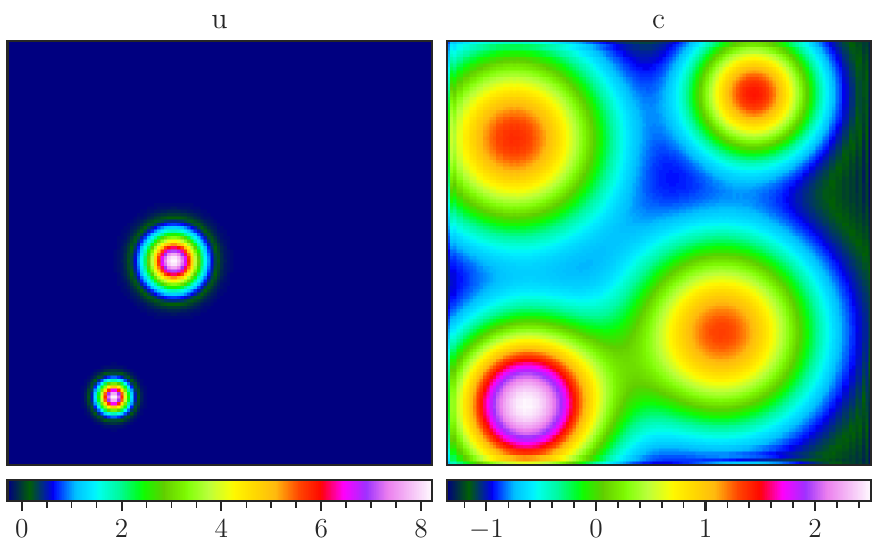}
        \caption{Inputs: displacement $u$ and propagation field $c$.}
    \end{subfigure}
    \hfill
    \begin{subfigure}[b]{\textwidth}
        \centering
        \includegraphics[width=0.55\textwidth]{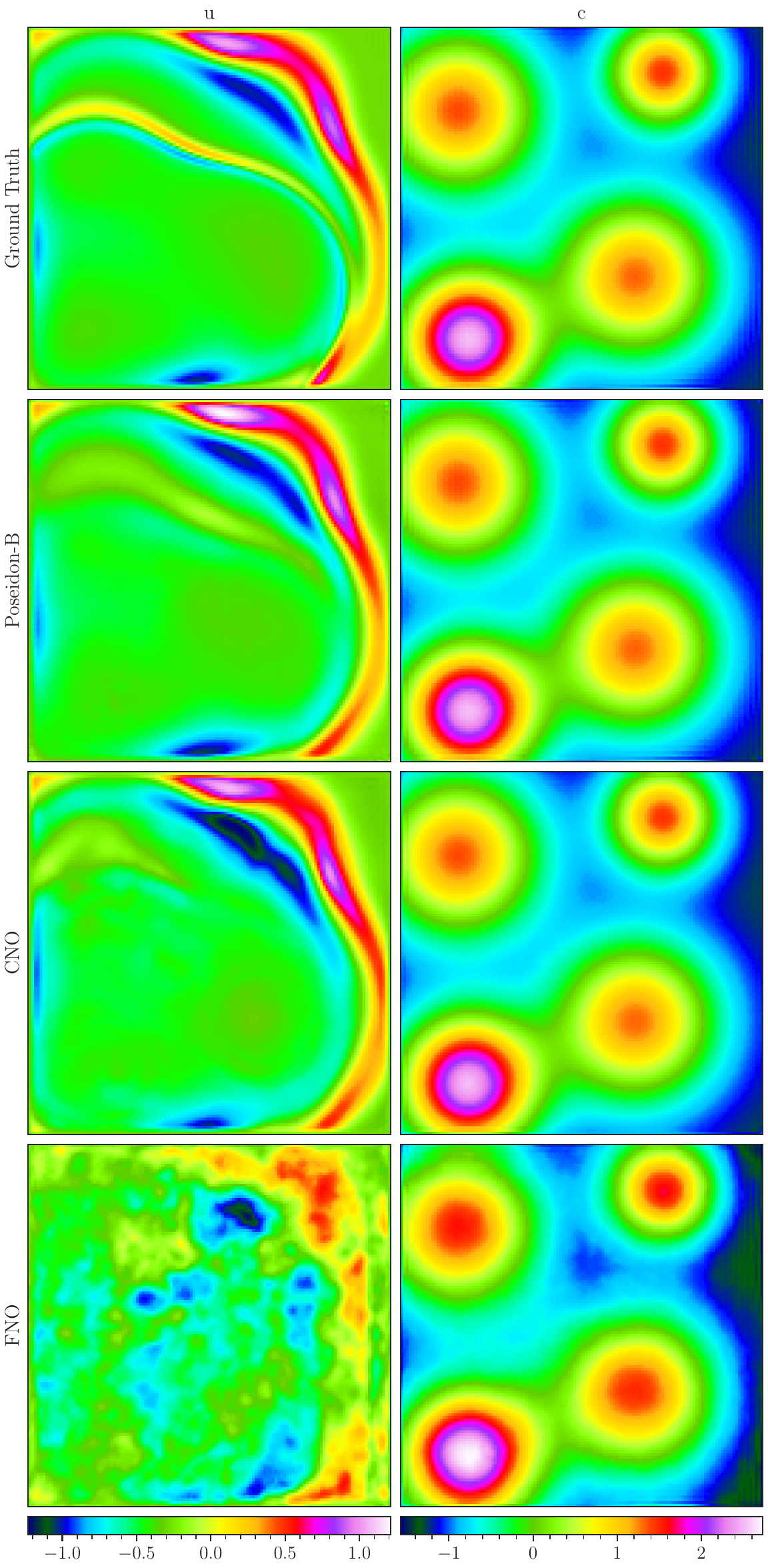}
        \caption{(Top) Ground truth. (From second row onwards) Samples predicted by the finetuned \textsc{Poseidon}-B, CNO, and FNO at the 14-th time step.}
    \end{subfigure}
    \caption{Wave-Gauss. Visualization of a random sample.}
    \label{fig:wave_gaussians}
\end{figure}

\begin{figure}[p]
    \centering
    \begin{subfigure}[t]{\textwidth}
        \centering
        \includegraphics[width=0.55\textwidth]{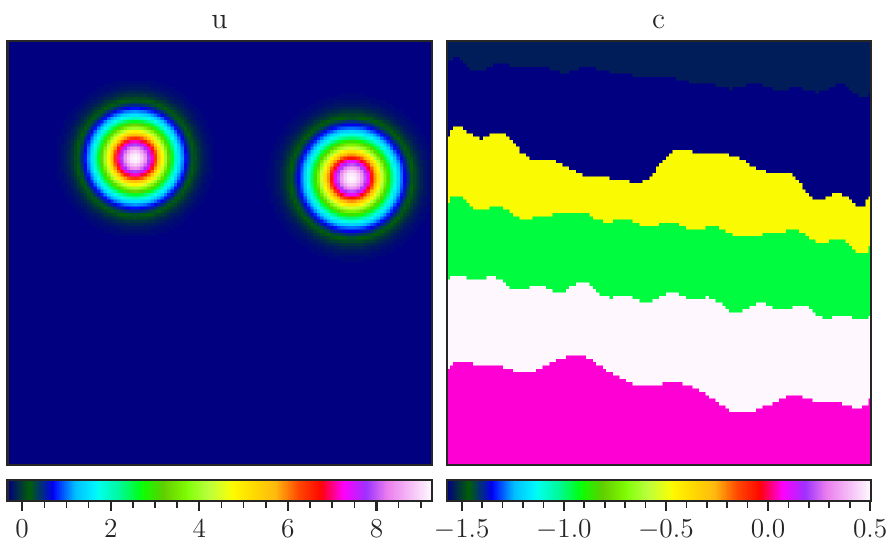}
        \caption{Inputs: displacement $u$ and propagation field $c$.}
    \end{subfigure}
    \hfill
    \begin{subfigure}[b]{\textwidth}
        \centering
        \includegraphics[width=0.55\textwidth]{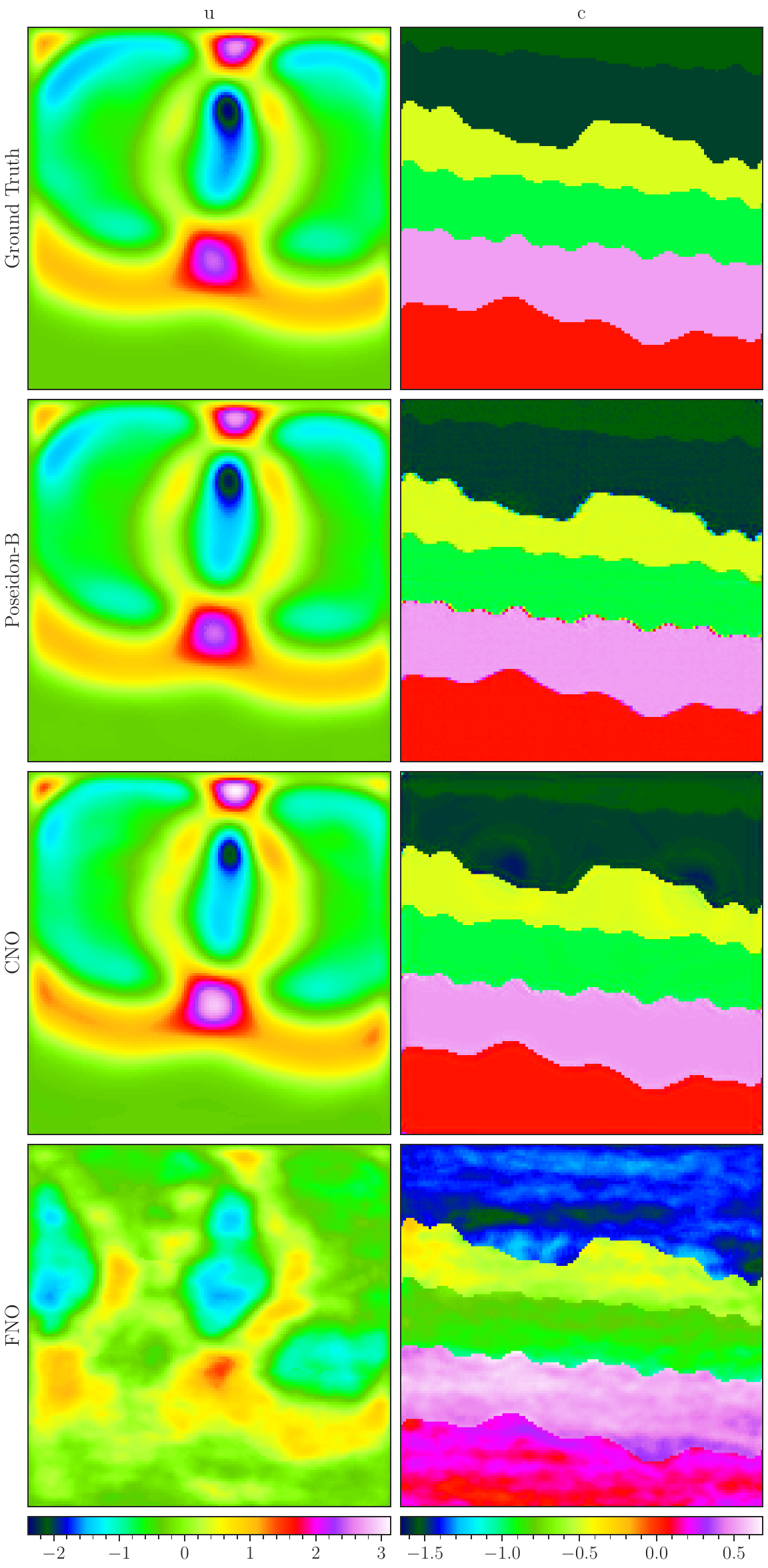}
        \caption{(Top) Ground truth. (From second row onwards) Samples predicted by the finetuned \textsc{Poseidon}-B, CNO, and FNO at $T = 0.7$.}
    \end{subfigure}
    \caption{Wave-Layer. Visualization of a random sample.}
    \label{fig:wave_layer}
\end{figure}

\begin{figure}[p]
    \centering
    \begin{subfigure}[t]{\textwidth}
        \centering
        \includegraphics[width=0.28\textwidth]{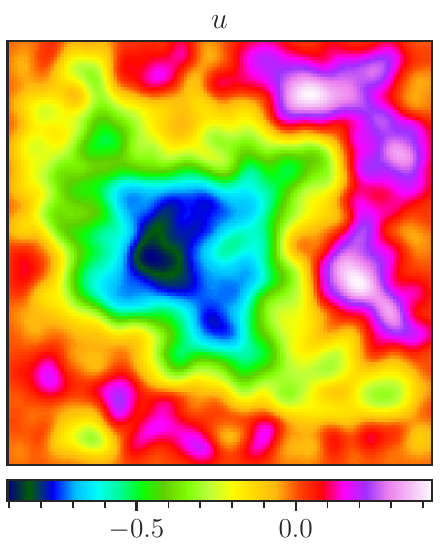}
        \caption{Inputs: concentration $u$.}
    \end{subfigure}
    \hfill
    \begin{subfigure}[b]{\textwidth}
        \centering
        \includegraphics[width=0.28\textwidth]{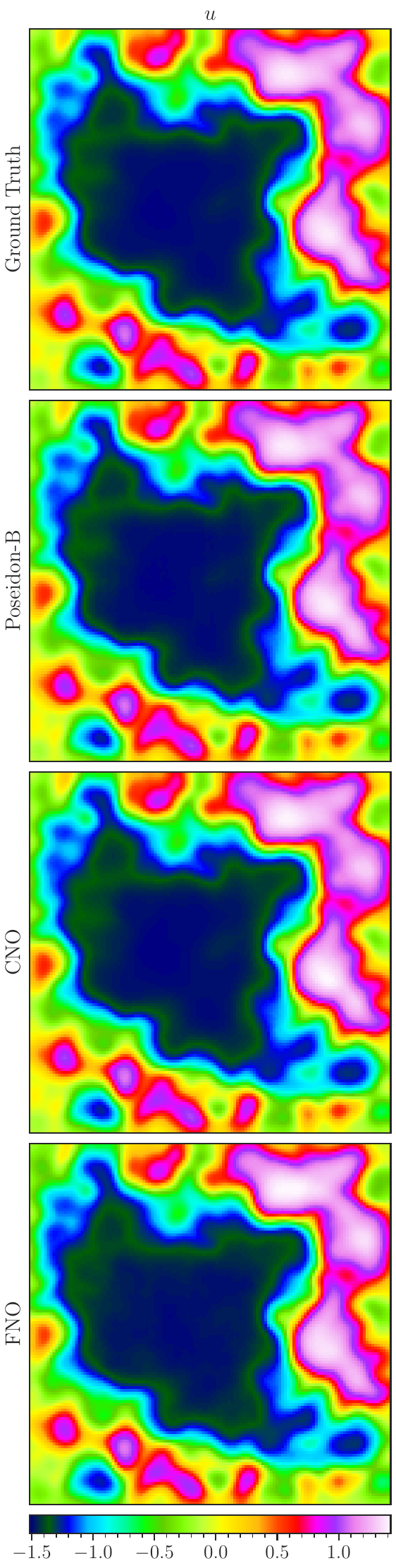}
        \caption{(Top) Ground truth. (From second row onwards) Samples predicted by the finetuned \textsc{Poseidon}-B, CNO, and FNO at the 14-th time step.}
    \end{subfigure}
    \caption{ACE. Visualization of a random sample.}
    \label{fig:ace}
\end{figure}

\begin{figure}[p]
    \centering
    \begin{subfigure}[t]{\textwidth}
        \centering
        \includegraphics[width=0.28\textwidth]{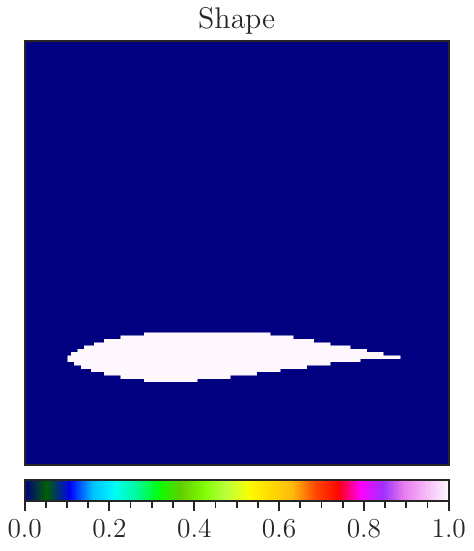}
        \caption{Inputs: airfoil shape function}
    \end{subfigure}
    \hfill
    \begin{subfigure}[b]{\textwidth}
        \centering
        \includegraphics[width=0.28\textwidth]{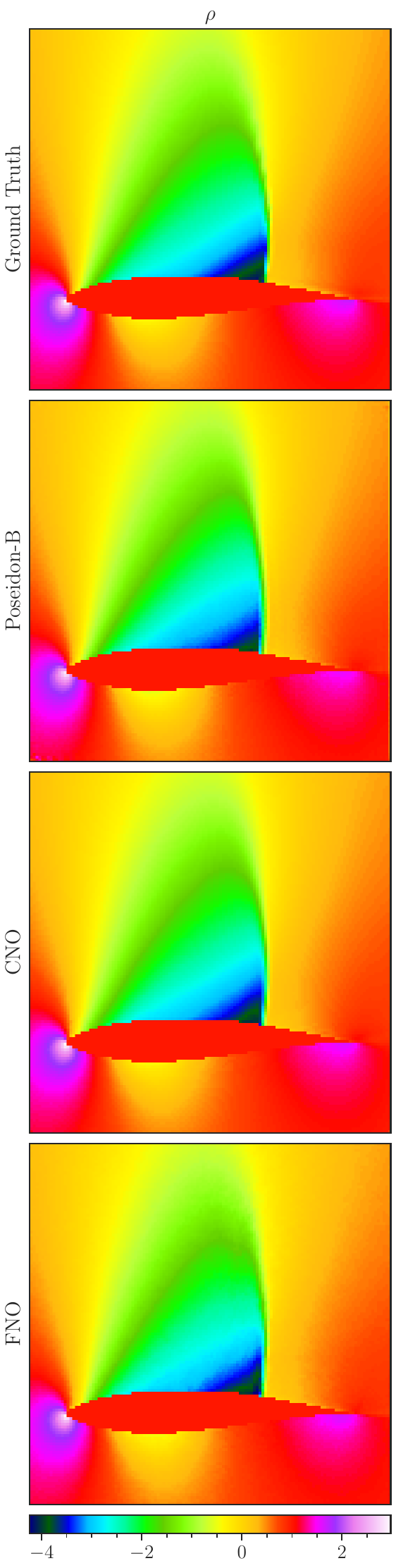}
        \caption{(Top) Ground truth. (From second row onwards) Samples (density $\rho$) predicted by the finetuned \textsc{Poseidon}-B, CNO, and FNO.}
    \end{subfigure}
    \caption{SE-AF. Visualization of a random sample.}
    \label{fig:se_af}
\end{figure}

\begin{figure}[p]
    \centering
    \begin{subfigure}[t]{\textwidth}
        \centering
        \includegraphics[width=0.28\textwidth]{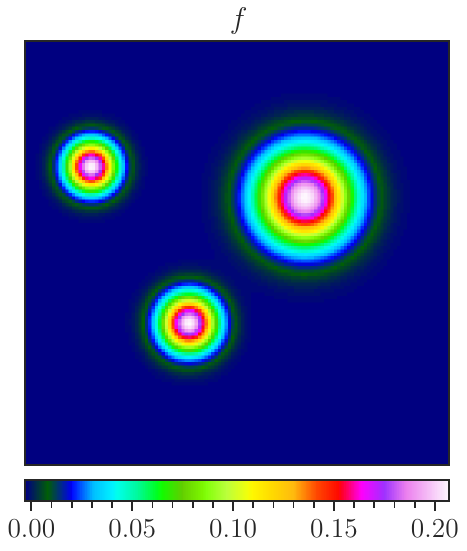}
        \caption{Inputs: source term $f$.}
    \end{subfigure}
    \hfill
    \begin{subfigure}[b]{\textwidth}
        \centering
        \includegraphics[width=0.28\textwidth]{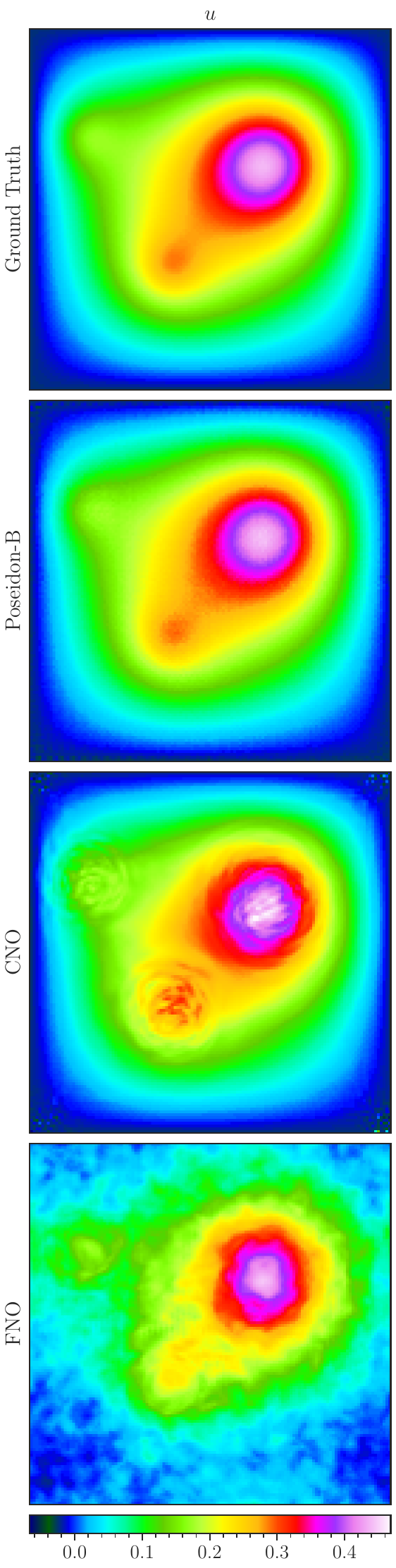}
        \caption{(Top) Ground truth. (From second row onwards) Samples (solution $u$) predicted by the finetuned \textsc{Poseidon}-B, CNO, and FNO.}
    \end{subfigure}
    \caption{Poisson-Gauss. Visualization of a random sample.}
    \label{fig:poisson_gaussians}
\end{figure}

\begin{figure}[p]
    \centering
    \begin{subfigure}[t]{\textwidth}
        \centering
        \includegraphics[width=0.28\textwidth]{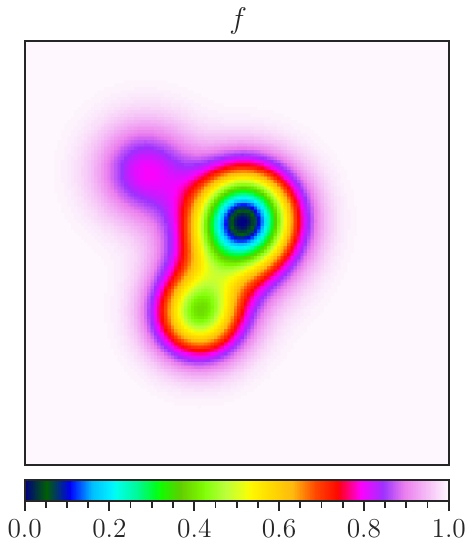}
        \caption{Inputs: propagation speed $f$.}
    \end{subfigure}
    \hfill
    \begin{subfigure}[b]{\textwidth}
        \centering
        \includegraphics[width=0.28\textwidth]{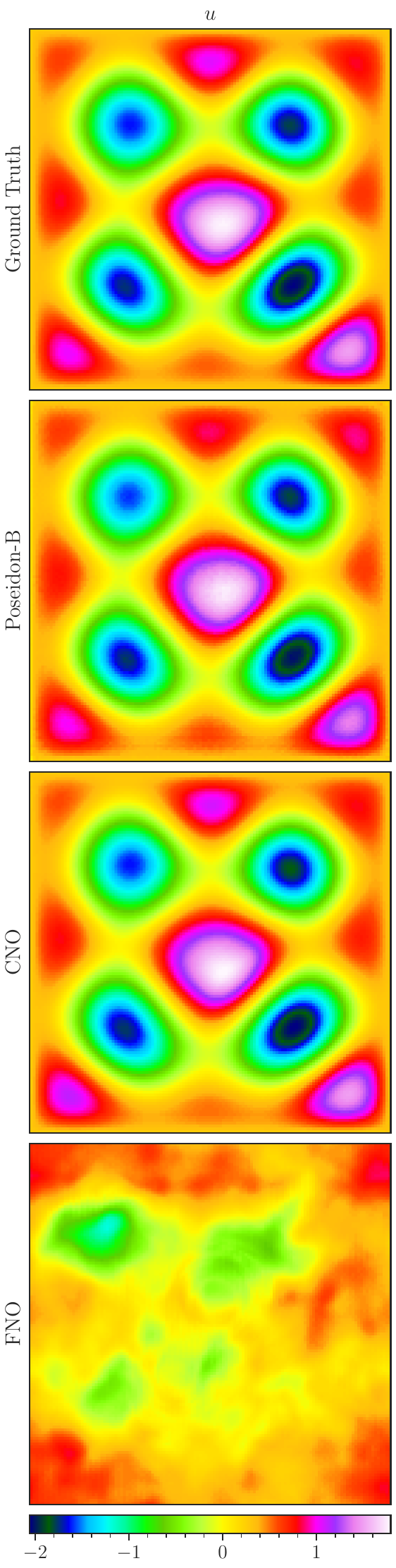}
        \caption{(Top) Ground truth. (From second row onwards) Samples (solution $u$) predicted by the finetuned \textsc{Poseidon}-B, CNO, and FNO.}
    \end{subfigure}
    \caption{Helmholtz. Visualization of a random sample.}
    \label{fig:helmholtz}
\end{figure}

In Figure \ref{fig:error_histograms}, we plot the distribution of errors across the test set for all downstream tasks with the \textsc{Poseidon}-B model, finetuned with 128 trajectories (samples).
\clearpage

\section{Computational Resources}
\label{sec:computational_resources}
All experiments were run on different types of GPUs, on the Euler cluster of ETH Zurich. Depending on the experiment, we use between 8 and 128 CPU cores and up to 512GB of RAM, with pretrainings using the most CPU cores and RAM. However, we note that this is more than is actually needed, as we tried to minimize being bottlenecked by dataloading. For all our models and baselines, we used consumer-grade GPUs with 24GB of VRAM. All our pretrainings were performed in (data-)parallel on 8 NVIDIA GeForce RTX 4090 GPUs. All finetuning experiments and most scratch trainings were performed on a single GPU, while some scratch training runs with a lot of data were performed in (data-)parallel. Pretraining times can be read off from Table \ref{tab:training_time}.
\begin{table}[htb]
    \centering
    \caption{Approximate pretraining times on 8 NVIDIA GeForce RTX 4090 GPUs. Batch sizes are given in parentheses.}
    \begin{tabular}{c c c c}
    \toprule
     \textsc{Poseidon}-L &  \textsc{Poseidon}-B &  \textsc{Poseidon}-T & CNO-FM\\
     \midrule\midrule
     165h (16) & 118h (40) & 22h (80) & 178h (32) \\
       \bottomrule
    \end{tabular}
    \label{tab:training_time}
\end{table}

In Table \ref{tab:inference_time}, we provide an overview over the inference times of each model for a single call to it. We observe from this table that even the biggest \textsc{Poseidon}-L has an (average) inference time of less than $10^{-2}$ secs. This is contrast to the PDE solvers that were used to generate the data in this paper. Their run times, for a resolution of $128^2$ ranged from anywhere between $0.1$ sec (for highly optimized GPU solver \cite{azeban} for the NS datasets to $10$ secs for FENICS \cite{fenics} FEM solver for the Poisson-Gauss dataset to approx $100$ secs for NEWTUN \cite{LMR1} for generating the airfoils datasets to $500$ secs for the well-balanced scheme to generate the GCE-RT. Thus, we observe a gain in inference time from anywhere between $1-5$ orders of magnitude.

\begin{table}[htb]
    \centering
    \caption{Approximate inference times (per call and normalized to a single sample) for different models, all reported on a NVIDIA GeForce RTX 4090 GPU for the FNS-KF experiment. Batch sizes are given in parentheses. We note that the values given here are just proxies as this was not tested in a controlled environment.}
    \begin{tabular}{r c}
    \toprule
     Model & Approximate inference time\\
     \midrule\midrule
      \textsc{Poseidon}-L   &  4 ms (16) \\\midrule
       \textsc{Poseidon}-B  & 2.9ms (40)\\\midrule
       \textsc{Poseidon}-T & 1.6ms (40)\\\midrule
       CNO-FM & 1.8ms (32)\\\midrule
       MPP-B & 10ms (4)\\\midrule
       CNO & 0.9ms (32)\\\midrule
       scOT & 3ms (40)\\\midrule
       FNO & 2ms (40)\\
       \bottomrule
    \end{tabular}
    \label{tab:inference_time}
\end{table}

\section{Pretrained Models, Datasets, and Source Code}
\label{sec:models_datasets_code}

The source code corresponding to this work is available on Github (\href{https://github.com/camlab-ethz/poseidon}{https://github.com/camlab-ethz/poseidon}). Everything is tightly integrated into Huggingface Transformers \cite{Wolf_Transformers_State-of-the-Art_Natural_2020} and we make heavy use of Huggingface Accelerate for distributed training.

In addition to the code, we make (pretrained) models and datasets available on the Huggingface Hub (\href{https://huggingface.co/camlab-ethz}{https://huggingface.co/camlab-ethz}), see the \href{https://huggingface.co/collections/camlab-ethz/poseidon-664fa125729c53d8607e209a}{Poseidon collection} for pretrained models and pretraining datasets, the \href{https://huggingface.co/collections/camlab-ethz/poseidon-downstream-tasks-664fa237cd6b0c097971ef14}{Poseidon -- Downstream Tasks collection} for all downstream tasks, or the \href{https://huggingface.co/collections/camlab-ethz/pdegym-665472c2b1181f7d10b40651}{\textsc{PDEgym} collection} for all datasets in \textsc{PDEgym}.

\clearpage
\section{Visualizations}

\end{document}